%% file: main.tex
\RequirePackage{snapshot}

\documentclass[10pt,twocolumn,letterpaper]{article}

\usepackage[accsupp]{axessibility}  %
\usepackage{cvpr}           

\makeatletter
\@namedef{ver@everyshi.sty}{}
\makeatother
\usepackage{tikz}

\usepackage{graphicx}
\usepackage{amsmath}
\usepackage{amssymb}
\usepackage{booktabs}
\input{math_commands}

\usepackage[pagebackref,breaklinks,colorlinks]{hyperref}

\usepackage{amsfonts}       %
\usepackage{microtype}      %
\usepackage{makecell}
\usepackage{caption}
\usepackage{subcaption}
\usepackage{multirow}
\usepackage[section]{placeins}

\usepackage{graphicx}
\usepackage{svg}
\usepackage{algorithm} 
\usepackage{algpseudocode}

\usepackage{float}

\newcommand{\todo}[1]{\textcolor{red}{TODO: #1}}

\newcommand{\chenlin}[1]{\textcolor{cyan}{[CM: #1]}}

\usepackage[capitalize]{cleveref}
\crefname{section}{Sec.}{Secs.}
\Crefname{section}{Section}{Sections}
\Crefname{table}{Table}{Tables}
\crefname{table}{Tab.}{Tabs.}

\def\cvprPaperID{10027} %
\def\confName{CVPR}
\def\confYear{2023}

\begin{document}

\title{On Distillation of Guided Diffusion Models}

\author{Chenlin Meng$^{1}$\\
{\tt\small chenlin@cs.stanford.edu}
\and
Robin Rombach$^2$\\
{\tt\small robin@stability.ai}
\and
Ruiqi Gao$^3$\\
{\tt\small ruiqig@google.com}
\and
Diederik Kingma$^3$\\
{\tt\small durk@google.com}
\and
Stefano Ermon$^1$\\
{\tt\small ermon@cs.stanford.edu}
\and
Jonathan Ho$^3$\\
{\tt\small jonathanho@google.com}
\and
Tim Salimans$^3$\\
{\tt\small salimans@google.com}
\and
$^1$Stanford University \qquad $^2$Stability AI \& LMU Munich \qquad $^3$ Google Research, Brain Team
}

\input{figures}

\twocolumn[{%
\vspace{-5em}
\maketitle%
\newteaser%
}]

\maketitle

\renewcommand{\thefootnote}{\fnsymbol{footnote}}
\footnotetext[1]{Work partially done during an internship at Google}

\input{0_abstract}
\input{1_intro}

\input{2_background}

\input{3_method}

\input{4_experiment}

\input{5_related}
\input{6_conclusion}

{\small
\bibliographystyle{ieee_fullname}
\bibliography{egbib}
}
\clearpage 
\input{appendix}

\end{document}

%% file: math_commands.tex
\usepackage{amsmath}
\usepackage{bm}
\usepackage{caption}
\usepackage{adjustbox}

\def\eqref#1{equation~\ref{#1}}

\def\1{\bm{1}}

\def\eps{{\epsilon}}

\def\rvepsilon{{\mathbf{\epsilon}}}

\def\rvv{{\mathbf{v}}}

\def\rvx{{\mathbf{x}}}

\def\rvz{{\mathbf{z}}}

\DeclareMathAlphabet{\mathsfit}{\encodingdefault}{\sfdefault}{m}{sl}
\SetMathAlphabet{\mathsfit}{bold}{\encodingdefault}{\sfdefault}{bx}{n}

\newcommand{\x}{\textbf{x}}

%% file: figures.tex
\newcommand{\teaser}{
  \vspace{-1em}
	\centering
	\includegraphics[width=1.00\textwidth, trim=0em 0em 0em 0em, clip]{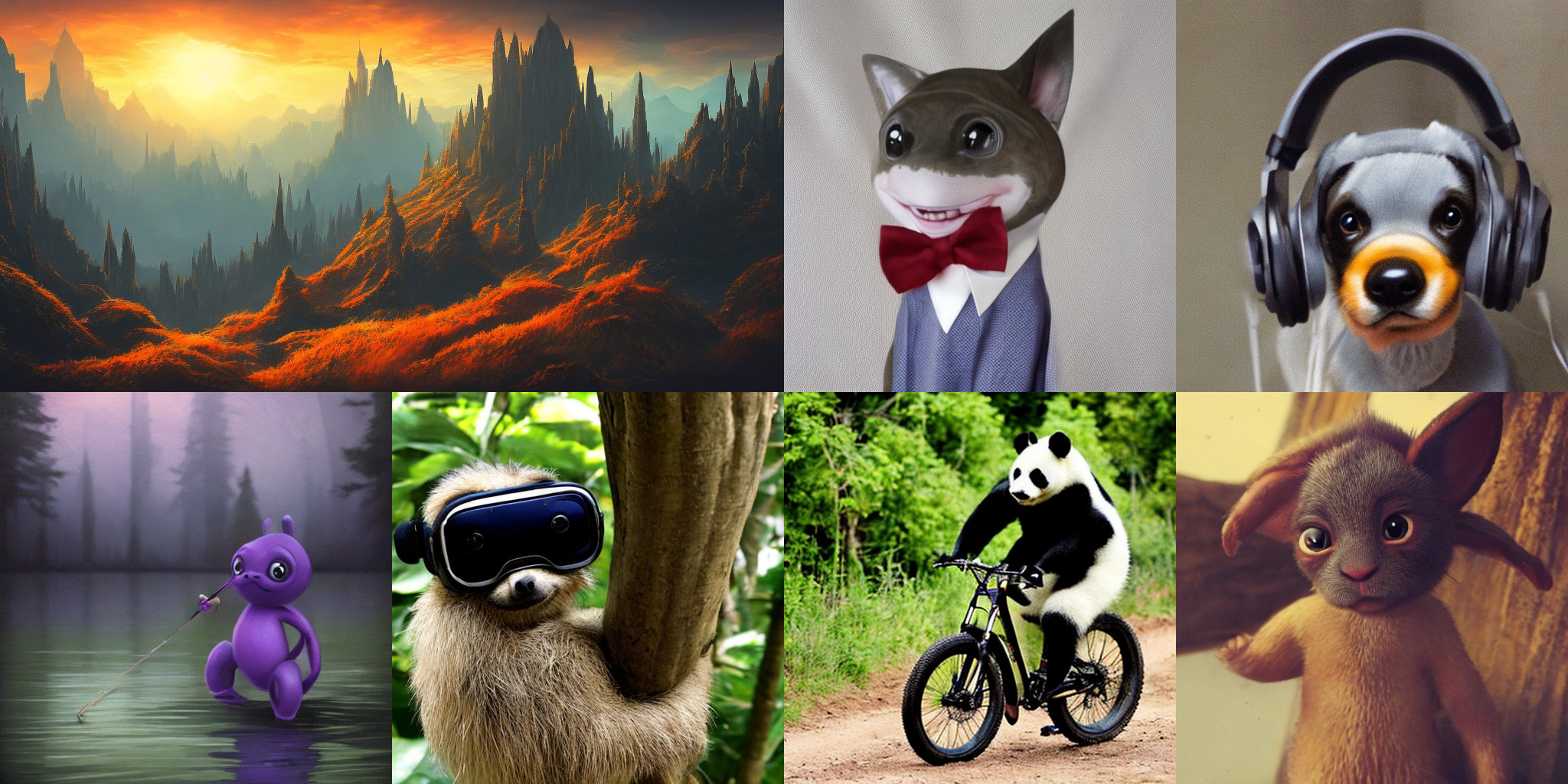}
  \captionof{figure}{Distilled Stable Diffusion samples produced in 4 sampling steps. \todo{A caption}\chenlin{perhaps we can add one or two more images per row? now the images look really large}}
	\label{fig:teaser}
  \vspace{1.5em}
}

\newcommand{\inpainting}{
\begin{figure*}
\centering
\includegraphics[width=\textwidth]{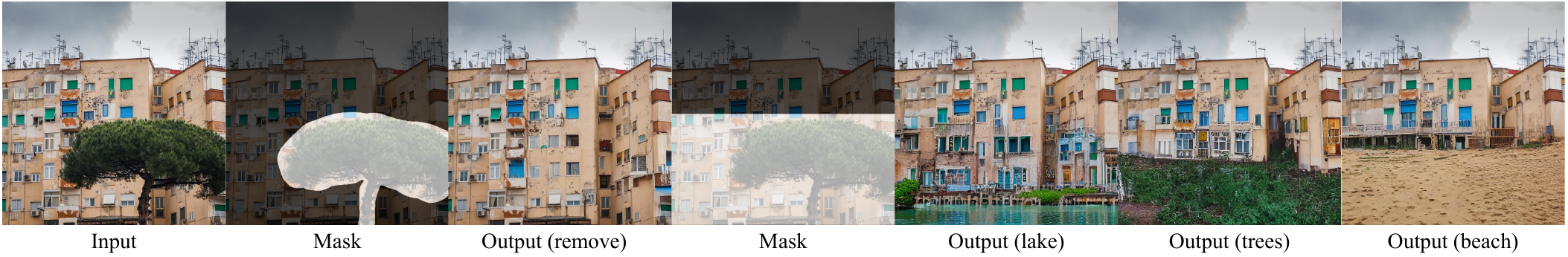}
\caption{Image inpainting with our distilled Stable Diffusion model (4 denoising steps). Our model is able to generate high-quality image inpainting results using 4  denoising steps on unseen data.
}
\label{fig:image_inpainting}
\end{figure*}
}

\newcommand{\imgimg}{
\begin{figure*}%
  \vspace{-15pt}
    \centering
    \includegraphics[width=\linewidth]{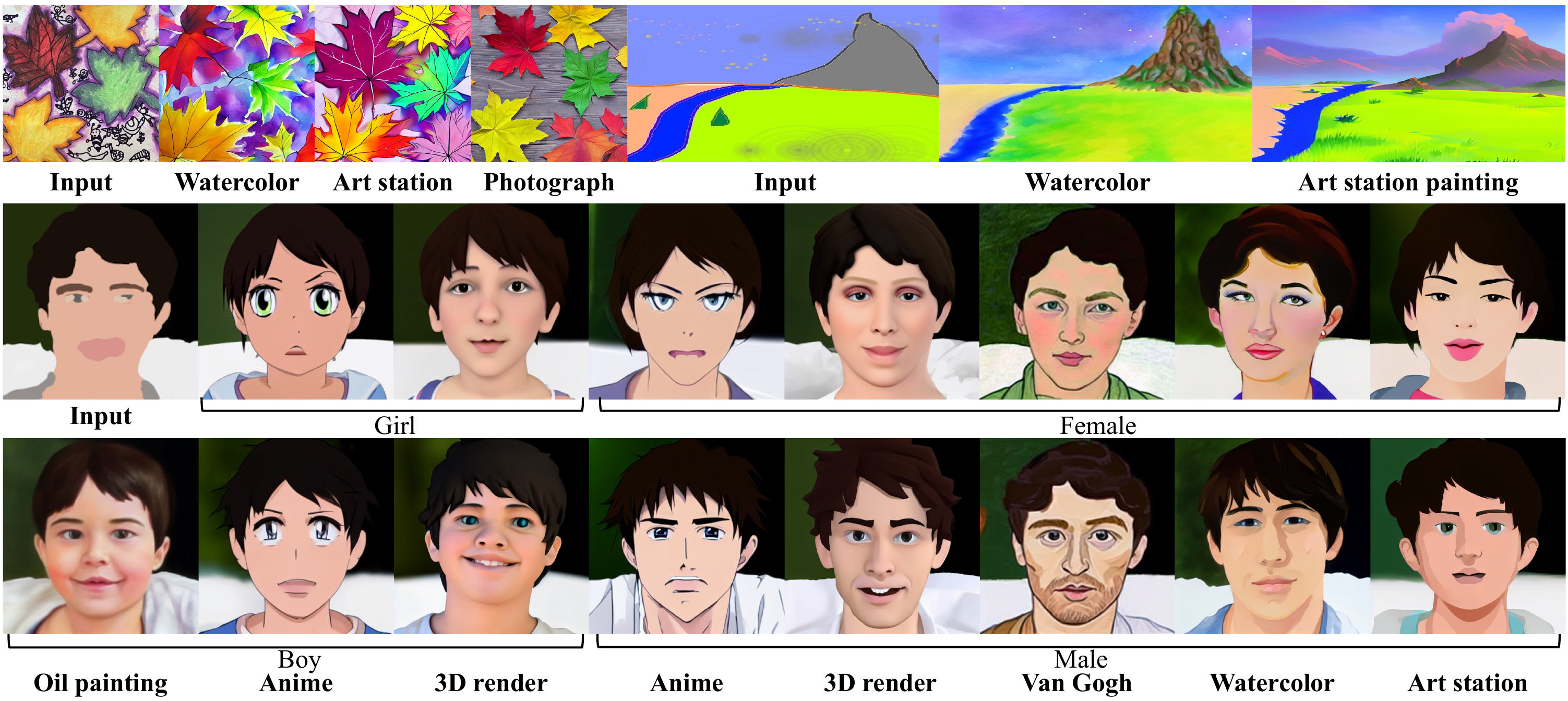}
    \caption{Text-guided image-to-image translation~\cite{meng2021sdedit} with the distilled Stable Diffusion model (3 denoising steps). We observe that our model is able to generate high-quality and faithful outputs using only 3 denoising steps.
    \vspace{-1em}}
    \label{fig:img2img}
\end{figure*}
}

\newcommand{\newteaser}{
  \vspace{-1em}
	\centering
	\includegraphics[width=1.00\textwidth, trim=0em 0em 0em 0em, clip]{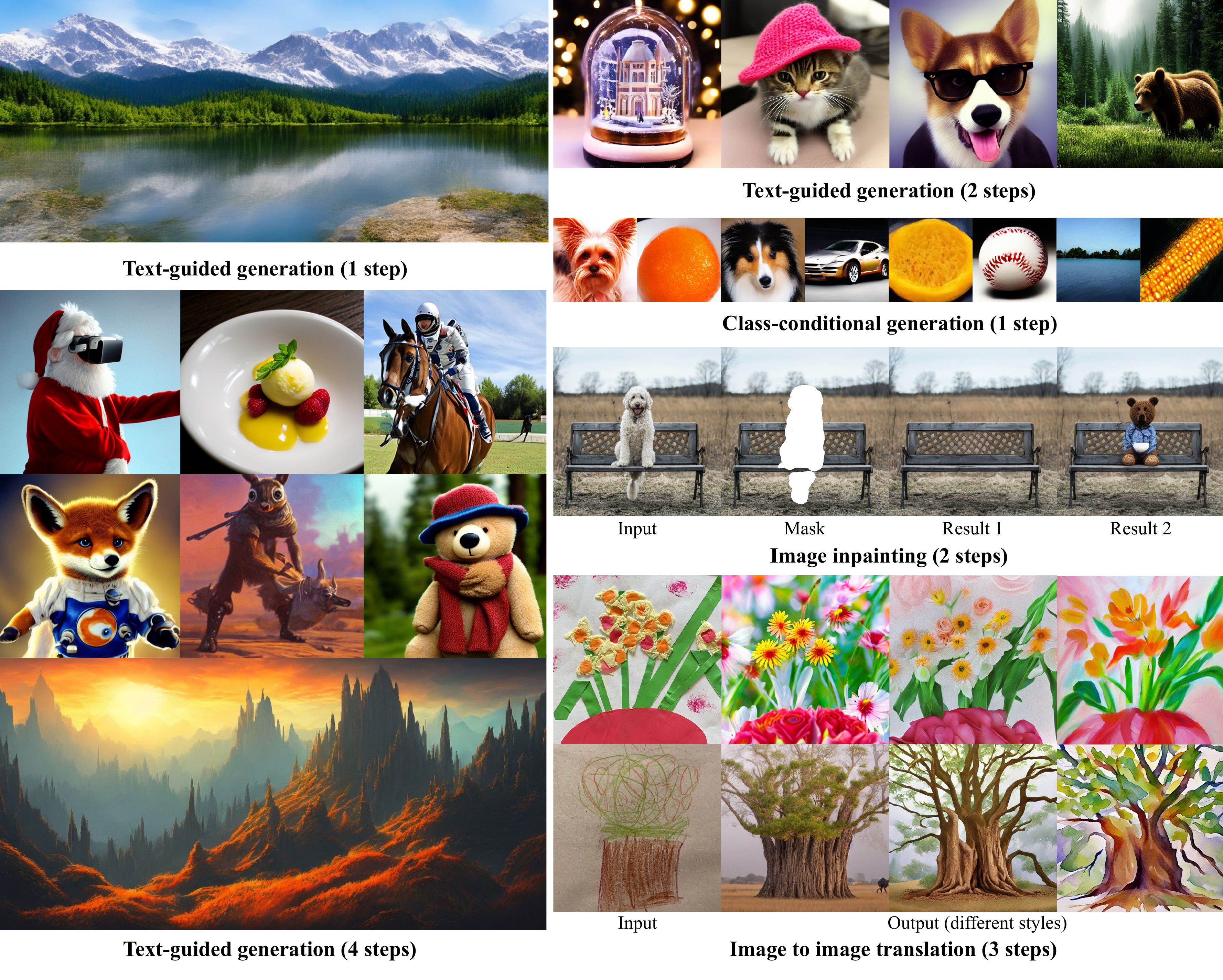}
    \vspace{-2em}
  \captionof{figure}{Distilled Stable Diffusion samples generated by our method. Our two-stage distillation approach is able to generate realistic images using only 1 to 4 denoising steps on various tasks. Compared to standard classifier-free guided diffusion models, we reduce the total number of sampling steps by at least 20$\times$.
  \vspace{-1em}
  }
	\label{fig:teaser}
  \vspace{1.5em}
}

\newcommand{\distilledvsbase}{
\begin{figure*}%
	\centering
	\includegraphics[width=1.00\textwidth, trim=0em 0em 0em 0em, clip]{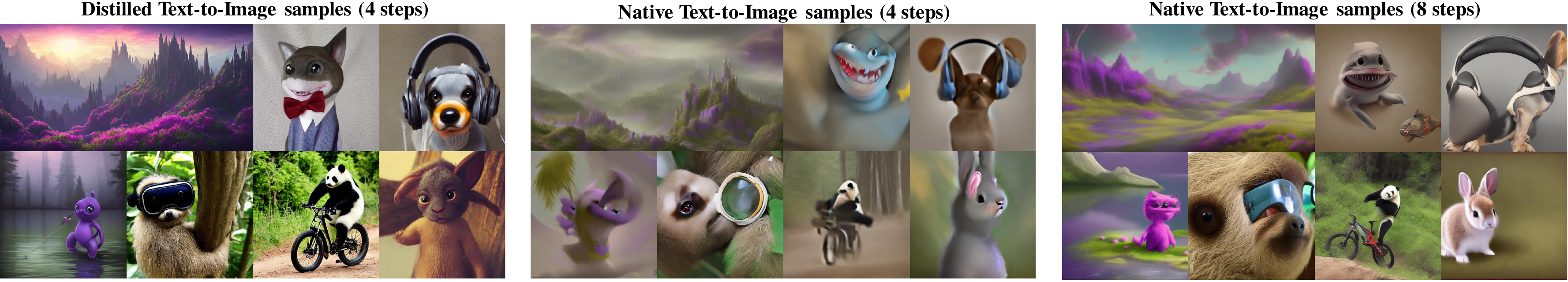}
\caption{Text-guided Stable Diffusion results. We distill the public \emph{Stable Diffusion} model using the proposed pipeline, arriving at a model that achieves high sample quality using only four denoising steps (\emph{left}). When sampling from the original model using four DDIM steps, the generated samples have clear artifacts (\emph{middle}). When using eight DDIM steps, the results get better  (\emph{right}), but are still blurry and less consistent than the distilled results using fewer steps. More samples are provided in \cref{fig:text2img_demo}. 
}.
\label{fig:distilledvsbase}
\vspace{-2em}
\end{figure*}
}

\newcommand{\cinldmresults}{
\begin{figure} %
     \centering
     \begin{subfigure}[b]{0.49\linewidth}
         \centering
         \includegraphics[width=\textwidth]{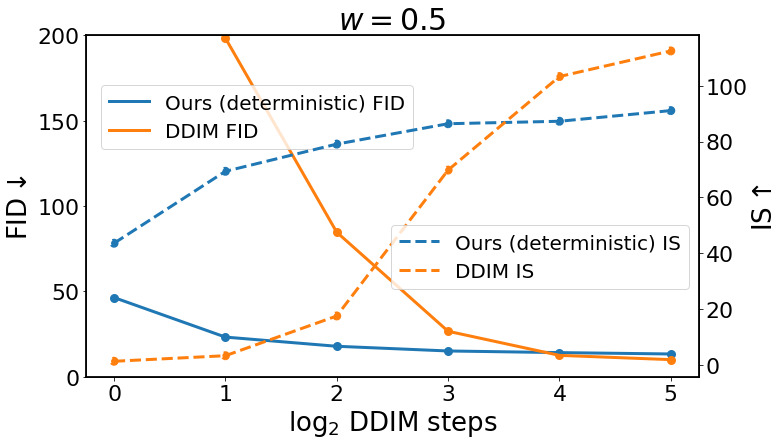}
     \end{subfigure}
     \begin{subfigure}[b]{0.49\linewidth}
         \centering
         \includegraphics[width=\textwidth]{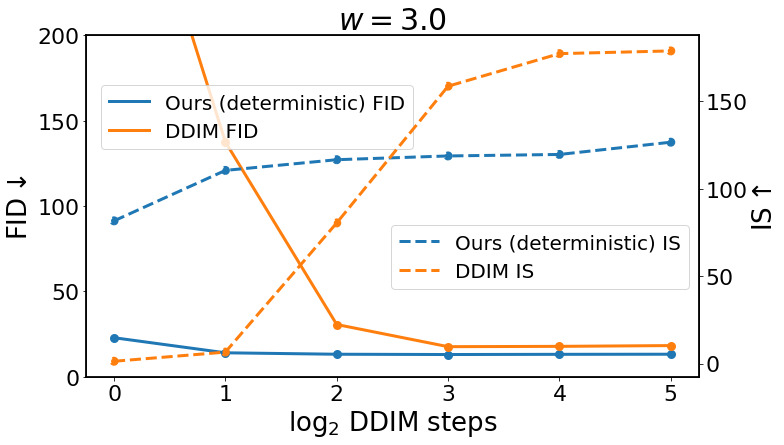}
     \end{subfigure}
    \caption{FID and Inception Score for class-conditional image generation on ImageNet ($256\times 256$) with distilled latent diffusion. The results are evaluated on 5000 samples. Our distilled latent diffusion model is able to generate high-quality image samples using significantly less sampling steps (up to a factor of 16) than the original model while achieving similar or better FID scores. 
    }
    \label{fig:cinldmresults}
\end{figure}
}

\newcommand{\stableplot}{
\begin{figure}%
     \centering
     \begin{subfigure}[b]{0.49\linewidth}
         \centering
         \includegraphics[width=\textwidth]{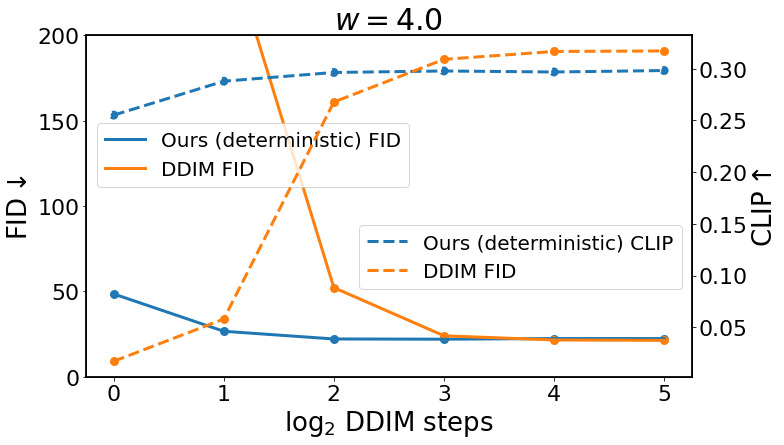}
     \end{subfigure}
     \begin{subfigure}[b]{0.49\linewidth}
         \centering
         \includegraphics[width=\textwidth]{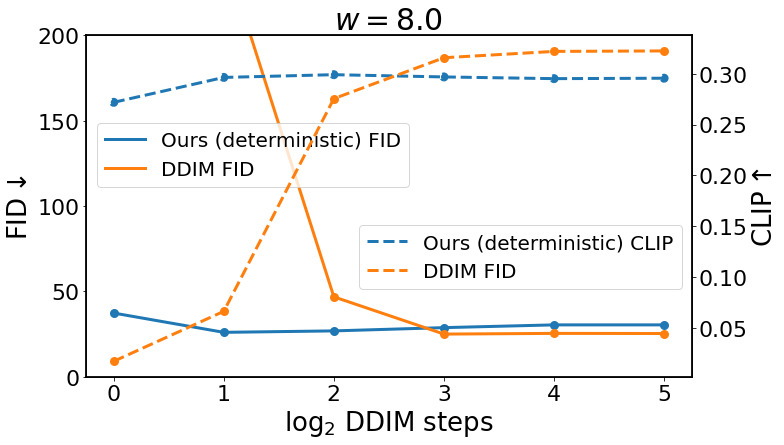}
     \end{subfigure}
    \caption{FID and CLIP ViT-g/14 score for text-to-image generation at $512\times 512$ px using the distilled Stable Diffuion model. The results are evaluated on 5000 captions from the COCO2017~\cite{lin2014microsoft} validation set. Our distilled latent diffusion model is able to generate high-quality image samples using significantly less sampling steps than the original model while achieving similar or better FID and CLIP scores, especially in the low-step regime.
    \vspace{-1.5em}}
    \label{fig:stableplot}
\end{figure}
}

\newcommand{\laionconvergence}{
\begin{figure} %
     \centering
     \begin{subfigure}[b]{0.49\linewidth}
         \centering
         \includegraphics[width=\textwidth]{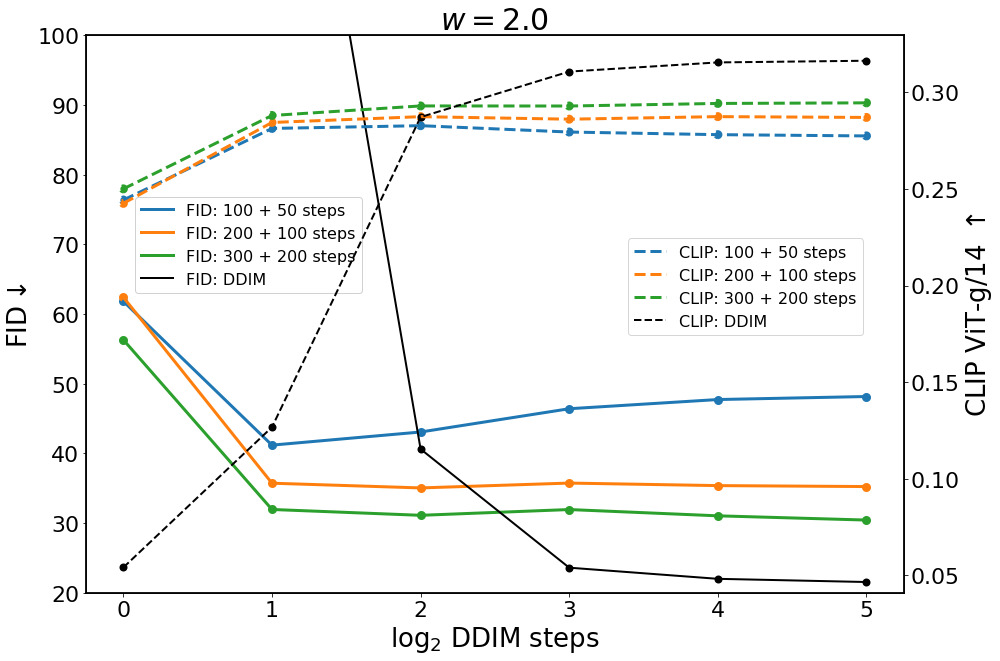}
     \end{subfigure}
    \begin{subfigure}[b]{0.49\linewidth}
         \centering
         \includegraphics[width=\textwidth]{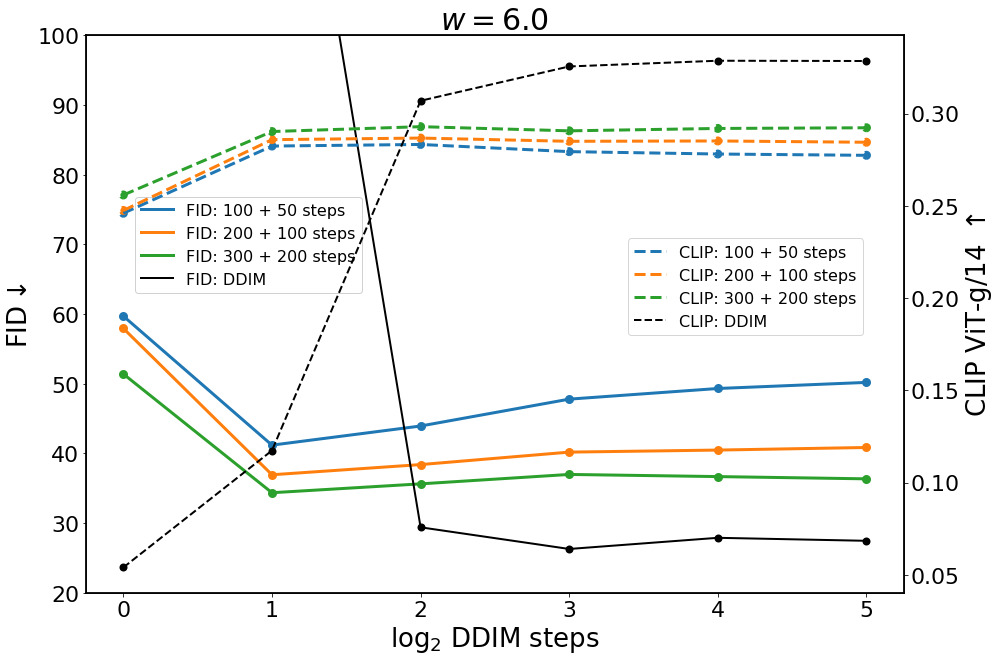}
     \end{subfigure}
    \caption{FID and Inception Score for text-guided image generation on LAION ($256\times 256$) with distilled latent diffusion. The results are evaluated on 5000 captions from COCO2017. 
    We observe that our distillation method approaches DDIM sampling after only a few thousand training steps, see \cref{sec:app:text_guided}.
    }
    \label{fig:laionconvergence}
     \vspace{-1em}
\end{figure}
}

\newcommand{\cinldmprresults}{
\begin{figure} %
     \centering
     \begin{subfigure}[b]{0.49\linewidth}
         \centering
         \includegraphics[width=\textwidth]{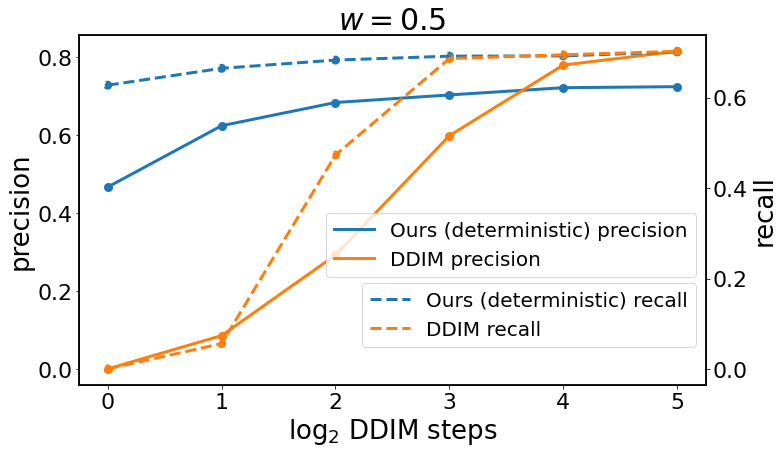}
     \end{subfigure}
     \begin{subfigure}[b]{0.49\linewidth}
         \centering
         \includegraphics[width=\textwidth]{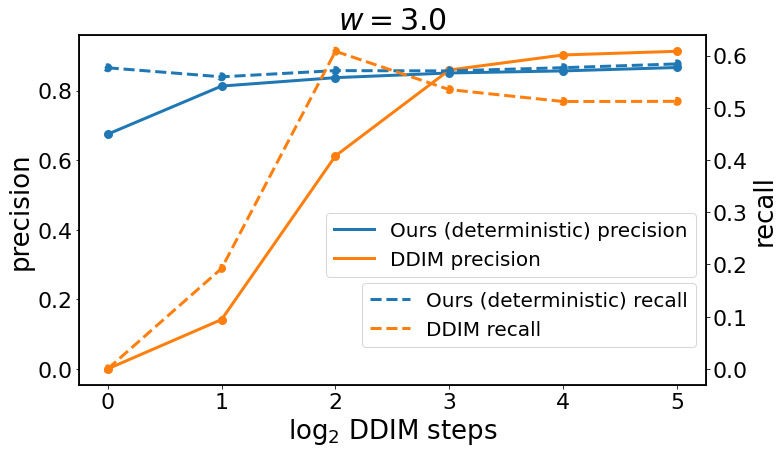}
     \end{subfigure}
    \caption{Precision and recall~\cite{kynkaanniemi2019improved} for class-conditional image generation on ImageNet ($256\times 256$) with distilled latent diffusion. The results are evaluated on 5000 samples. Our distilled latent diffusion model for 2- and 4-step sampling nearly matches DDIM performance at 32$\times$2 steps in terms of precision, and strictly outperforms it in terms of recall. 
    }
    \label{fig:cinldmprresults}
\end{figure}
}

\newcommand{\dpmddimfidtable}{
    \begin{table}
    \centering
    \resizebox{0.9\linewidth}{!}{
    \begin{tabular}{ccccccc}
    \Xhline{1\arrayrulewidth}
    Setting & vs. DDIM (FID) & vs. DPM$++$ (FID)\\
    \Xhline{2\arrayrulewidth}
    2-step, $w=2.0$ & $+89.8\%$ & $+69.4\%$ \\
    4-step, $w=2.0$ & $+68.9\%$ & $+32.5\%$ \\
    2-step, $w=8.0$ & $+89.5\%$ & $+73.7\%$ \\
    4-step, $w=8.0$ & $+42.6\%$ & $+21.6\%$ \\
    \Xhline{2\arrayrulewidth}
    \end{tabular}
    }
    \caption{\label{tab:dpmddimfidtable} Relative performance of our distilled $512 \times 512$ LAION model compared to DDIM~\cite{song2020denoising} and DPM$++$~\cite{dpmpp} sampling of the base model. Note that DDIM and DPM-Solver use 2$\times$ more steps than the one listed under ``Setting", as they rely on classifier-free guidance instead of $w$-conditioning. This requires DDIM and DPM-Solver to evaluate both an unconditional and a conditional diffusion model at each denoising step, giving rise to the $\times$2 overhead.
    }
    \end{table}
}

\newcommand{\dpmddimcliptable}{
    \begin{table}
    \centering
    \resizebox{0.9\linewidth}{!}{
    \begin{tabular}{ccccccc}
    \Xhline{1\arrayrulewidth}
    Setting & vs. DDIM (CLIP) & vs. DPM (CLIP)\\
    \Xhline{2\arrayrulewidth}
    2-step, $w=2.0$ & $+550\%$ & $+27.9\%$ \\
    4-step, $w=2.0$ & $+19.2\%$ & $+0.1\%$\\
    2-step, $w=8.0$ & $+348\%$ & $+47.5\%$ \\
    4-step, $w=8.0$ & $+8.6\%$ & $+0.6\%$ \\
    \Xhline{2\arrayrulewidth}
    \end{tabular}
    }
    \caption{\label{tab:dpmddimcliptable} Relative performance of our distilled $512 \times 512$ LAION model compared to DDIM~\cite{song2020denoising} and DPM-Solver (DPM$++$)~\cite{dpmsolver,dpmpp} sampling of the base model. 
    Note that DDIM and DPM use 2$\times$ more steps than the one listed under ``Setting", as they rely on classifier-free guidance instead of $w$-conditioning. This requires  DDIM and DPM to evaluate both an unconditional and a conditional diffusion model at each denoising step, giving rise to the $\times$2 overhead.
    We use CLIP ViT-g/14 for evaluation~\cite{ilharco_gabriel_2021_5143773,radford2021learning}.
    }
    \end{table}
}

\newcommand{\stabledpmplot}{
\begin{figure}%
     \centering
     \begin{subfigure}[b]{0.49\linewidth}
         \centering
         \includegraphics[width=\textwidth]{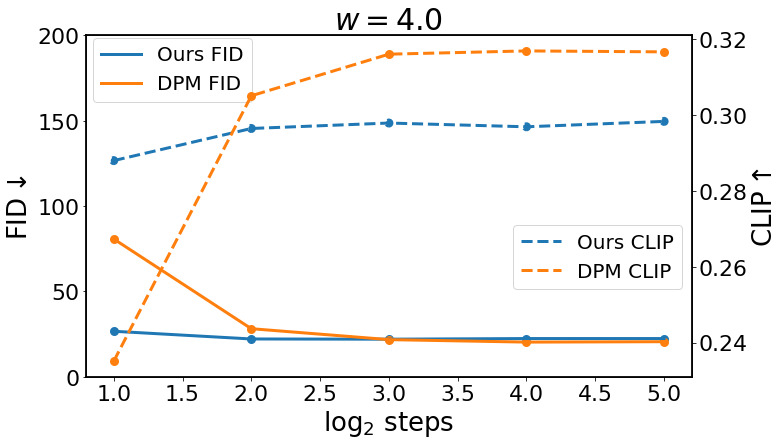}
     \end{subfigure}
     \begin{subfigure}[b]{0.49\linewidth}
         \centering
         \includegraphics[width=\textwidth]{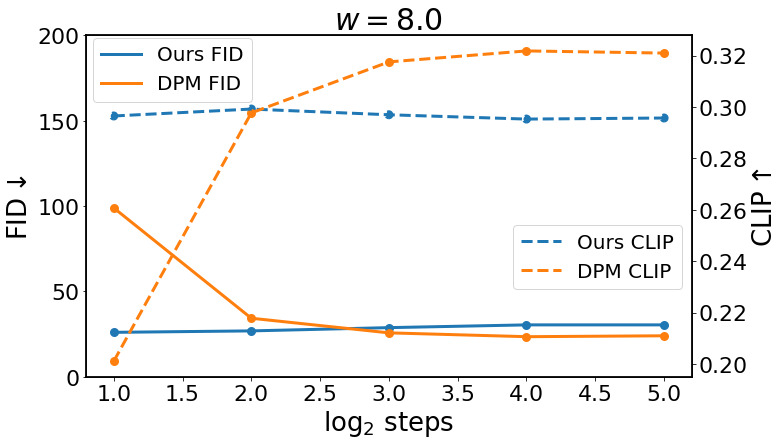}
     \end{subfigure}
    \caption{FID and CLIP ViT-g/14 score for text-to-image generation at $512\times 512$ px using the distilled \emph{Stable Diffusion} model. The results are evaluated on 5000 captions from the COCO2017~\cite{lin2014microsoft} validation set. Our distilled model outperforms the state-of-the-art accelerated sampler \emph{DPM-Solver}(DPM$++$)~\cite{dpmsolver,dpmpp} in the 2- and 4- step regime.
    We believe the difference in CLIP scores for $>10$-step sampling can be closed by longer training.
    We stress that DPM-Solver, as DDIM, uses classifier-free guidance during sampling, which requires evaluating both an unconditional and a conditional diffusion model at each denoising step, giving rise to an extra $\times$2 overhead compared to our method.
    }
    \label{fig:stabledpmplot}
\end{figure}
}

\newcommand{\inpaintingtable}{
    \begin{table}
    \centering
    \resizebox{0.85\linewidth}{!}{
    \begin{tabular}{ccccccc}
    \Xhline{1\arrayrulewidth}
    Setting & Ours (FID $\downarrow$) & DDIM (FID $\downarrow$) \\
    \Xhline{2\arrayrulewidth}
    2-step, $w=4.0$ & 29.50 & 109.35 \\
    4-step, $w=4.0$ & 24.90 & 26.89 \\
    2-step, $w=11.0$ & 31.43 & 105.71 \\
    4-step, $w=11.0$ & 24.36 & 27.22 \\
    \Xhline{2\arrayrulewidth}
    \end{tabular}
    }
    \caption{\label{tab:inpaintingtable} Quantitative inpainting results as evaluated by FID. We evaluate on 2000 examples from COCO2017. Note that DDIM, which is evaluated with classifier-free guidance, uses two times more function evaluations than the one listed under ``Setting".}
    \end{table}
}

\newcommand{\inpaintingtablewithclip}{
    \begin{table}[!ht]
    \centering
    \resizebox{0.99\linewidth}{!}{
    \begin{tabular}{ccccccc}
    \Xhline{1\arrayrulewidth}
    Setting & ours [FID] & DDIM [FID] & \phantom{x} & ours [CLIP] & DDIM [CLIP] \\
    \Xhline{2\arrayrulewidth}
    2-step, $w=2.0$ & X & X & & Y & Y \\
    \Xhline{2\arrayrulewidth}
    \end{tabular}
    }
    \caption{\label{tab:inpaintingtablewithclip} Quantitative inpainting results. We evaluate on 2000 examples from COCO2017 and use CLIP ViT-g/14 for evaluation~\cite{ilharco_gabriel_2021_5143773,radford2021learning}. }
    \end{table}
}

\newcommand{\cinqualitative}{
\begin{figure*} %
     \centering
     
     \begin{subfigure}[b]{0.49\textwidth}
         \centering
         \includegraphics[width=\textwidth]{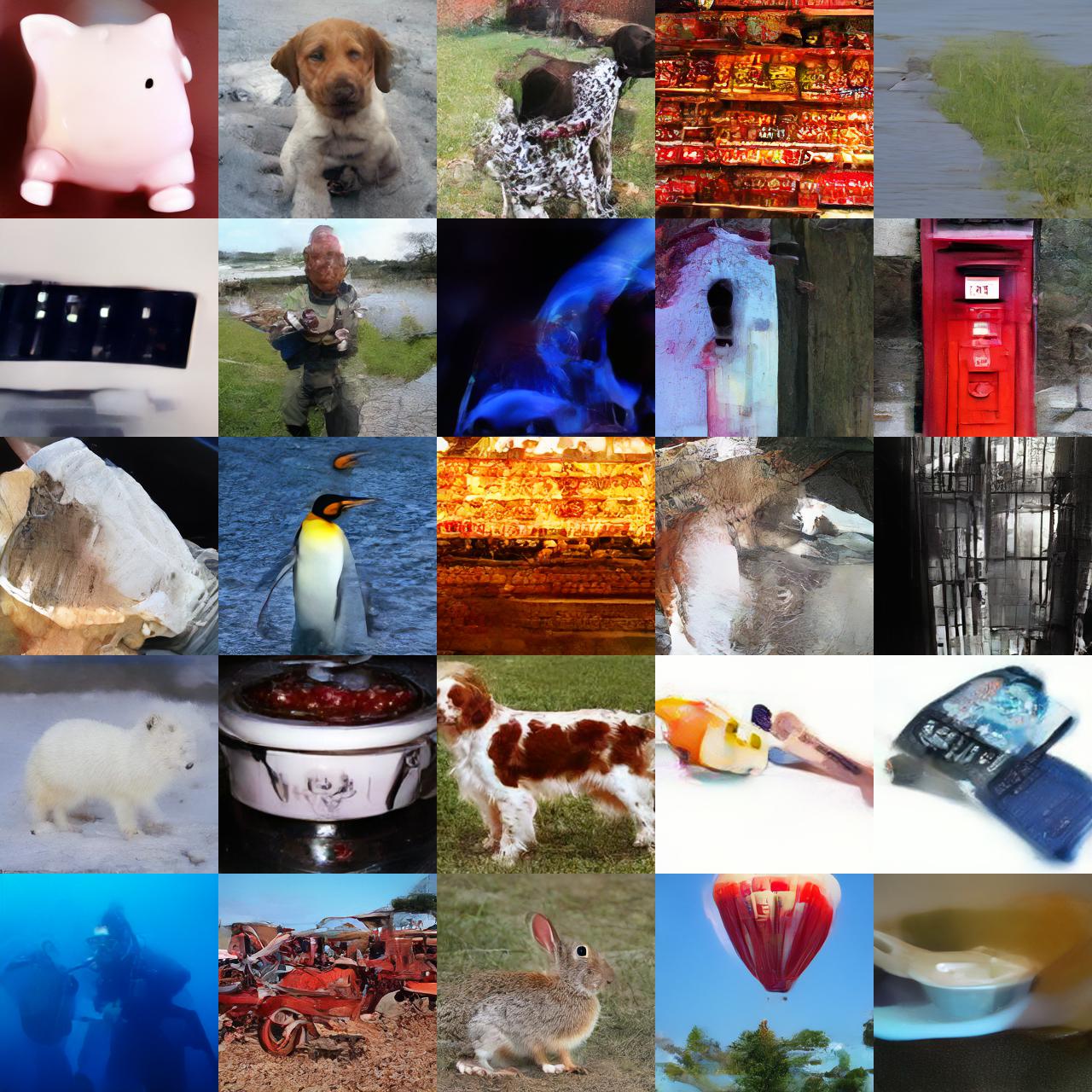}
         \caption{1 denoising step, ours}
     \end{subfigure}
     \begin{subfigure}[b]{0.49\textwidth}
         \centering
         \includegraphics[width=\textwidth]{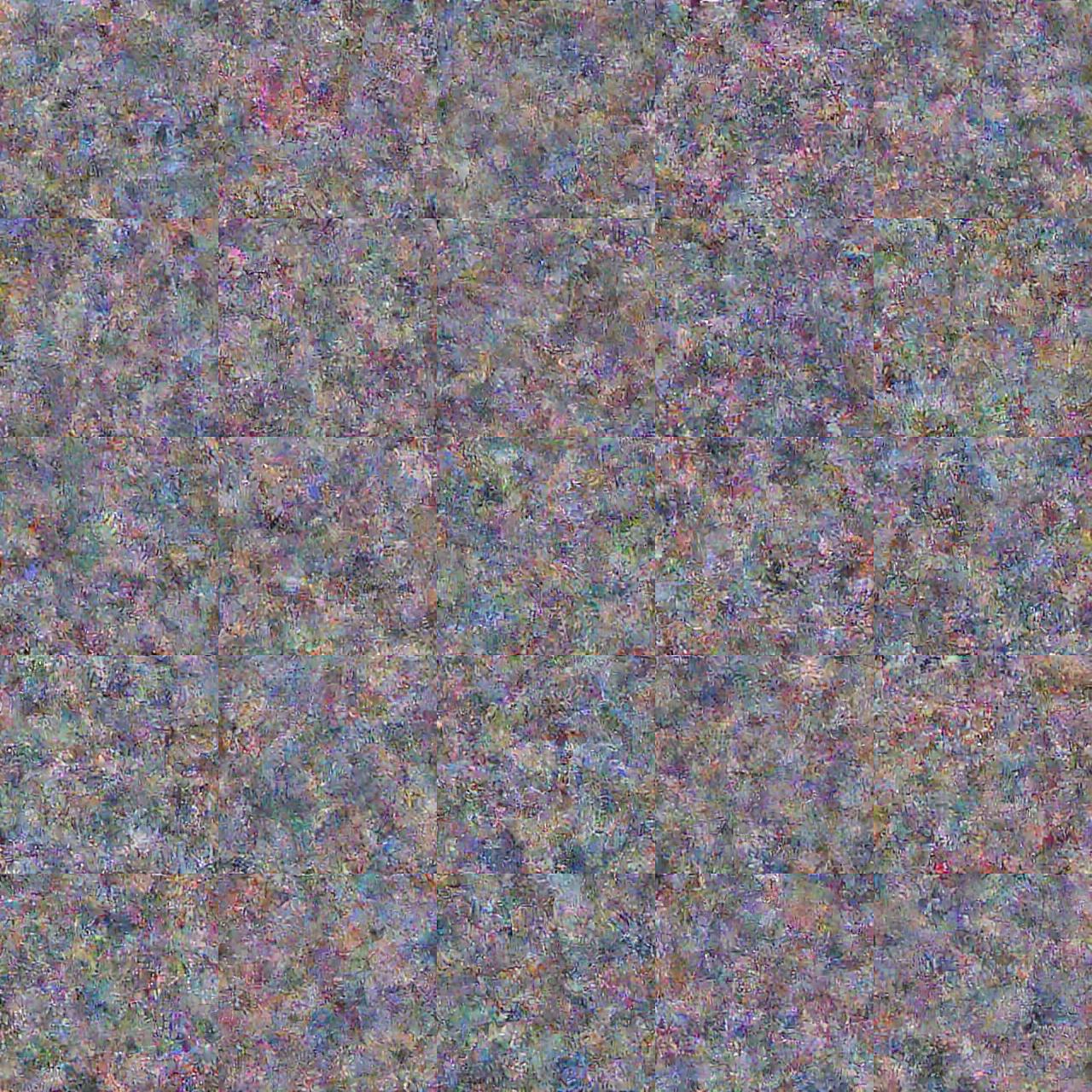}
         \caption{2$\times$ 1 denoising step, DDIM}
     \end{subfigure}
     \begin{subfigure}[b]{0.49\textwidth}
         \centering
         \includegraphics[width=\textwidth]{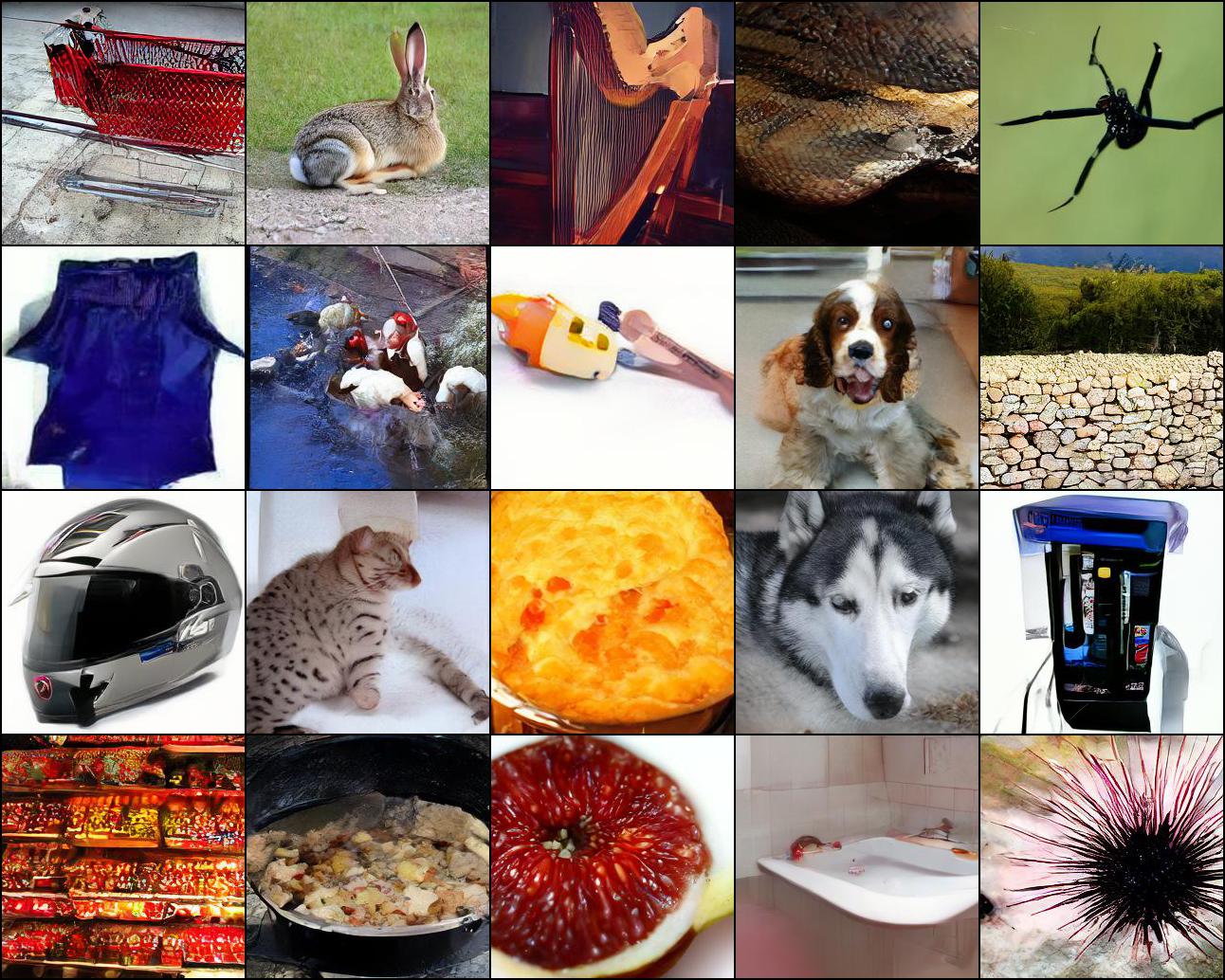}
         \caption{2 denoising steps, ours}
     \end{subfigure}
     \begin{subfigure}[b]{0.49\textwidth}
         \centering
         \includegraphics[width=\textwidth]{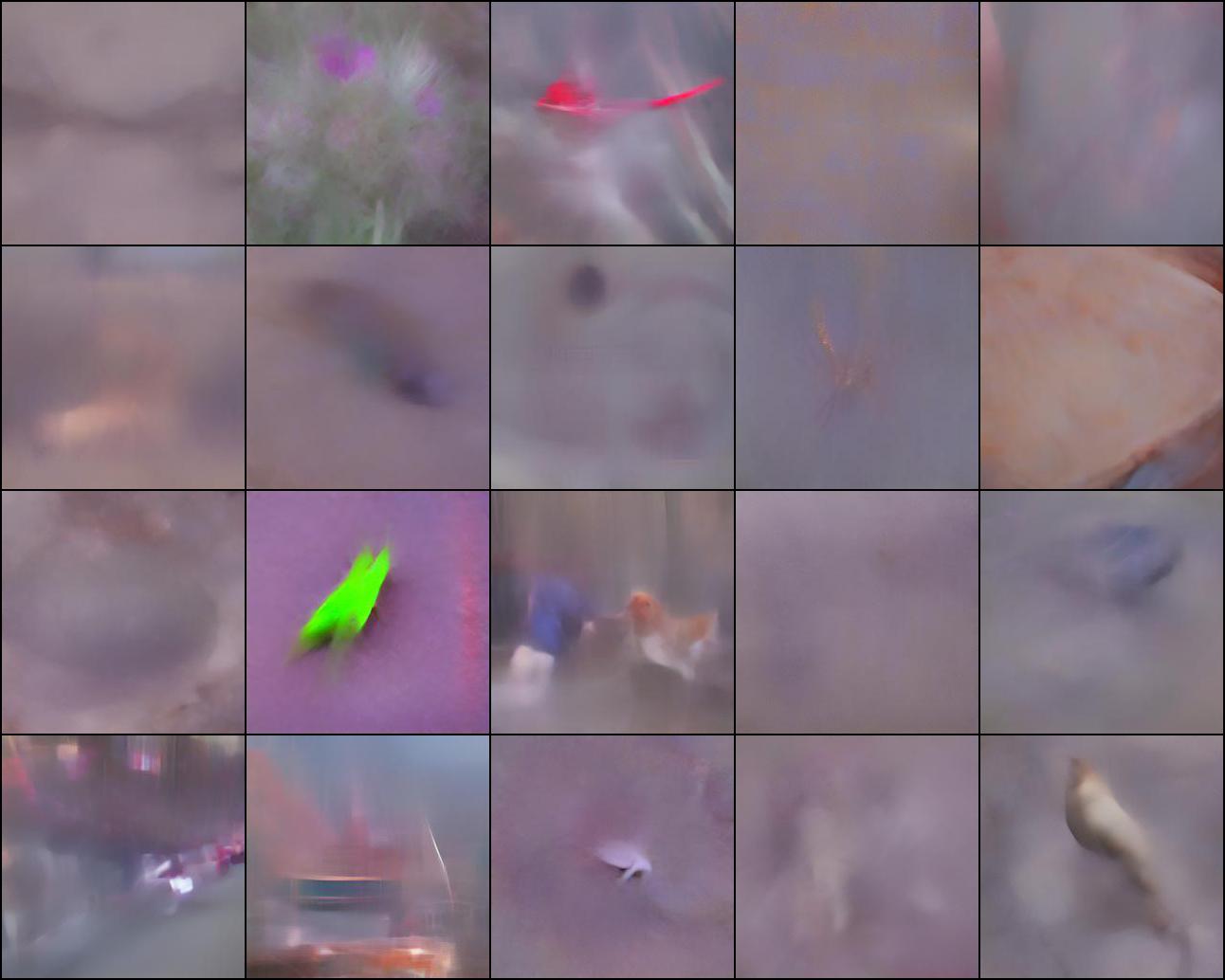}
         \caption{2$\times$2 denoising steps, DDIM}
     \end{subfigure}
    \caption{Random 256$\times$256 class-conditional samples from our distilled model and from the DDIM teacher for 1 and 2 denoising steps for $w=3.0$.
    }
    \label{fig:cinqualitative}
\end{figure*}
}

\newcommand{\laionqualitativesupp}{
\begin{figure*} %
     \centering
     
     \begin{subfigure}[b]{0.32\textwidth}
         \centering
         \includegraphics[width=\textwidth]{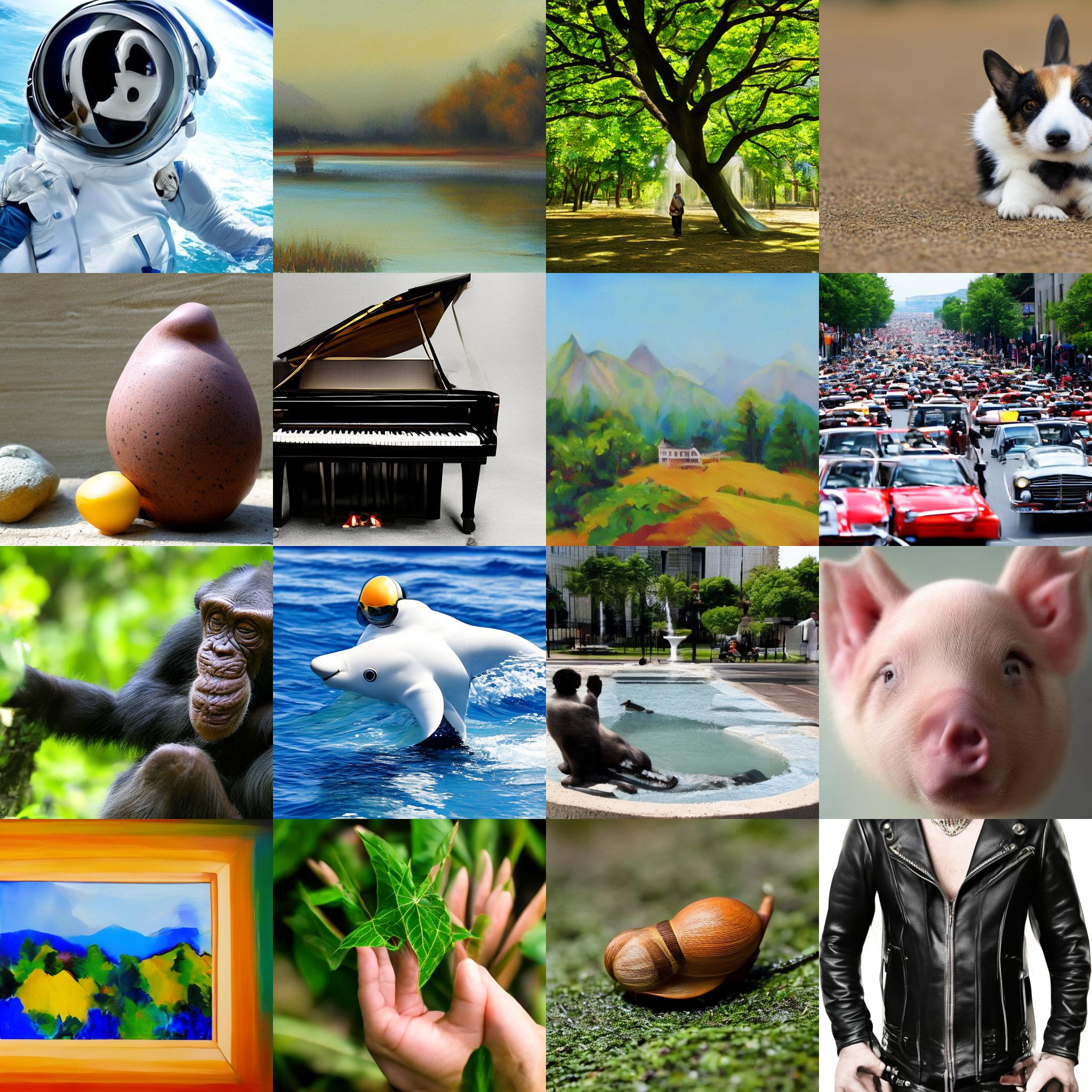}
         \caption{2 denoising steps, ours}
     \end{subfigure}
     \begin{subfigure}[b]{0.32\textwidth}
         \centering
         \includegraphics[width=\textwidth]{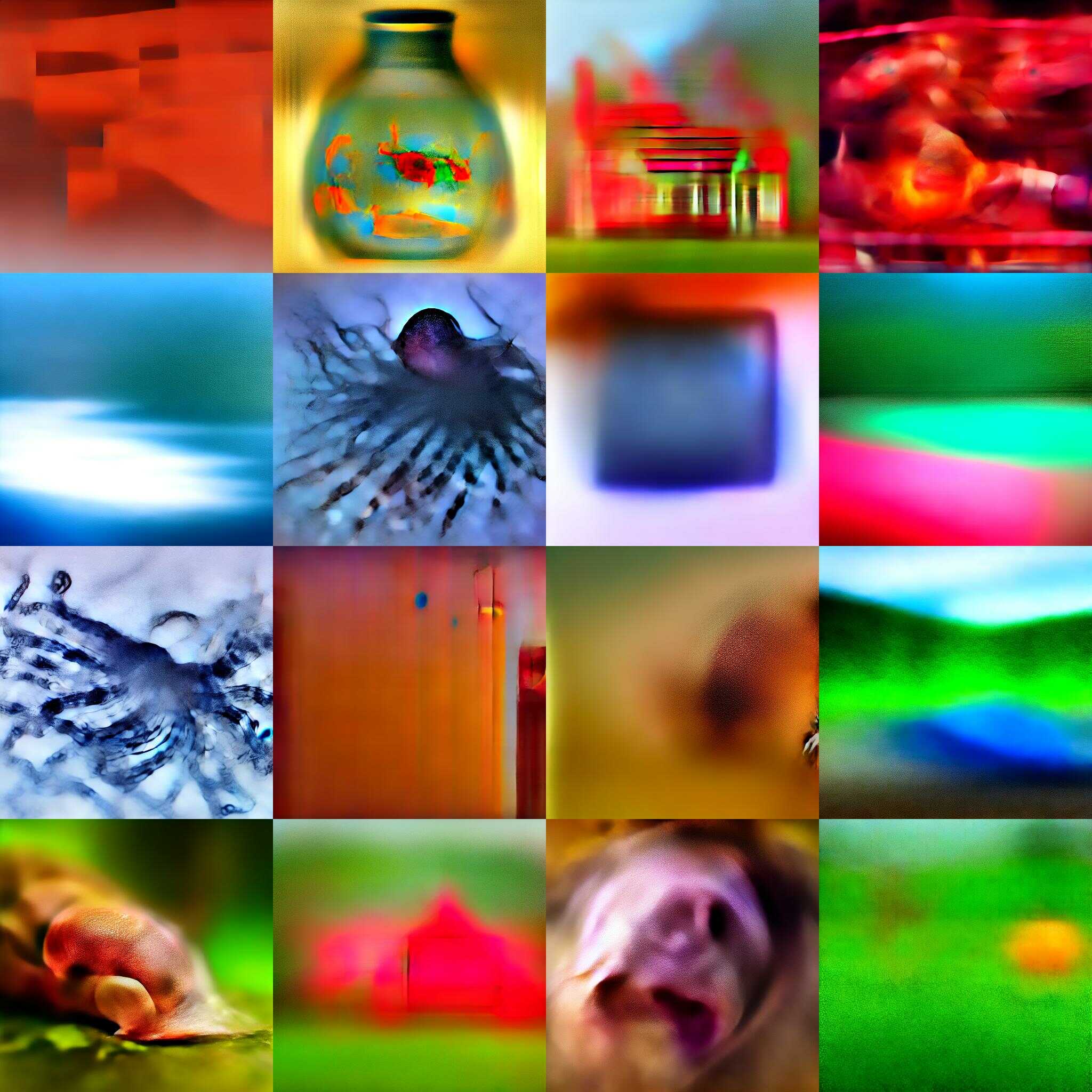}
         \caption{2$\times$2 denoising steps, DPM}
     \end{subfigure}
     \begin{subfigure}[b]{0.32\textwidth}
         \centering
         \includegraphics[width=\textwidth]{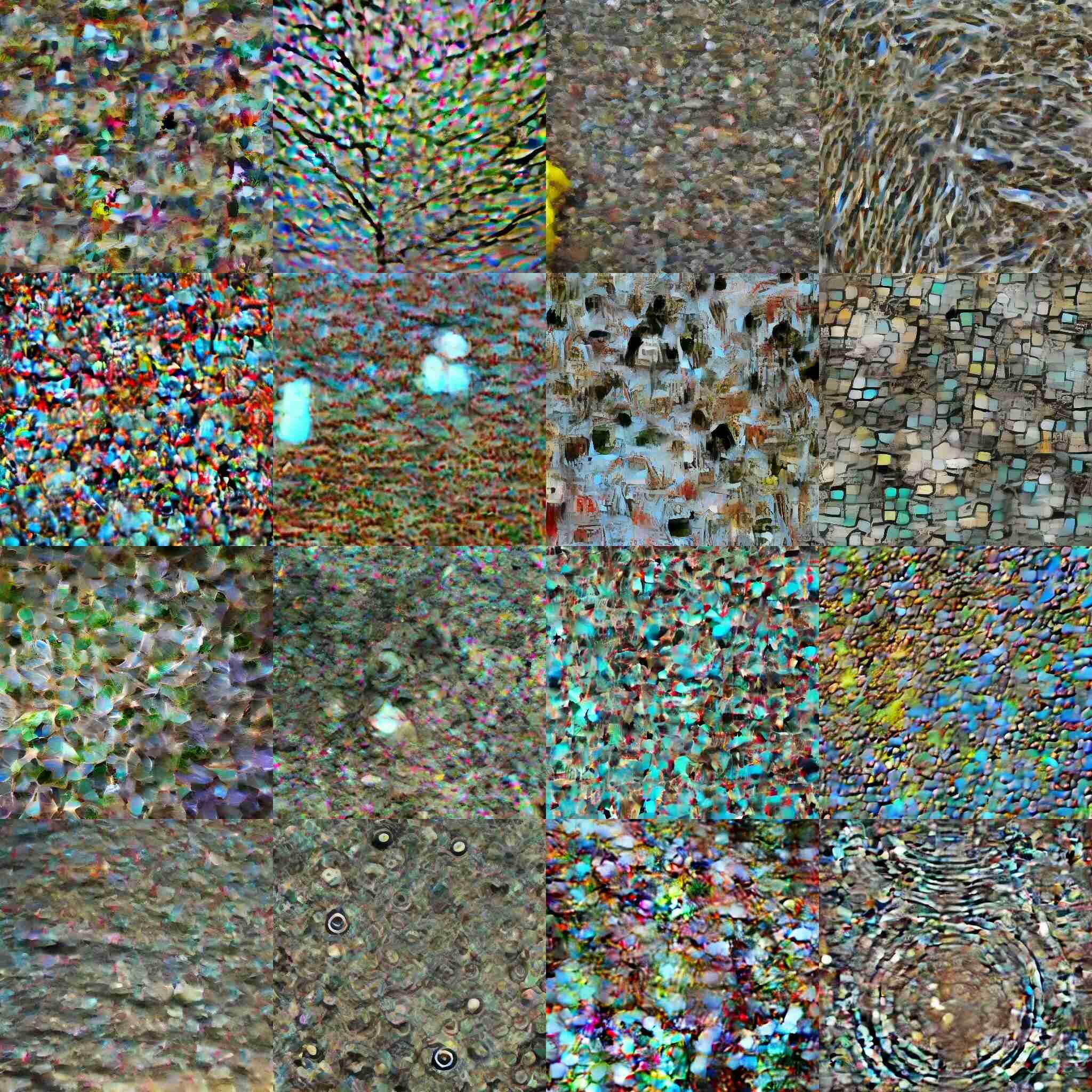}
         \caption{2$\times$2 denoising steps, DDIM}
     \end{subfigure}
     
     \begin{subfigure}[b]{0.32\textwidth}
         \centering
         \includegraphics[width=\textwidth]{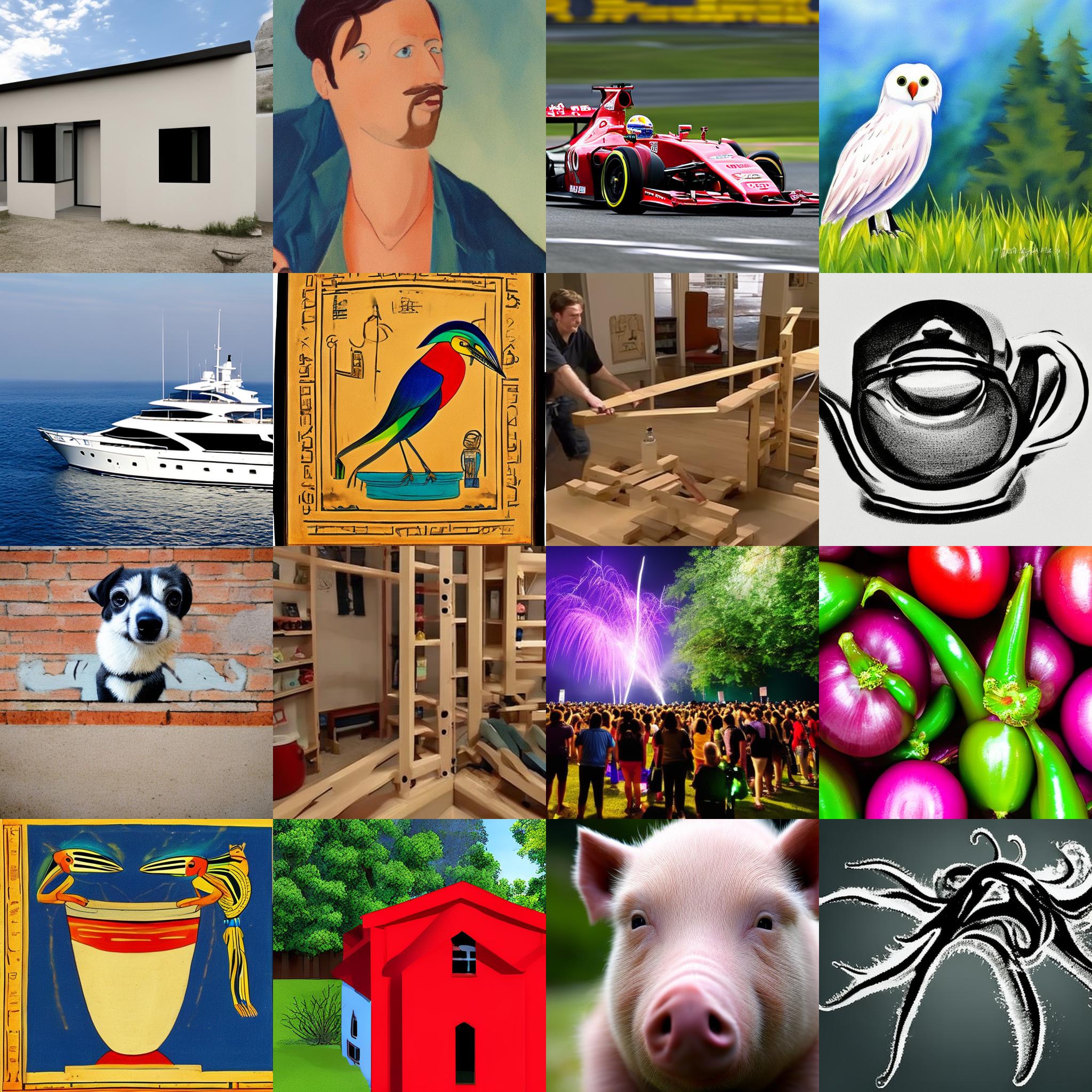}
         \caption{4 denoising steps, ours}
     \end{subfigure}
     \begin{subfigure}[b]{0.32\textwidth}
         \centering
         \includegraphics[width=\textwidth]{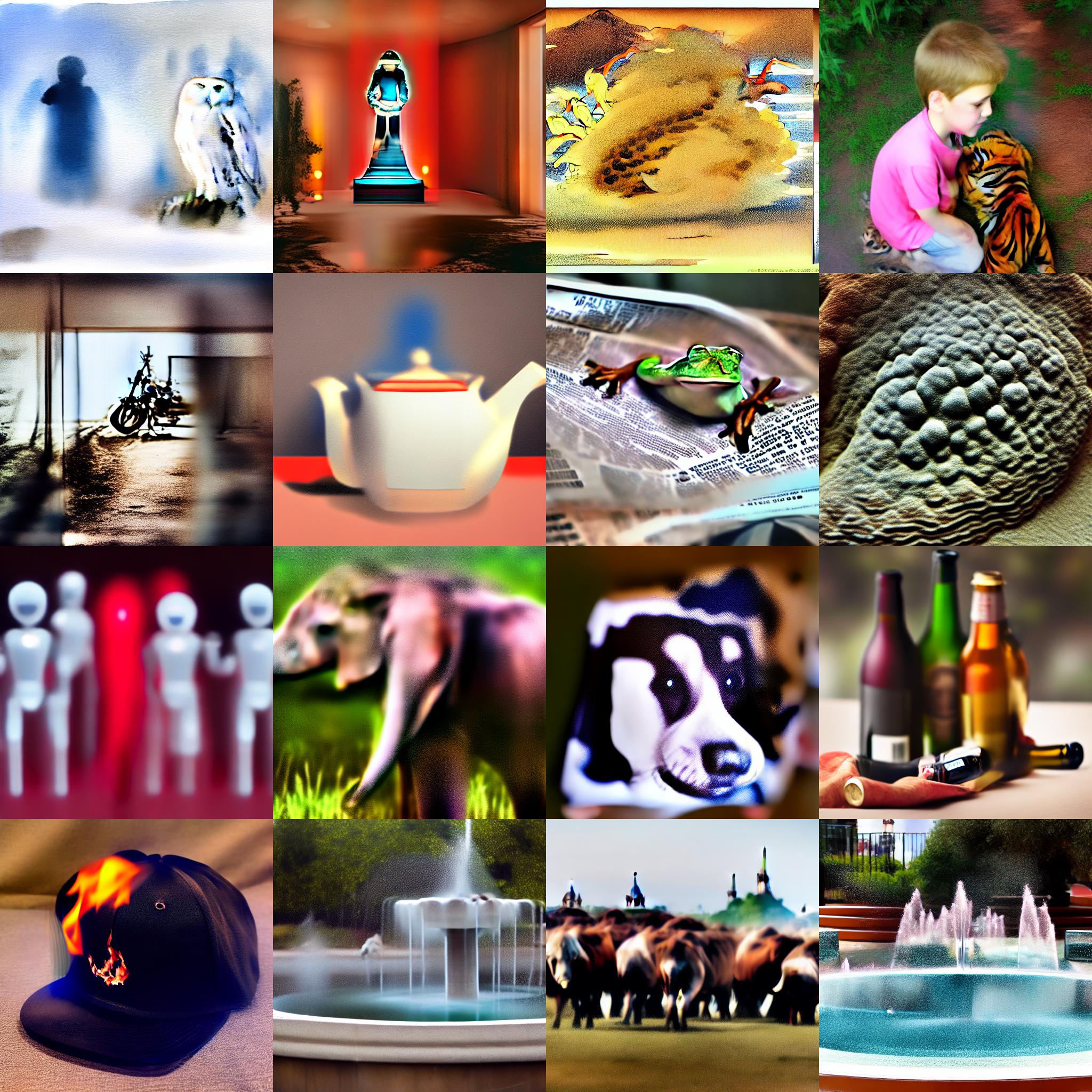}
         \caption{4$\times$2 denoising steps, DPM}
     \end{subfigure}
     \begin{subfigure}[b]{0.32\textwidth}
         \centering
         \includegraphics[width=\textwidth]{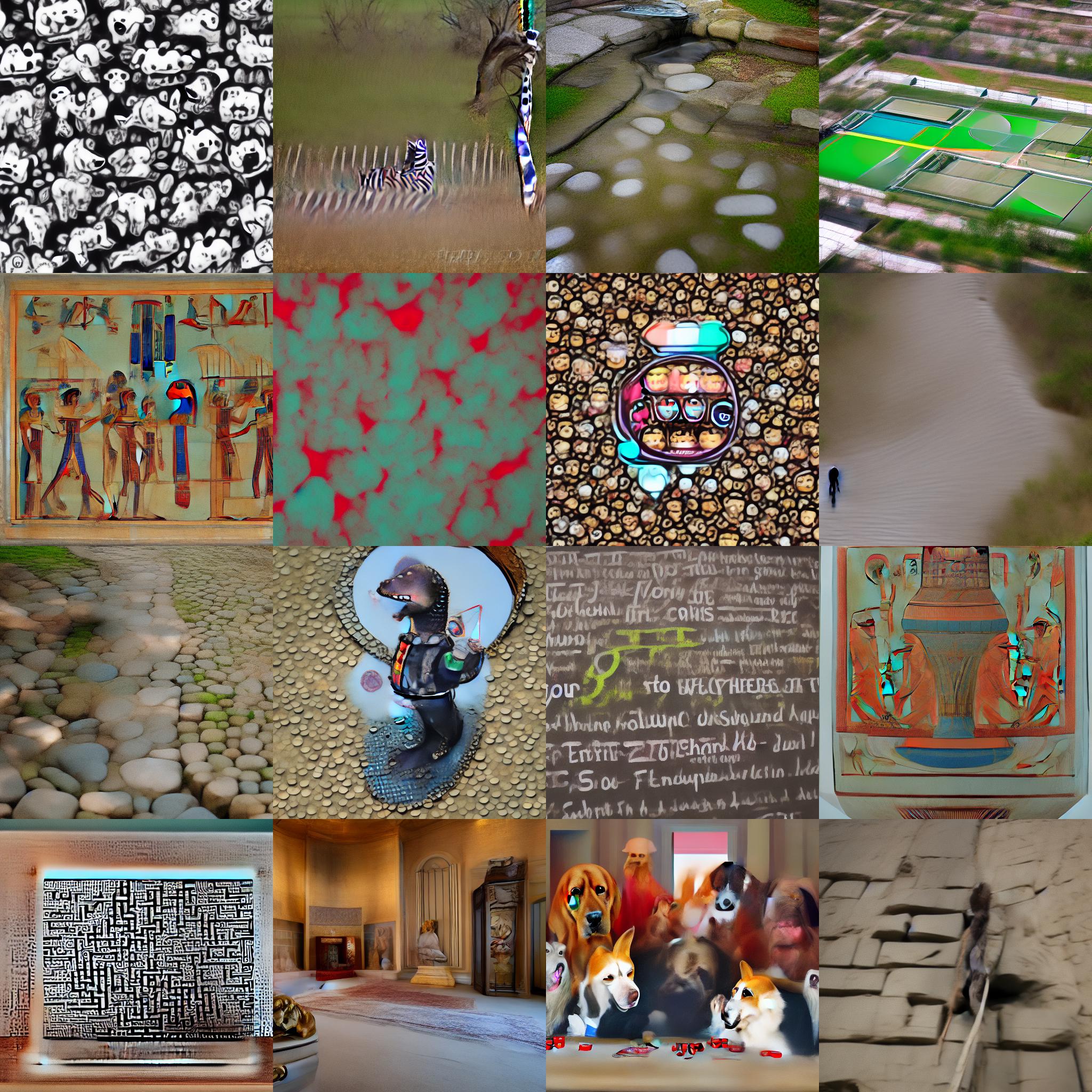}
         \caption{4$\times$2 denoising steps, DDIM}
     \end{subfigure}
     
    \caption{Random 512$\times$512 text-guided samples from our distilled \emph{Stable Diffusion} model compared to the DDIM teacher and DPM-solver for 2 and 4 denoising steps for $w=11.5$.
    }
    \label{fig:laionqualitativesupp}
\end{figure*}
}

\newcommand{\inpaintingqualitative}{
\begin{figure*} [!ht]
     \centering
     \begin{subfigure}[b]{0.49\textwidth}
         \centering
         \includegraphics[width=\textwidth]{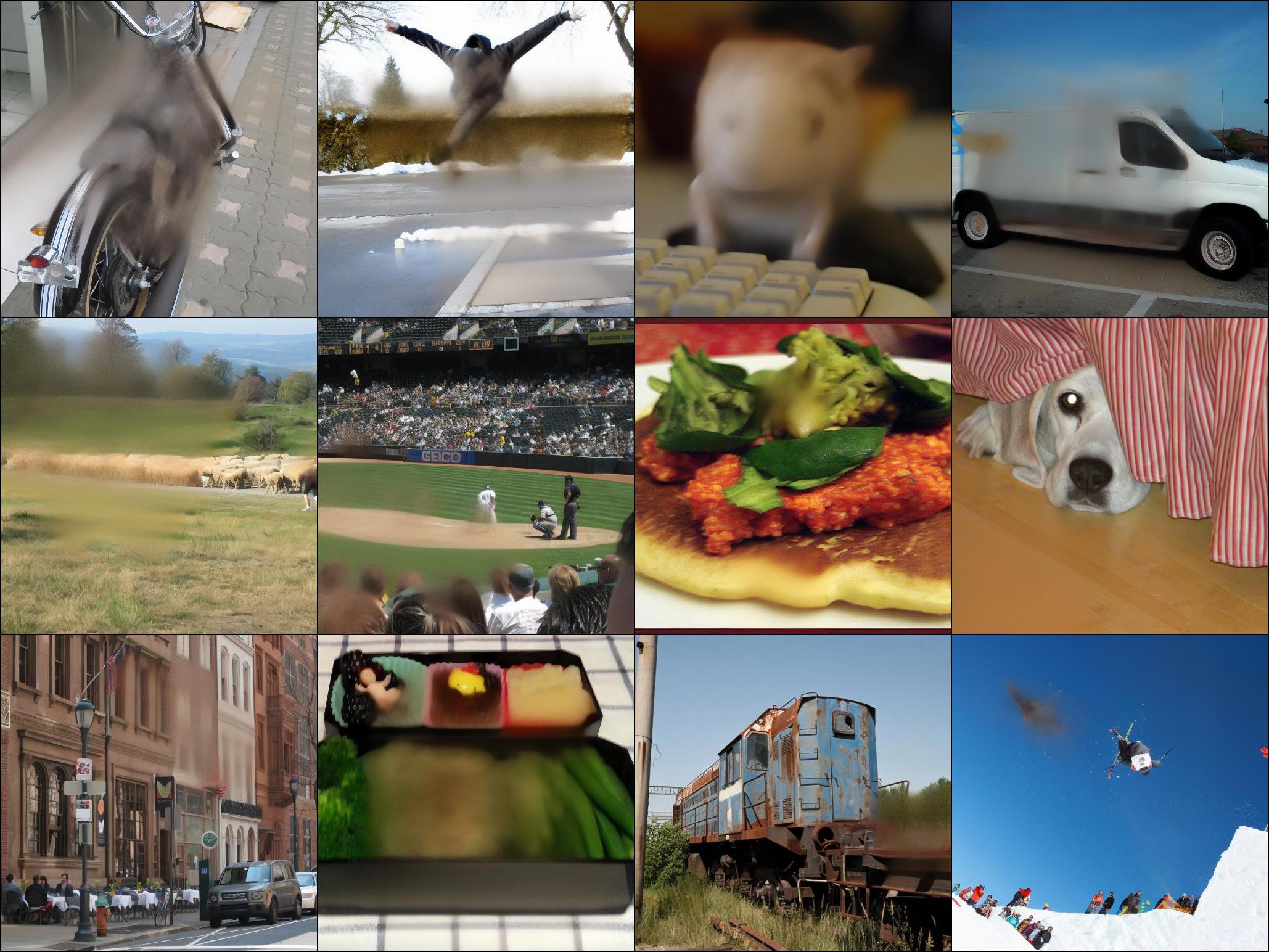}
         \caption{2 denoising steps, ours}
     \end{subfigure}
     \begin{subfigure}[b]{0.49\textwidth}
         \centering
         \includegraphics[width=\textwidth]{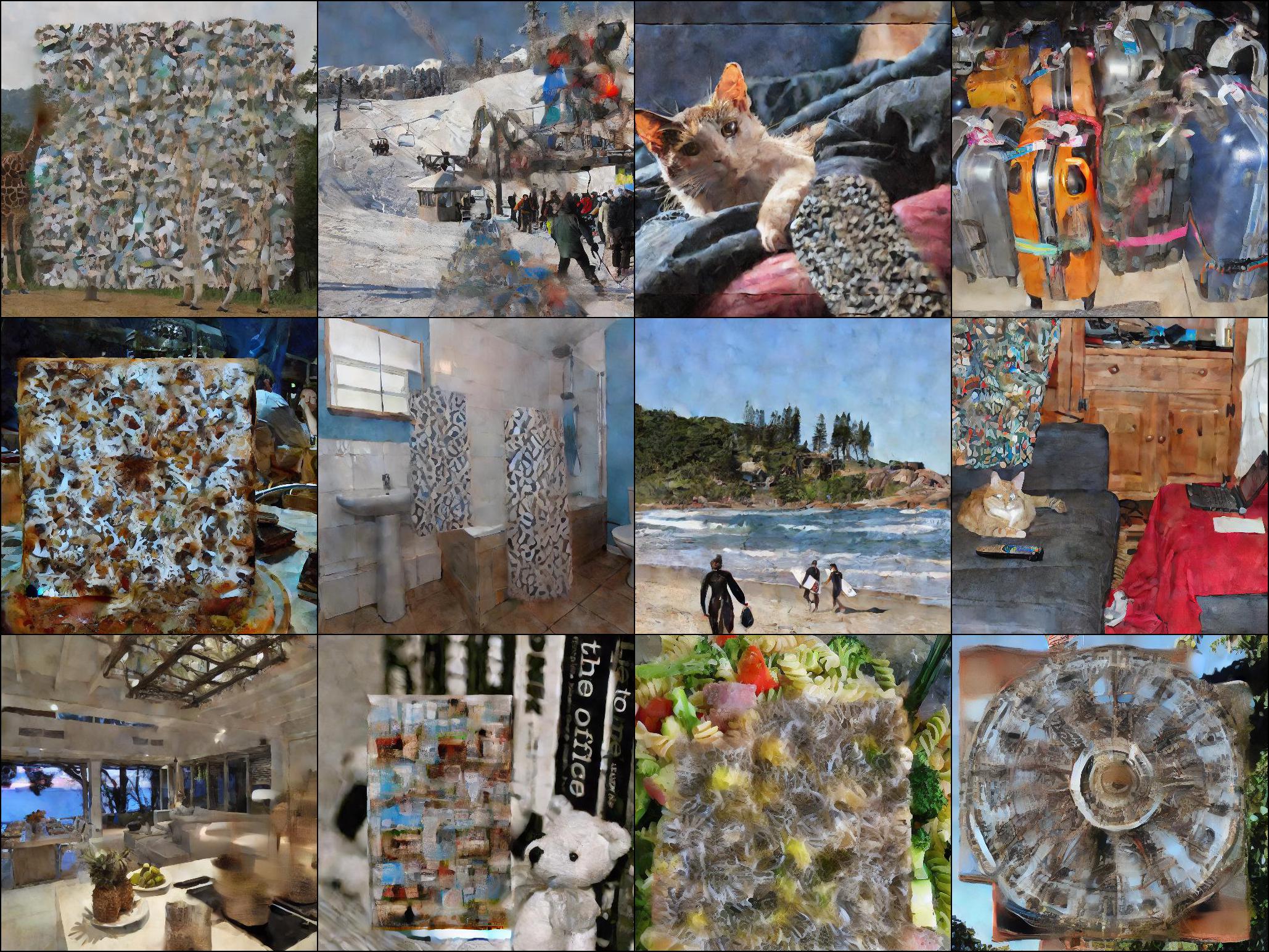}
         \caption{2$\times$ 2 denoising steps, DDIM}
     \end{subfigure}
    \caption{Random 512$\times$512 inpainting samples from our distilled model and from the DDIM teacher for 2 denoising steps for $w=11.0$. 
    }
    \label{fig:inpaintingqualitative}
\end{figure*}
}

%% file: 0_abstract.tex
\begin{abstract}
Classifier-free guided diffusion models have recently been shown to be highly effective at high-resolution image generation, and they have been widely used in large-scale diffusion frameworks including DALL$\cdot$E 2, Stable Diffusion and Imagen. However, a downside of classifier-free guided diffusion models is that they are computationally expensive at inference time since they require evaluating two diffusion models, a class-conditional model and an unconditional model, tens to hundreds of times. To deal with this limitation, we propose an approach to distilling classifier-free guided diffusion models into models that are fast to sample from: Given a pre-trained classifier-free guided model, we first learn a single model to match the output of the combined conditional and unconditional models, and then we progressively distill that model to a diffusion model that requires much fewer sampling steps. 
For standard diffusion models trained on the pixel-space, our approach is able to generate images visually comparable to that of the original model using as few as 4 sampling steps on ImageNet 64x64 and CIFAR-10, achieving FID/IS scores comparable to that of the original model while being up to 256 times faster to sample from. 
For diffusion models trained on the latent-space (\eg, Stable Diffusion), our approach is able to generate high-fidelity images using as few as 1 to 4 denoising steps,
accelerating inference by at least 10-fold compared to existing methods on ImageNet 256x256 and LAION datasets.
We further demonstrate the effectiveness of our approach on text-guided image editing and inpainting, where our distilled model is able to generate high-quality results using as few as 2-4 denoising steps.

\end{abstract}

%% file: 1_intro.tex
\section{Introduction}
Denoising diffusion probabilistic models (DDPMs)~\cite{sohl-dickstein2015deep,ho2020denoising,song2019generative,song2020score} have achieved state-of-the-art performance on image generation~\cite{nichol2021improved,rombach2022high,ramesh2021zero,ramesh2022hierarchical,saharia2022photorealistic}, audio synthesis~\cite{kong2020diffwave}, molecular generation~\cite{xu2022geodiff}, and likelihood estimation~\cite{kingma2021variational}. 
Classifier-free guidance~\cite{ho2022classifier} further improves the sample quality of diffusion models and has been widely used in large-scale diffusion model frameworks including GLIDE~\cite{nichol2021glide}, 
Stable Diffusion~\cite{rombach2022high}, 
DALL$\cdot$E 2~\cite{ramesh2022hierarchical}, and Imagen~\cite{saharia2022photorealistic}.
However, one key limitation of classifier-free guidance is its low sampling efficiency---it requires evaluating two diffusion models 
tens to hundreds of times to generate one sample. 
This limitation has hindered the application of classifier-free guidance models in real-world settings.
Although distillation approaches have been proposed for diffusion models~\cite{salimans2022progressive,song2020denoising}, these approaches are not directly applicable to classifier-free guided diffusion models. 
To deal with this issue, we propose a two-stage distillation approach to improving the sampling efficiency of classifier-free guided models. In the first stage, we introduce a single student model to match the combined output of the two diffusion models of the teacher. In the second stage, we \emph{progressively distill} the model learned from the first stage to a fewer-step model using the approach introduced in~\cite{salimans2022progressive}. 
Using our approach, a \emph{single} distilled model
is able to handle a wide range of different guidance strengths, allowing for the trade-off between sample quality and diversity efficiently. To sample from our model, we consider existing deterministic samplers in the literature~\cite{song2020denoising,salimans2022progressive} and further propose a stochastic sampling process. 

\begin{figure}[h]
    \centering
    \includegraphics[width=\linewidth]{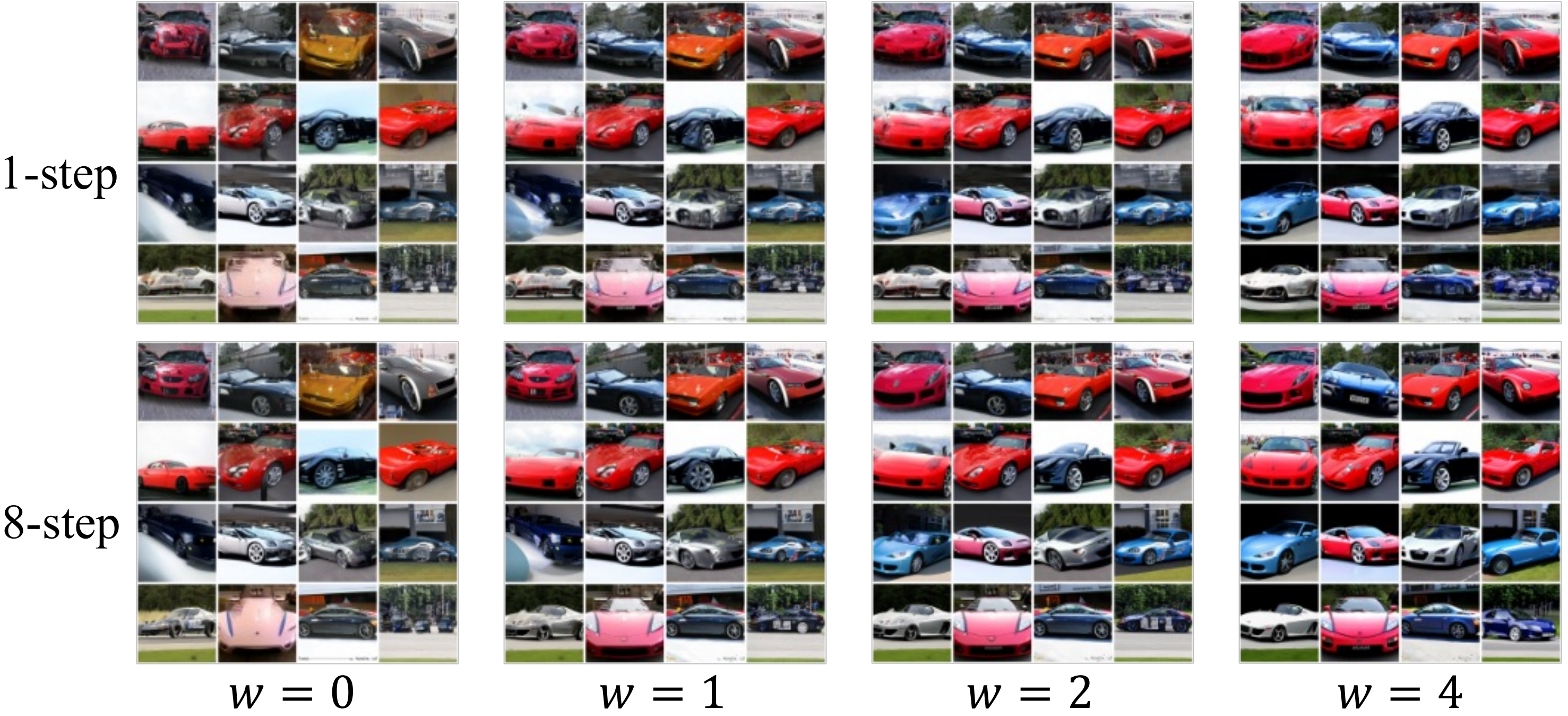}
    \caption{
    Class-conditional samples from our two-stage (deterministic) approach on ImageNet 64x64 for diffusion models trained on the pixel-space. By varying the guidance weight $w$, our distilled model is able to trade-off between sample diversity and quality, while achieving good results using as few as \emph{one} sampling step.
\vspace{-1em}
    }
    \label{fig:ours_d_sweep}
    \vspace{-10pt}
\end{figure}

Our distillation framework can not only be applied to standard diffusion models trained on the pixel-space~\cite{ho2020denoising, sohl2015deep,song2019generative}, but also diffusion models trained on the latent-space of an autoencoder~\cite{sinha2021d2c,rombach2022high} (\eg, Stable Diffusion~\cite{rombach2022high}).
For diffusion models directly trained on the pixel-space, our experiments on ImageNet 64x64 and CIFAR-10 show that the proposed distilled model can generate samples visually comparable to that of the teacher using only 4 steps and is able to achieve comparable FID/IS scores as the teacher model using as few as 4 to 16 steps on a wide range of guidance strengths (see \cref{fig:ours_d_sweep}).
For diffusion model trained on the latent-space of an encoder~\cite{sinha2021d2c, rombach2022high},  
our approach is able to achieve comparable visual quality to the base model using as few as 1 to 4 sampling steps (at least 10$\times$ fewer steps than the base model) on ImageNet 256$\times$256 and LAION 512$\times$512, matching the performance of the teacher (as evaluated by FID) with only 2-4 sampling steps.
To the best of our knowledge, our work is the first to demonstrate the effectiveness of distillation for both pixel-space and latent-space classifier-free diffusion models. 
Finally, we apply our method to text-guided image inpainting and text-guided image editing tasks~\cite{meng2021sdedit}, where we reduce the total number of sampling steps to as few as 2-4 steps, demonstrating the potential of the proposed framework in style-transfer and image-editing applications~\cite{su2022dual,meng2021sdedit}.

%% file: 2_background.tex
\section{Background on diffusion models}
\label{sec:background}
Given samples $\rvx$ from a data distribution $p_\text{data}(\rvx)$, noise scheduling functions $\alpha_t$ and $\sigma_t$,
we train a diffusion model $\hat{\rvx}_{{\bm\theta}}$, with parameter ${\bm\theta}$, via minimizing the weighted mean squared error~\cite{ho2020denoising,sohl2015deep,song2019generative,song2020score}
\begin{equation}
\mathbb{E}_{t\sim U[0,1], \rvx\sim p_{\text{data}}(\rvx),\rvz_t\sim q(\rvz_t|\rvx) }[\omega(\lambda_t)||\hat{\rvx}_{{\bm\theta}}(\rvz_t)-\rvx||_2^2],
\label{eq:dmobjetive}
\end{equation}
where $\lambda_t=\log[\alpha_t^2/\sigma_t^2]$ is a signal-to-noise ratio \cite{kingma2021variational}, $q(\rvz_t|\rvx)=\mathcal{N}(\rvz_t; \alpha_t \rvx, \sigma_t^2 \mathbf{I})$ and $\omega(\lambda_t)$ is a pre-specified weighting function~\cite{kingma2021variational}.

Once the diffusion model $\hat{\rvx}_{{\bm\theta}}$ is trained, one can use discrete-time DDIM sampler~\cite{song2020denoising} to sample from the model.
Specifically, 
the DDIM sampler starts with $\rvz_1\sim \mathcal{N}(\bm{0}, \textbf{I})$ and updates as follows
\begin{align}
\label{eq:ddim_sampling}
    \rvz_s=\alpha_s\hat{\rvx}_{{\bm\theta}}(\rvz_t)+\sigma_s\frac{\rvz_t-\alpha_t \hat{\rvx}_{{\bm\theta}}(\rvz_t)}{\sigma_t},\; s = t - 1/N
\end{align}
with $N$ the total number of sampling steps.
The final sample will then be generated using $\hat{\rvx}_{\bm\theta}(\rvz_0)$.

\begin{figure*}[H]
\vspace{-15pt}
    \centering
    \includegraphics[width=\linewidth]{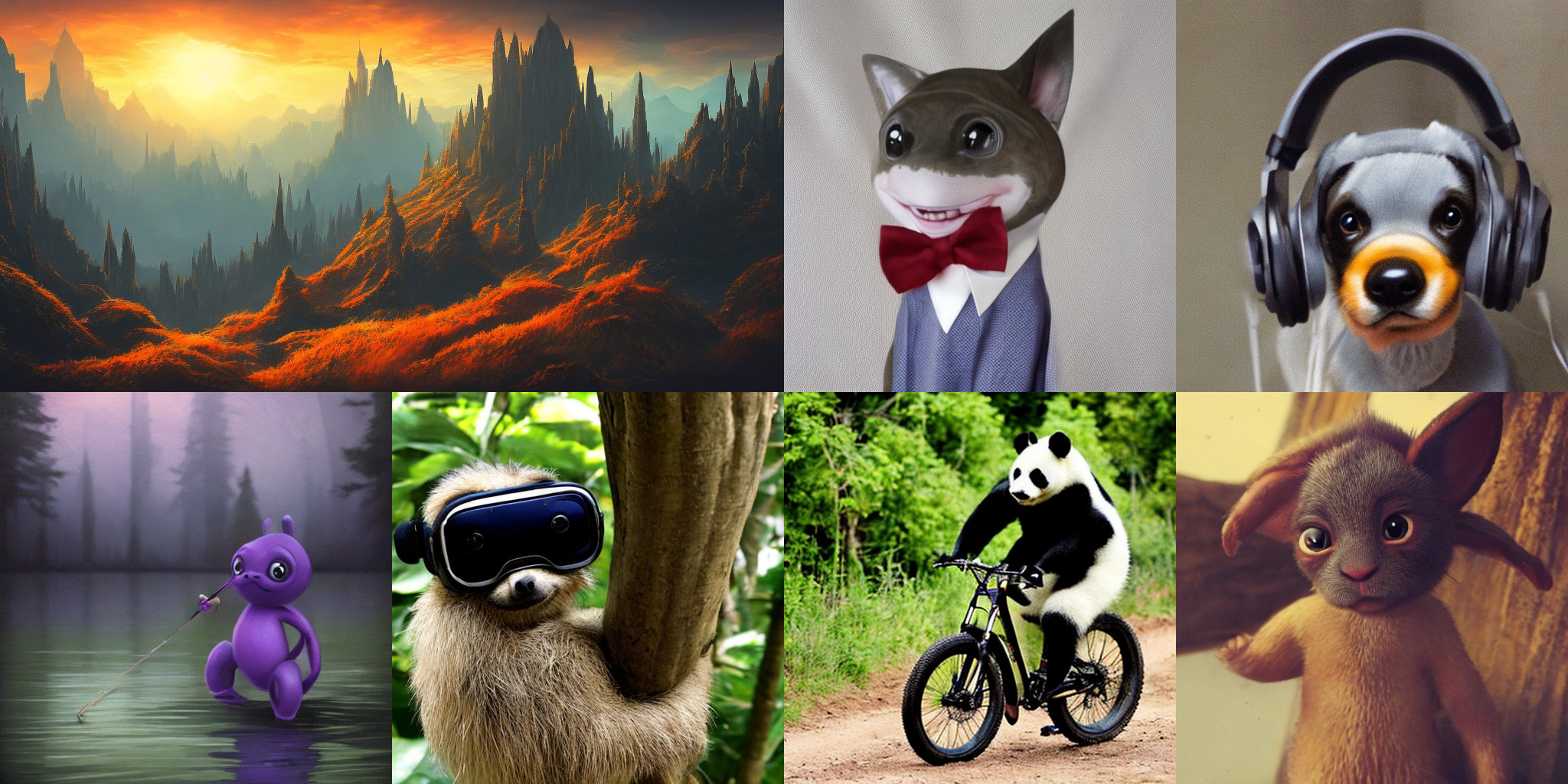}
    \caption{
    Distilled Stable Diffusion samples produced in 4 sampling steps. \todo{A caption}\chenlin{perhaps we can add one or two more images per row? now the images look really large}
    }
    \label{fig:teaser_1}
\end{figure*}

\textbf{Classifier-free guidance }
Classifier-free guidance~\cite{ho2022classifier} is an effective approach shown to significantly improve the sample quality of class-conditioned diffusion models, and has been widely used in large-scale diffusion models including GLIDE~\cite{nichol2021glide}, Stable Diffusion~\cite{rombach2022high}, DALL$\cdot$E 2~\cite{ramesh2022hierarchical} and Imagen~\cite{saharia2022photorealistic}. 
Specifically, it introduces a guidance weight parameter $w\in \mathbb{R}^{\ge 0}$ to trade-off between sample quality and diversity. 
To generate a sample, classifier-free guidance evaluates both a conditional diffusion model $\hat{\rvx}_{c, {\bm\theta}}$, where $c$ is the context (\eg, class label, text prompt) to be conditioned on, and a jointly trained unconditional diffusion model $\hat{\rvx}_{{\bm\theta}}$ at each update step, using
$\hat{\rvx}_{{\bm\theta}}^{w}=(1+w)\hat{\rvx}_{c, {\bm\theta}}-w\hat{\rvx}_{{\bm\theta}}$ 
as the model prediction in \cref{eq:ddim_sampling}. 
As each sampling update requires evaluating two diffusion models, sampling with classifier-free guidance is often expensive~\cite{ho2022classifier}. 

\textbf{Progressive distillation }
Our approach is inspired by \emph{progressive distillation}~\cite{salimans2022progressive}, an effective method for improving the sampling speed of (unguided) diffusion models by repeated distillation. 
Until now, this method could not be directly applied to distilling classifier-free guided models or studied for samplers other than the deterministic DDIM sampler~\cite{song2020denoising,salimans2022progressive}. 
In this paper we resolve these shortcomings.

\textbf{Latent diffusion models (LDMs)}~\cite{rombach2022high,sinha2021d2c,preechakul2022diffusion,mittal2021symbolic} increase the training and inference efficiency of diffusion models (directly learned on the pixel-space) by modeling images in the latent space of a pre-trained regularized autoencoder, where the latent representations are usually of lower dimensionality than the pixel-space. Latent diffusion models can be considered as an alternative to cascaded diffusion approaches~\cite{ho2021cascaded}, which rely on one or more super-resolution diffusion models to scale up a low-dimensional image to the desired target resolution.

In this work, we will apply our distillation framework to classifier-free guided diffusion models learned on both pixel-space~\cite{ho2020denoising,sohl2015deep,song2019generative} and latent-space~\cite{rombach2022high,sinha2021d2c,mittal2021symbolic,preechakul2022diffusion}.

%% file: 3_method.tex
\section{Distilling a guided diffusion model}
\label{sec:method}
In the following, we discuss our approach for distilling a classifier-free guided diffusion model~\cite{ho2022classifier} into a student model that requires fewer steps to sample from. Using a \emph{single} distilled model conditioned on the guidance strength, our model can capture a wide range of classifier-free guidance levels, allowing for the trade-off between sample quality and diversity efficiently. 

Given a trained guided model $[\hat{\rvx}_{c, \bm\theta}, \hat{\rvx}_{\bm\theta}]$ (teacher) either on the pixel-space or latent-space, our approach can be decomposed into two stages.

\begin{figure*}[!ht]
     \centering
     \begin{subfigure}[b]{0.32\textwidth}
         \centering
         \includegraphics[width=\textwidth]{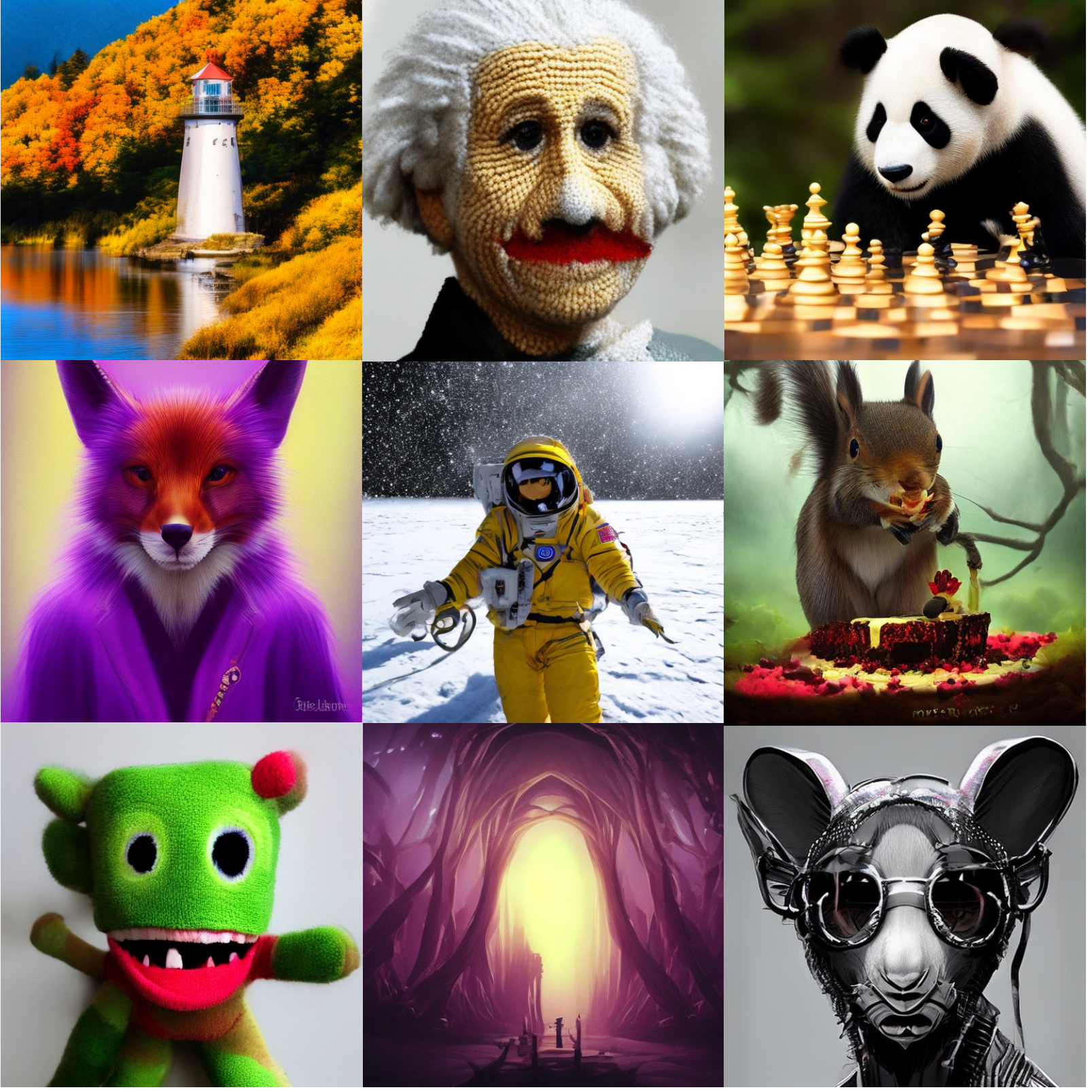}
         \caption{2 denoising steps}
     \end{subfigure}
     \begin{subfigure}[b]{0.32\textwidth}
         \centering
         \includegraphics[width=\textwidth]{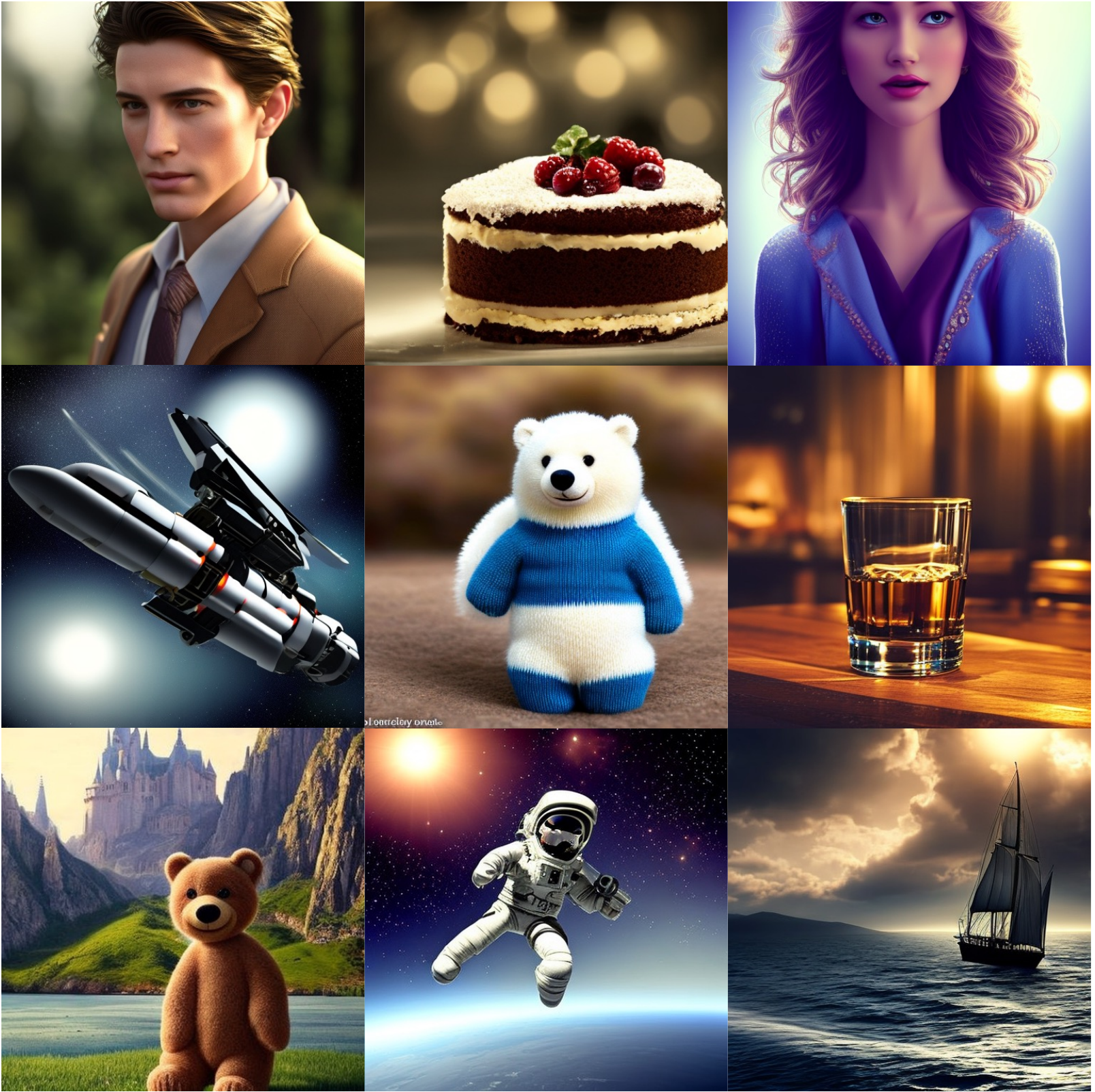}
         \caption{4 denoising steps}
     \end{subfigure}
     \begin{subfigure}[b]{0.32\textwidth}
         \centering
         \includegraphics[width=\textwidth]{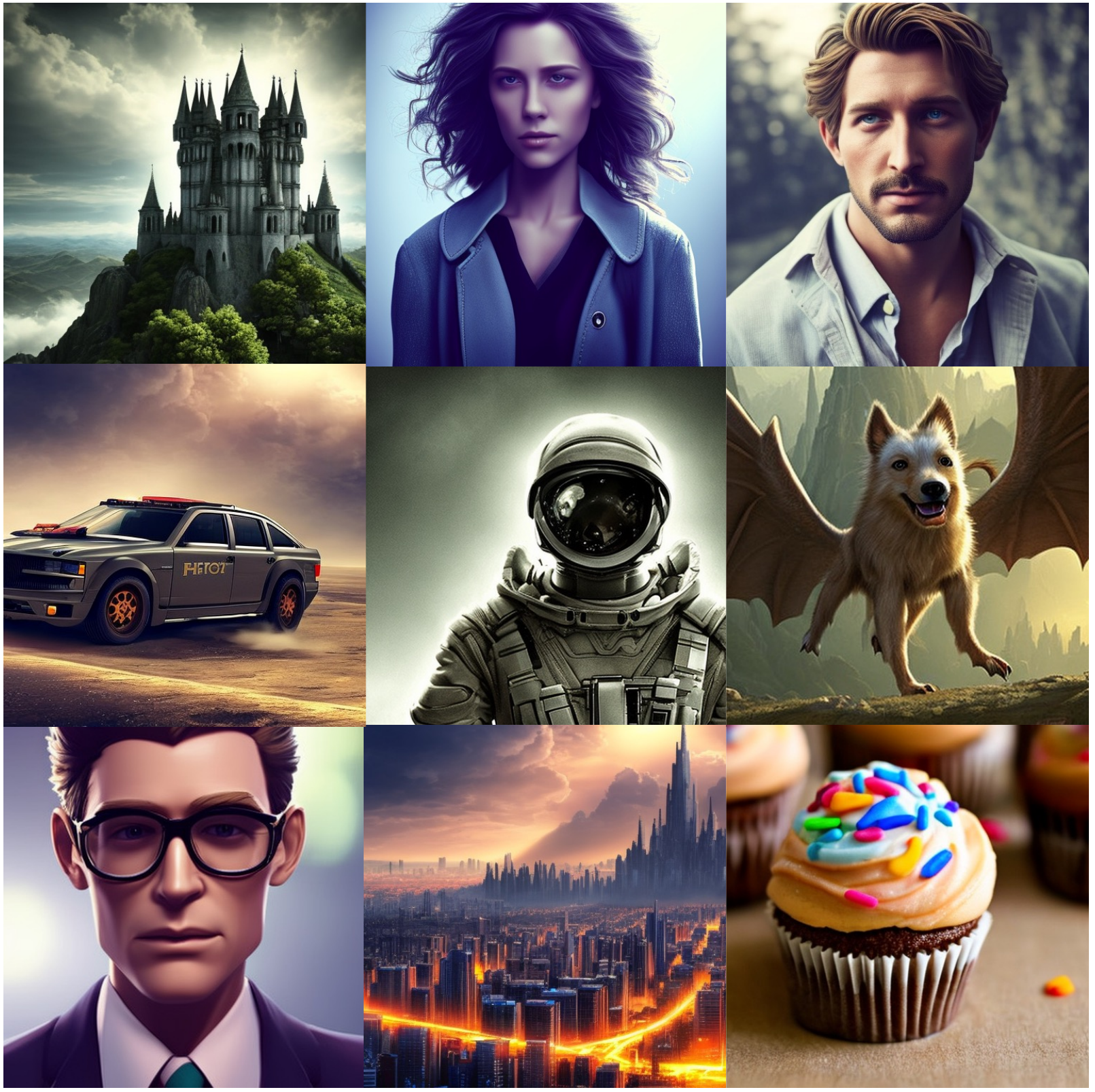}
         \caption{8 denoising steps}
     \end{subfigure}
    \caption{Text-guided generation on LAION (512x512) using our distilled Stable Diffusion model. Our model is able to generate high-quality image samples using 2, 4 or 8 denoising steps, significantly improving the inference efficiency of Stable Diffusion.}
    \label{fig:text2img_demo}
    \vspace{-10pt}
\end{figure*}

\subsection{Stage-one distillation}
\label{app:sec:step_one}

\input{table_main}

In the first stage, 
we introduce a student model $\hat{\rvx}_{{\bm\eta}_1}(\rvz_t, w)$, with learnable parameter ${\bm\eta}_1$, to match the output of the teacher at any time-step $t\in [0,1]$. The student model can either be a continuous-time model~\cite{song2020score} or a discrete-time model~\cite{ho2020denoising,song2020denoising} depending on whether the teacher model is discrete or continuous. 
For simplicity, in the following discussion, we assume both the student and teacher models are continuous  as the algorithm for discrete models is almost identical.

A key functionality of classifier-free guidance~\cite{ho2022classifier} is its ability to easily trade-off between sample quality and diversity, which is controlled by a ``guidance strength" parameter. This property has demonstrated utility in real-world applications~\cite{ho2022classifier,saharia2022photorealistic,rombach2022high,nichol2021glide,ramesh2022hierarchical}, where the optimal ``guidance strength" is often a user preference.
Thus, we would also want our distilled model to maintain this property.
Given a range of guidance strengths $[w_{\text{min}}, w_{\text{max}}]$ we are interested in, 
we optimize the student model using the following objective
\begin{equation}
\small
\mathbb{E}_{w\sim p_w, t\sim U[0,1], \rvx\sim p_{\text{data}}(\rvx)} \bigg[ \omega(\lambda_t) \lVert \hat{\rvx}_{{\bm\eta}_1}(\rvz_t, w) -  \hat\rvx^{w}_{\bm\theta}(\rvz_t)\rVert_{2}^{2}\bigg],
\end{equation}
where $ \; \hat{\rvx}_{\bm\theta}^{w}(\rvz_t)=(1+w)\hat{\rvx}_{c, \bm\theta}(\rvz_t)-w\hat{\rvx}_{\bm\theta}(\rvz_t)$, 
$\rvz_t \sim q(\rvz_t | \rvx)$ and $p_w(w) = U[w_{\text{min}}, w_{\text{max}}]$. 
Note that here, our distilled model $\hat{\rvx}_{{\bm\eta}_1}(\rvz_t, w)$ is also conditioned on the context $c$ (\eg, text prompt), but we drop the notation $c$ in the paper for simplicity.
We provide the detailed training algorithm in \cref{alg:stage1} in the supplement. 

To incorporate the guidance weight $w$, we introduce a $w$-conditioned model, where $w$ is fed as an input to the student model. To better capture the feature, we apply Fourier embedding to $w$, which is then incorporated into the diffusion model backbone in a way similar to how the time-step was incorporated in ~\cite{kingma2021variational,salimans2022progressive}. 
As initialization plays a key role in the performance~\cite{salimans2022progressive}, we initialize the student model with the same parameters as the conditional model of the teacher, except for the newly introduced parameters related to $w$-conditioning. 
The model architecture we use is a  U-Net model similar to the ones used in \cite{ho2022classifier} for pixel-space diffusion models and \cite{dhariwal2021diffusion,rombach2022high} for latent-space diffusion models.
We use the same number of channels and attention as used in \cite{ho2022classifier} and the open-sourced Stable Diffusion repository\footnote{\url{https://github.com/CompVis/stable-diffusion}} for our experiments. We provide more details
in the supplement.

\subsection{Stage-two distillation}
\label{app:sec:step_two}
In the second stage, we consider a discrete time-step scenario and progressively distill the learned model from the first-stage $\hat{\rvx}_{{\bm\eta}_1}(\rvz_t, w)$ into an fewer-step student model $\hat{\rvx}_{{\bm\eta}_2}(\rvz_t, w)$ with learnable parameter ${\bm\eta}_2$, by halving the number of sampling steps each time.
Letting $N$ denote the number of sampling steps, given $w\sim U[w_{\text{min}} , w_{\text{max}}]$ and $t\in \{1,...,N\}$, we train the student model to match the output of two-step DDIM sampling of the teacher (i.e., from $t/N$ to $t-0.5/N$ and from $t-0.5/N$ to $t-1/N$) in one step, following the approach of \cite{salimans2022progressive}.
After distilling the $2N$ steps in the teacher model to $N$ steps in the student model, we can use the $N$-step student model as the new teacher model, repeat the same procedure, and distill the teacher model into a $N/2$-step student model.
At each step, we initialize the student model with the parameters of the teacher. We provide the training algorithm and extra details in the supplementary material.

\subsection{$N$-step deterministic and stochastic sampling }
\begin{figure}%
\centering
\vspace{-1em}
\includegraphics[width=\linewidth]{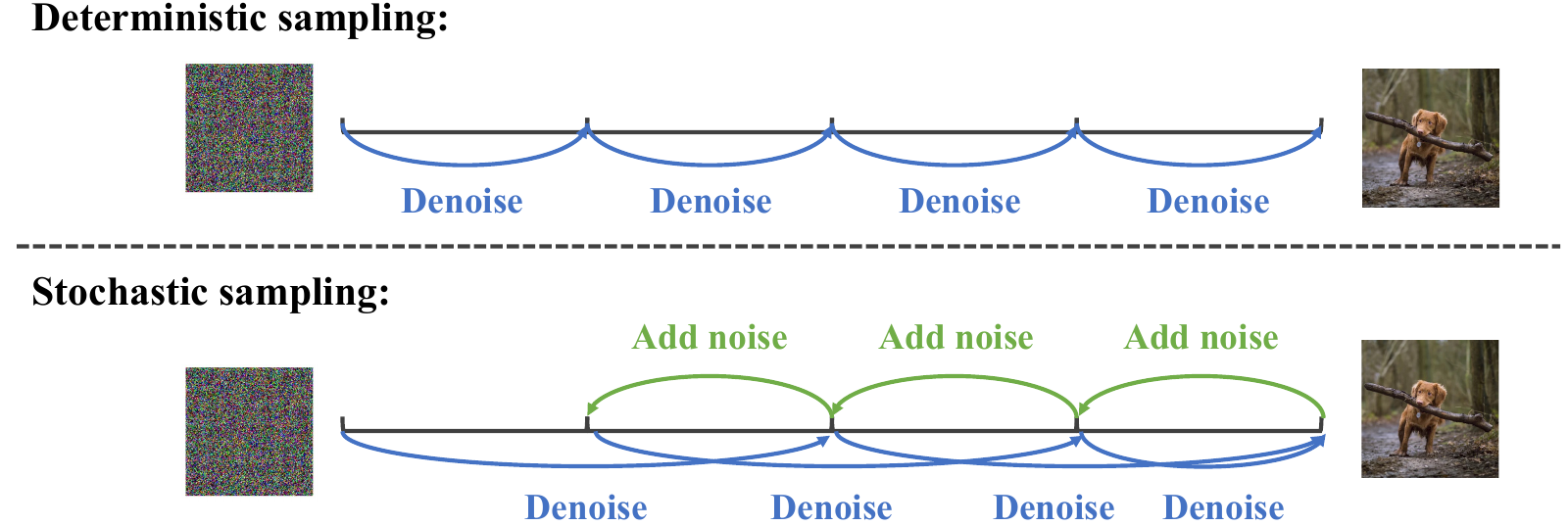}
\caption{Sampling procedures of the distilled model where the number of denoising steps is 4.}
\label{fig:sampling_illustration}
\vspace{-1.5em}
\end{figure}

Once the model $\hat{\rvx}_{{\bm\eta}_2}$ is trained, given a specified guidance strength $w\in[w_{\text{min}} , w_{\text{max}}]$, we can perform sampling via the DDIM update rule in \cref{eq:ddim_sampling}.
We note that given the distilled model $\hat{\rvx}_{{\bm\eta}_2}$, this sampling procedure is \textit{deterministic} given the initialization $\rvz_{1}^w$. 
In fact, we can also perform $N$-step \textit{stochastic} sampling: We apply one deterministic sampling step with two-times the original step-length (i.e., the same as a $N/2$-step deterministic sampler) and then perform one stochastic step backward (i.e., perturb with noise) using the original step-length, a process inspired by \cite{karras2022elucidating}.
With  $\rvz_1^w\sim \mathcal{N}(\textbf{0}, \textbf{I})$, we use the following update rule when $t>1/N$
\begin{align}
\label{eq:distilled_sampling_s}
    \rvz^w_{k}&=\alpha_k \hat{\rvx}_{{\bm\eta}_2}(\rvz^w_{t})+
    \sigma_k\frac{\rvz^w_t-\alpha_t \hat{\rvx}^w_{{\bm\eta_2}}(\rvz_t)}{\sigma_t},\\
    \text{where }\rvz^w_{s}&=(\alpha_{s}/\alpha_{k})\rvz^w_{k}+\sigma_{s|k}\bm\eps, \;
    \bm\eps \sim \mathcal{N}(\textbf{0}, \textbf{I}); \\
    \rvz^w_{h}&=\alpha_h \hat{\rvx}_{{\bm\eta}_2}(\rvz^w_{s})+
    \sigma_h\frac{\rvz^w_s-\alpha_s \hat{\rvx}^w_{{\bm\eta_2}}(\rvz^w_s)}{\sigma_s},\\
    \text{where }\rvz^w_{k}&=(\alpha_{k}/\alpha_{h})\rvz^w_{h}+\sigma_{k|h}\bm\eps, \; \bm\eps \sim \mathcal{N}(\textbf{0}, \textbf{I}).
\end{align}
In the above equations, $h=t-3/N$, $k=t-2/N$,  $s=t-1/N$ and
$\sigma^2_{a|b}=(1-e^{\lambda_a-\lambda_b})\sigma_a^2$.
When $t=1/N$, we use deterministic update \cref{eq:ddim_sampling} to obtain $\rvz_{0}^w$ from $\rvz_{1/N}^w$. We provide an illustration of the process in \cref{fig:sampling_illustration}, where the number of denoising steps is 4.
We note that compared to the \emph{deterministic} sampler,  performing \emph{stochastic} sampling requires evaluating the model at slightly different time-steps, and would require small modifications to training algorithm for the edge cases. We provide the algorithm and more details in the supplementary material.

%% file: table_main.tex
\begin{table*}%
\centering
\resizebox{\linewidth}{!}{
\begin{tabular}{cccccccccc}
\Xhline{2\arrayrulewidth}
& \multicolumn{2}{c}{$w=0$} &\multicolumn{2}{c}{$w=0.3$} &\multicolumn{2}{c}{$w=1$} &\multicolumn{2}{c}{$w=4$}\\
\Xhline{1\arrayrulewidth}
 Method & FID ($\downarrow$) & IS ($\uparrow$) &{FID ($\downarrow$)} & IS ($\uparrow$) &{FID ($\downarrow$)} & IS ($\uparrow$) &{FID ($\downarrow$)} & IS ($\uparrow$) \\
\Xhline{1\arrayrulewidth}
Ours 1-step (D/S)     &22.74 / 26.91  &25.51 / 23.55 &14.85 / 18.48   &37.09 / 33.30  &7.54  / 8.92  & 75.19 / 67.80 &18.72  / \textbf{17.85}  &157.46 / 148.97\\
Ours 4-step (D/S)     &4.14 / 3.91 &46.64 / 48.92 &2.17  / 2.24  &69.64 / 73.73 &7.95 / 8.51 & 128.98  / 135.36  &26.45 / 27.33  &207.45 / 216.56\\
Ours 8-step (D/S)     &2.79 / 2.44 &50.72 / 55.03  &\textbf{2.05} / 2.31  &76.01 / 83.00  &9.33  / 10.56   &136.47  / 147.39  &26.62 / 27.84  &203.47 / \textbf{219.89} \\
Ours 16-step (D/S)     &2.44 / \textbf{2.10}  &52.53 / \textbf{57.81}  &2.20 / 2.56  &79.47 / \textbf{87.50}  &9.99 / 11.63 &139.11 / \textbf{153.17} &26.53 / 27.69  &204.13 / 218.70 \\
Single-$w$ 1-step   &19.61  &24.00  &11.70  &36.95  &\textbf{6.64}  &74.41  &19.857  &170.69  \\
Single-$w$ 4-step   &4.79  &38.77  &2.34  &62.08  &8.23   &118.52  &27.75  &219.64\\
Single-$w$ 8-step   &3.39 &42.13 &2.32  &68.76 &9.69  &125.20 &27.67 &218.08 \\
Single-$w$ 16-step   &2.97  &43.63  &2.56   &70.97  &10.34  &127.70 &27.40 &216.52 \\
DDIM 16x2-step~\cite{song2020denoising}     &7.68 &37.60   &5.33 &60.83  &9.53   &112.75  &21.56 &195.17\\
DDIM 32x2-step~\cite{song2020denoising}     &5.03 &40.93 &7.47 &9.33 &9.26   &126.22 &23.03  &213.23 \\
DDIM 64x2-step~\cite{song2020denoising}     &3.74 &43.16 &5.52  &9.51 &9.53  &133.17 &23.64 &217.88\\
\Xhline{1\arrayrulewidth}
Teacher (DDIM 1024x2-step) &2.92 &44.81 &2.36  &74.83 &9.84 &139.50 &23.94 & 224.74\\

\Xhline{2\arrayrulewidth}
\end{tabular}
}
\caption{ 
ImageNet 64x64 distillation results for pixel-space diffusion models ($w=0$ refers to non-guided models). For our method, $D$ and $S$ stand for deterministic and stochastic sampler respectively. We observe that training the model conditioned on a guidance interval $w\in [0,4]$ performs comparably with training a model on a fixed $w$ (see {Single-}$w$). Our approach significantly outperforms DDIM when using fewer steps, and is able to match the teacher performance using as few as 8 to 16 steps.
}
\label{table:i64_main}
\vspace{-1em}
\end{table*}

%% file: 4_experiment.tex
\section{Experiments}
In this section, we evaluate the performance of our distillation approach on pixel-space diffusion models (\ie DDPM~\cite{ho2020denoising}) and latent-space diffusion models (\ie Stable Diffusion~\cite{rombach2022high}). We further apply our approach to text-guided image editing and inpainting tasks.
Experiments show that our approach is able to achieve competitive performance while using as few as 2-4 steps on all tasks.

\subsection{Distillation for pixel-space guided models }
\label{sec:pixelspace}
In this experiment, we consider class-conditional diffusion models directly trained on the pixel-space~\cite{ho2020denoising,ho2022classifier,salimans2022progressive}.

\textbf{Settings }
We focus on ImageNet 64x64~\cite{russakovsky2015imagenet} and CIFAR-10~\cite{krizhevsky2009learning}  as higher-resolution image generation in this scenario often relies on combining with other super-resolution techniques~\cite{ho2021cascaded,saharia2022photorealistic}. We explore different ranges for the guidance weight and observe that all ranges work comparably and therefore use $[w_{min}, w_{max}]=[0,4]$ for the experiments. 
The baselines we consider include DDPM ancestral sampling~\cite{ho2020denoising} and DDIM~\cite{song2020denoising}. The teacher model we use is a 1024x2-step DDIM model, where the conditional and unconditional components both use 1024 DDIM denoising steps. To better understand how the guidance weight $w$ should be incorporated, we also include models trained using a single fixed $w$ as a baseline. 
We use the same pre-trained teacher model for all the methods for fair comparisons. 
Following~\cite{ho2020denoising,ho2022classifier,song2019generative}, we use a U-Net~\cite{ronneberger2015u,song2019generative} architecture for the baselines, and the same U-Net backbone with the introduced $w$-embedding for our two-step student models (see \cref{sec:method}). 
Following \cite{salimans2022progressive}, we use a
$\rvv$-prediction model for both datasets. 

\begin{figure*}[!ht]
 \vspace{-5pt}
    \centering
    \includegraphics[width=0.95\linewidth]{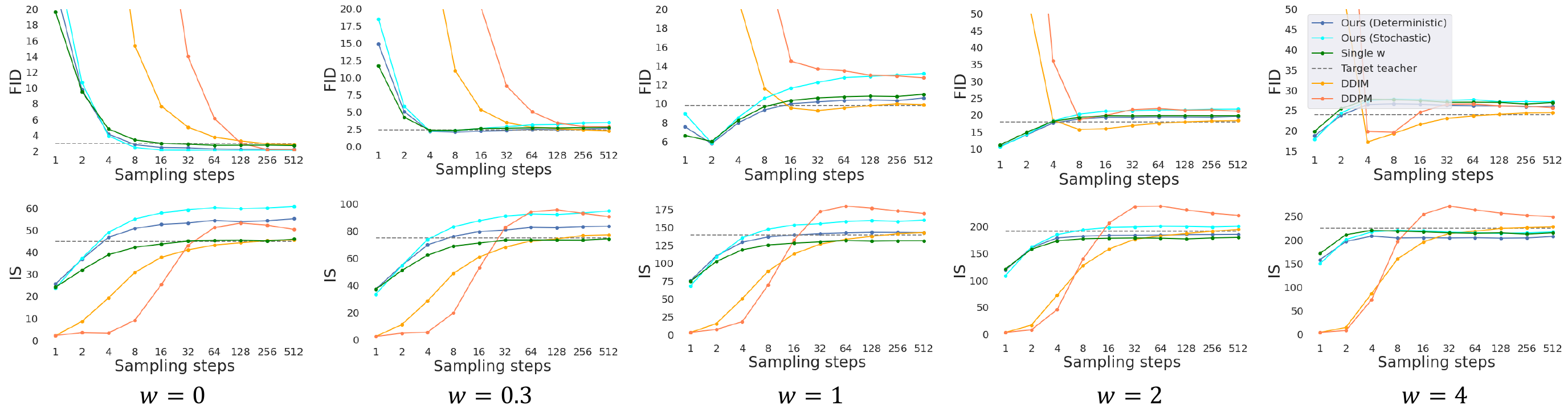}
    \caption{ImageNet 64x64 sample quality evaluated by FID and IS scores. Our distilled model significantly outperform the DDPM and DDIM baselines, and is able to match the performance of the teacher using as few as 8 to 16 steps. By varying $w$, a \emph{single} distilled model is able to capture the trade-off between sample diversity and quality.
    }
    \label{fig:i64_curve}
    \vspace{-6pt}
\end{figure*}

\textbf{Results }
We report the performance as evaluated in FID~\cite{heusel2017gans} and Inception scores (IS)~\cite{salimans2016improved} for all approaches on ImageNet 64x64
in \cref{fig:i64_curve} and \cref{table:i64_main} and provide extended ImageNet 64x64 and CIFAR-10 results in the supplement. 
We observe that our distilled model is able to match a teacher guided DDIM model with 1024x2 sampling steps using only 4-16 steps, achieving a speedup for up to $256\times$. We emphasize that, using our approach, a \emph{single} distilled model is able to match the teacher performance on a wide range of guidance strengths. This has not been achieved by any previous methods.

\subsection{Distillation for latent-space guided models}
\label{sec:sd}
\distilledvsbase

After demonstrating the effectiveness of our method on pixel-space class-guided diffusion models in \cref{sec:pixelspace}, we now expand its scope to latent-space diffusion models. In the following sections, we show the effectiveness of our approach on Latent Diffusion~\cite{rombach2022high} %
on a variety of tasks, including class-conditional generation, text-to-image generation, image inpainting and text-guided style-transfer~\cite{meng2021sdedit}.

 \imgimg

In the following experiments, we use the open-sourced latent-space diffusion models~\cite{rombach2022high} as the teacher models. 
As $\rvv$-prediction teacher model tends to perform better than $\epsilon$-prediction model, we
\emph{fine-tune} the open-sourced $\rvepsilon$-prediction models into $\rvv$-prediction teacher models. We provide more details in the supplementary material.

\subsubsection{Class-conditional generation}
In this section, we apply our method to a class-conditional latent diffusion model pre-trained on ImageNet $256 \times 256$. 
We start from the DDIM teacher model with 512 sampling steps, and use the output as the target to train our distilled model.
We use a batch size of 512 and uniformly sample the guidance strength $w \in [w_{min}=0, w_{max}=14]$ during training. %

\textbf{Results }
Empirically, we find that our distilled model is able to match the performance of the teacher model (originally trained on 1000 steps) in terms of FID scores while using only 2 or 4 sampling steps. %
We also achieve significantly better performance than DDIM when using 1-4 sampling steps (see \cref{fig:cinldmresults}). Qualitatively, we find that samples synthesized using a single denoising step still yield satisfying results, while the baseline fails to generate images with meaningful contents. We provide extra samples in the supplementary material. 

Similar to the pixel-based results in \cref{fig:i64_curve}, we also observe the trade-off between sampling quality and diversity as measured by FID and Inception Score for our distilled latent diffusion model. Following \emph{Kynk{\"a}{\"a}nniemi et al}~\cite{kynkaanniemi2019improved}, we further compute improved precision and recall metrics for this experiment in the appendix.

\subsubsection{Text-guided image generation}
\label{sec:text_guided}

In this section, we focus on the text-guided Stable Diffusion model pretrained on  subsets~\footnote{\url{https://github.com/CompVis/stable-diffusion/blob/main/Stable_Diffusion_v1_Model_Card.md}} of LAION-5B~\cite{schuhmann2022laion} at a resolution $512\times 512$.
We then follow our two-stage approach introduced in \cref{sec:method} and distill the guided model in 3000 gradient updates into a $w$-conditioned model using $w \in [w_{min}=2, w_{max}=14]$, %
and a batch size of 512. 
Although we can condition on a broader range of $w$ for the distilled (student) model, the utility remains unclear as we typically do not exceed the normal guidance range when sampling with the teacher model.
The final model is obtained by applying progressive distillation for 2000 training steps per stage, except when for the low-step regime of 1,2, and 4 steps, where we train for 20000 gradient updates. A detailed analysis of the convergence properties of this model in the supplement.

\begin{table}[!h]
    \centering
    \vspace{-1em}
    \resizebox{0.7\linewidth}{!}{%
    \begin{tabular}{l c c c}
        \toprule
        \textbf{Method} & 2-step & 4-step & 8-step \\
         \midrule
         DPM~\cite{dpmsolver} & 98.9/0.20 & 34.3/0.29 & 31.7/0.32 \\
         DPM++~\cite{dpmpp} & 98.8/0.20 & 34.1/0.29 & 25.6/0.32 \\
        \textbf{Ours} & 37.3/0.27 & 26.0/0.30 & 26.9/0.30 \\
        \bottomrule
        \vspace{-2.2em}
    \end{tabular}
    }
     \caption{
     FID/CLIP scores on LAION 512X512 ($w=8.0$). 
        We point out that DPM and DPM++ use both the conditional and unconditional components for sampling. Depending on the implementation, this either requires higher peak memory or two times more sampling steps.
        \vspace{-1em}}
    \label{table:dpm_dpmpp}
\end{table}

\textbf{Results } 
We present samples in \cref{fig:text2img_demo}. We evaluate the resulting model both qualitatively and quantitatively. For the latter analysis, we follow \cite{saharia2022photorealistic} and evaluate CLIP~\cite{radford2021learning} and FID scores to asses text-image alignment and quality, respectively. We use the open-sourced ViT-g/14~\cite{ilharco_gabriel_2021_5143773} CLIP model for evaluation.
The quantitative results in \cref{fig:stableplot} show that our method can significantly increase the performance in both metrics over DDIM sampling on the base model for 2 and 4 sampling steps. 
For 8 steps, these metrics do not show a significant difference.
However, when considering the corresponding samples in \cref{fig:distilledvsbase} we can observe a stark difference in terms of visual image quality. In contrast to the 8-step DDIM samples from the original model, the distilled samples are sharper and more coherent. We hypothesize that FID and CLIP do not fully capture these differences in our evaluation setting on COCO2017~\cite{lin2014microsoft}, where we used 5000 random captions from the validation set.
We further compute the FID and CLIP scores for our distilled LAION 512x512 model and compare them with the DPM~\cite{dpmsolver} and DPM++~\cite{dpmpp} solver in \cref{table:dpm_dpmpp}. We observe that our method is able to achieve significantly better performance when the denoising step is 2 or 4.
Furthermore, we stress that stage-one of our method already decreases the number of function evaluations by a factor of 2, as we distill the classifier-free guidance step into a single model. Depending on the exact implementation (batched vs. sequential network evaluation), this either decreases peak memory or sampling time compared to existing solvers~\cite{song2020denoising,dpmsolver,dpmpp}.

 \inpainting

  \stableplot
  \cinldmresults

\begin{figure}%
    \centering
    \includegraphics[width=\linewidth]{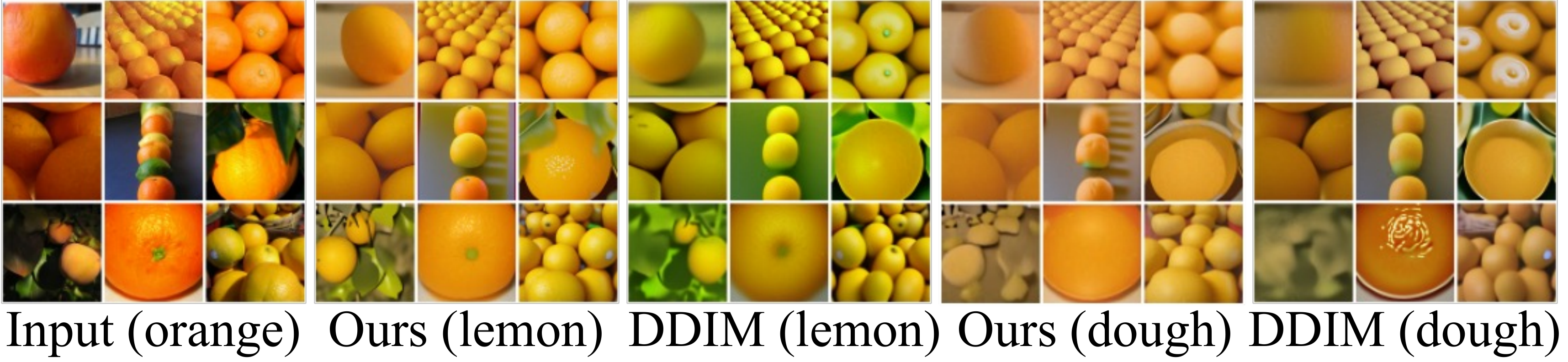}
    \caption{Style transfer comparison on ImageNet 64x64 for pixel-space models. For our approach, we use a distilled encoder and decoder. For the baseline, we encode and decode using DDIM. We use $w=0$ and $16$ sampling steps for both the encoder and decoder. We observe that our method achieves more realistic outputs.
    }
    \label{fig:style_comparison}
\end{figure}

\subsubsection{Text-guided image-to-image translation}
In this section, we perform experiments on text-guided image-to-image translation with SDEdit~\cite{meng2021sdedit} using our distilled model from \cref{sec:text_guided}. Following SDEdit~\cite{meng2021sdedit}, we perform stochastic encoding in the latent space, but instead use the deterministic sampler of the distilled model to perform deterministic decoding.
We consider input image and text of various kinds and  provide qualitative results in \cref{fig:img2img}.
We observe that our distilled model generates high-quality style-transfer results using as few as 3 denoising steps. We provide more analysis on the trade-off between sample quality, controllability and efficiency in the supplement.

\subsubsection{Image inpainting}
In this section, we apply our approach to a pre-trained image inpainting latent diffusion model. We use the open-source \emph{Stable Diffusion Inpainting}\footnote{\url{https://huggingface.co/runwayml/stable-diffusion-inpainting}} image-inpainting model. 
This model is a fine-tuned version of the pure text-to-image \emph{Stable Diffusion} model from above, where additional input channels were added to process masks and masked images.

We use the same distillation algorithm as used in the previous section. For training, we start from the $\rvv$-prediction teacher model sampled with 512 DDIM steps, and use the output as the target to optimize our student model.
We present qualitative results in \cref{fig:image_inpainting}, demonstrating the potential of our method for fast, real-world image editing applications.
For additional training details and a quantitative evaluation, see the supplementary.

\subsection{Progressive distillation for encoding }
In this experiment, we explore distilling the encoding process for the teacher model and perform experiments on style-transfer in a setting similar to \cite{su2022dual}.
We focus on pixel-space diffusion models pre-trained on ImageNet 64$\times$ 64.
Specifically, to perform style-transfer between two domains $A$ and $B$, we encode the image from domain-$A$ using a diffusion model trained on domain-$A$, and then decode with a diffusion model trained on domain-$B$. As the encoding process can be understood as reversing the DDIM sampling process, 
we perform distillation for both the encoder and decoder with classifier-free guidance, and compare with a DDIM encoder and decoder in \cref{fig:style_comparison}. We also explore how modifying the guidance strength $w$ can impact the performance and provide more details in the supplementary material.

%% file: 5_related.tex
\section{Related Work}

Our approach is related to existing works on improving the sampling speed of diffusion models~\cite{sohl-dickstein2015deep,ho2020denoising,song2020score}.
For instance, denoising diffusion implicit models (DDIM~\cite{song2020denoising}), probability flow sampler~\cite{song2020score}, fast SDE integrators~\cite{jolicoeur2021gotta} have been proposed to improve the sampling speed of diffusion models. 
Other works develop higher-order solvers~\cite{lu2022dpm}, exponential integrators \cite{liu2022pseudo}, and  dynamic programming based approach~\cite{watson2022learning}  to accelerating sampling speed.
However, none of these approaches have achieved comparable performance as our method on distilling classifier-free guided diffusion models.

Existing distillation-based methods for diffusion models are mainly designed for non-classifier-free guided diffusion models.
For instance, \cite{luhman2021knowledge} proposes to predict the data from noise in one single step by inverting a deterministic encoding of DDIM, \cite{dockhorn2022genie} proposes to achieve faster sampling speed by distilling higher order solvers into an additional prediction head of the neural network backbone~\cite{dockhorn2022genie}.
\emph{Progressive distillation}~\cite{salimans2022progressive} is perhaps the most relevant work. Specifically, it proposes to progressively distill a pre-trained diffusion model into a fewer-step student model with the same model architecture. 
However, none of these approaches are directly applicable or have been applied to classifier-free guided diffusion models. They are also unable to capture a range of different guidance strengths using one single distilled model.
On the contrary, by incorporating the guidance strength into the model architecture and training the model using a two-stage procedure, our approach is able to match the performance of the teacher model on a wide range of guidance strength using one \emph{single} model. Using our method, one single model can capture the trade-off between sample quality and diversity, enabling the real-world application of classifier-free guided diffusion models, where the guidance strength is often specified by users. 
Moreover, none of the above distillation approaches have been applied to or shown effectiveness for latent-space text-to-image models.
Finally, most fast sampling approaches~\cite{song2020denoising,song2020score,salimans2022progressive} only consider using deterministic sampling schemes to improve the sampling speed. In this work, we further develop an effective stochastic sampling approach to sample from the distilled models.

%% file: 6_conclusion.tex
\section{Conclusion}
In this paper, we propose a distillation approach for guided diffusion models~\cite{ho2022classifier}. Our two-stage approach allows us to significantly speed up popular but relatively inefficient guided diffusion models. 
We show that our approach can reduce the inference cost of classifier-free guided pixel-space and latent-space diffusion models by at least an order of magnitude. Empirically, we show that our approach is able to produce visually appealing results with only 2 steps, achieving a comparable FID score to the teacher with as few as 4 to 8 steps.
We further demonstrate practical applications of our distillation approach to text-guided image-to-image translation and inpainting tasks. 
We hope that by significantly reducing the inference cost of classifier-free guided diffusion models, our method will promote creative applications as well as the wider adoption of image generation systems. In the future work, we aim to further improve the performance in the two and one sampling step regimes.

\section*{Acknowledgements}
We thank the anonymous reviewers for their insightful discussions and feedback.
All experiments on Stable Diffusion are supported by Stability AI.

%% file: appendix.tex
\appendix

\section{Results overview}

\begin{figure*}
     \centering
     \begin{subfigure}[b]{0.9\linewidth}
         \centering
         \includegraphics[width=\textwidth]{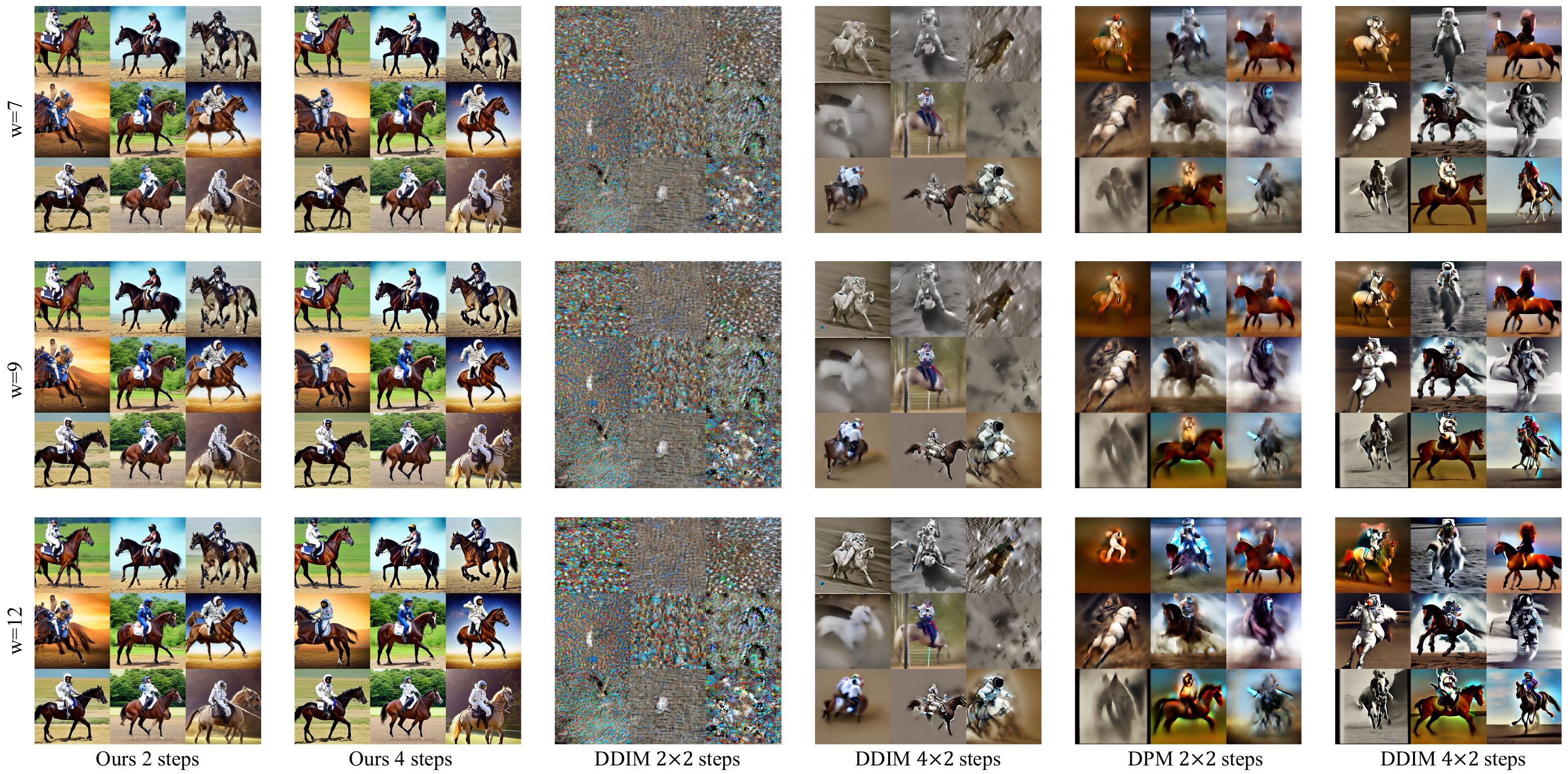}
        \caption*{Prompt: `\textit{`A photograph of an astronaut riding a horse}."}
     \end{subfigure}
     \begin{subfigure}[b]{0.9\linewidth}
         \centering
         \includegraphics[width=\textwidth]{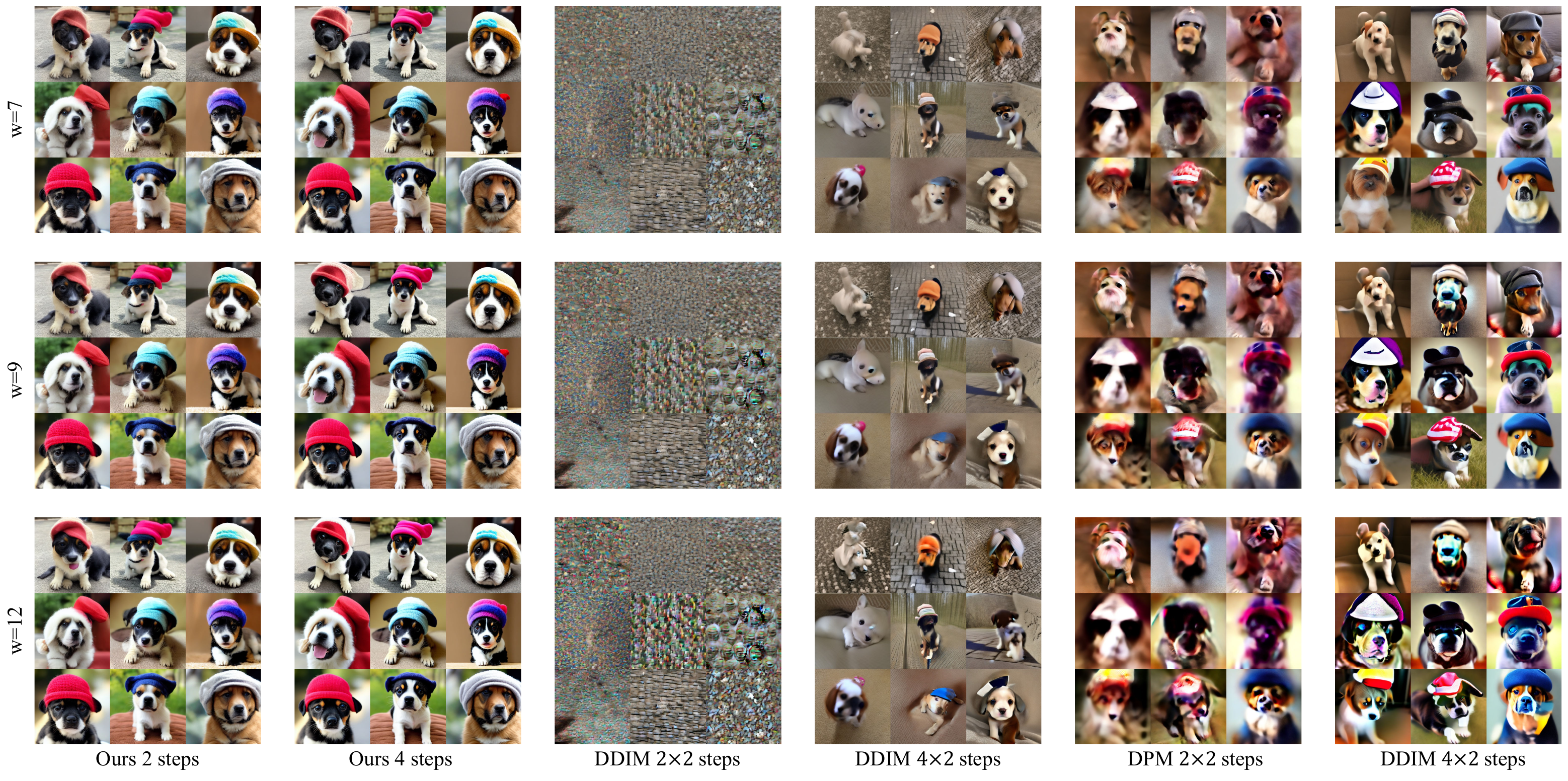}
        \caption*{Prompt: ``\textit{A puppy wearing a hat}
."}
     \end{subfigure}
    \caption{Text-guided image generation on LAION-5B (512$\times$ 512). We compare our distilled model with the original model sampled with DDIM~\cite{song2020denoising} and DPM$++$~\cite{dpmpp}. 
    We observe that our model, when using only two steps, is able to generate more realistic and higher quality images compared to the baselines using more steps. 
    We note that both DDIM and DPM-Solver require evaluating both a conditional and an unconditional diffusion model at each denoising step, while we distill the two models into one model at our stage-one distillation and only require evaluating one model at each denoising step. Depending on the implementation, DDIM and DPM-Solver require either extra $\times 2$ peak memory or $\times 2$ sampling steps compared to our approach.}
    \label{fig:app:extra_txt2img}
\end{figure*}

\begin{figure*}
     \centering
     \begin{subfigure}[b]{0.9\linewidth}
         \centering
         \includegraphics[width=\textwidth]{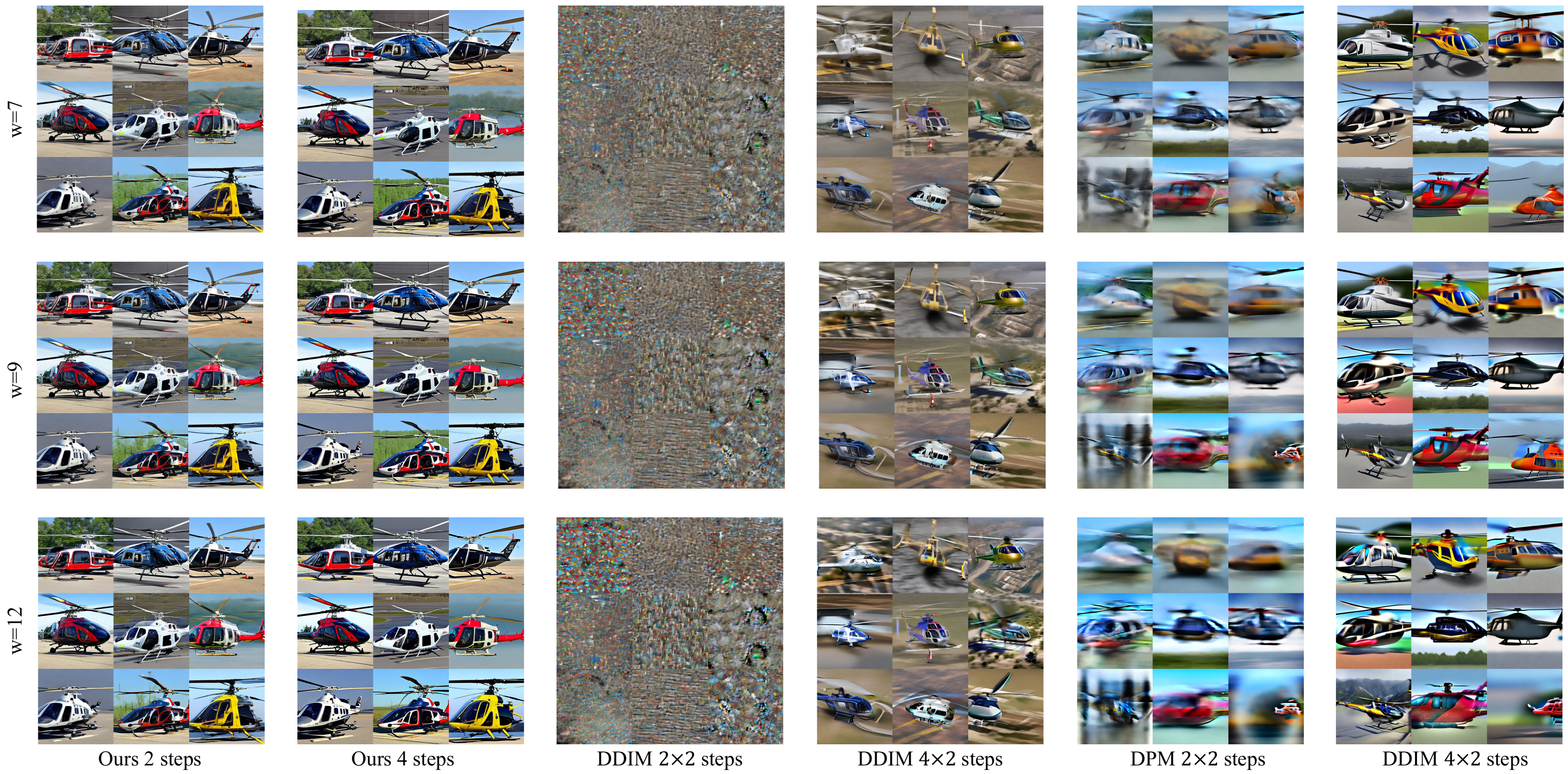}
        \caption*{Prompt: ``\textit{A brand new helicopter}."}
     \end{subfigure}
     \begin{subfigure}[b]{0.9\linewidth}
         \centering
         \includegraphics[width=\textwidth]{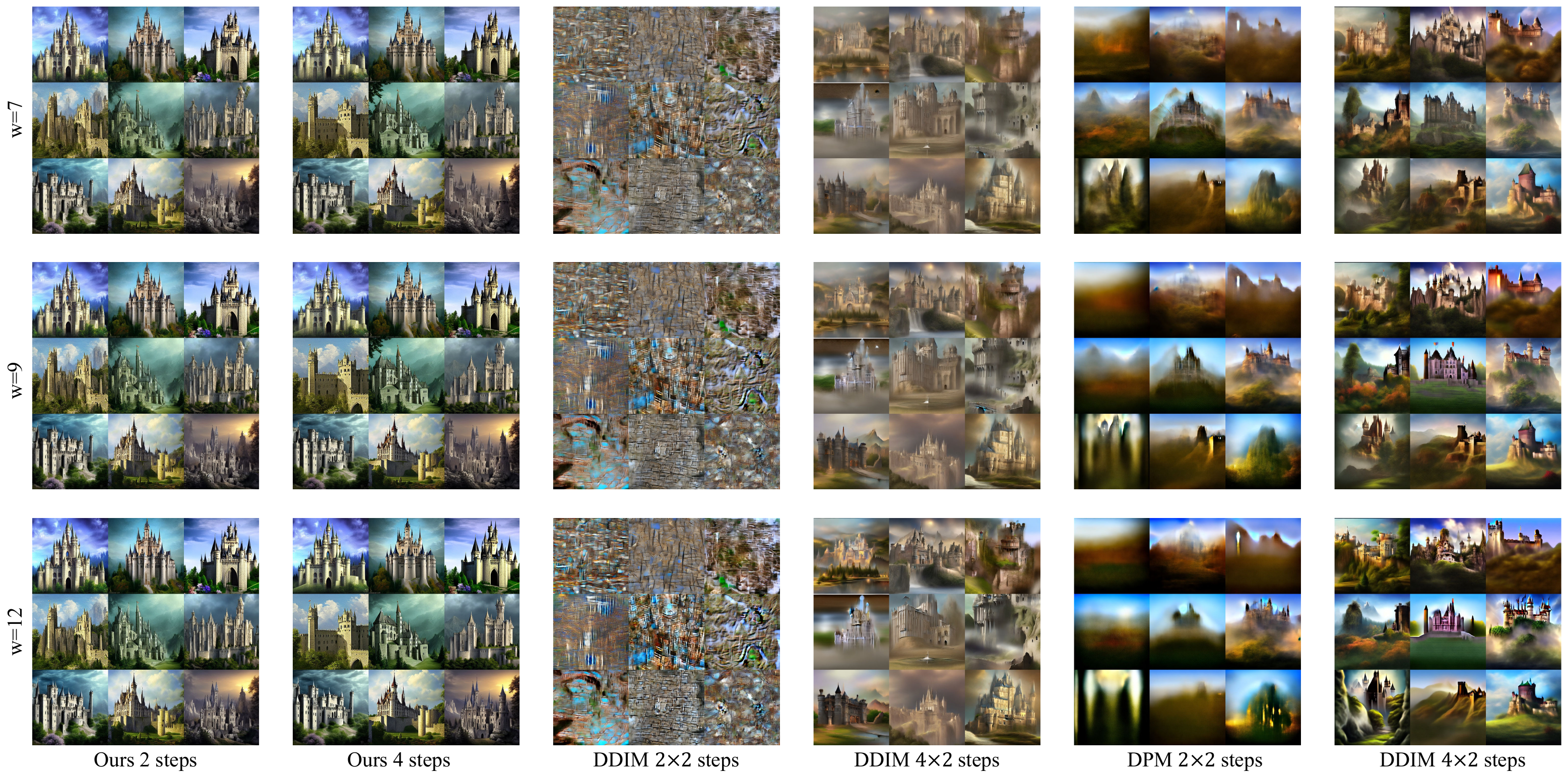}
        \caption*{Prompt: ``\textit{A beautiful castle, matte painting}."}
     \end{subfigure}
    \caption{Text-guided image generation on LAION-5B (512$\times$ 512). We compare our distilled model with the original model sampled with DDIM~\cite{song2020denoising} and DPM$++$\cite{dpmpp}. 
    We observe that our model, when using only two steps, is able to generate more realistic and higher quality images compared to the baselines using more steps. 
    We note that both DDIM and DPM-Solver require evaluating both a conditional and an unconditional diffusion model at each denoising step, while we distill the two models into one model at our stage-one distillation and only require evaluating one model at each denoising step. Depending on the implementation, DDIM and DPM-Solver require either extra $\times 2$ peak memory or $\times 2$ sampling steps compared to our approach.
    }
    \label{fig:app:extra_txt2img2}
\end{figure*}

In this section, we provide an overview table for the speed-up we achieved for pixel-space and latent-space diffusion models (see \cref{table:speedup}).
We also provide extra samples from the text-guided image generation model as well as comparison with DDIM~\cite{song2020denoising}, DPM~\cite{lu2022dpm} and DPM++~\cite{dpmpp} solvers in \cref{fig:app:extra_txt2img} and \cref{fig:app:extra_txt2img2}. We provide more experimental details on pixel-space distillation in \cref{app:sec:pixel_extra_details} and latent-space distillation in \cref{app:sec:latent_space}.

\section{Pixel-space distillation}
\label{app:sec:pixel_extra_details}

\subsection{Teacher model}
The model architecture we use is a  U-Net model similar to the ones used in \cite{ho2022classifier}. The model is parameterized to predict $\mathbf{v}$ as discussed in \cite{salimans2022progressive}. We use the same training setting as \cite{ho2022classifier}.

\subsection{Stage-one distillation}
\label{app:sec:step_one}
The model architecture we use is a  U-Net model similar to the ones used in \cite{ho2022classifier}.
We use the same number of channels and attention as used in \cite{ho2022classifier} for both ImageNet 64x64 and CIFAR-10.
As mentioned in Section 3, we also make the model take $w$ as input. Specifically, we apply Fourier embedding to $w$ before combining with the model backbone. The way we incorporate $w$ is the same as how time-step is incorporated to the model as used in~\cite{kingma2021variational,salimans2022progressive}.
We parameterize the model to predict $\mathbf{v}$ as discussed in \cite{salimans2022progressive}.
We train the distilled model using \cref{alg:stage1}.
We train the model using SNR loss~\cite{kingma2021variational,salimans2022progressive}. For ImageNet 64x64, we use learning rate $3e-4$, with EMA decay $0.9999$; for CIFAR-10, we use learning rate $1e-3$, with EMA decay $0.9999$.
We initialize the student model with parameters from the teacher model except for the parameters related to $w$-embedding.

\begin{table*}[!ht]
\centering
\resizebox{\linewidth}{!}{
\begin{tabular}{ccccccc}
\Xhline{1\arrayrulewidth}
Space &Task &Dataset &Metric &Student diffusion step &Comparable teacher diffusion step & Speed-up\\
\Xhline{2\arrayrulewidth}
\multirow{4}{*}{Pixel-space} & class-conditional generation &CIFAR-10 &FID &4 &1024 DDIM$\times 2$ &$\times$512\\
&class-conditional generation &CIFAR-10 &IS &4 &1024 DDIM$\times 2$ &$\times$512\\
&class-conditional generation &ImageNet 64$\times$64 &FID &8 &1024 DDIM$\times 2$ &$\times$256 \\
&class-conditional generation &ImageNet 64$\times$64 &IS &8 &1024 DDIM$\times 2$ &$\times$256\\
\Xhline{1\arrayrulewidth}
\multirow{4}{*}{Latent-space} & class-conditional generation & ImageNet 256$\times$256 & FID & 2 & 16 DDIM $\times$2 & $\times$16\\
& class-conditional generation & ImageNet 256$\times$256 & Recall & 2 & 16 DDIM $\times$2 &  $\times$16 \\

& text-guided generation & LAION-5B 512$\times$ 512 & FID & 2 & 16 DDIM / 8 DPM$++$  $\times$2 & $\times$16 / $\times$8 \\
& text-guided generation & LAION-5B 512$\times$ 512 & CLIP & 4 & 8 DDIM / 4 DPM$++$  $\times$2 & $\times$8 / $\times$4 \\

\Xhline{2\arrayrulewidth}
\end{tabular}
}
\caption{
Speed-up overview for pixel-space diffusion and latent-space diffusion. We note that the original model (without distillation) requires evaluating both the unconditional and the conditional diffusion model at each denoising step. Our model, on the other hand, only requires evaluating one diffusion model at each denoising step. This is because in our stage-one distillation, we distill the output of the unconditional and conditional models into the output of one model. Thus our method further decreases either the peak memory or sampling time by a half compared to the original model. 
}
\label{table:speedup}
\end{table*}

\begin{algorithm}
\centering
\caption{Stage-one distillation}\label{alg:stage1}
\begin{algorithmic}
\Require Trained classifier-free guidance teacher model $[\hat{\rvx}_{c, \bm\theta}, \hat{\rvx}_{\bm\theta}]$
\Require Data set $\mathcal{D}$
\Require Loss weight function $\omega()$
\While{not converged}
\State $\rvx \sim \mathcal{D}$ \Comment{Sample data}
\State $t \sim U[0, 1]$ \Comment{Sample time}
\State $w \sim U[w_{\text{min}}, w_{\text{max}}]$ \Comment{Sample guidance}
\State $\rvepsilon \sim N(0, I)$ \Comment{Sample noise}
\State $\rvz_t = \alpha_t \rvx + \sigma_t \rvepsilon$ \Comment{Add noise to data}
\State $\lambda_t = \log[\alpha^{2}_t / \sigma^{2}_t]$ \Comment{log-SNR}
\State $\hat{\rvx}_{\bm\theta}^{w}(\rvz_t)=(1+w)\hat{\rvx}_{c, \bm\theta}(\rvz_t)-w\hat{\rvx}_{\bm\theta}(\rvz_t)$ \Comment{Compute target}
\State $L_{{\bm\eta}_1} = \omega(\lambda_t) \lVert   \hat\rvx^{w}_{\bm\theta}(\rvz_t)-\hat{\rvx}_{{\bm\eta}_1}(\rvz_t, w)\rVert_{2}^{2}$ \Comment{Loss}
\State ${\bm\eta}_1 \leftarrow {\bm\eta_1} - \gamma\nabla_{{\bm\eta}_1}L_{{\bm\eta}_1}$ \Comment{Optimization}
\EndWhile
\end{algorithmic}
\end{algorithm}
\subsection{Stage-two distillation for deterministic sampler}
\label{app:sec:step_two}
We use the same model architectures as the ones used in Stage-one (see \cref{app:sec:step_one}).
We train the distilled model using \cref{alg:app:stage2}.
We first use the student model from Stage-one as the teacher model.
We start from $1024$ DDIM sampling steps and progressively distill the student model from Stage-one to a one step model. 
We train the student model for 50,000 parameter updates, except for sampling step equals to one or two where we train the model for 100,000 parameter updates, before the number of sampling step is halved and the student model becomes the new teacher model.
At each sampling step, we initialize the student model with the parameters from the teacher model. 
We train the model using SNR truncation loss~\cite{kingma2021variational,salimans2022progressive}. For each step, we linearly anneal the learning rate from $1e-4$ to $0$ during each parameter update. We do not use EMA decay for training. Our training setting follows the setting in \cite{salimans2022progressive} closely.

\begin{algorithm}
\centering
\caption{Stage-two distillation for deterministic sampler}\label{alg:app:stage2}
\begin{algorithmic}
\Require {\setlength{\fboxsep}{0pt}{Trained teacher model $\hat\rvx_{{\bm\eta}}(\rvz_t, w)$}}
\Require Data set $\mathcal{D}$
\Require Loss weight function $\omega()$
\Require {\setlength{\fboxsep}{0pt}{Student sampling steps $N$}}
\For{{\setlength{\fboxsep}{0pt}{$K$ iterations}}}
\State {\setlength{\fboxsep}{0pt}{${\bm\eta_2} \leftarrow {\bm\eta}$}} \Comment{Init student from teacher}
\While{not converged}
\State $\rvx \sim \mathcal{D}$
\State {\setlength{\fboxsep}{0pt}{$t=i/N, ~~  i \sim Cat[1, 2, \ldots, N]$}}
\State $w \sim U[w_{\text{min}}, w_{\text{max}}]$ \Comment{Sample guidance}
\State $\rvepsilon \sim N(0, I)$
\State $\rvz_t = \alpha_t \rvx + \sigma_t \rvepsilon$
\State {\setlength{\fboxsep}{0pt}{\texttt{\# 2 steps of DDIM with teacher}}}
\State {\setlength{\fboxsep}{0pt}{$t' = t-0.5/N$,~~~ $t'' = t-1/N$}}
\State {\setlength{\fboxsep}{0pt}{$\rvz^w_{t'}=\alpha_{t'}\hat\rvx_{{\bm\eta}}(\rvz_t, w) + \frac{\sigma_{t'}}{\sigma_{t}}(\rvz_t - \alpha_t\hat\rvx_{{\bm\eta}}(\rvz_t, w))$
}}
\State {\setlength{\fboxsep}{0pt}{$\rvz^w_{t''}=\alpha_{t''}\hat\rvx_{{\bm\eta}}(\rvz^w_{t'}, w) + \frac{\sigma_{t''}}{\sigma_{t'}}(\rvz^w_{t'} - \alpha_{t'}\hat\rvx_{{\bm\eta}}(\rvz^w_{t'}, w))$
}}
\State {\setlength{\fboxsep}{0pt}{$\tilde\rvx^w = \frac{\rvz^w_{t''}-(\sigma_{t''}/\sigma_{t})\rvz_t}{\alpha_{t''}-(\sigma_{t''}/\sigma_{t})\alpha_t}$}} \Comment{Teacher $\hat\rvx$ target}
\State $\lambda_t = \log[\alpha^{2}_t / \sigma^{2}_t]$
\State $L_{{\bm\eta}_2} = \omega(\lambda_t) \lVert \tilde\rvx^w -  \hat\rvx_{{\bm\eta}_2}(\rvz_t, w)\rVert_{2}^{2}$
\State ${\bm\eta}_2 \leftarrow {\bm\eta_2} - \gamma\nabla_{{\bm\eta}_2}L_{{\bm\eta}_2}$
\EndWhile
\State {\setlength{\fboxsep}{0pt}{${\bm\eta} \leftarrow {\bm\eta}_2$}} \Comment{Student becomes next teacher}
\State {\setlength{\fboxsep}{0pt}{$N \leftarrow N/2$}} \Comment{Halve number of sampling steps}
\EndFor
\end{algorithmic}
\end{algorithm}

\subsection{Stage-two distillation for stochastic sampling}
\begin{figure*}[!ht]
    \centering
    \includegraphics[width=0.8\linewidth]{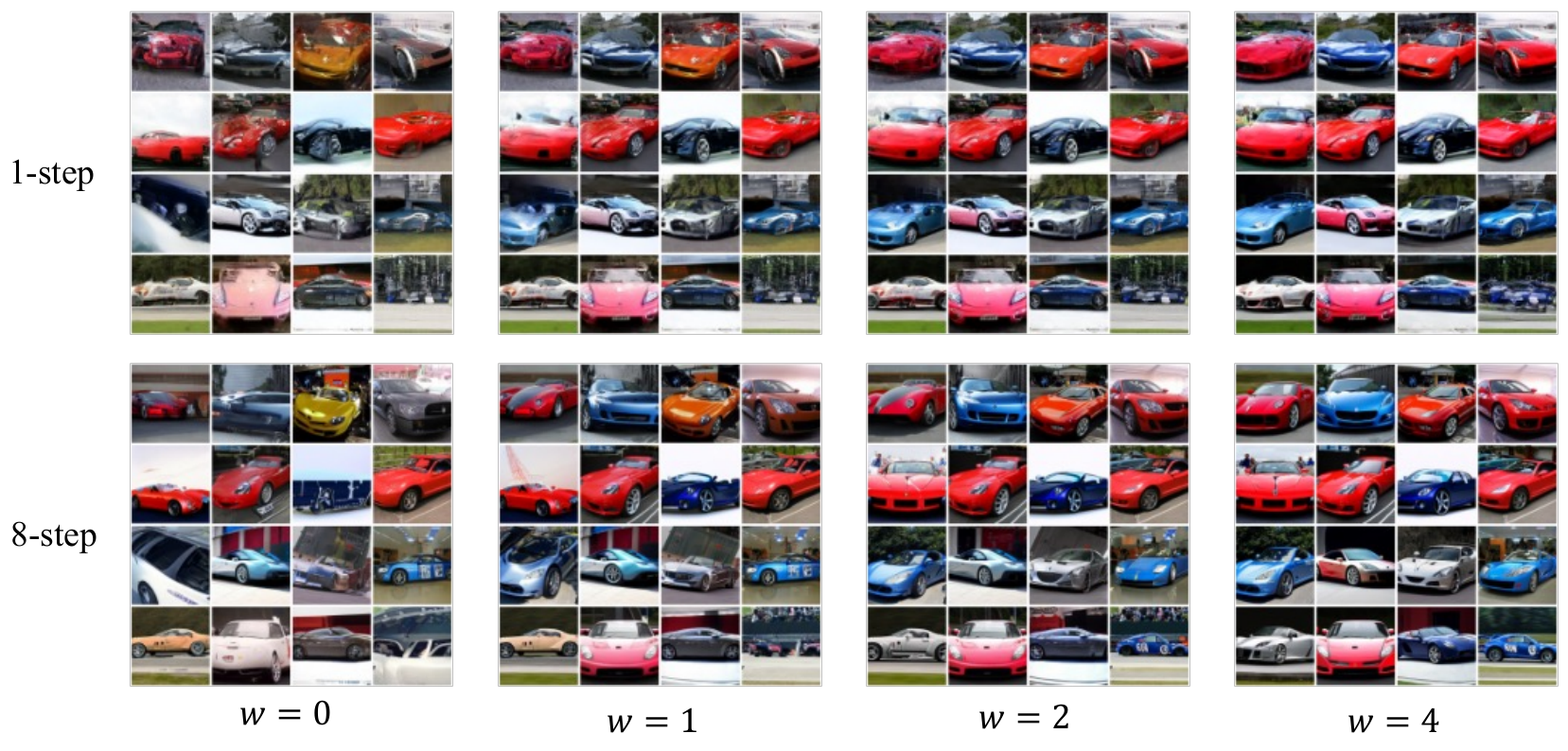}
    \caption{Class-conditional samples from our two-step (stochastic) approach on ImageNet 64x64. By varying the guidance weight $w$, our distilled model is able to trade-off between sample diversity and quality, while achieving visually pleasant results using as few as \emph{one} sampling step.}
    \label{fig:ours_s_sweep}
\end{figure*}
We train the distilled model using \cref{alg:stage2_stochastic}. We use the same model architecture and training setting as Stage-two distillation described in \cref{app:sec:step_two} for both ImageNet 64x64 and CIFAR-10: The main difference here is that our distillation target corresponds to taking a sampling step that is twice as large as for the deterministic sampler. We provide visualization for samples with varying guidance strengths $w$ in \cref{fig:ours_s_sweep}.

\begin{algorithm}
\centering
\caption{Stage-two distillation for stochastic sampler
}\label{alg:stage2_stochastic}
\begin{algorithmic}
\Require {\setlength{\fboxsep}{0pt}{Trained teacher model $\hat\rvx_{{\bm\eta}}(\rvz_t, w)$}}
\Require Data set $\mathcal{D}$
\Require Loss weight function $\omega()$
\Require {\setlength{\fboxsep}{0pt}{Student sampling steps $N$}}
\For{{\setlength{\fboxsep}{0pt}{$K$ iterations}}}
\State {\setlength{\fboxsep}{0pt}{${\bm\eta_2} \leftarrow {\bm\eta}$}} \Comment{Init student from teacher}
\While{not converged}
\State $\rvx \sim \mathcal{D}$
\State {\setlength{\fboxsep}{0pt}{$t=i/N, ~~  i \sim Cat[1, 2, \ldots, N]$}}
\State $w \sim U[w_{\text{min}}, w_{\text{max}}]$ \Comment{Sample guidance}
\State $\rvepsilon \sim N(0, I)$
\State $\rvz_t = \alpha_t \rvx + \sigma_t \rvepsilon$
\If{$t>1/N$}
\State {\setlength{\fboxsep}{0pt}{\texttt{\# 2 steps of DDIM with teacher}}}
\State {\setlength{\fboxsep}{0pt}{$t' = t-1/N$,~~~ $t'' = t-2/N$}}
\State {\setlength{\fboxsep}{0pt}{$\rvz^w_{t'}=\alpha_{t'}\hat\rvx_{{\bm\eta}}(\rvz_t, w) + \frac{\sigma_{t'}}{\sigma_{t}}(\rvz_t - \alpha_t\hat\rvx_{{\bm\eta}}(\rvz_t, w))$
}}
\State {\setlength{\fboxsep}{0pt}{$\rvz^w_{t''}=\alpha_{t''}\hat\rvx_{{\bm\eta}}(\rvz^w_{t'}, w) + \frac{\sigma_{t''}}{\sigma_{t'}}(\rvz^w_{t'} - \alpha_{t'}\hat\rvx_{{\bm\eta}}(\rvz^w_{t'}, w))$
}}
\State {\setlength{\fboxsep}{0pt}{$\tilde\rvx^w = \frac{\rvz^w_{t''}-(\sigma_{t''}/\sigma_{t})\rvz_t}{\alpha_{t''}-(\sigma_{t''}/\sigma_{t})\alpha_t}$}} \Comment{Teacher $\hat\rvx$ target}
\Else\Comment{Edge case}
\State {\setlength{\fboxsep}{0pt}{\texttt{\# 1 step of DDIM with teacher}}}
\State {\setlength{\fboxsep}{0pt}{$t' = t-1/N$}}
\State {\setlength{\fboxsep}{0pt}{$\rvz^w_{t'}=\alpha_{t'}\hat\rvx_{{\bm\eta}}(\rvz_t, w) + \frac{\sigma_{t'}}{\sigma_{t}}(\rvz_t - \alpha_t\hat\rvx_{{\bm\eta}}(\rvz_t, w))$
}}
\State {\setlength{\fboxsep}{0pt}{$\tilde\rvx^w = \frac{\rvz^w_{t'}-(\sigma_{t'}/\sigma_{t})\rvz_t}{\alpha_{t'}-(\sigma_{t'}/\sigma_{t})\alpha_t}$}} \Comment{Teacher $\hat\rvx$ target}
\EndIf
\State $\lambda_t = \log[\alpha^{2}_t / \sigma^{2}_t]$
\State $L_{{\bm\eta}_2} = \omega(\lambda_t) \lVert \tilde\rvx^w -  \hat\rvx_{{\bm\eta}_2}(\rvz_t, w)\rVert_{2}^{2}$
\State ${\bm\eta}_2 \leftarrow {\bm\eta_2} - \gamma\nabla_{{\bm\eta}_2}L_{{\bm\eta}_2}$
\EndWhile
\State {\setlength{\fboxsep}{0pt}{${\bm\eta} \leftarrow {\bm\eta}_2$}} \Comment{Student becomes next teacher}
\State {\setlength{\fboxsep}{0pt}{$N \leftarrow N/2$}} \Comment{Halve number of sampling steps}
\EndFor
\end{algorithmic}
\end{algorithm}

\input{table}

\subsection{Baseline samples}
\begin{figure*}
    \centering
    \includegraphics[width=0.8\linewidth]{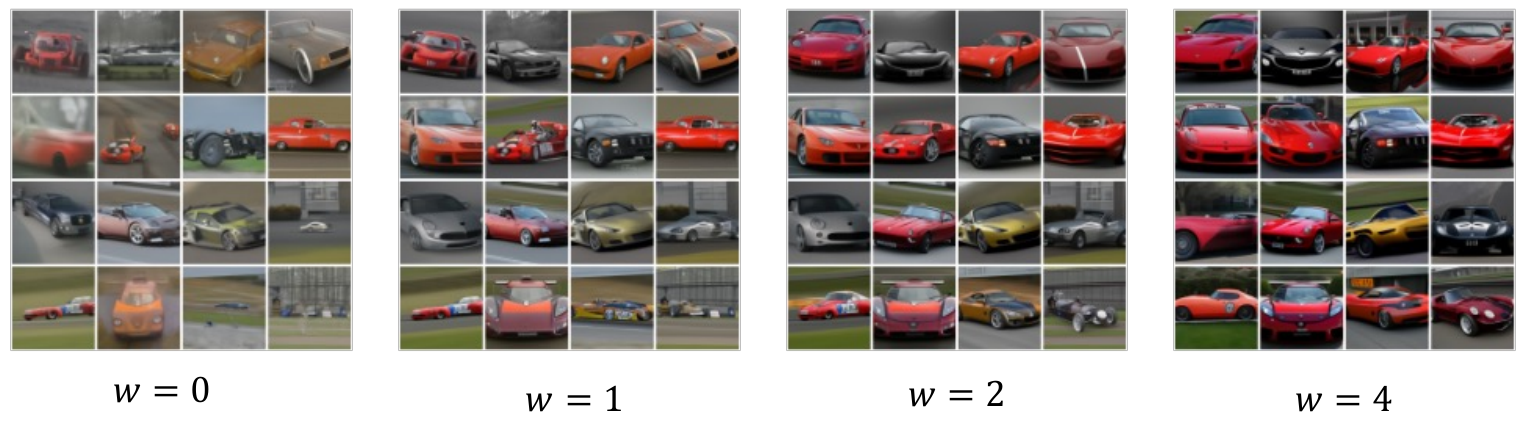}
    \caption{ImageNet 64x64 class-conditional generation using DDIM (baseline) 8$\times$2 sampling steps. We observe clear artifacts when $w=0$.}
    \label{fig:ddim_8step_sweep}
\end{figure*}

\begin{figure*}
    \centering
    \includegraphics[width=0.8\linewidth]{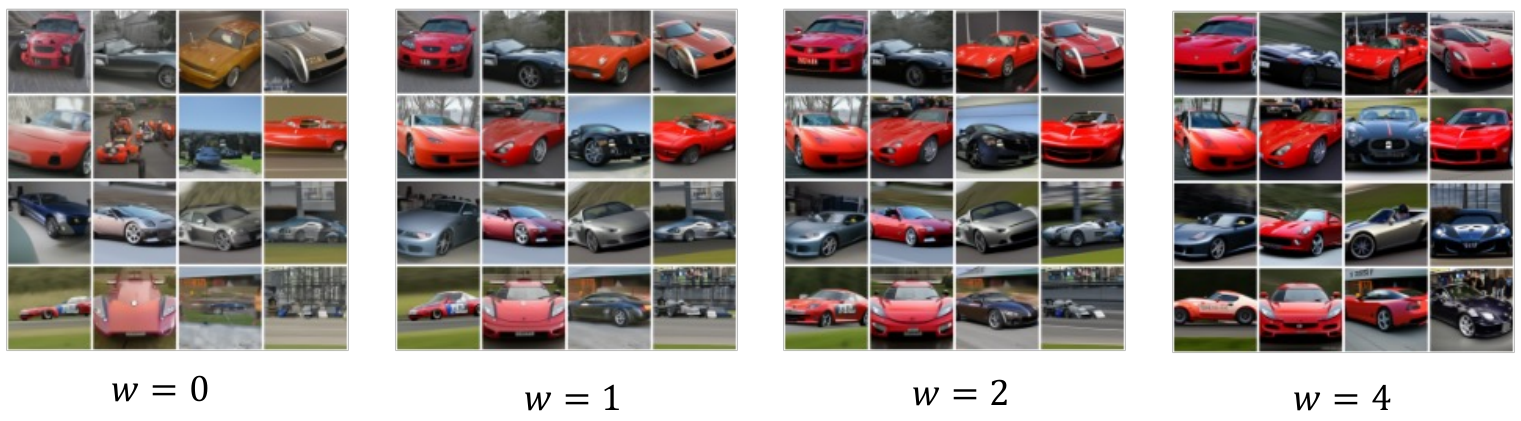}
    \caption{ImageNet 64x64 class-conditional generation using DDIM (baseline) 16$\times$2 sampling steps.}
    \label{fig:ddim_16step_sweep}
\end{figure*}

We provide extra samples for the DDIM baseline in \cref{fig:ddim_8step_sweep} and \cref{fig:ddim_16step_sweep}.

\begin{figure*}
    \centering
    \includegraphics[width=0.8\linewidth]{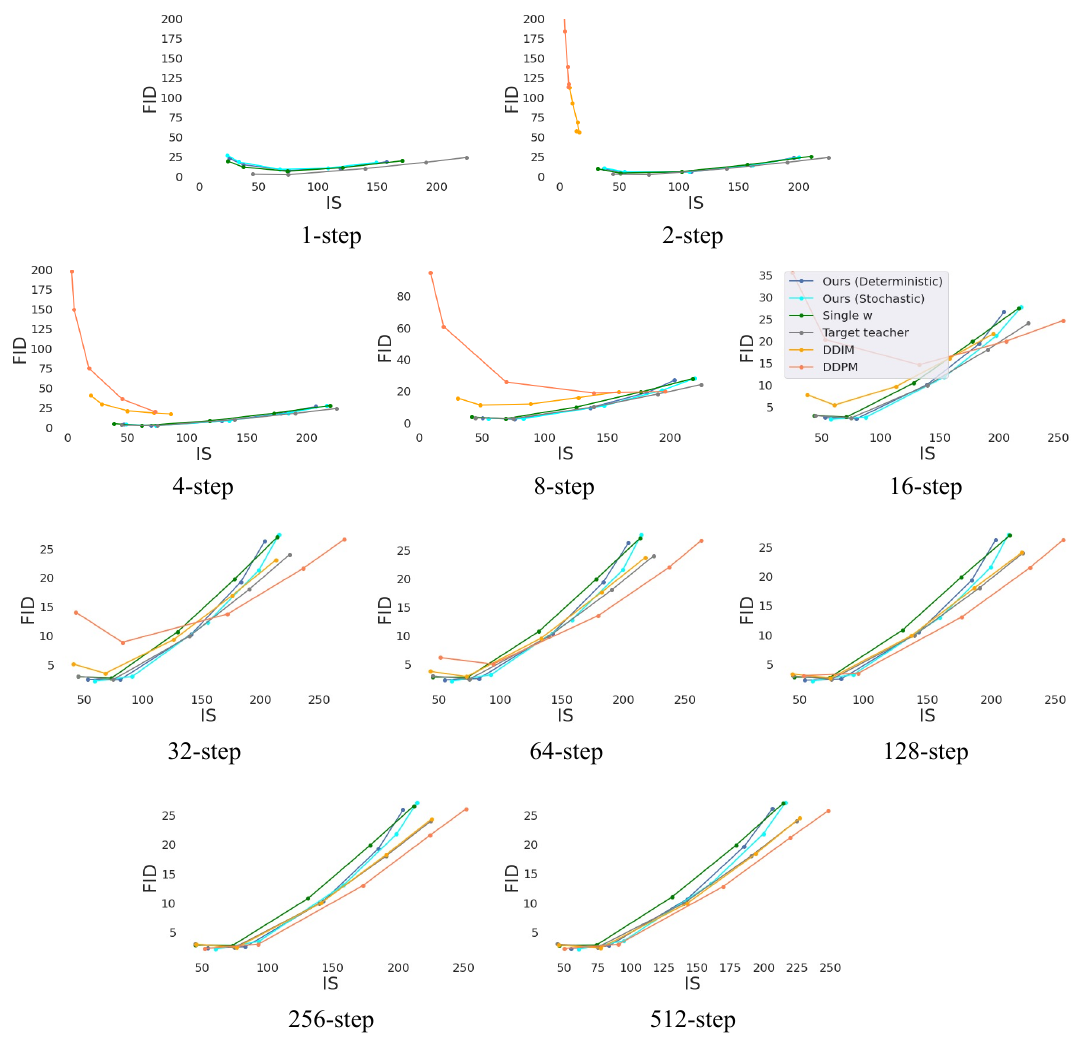}
    \caption{FID and IS score trade-off on ImageNet 64x64. We plot the results using guidance strength  $w=\{0, 0.3, 1, 2, 4\}$. For the \texttt{1-step} plot, the curves of DDIM and DDPM are too far away to be visualized.}
    \label{fig:i64_tradeoff}
\end{figure*}

\begin{figure*}
    \centering
    \includegraphics[width=0.8\linewidth]{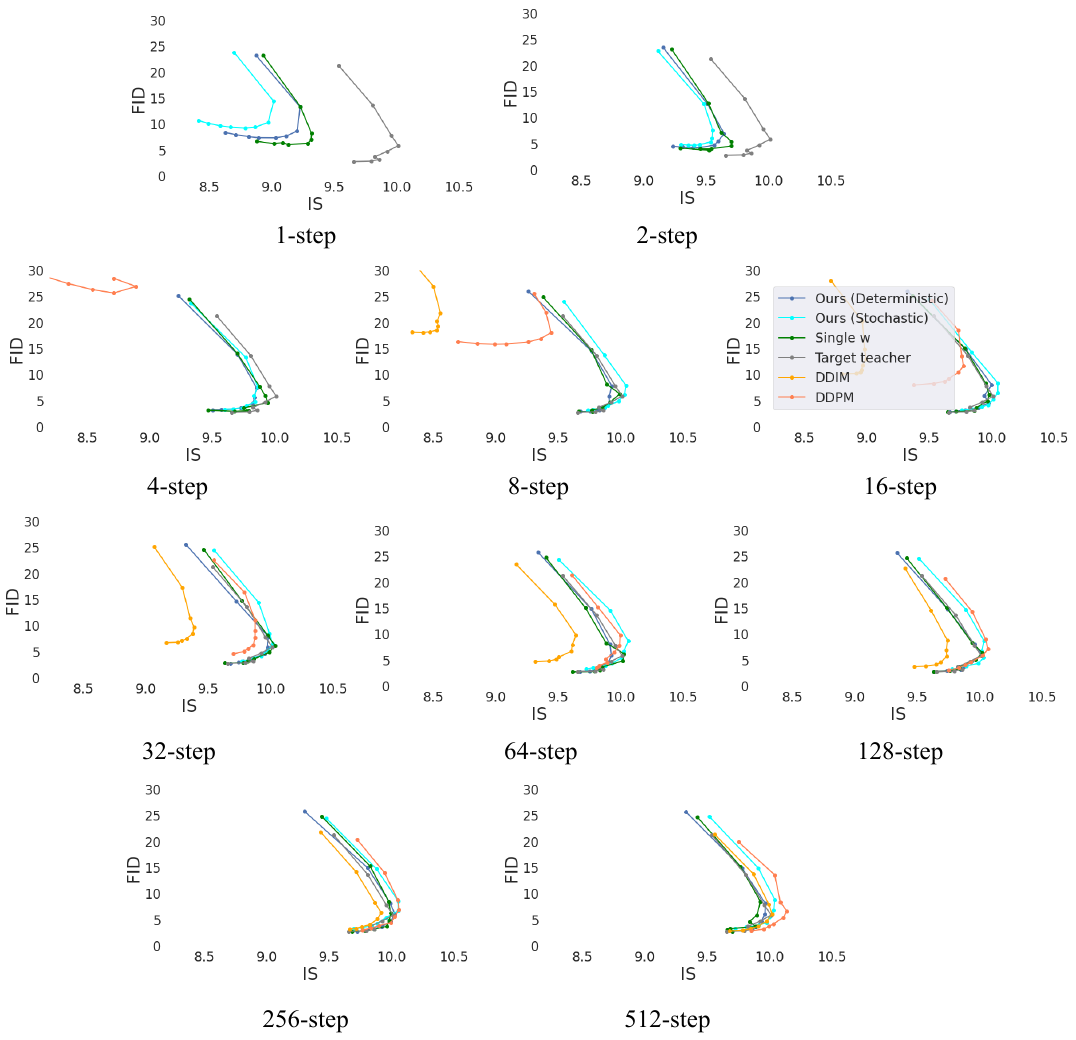}
    \caption{
    FID and IS score trade-off on CIFAR-10. We plot the results using guidance strength $w=\{0, 0.1, 0.2, 0.3, 0.5, 0.7, 1, 2, 4\}$. For the \texttt{1-step} and \texttt{2-step} plots, the curves of DDIM and DDPM are too far away to be visualized. For the \texttt{4-step} plot, the curve of DDIM is too far away to be visualized.}
    \label{fig:cifar_tradeoff}
\end{figure*}

\subsection{Extra distillation results}
\label{app:sec:extra_result}
\begin{figure*} %
     \centering
     \begin{subfigure}[b]{\textwidth}
         \centering
         \includegraphics[width=\textwidth]{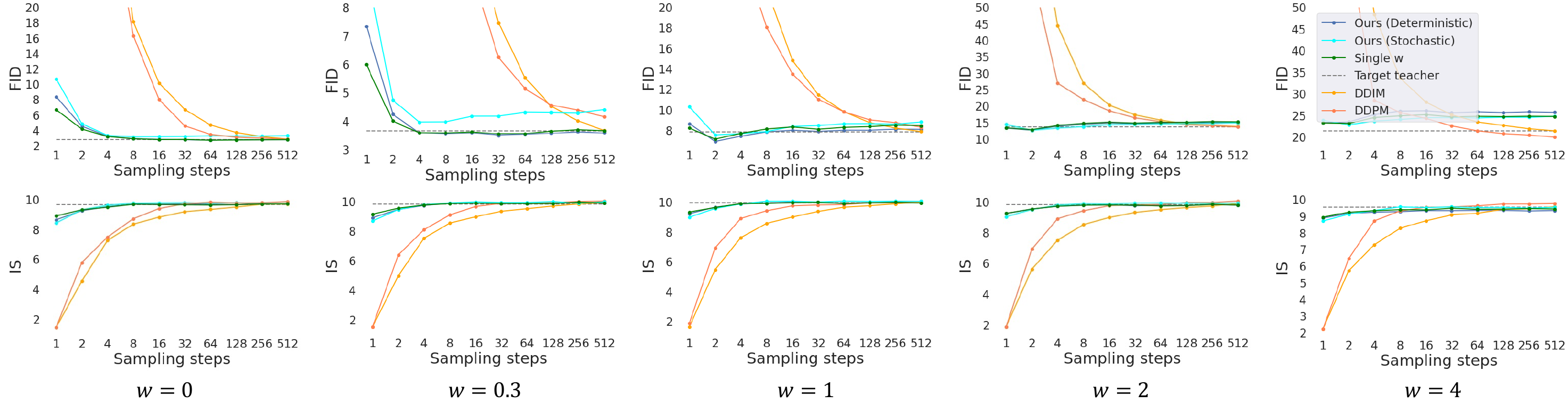}
         \caption{CIFAR-10}
         \label{fig:app:cifar_curve}
     \end{subfigure}
     \begin{subfigure}[b]{\textwidth}
         \centering
         \includegraphics[width=\textwidth]{images/i64_fid_is.pdf}
         \caption{ImageNet 64x64}
         \label{fig:i64_curve_complete}
     \end{subfigure}
    \caption{CIFAR-10 and ImageNet sample quality evaluated by FID and IS scores for pixel-space diffusion models. We follow the setting of \cite{salimans2022progressive} for our evaluation. 
    We note that, the DDPM and DDIM baseline require evaluating both an unconditional and a conditional diffusion model at each denoising step for classifier-free guidance, giving rise to either an extra $\times$2 overhead for peak memory or an extra $\times$2 sampling steps than the ``Sampling steps" value shown in the plot.
    Our distilled model significantly outperform the DDPM and DDIM baselines, and is able to match the performance of the teacher using as few as 4 to 16 steps. By varying $w$, a \emph{single} distilled model is able to capture the trade-off between sample diversity and quality.}
    \label{fig:app:curve}
\end{figure*}

We provide the  FID and IS results for our method and the baselines on ImageNet 64x64 and CIFAR-10 in \cref{fig:i64_curve_complete}, \cref{fig:app:cifar_curve} and  \cref{table:i64_cifar10}.
We also visualize the FID and IS trade-off curves for both datasets in \cref{fig:i64_tradeoff} and \cref{fig:cifar_tradeoff}, where we select guidance strength $w=\{0, 0.3, 1, 2, 4\}$ for ImageNet 64x64 and $w=\{0, 0.1, 0.2, 0.3, 0.5, 0.7, 1, 2, 4\}$ for CIFAR-10.

\subsection{Style transfer}
We focus on ImageNet 64x64 for this experiment.
As discussed in \cite{su2022dual}, one can perform style-transfer between domain A and B by encoding (performing reverse DDIM) an image using a diffusion model train on domain A and then decoding using DDIM with a diffusion model trained on domain B.
We train the model using \cref{alg:encoder}.
We use the same $w$-conditioned model architecture and training setting as discussed in \cref{app:sec:step_two}. 

\begin{figure*}%
    \centering
    \includegraphics[width=0.85\linewidth]{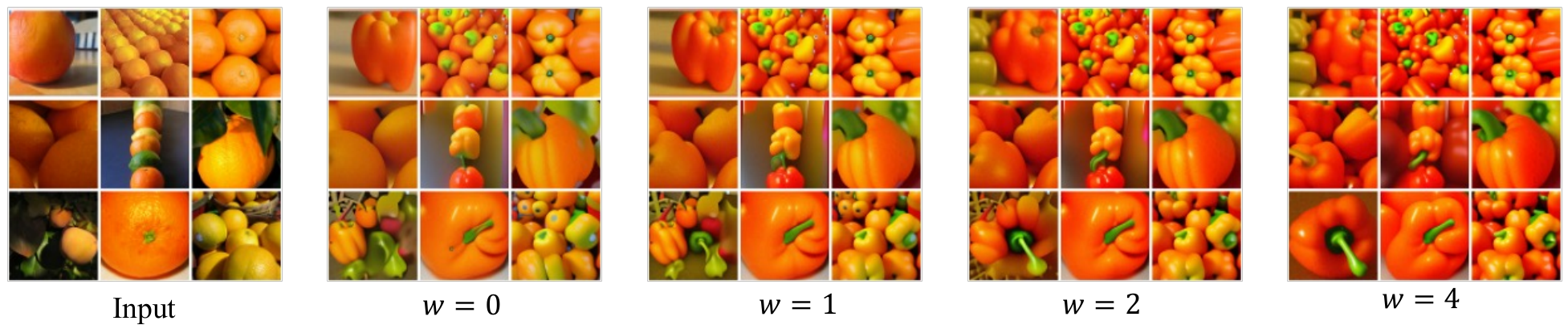}
    \caption{
    Style transfer on ImageNet 64x64 for pixel-space models (\texttt{orange} to \texttt{bell pepper}).
    We use a distilled 
    16-step encoder and decoder. We fix the encoder guidance strength to be $0$ and vary the decoder guidance strength from $0$ to $4$. As we increase $w$, we notice a trade-off between sample diversity and sharpness. 
    }
    \label{fig:style_2}
\end{figure*}

\begin{figure*}
    \centering
    \includegraphics[width=0.85\linewidth]{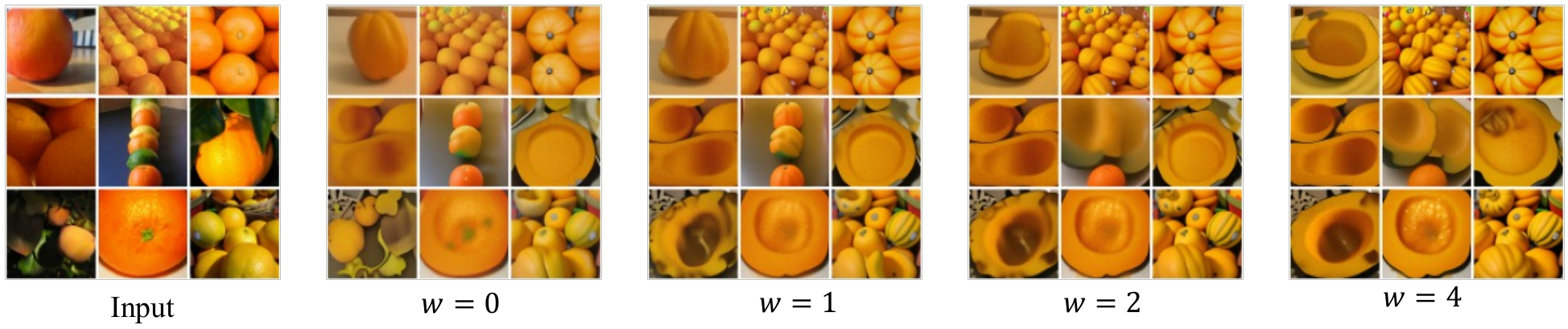}
    \caption{Style transfer on ImageNet 64x64 (\texttt{orange} to \texttt{acorn squash}). We use a distilled 16-step encoder and decoder. We fix the encoder guidance strength to be $0$ and vary the decoder guidance strength from $0$ to $4$. As we increase the guidance strength $w$, we notice a trade-off between sample diversity and sharpness.}
    \label{fig:style_1}
\end{figure*}

\begin{algorithm}
\centering
\caption{Encoder distillation
}\label{alg:encoder}
\begin{algorithmic}
\Require {\setlength{\fboxsep}{0pt}{Trained teacher model $\hat\rvx_{{\bm\eta}}(\rvz_t, w)$}}
\Require Data set $\mathcal{D}$
\Require Loss weight function $\omega()$
\Require {\setlength{\fboxsep}{0pt}{Student sampling steps $N$}}
\For{{\setlength{\fboxsep}{0pt}{$K$ iterations}}}
\State {\setlength{\fboxsep}{0pt}{${\bm\eta_2} \leftarrow {\bm\eta}$}} \Comment{Init student from teacher}
\While{not converged}
\State $\rvx \sim \mathcal{D}$
\State {\setlength{\fboxsep}{0pt}{$t=i/N, ~~  i \sim Cat[0, 1, \ldots, N-1]$}}
\State $w \sim U[w_{\text{min}}, w_{\text{max}}]$ \Comment{Sample guidance}
\State $\rvepsilon \sim N(0, I)$
\State $\rvz_t = \alpha_t \rvx + \sigma_t \rvepsilon$
\State {\setlength{\fboxsep}{0pt}{\texttt{\# 2 steps of reversed DDIM with teacher}}}
\State {\setlength{\fboxsep}{0pt}{$t' = t+0.5/N$,~~~ $t'' = t+1/N$}}
\State {\setlength{\fboxsep}{0pt}{$\rvz^w_{t'}=\alpha_{t'}\hat\rvx_{{\bm\eta}}(\rvz_t, w) + \frac{\sigma_{t'}}{\sigma_{t}}(\rvz_t - \alpha_t\hat\rvx_{{\bm\eta}}(\rvz_t, w))$
}}
\State {\setlength{\fboxsep}{0pt}{$\rvz^w_{t''}=\alpha_{t''}\hat\rvx_{{\bm\eta}}(\rvz^w_{t'}, w) + \frac{\sigma_{t''}}{\sigma_{t'}}(\rvz^w_{t'} - \alpha_{t'}\hat\rvx_{{\bm\eta}}(\rvz^w_{t'}, w))$
}}
\State {\setlength{\fboxsep}{0pt}{$\tilde\rvx^w = \frac{\rvz^w_{t''}-(\sigma_{t''}/\sigma_{t})\rvz_t}{\alpha_{t''}-(\sigma_{t''}/\sigma_{t})\alpha_t}$}} \Comment{Teacher $\hat\rvx$ target}
\State $\lambda_t = \log[\alpha^{2}_t / \sigma^{2}_t]$
\State $L_{{\bm\eta}_2} = \omega(\lambda_t) \lVert \tilde\rvx^w -  \hat\rvx_{{\bm\eta}_2}(\rvz_t, w)\rVert_{2}^{2}$
\State ${\bm\eta}_2 \leftarrow {\bm\eta_2} - \gamma\nabla_{{\bm\eta}_2}L_{{\bm\eta}_2}$
\EndWhile
\State {\setlength{\fboxsep}{0pt}{${\bm\eta} \leftarrow {\bm\eta}_2$}} \Comment{Student becomes next teacher}
\State {\setlength{\fboxsep}{0pt}{$N \leftarrow N/2$}} \Comment{Halve number of sampling steps}
\EndFor
\end{algorithmic}
\end{algorithm}

\begin{algorithm}
\centering
\caption{Two-student progressive distillation
}\label{alg:two_student}
\begin{algorithmic}
\Require Trained classifier-free guidance teacher model $[\hat{\rvx}_{c, \bm\theta}, \hat{\rvx}_{\bm\theta}]$
\Require Data set $\mathcal{D}$
\Require Loss weight function $\omega()$
\Require {\setlength{\fboxsep}{0pt}{Student sampling steps $N$}}
\For{{\setlength{\fboxsep}{0pt}{$K$ iterations}}}
\State {\setlength{\fboxsep}{0pt}{${\bm\eta} \leftarrow {\bm\theta}$}} \Comment{Init student from teacher}
\While{not converged}
\State $\rvx \sim \mathcal{D}$
\State {\setlength{\fboxsep}{0pt}{$t=i/N, ~~  i \sim Cat[1, 2, \ldots, N]$}}
\State $w \sim U[w_{\text{min}}, w_{\text{max}}]$ \Comment{Sample guidance}
\State $\rvepsilon \sim N(0, I)$
\State $\rvz_t = \alpha_t \rvx + \sigma_t \rvepsilon$
\State $\hat{\rvx}_{\bm\theta}^{w}(\rvz_t)=(1+w)\hat{\rvx}_{c, \bm\theta}(\rvz_t)-w\hat{\rvx}_{\bm\theta}(\rvz_t)$ \Comment{Compute target}
\State {\setlength{\fboxsep}{0pt}{\texttt{\# 2 steps of DDIM with teacher}}}
\State {\setlength{\fboxsep}{0pt}{$t' = t-0.5/N$,~~~ $t'' = t-1/N$}}
\State {\setlength{\fboxsep}{0pt}{$\rvz^w_{t'}=\alpha_{t'}\hat\rvx^w_{{\bm\theta}}(\rvz_t) + \frac{\sigma_{t'}}{\sigma_{t}}(\rvz_t - \alpha_t\hat\rvx^w_{{\bm\theta}}(\rvz_t))$
}}
\State {\setlength{\fboxsep}{0pt}{$\rvz^w_{c, t''}=\alpha_{t''}\hat\rvx_{c, {\bm\theta}}(\rvz^w_{t'}) + \frac{\sigma_{t''}}{\sigma_{t'}}(\rvz^w_{t'} - \alpha_{t'}\hat\rvx_{c,{\bm\theta}}(\rvz^w_{t'}))$
}}
\State {\setlength{\fboxsep}{0pt}{$\tilde\rvx^w_c = \frac{\rvz^w_{c,t''}-(\sigma_{t''}/\sigma_{t})\rvz_t}{\alpha_{t''}-(\sigma_{t''}/\sigma_{t})\alpha_t}$}} \Comment{Conditional teacher $\hat\rvx$ target}
\State {\setlength{\fboxsep}{0pt}{$\rvz^w_{t''}=\alpha_{t''}\hat\rvx_{{\bm\theta}}(\rvz^w_{t'}) + \frac{\sigma_{t''}}{\sigma_{t'}}(\rvz^w_{t'} - \alpha_{t'}\hat\rvx_{{\bm\theta}}(\rvz^w_{t'}))$
}}
\State {\setlength{\fboxsep}{0pt}{$\tilde\rvx^w = \frac{\rvz^w_{t''}-(\sigma_{t''}/\sigma_{t})\rvz_t}{\alpha_{t''}-(\sigma_{t''}/\sigma_{t})\alpha_t}$}} \Comment{Unconditional teacher $\hat\rvx$ target}
\State $\lambda_t = \log[\alpha^{2}_t / \sigma^{2}_t]$
\State $L_{{\bm\eta}} = \omega(\lambda_t) (\lVert \tilde\rvx^w_c -  \hat\rvx_{c, {\bm\eta}}(\rvz_t, w)\rVert_{2}^{2}+\lVert \tilde\rvx^w -  \hat\rvx_{{\bm\eta}}(\rvz_t, w)\rVert_{2}^{2})$
\State ${\bm\eta} \leftarrow {\bm\eta} - \gamma\nabla_{{\bm\eta}}L_{{\bm\eta}}$
\EndWhile
\State {\setlength{\fboxsep}{0pt}{${\bm\theta} \leftarrow {\bm\eta}$}} \Comment{Student becomes next teacher}
\State {\setlength{\fboxsep}{0pt}{$N \leftarrow N/2$}} \Comment{Halve number of sampling steps}
\EndFor
\end{algorithmic}
\end{algorithm}

\begin{table}[!ht]
\centering
\resizebox{0.8\linewidth}{!}{
\begin{tabular}{cccc}
\Xhline{2\arrayrulewidth}
Guidance $w$  &Number of step & FID ($\downarrow$) & IS ($\uparrow$) \\
\Xhline{1\arrayrulewidth}
$w=0.0$  &1$\times$2  &212.20 &3.66\\
  &16$\times$2  &42.02 &7.95\\
  &64$\times$2 &35.37 &8.47\\
  &128$\times$2  &29.74 &8.87\\
  &256$\times$2 &20.14 &9.50\\
\Xhline{1\arrayrulewidth}
$w=0.3$ &1$\times$2 &213.07 &3.62\\
&16$\times$2 &48.74 &7.70\\
&128$\times$2 &34.28 &8.57\\
&256$\times$2 &24.54 &9.21 \\
\Xhline{1\arrayrulewidth}
$w=1.0$  &1$\times$2 &214.88 &3.54 \\
&16$\times$2 &64.92 &7.21 \\
&64$\times$2 &48.54 &7.62\\
&128$\times$2 &42.56 &8.00\\
&256$\times$2 &32.20 &8.81\\
\Xhline{1\arrayrulewidth}
$w=2.0$  &1$\times$2 &217.37 &3.48\\
&16$\times$2 &87.19 &6.50\\
&64$\times$2 &57.15 &7.22 \\
&128$\times$2 &50.30 &7.53 \\
&256$\times$2 &39.76 &8.26 \\
\Xhline{1\arrayrulewidth}
$w=4.0$  &1$\times$2 &220.11 &3.45\\
&16$\times$2 &115.57 &6.16\\
&64$\times$2 &71.45 &6.78\\
&128$\times$2 &61.75 &7.02\\
&256$\times$2 &49.21 &7.69\\
\Xhline{2\arrayrulewidth}
\end{tabular}
}
\caption{Distillation results on CIFAR-10 using the naive approach mentioned in \cref{app:sec:two_student_baseline}. 
Note that the naive approach still requires evaluating both a conditional and an unconditional model at each denoising step, and thus requires $\times$2 more steps or peak memory than our method. From the evaluated FID/IS scores, we observe that the naive distillation approach is not able to achieve strong performance.
}
\label{table:naive_sampler}
\end{table}

\clearpage
\subsection{Naive distillation approach}
\label{app:sec:two_student_baseline}
\begin{figure}[ht]
\centering
\begin{subfigure}{0.2\textwidth}
    \adjustbox{width=\linewidth, trim={.0\width} {.5\height} {0.5\width} {0\height},clip}{\includegraphics[width=\linewidth]{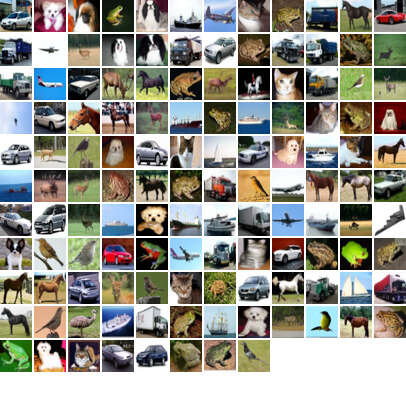}}
    \caption{256-step}
\end{subfigure}
\begin{subfigure}{0.2\textwidth}
    \adjustbox{width=\linewidth, trim={.0\width} {.5\height} {0.5\width} {0\height},clip}{\includegraphics[width=\linewidth]{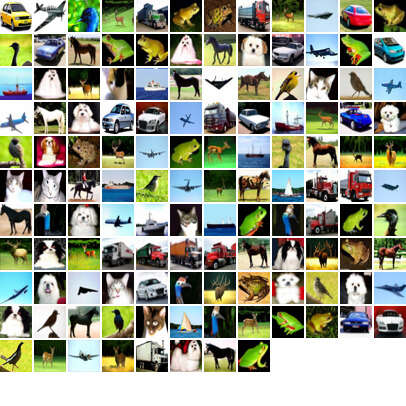}}
    \caption{64-step}
\end{subfigure}

\begin{subfigure}{0.2\textwidth}
     \adjustbox{width=\linewidth, trim={.0\width} {.5\height} {0.5\width} {0\height},clip}{\includegraphics[width=\linewidth]{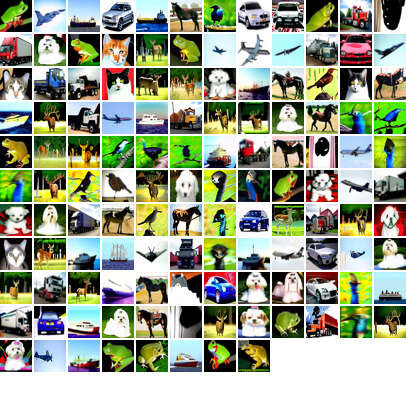}}
    \caption{16-step}
\end{subfigure}
\begin{subfigure}{0.2\textwidth}
    \adjustbox{width=\linewidth, trim={.0\width} {.5\height} {0.5\width} {0\height},clip}{\includegraphics[width=\linewidth]{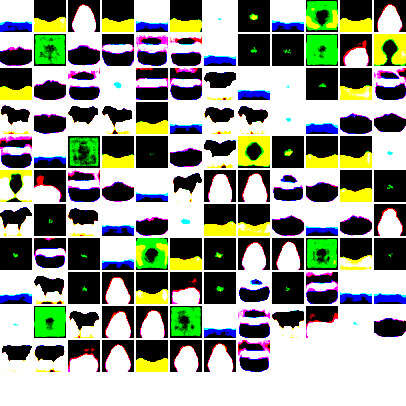}}
    \caption{1-step}
\end{subfigure}        
\caption{Samples using the distillation algorithm mentioned in \cref{app:sec:two_student_baseline}. The model is trained with guidance strength $w\in [0,4]$ on CIFAR-10. The samples are generated with DDIM (deterministic) sampler at $w=0$. We observe clear artifacts when the number of sampling step is small.}
\label{fig:naive_sample}
\end{figure}

A natural approach to progressively distill~\cite{salimans2022progressive}
a classifier-free guided model is to use a distilled student model that follows the same structure as the teacher---that is with a jointly trained distilled conditional and unconditional diffusion component. Denote the pre-trained teacher model $[\hat{\x}_{c,\bm\theta}, \hat{\x}_{\bm\theta}]$ and the student model $[\hat{\x}_{c,\bm\eta}, \hat{\x}_{\bm\eta}]$, we provide the training algorithm in \cref{alg:two_student}. To sample from the trained model, we can use DDIM deterministic sampler~\cite{song2020denoising} or the proposed stochastic sampler. 
We follow the training setting in \cref{app:sec:step_two}, use a $w$-conditioned model and train the model to condition on the guidance strength $[0,4]$.
We observe that the model distilled with \cref{alg:two_student} is not able to generate reasonable samples when the number of sampling is small. We provide the generated samples on CIFAR-10 with DDIM sampler in \cref{fig:naive_sample}, and the FID/IS scores in \cref{table:naive_sampler}.

\input{appendix_stable_diffusion}

\input{appendix_extra_samples}

%% file: table.tex
\begin{table*}[!ht]
\centering
\resizebox{0.65\linewidth}{!}{
\begin{tabular}{cccccc}
\Xhline{2\arrayrulewidth}
& & \multicolumn{2}{c}{ImageNet 64x64} &\multicolumn{2}{c}{CIFAR-10}\\
\Xhline{1\arrayrulewidth}
Guidance $w$ &Model & FID ($\downarrow$) & IS ($\uparrow$) &{FID ($\downarrow$)} & IS ($\uparrow$) \\
\Xhline{1\arrayrulewidth}
$w=0.0$ &Ours 1-step (D/S)     &22.74 / 26.91  &25.51 / 23.55 &8.34 / 10.65 &8.63 / 8.42 \\
&Ours 2-step (D/S) &9.75 /10.67 & 36.69 / 37.12 &4.48 / 4.81 &9.23 / 9.30 \\
&Ours 4-step (D/S)     &4.14 / 3.91 &46.64 / 48.92 &3.18 / 3.28  &9.50 / 9.60  \\
&Ours 8-step (D/S)     &2.79 / 2.44 &50.72 / 55.03  &2.86 / 3.11  &9.68 / 9.74\\
&Ours 16-step (D/S)     &2.44 / 2.10  &52.53 / 57.81  &2.78/3.12  &9.67 / 9.76 \\
&Single-$w$ 1-step   &19.61  &24.00 &6.64  & 8.88\\
&Single-$w$ 4-step   &4.79  &38.77 &3.14  &9.47\\
&Single-$w$ 8-step   &3.39 &42.13  &2.86 &9.67 \\
&Single-$w$ 16-step   &2.97  &43.63 &2.75 &9.65\\
&DDIM 16$\times$2-step~\cite{song2020denoising}     &7.68 &37.60  &10.11 &8.81\\
&DDIM 32$\times$2-step~\cite{song2020denoising}     &5.03 &40.93  &6.67  &9.17\\
&DDIM 64$\times$2-step~\cite{song2020denoising}     &3.74 &43.16 &4.64  &9.32\\
&Target (DDIM 1024$\times$2-step) &2.92 &44.81 &2.73 &9.66 \\
\Xhline{1\arrayrulewidth}
$w=0.3$  
&Ours 1-step (D/S)     &14.85 / 18.48   &37.09 / 33.30 &7.34 / 9.38 &8.90 / 8.67  \\
&Ours 2-step (D/S)  &5.052 / 5.81 &54.44 / 54.37 &4.23 / 4.74 & 9.45 / 9.45 \\
&Ours 4-step (D/S)     &2.17  / 2.24  &69.64 / 73.73 &3.58 / 3.95 &9.73 / 9.77 \\
&Ours 8-step (D/S)     &2.05 / 2.31  &76.01 / 83.00  &3.54 / 3.96 &9.87 / 9.90\\
&Ours 16-step (D/S)     &2.20 / 2.56  &79.47 / 87.50  &3.57 / 4.17 &9.89  / 9.97 \\
&Single-$w$ 1-step   &11.70  &36.95 &5.98 &9.13 \\
&Single-$w$ 4-step   &2.34  &62.08 &3.58  &9.75 \\
&Single-$w$ 8-step   &2.32  &68.76 &3.57 &9.85 \\
&Single-$w$ 16-step   &2.56   &70.97 &3.61 &9.88\\
&DDIM 16$\times$2-step     &5.33 &60.83 &10.83  &8.96\\
&DDIM 32$\times$2-step     &3.45  &68.03 &7.47 &9.33 \\
&DDIM 64$\times$2-step     &2.80   &72.55 &5.52  &9.51\\
&Target (DDIM 1024$\times$2-step) &2.36  &74.83 &3.65 	&9.83 \\
\Xhline{1\arrayrulewidth}
$w=1.0$   &Ours 1-step (D/S)     &7.54  / 8.92  & 75.19 / 67.80 &8.62 / 10.27 &9.21 / 8.97 \\
&Ours 2-step (D/S) & 5.77 /5.83 &109.97 / 108.38  &6.88 / 7.52 & 9.64 / 9.55\\
&Ours 4-step (D/S)     &7.95 / 8.51 & 128.98  / 135.36 & 7.39 / 7.64 &9.86 / 9.87\\
&Ours 8-step (D/S)     &9.33  / 10.56   &136.47  / 147.39 & 7.81 / 7.85 &9.9 / 10.05\\
&Ours 16-step (D/S)     &9.99 / 11.63 &139.11 / 153.17 &7.97 /  8.34   &10.00  / 10.05 \\
&Single-$w$ 1-step   &6.64  &74.41 &8.18  &9.32  \\
&Single-$w$ 4-step   &8.23   &118.52 &7.66  &9.88  \\
&Single-$w$ 8-step   &9.69  &125.20 &8.09  &9.89 \\
&Single-$w$ 16-step   &10.34  &127.70 &8.30 &9.95 \\
&DDIM 16$\times$2-step     &9.53   &112.75  &14.81 &8.98 \\
&DDIM 32$\times$2-step     &9.26   &126.22 & 11.44 &9.36\\
&DDIM 64$\times$2-step     &9.53  &133.17 & 9.79 &9.64 \\
&Target (DDIM 1024$\times$2-step) &9.84 &139.50 &7.80	&9.96 \\
\Xhline{1\arrayrulewidth}
$w=2.0$  &Ours 1-step (D/S)     & 10.71  / 10.55  &118.55  / 108.37 &13.23 / 14.33 &9.23 / 9.02\\
&Ours 2-step (D/S) &14.08 / 14.18 &160.04/ 161.43 &12.58 / 12.57 & 9.51 / 9.48  \\
&Ours 4-step (D/S)     &17.61 / 18.23  &178.29 / 184.45 &13.83 / 13.24 &9.70  / 9.77 \\
&Ours 8-step (D/S)     &18.80  / 20.25  &181.53  / 193.49 &14.41  / 13.67 &9.77 / 9.87 \\
&Ours 16-step (D/S)     & 19.25 / 21.11 &183.17  / 197.71 &14.80 / 14.28 &9.79  / 9.84 \\
&Single-$w$ 1-step   &11.12 &120.74 &13.31 & 9.23 \\
&Single-$w$ 4-step   &18.14   &172.74 &14.04 &9.70\\
&Single-$w$ 8-step   &19.24   &176.74 &14.67 &9.77\\
&Single-$w$ 16-step   &19.81  &177.69 &15.04 &9.79\\
&DDIM 16$\times$2-step     &15.92  &157.67 &20.25 &8.97\\
&DDIM 32$\times$2-step     &16.85  &175.72 &17.27  &9.29 \\
&DDIM 64$\times$2-step     &17.53  &182.11 &15.66 &9.48\\
&Target (DDIM 1024-step) &17.97 &190.56 &13.60	&9.81\\
\Xhline{1\arrayrulewidth}
$w=4.0$   &Ours 1-step (D/S)     &18.72  /  17.85  &157.46 / 148.97 &23.20  / 23.79  &8.88 / 8.70 \\
&Ours 2-step (D/S) &23.74 / 24.34 &196.05 / 200.11 &23.41 / 22.75  &9.16 / 9.11 \\
&Ours 4-step (D/S)     &26.45 / 27.33  &207.45 / 216.56 &25.11 / 23.62 &9.23  / 9.33 \\
&Ours 8-step (D/S)     &26.62 / 27.84  &203.47 / 219.89  &25.94 / 23.98  &9.26  / 9.55\\
&Ours 16-step (D/S)     &26.53 / 27.69  &204.13 / 218.70 &26.01 / 24.40 &9.33 / 9.50\\
&Single-$w$ 1-step   &19.857  &170.69 &23.17  &8.93 \\
&Single-$w$ 4-step   &27.75  &219.64 &24.45 &9.32\\
&Single-$w$ 8-step   &27.67 &218.08 &24.83 &9.38\\
&Single-$w$ 16-step   &27.40 &216.52 &25.11 &9.37\\
&DDIM 16$\times$2-step     &21.56 &195.17 &27.99   &8.71 \\
&DDIM 32$\times$2-step     &23.03  &213.23 &25.07  &9.07\\
&DDIM 64$\times$2-step     &23.64 &217.88 &23.41  &9.17\\
&Target (DDIM 1024$\times$2-step) &23.94 & 224.74 &21.28	&9.54\\
\Xhline{2\arrayrulewidth}
\end{tabular}
}
\caption{Distillation results on ImageNet 64x64 and CIFAR-10 ($w=0$ refers to non-guided models). For our method, $D$ and $S$ stand for deterministic and stochastic sampler respectively. We observe that training the model conditioned on an guidance interval $w\in [0,4]$ performs comparably with training a model on a fixed $w$ (see {Single-}$w$). Our approach significantly outperforms DDIM when using fewer steps, and is able to match the teacher performance using as few as 8 to 16 steps.
We also note that DDIM and DDPM evaluates both an unconditional and a conditional diffusion model at each denoising step, giving rise to the $\times$2 overhead either for peak memory or sampling steps. 
}
\label{table:i64_cifar10}
\end{table*}

%% file: appendix_stable_diffusion.tex
\FloatBarrier
\section{Latent-space distillation}
\label{app:sec:latent_space}
\subsection{Class-conditional generation}
\label{sec:app:class_condition}
\subsubsection{Training details}
In this experiment, we consider class-conditional generation on ImageNet 256$\times$256.
We first fine-tune the original $\epsilon$-prediction model to a $\rvv$-prediction model, and then start from the DDIM teacher model with 512 sampling steps, where we use the output as the target to train our distilled model.
For stage-one, we train the model for 2000 gradient updates with constant loss~\cite{kingma2021variational,salimans2022progressive}. For stage-two, we train the model with 2000 gradient updates except when the sampling size equals to 1,2, or 4, where we train for 20000 gradient updates. We train the second stage model with SNR-trunction loss~\cite{kingma2021variational,salimans2022progressive}. For both stages, we train with extra 500 learning rate warm-up steps, where we linearly increase the learning rate from zero to the target learning rate.
We use a batch size of 2048 and uniformly sample the guidance strength $w \in [w_{min}=0, w_{max}=14]$ during training. %
\cinqualitative

\paragraph{Additional results}
We provide quantitative results evaluated by precision and recall in \cref{fig:cinldmprresults}. These results confirm a significant performance boost of our method in the small-step regime, especially for 1-4 sampling steps. Our distilled latent diffusion model for 2- and 4-step sampling nearly matches DDIM performance at 32 steps in terms of precision and significantly outperforms it in terms of recall for low numbers of steps.
For more qualitative results, see \cref{fig:cinldmprresults}, where we depict random samples for the 1- and 2-step model and contrast them to DDIM sampling.
\cinldmprresults

\subsection{Text-guided image generation}
\label{sec:app:text_guided}
\subsubsection{Training details}
We consider the LAION-5B datasets with resolution 256$\times$256 and $512\times512$ in this experiment.
\paragraph{LAION-5B 256$\times$256}
\laionconvergence
Similar to \cref{sec:app:class_condition}, we first fine-tune the original $\epsilon$-prediction model to a $\rvv$-prediction model.
We start from the DDIM teacher model with 512 sampling steps, and use the output as the target to train our distilled model.
For stage-one, we train the model for 2000-5000 gradient updates with constant loss~\cite{kingma2021variational,salimans2022progressive}. For stage-two, we train the model with 2000-5000 gradient updates except when the sampling size equals to 1,2, or 4, where we train for 10000-50000 gradient updates. We train the second stage model with SNR-trunction loss~\cite{kingma2021variational,salimans2022progressive}. For both stages, we train with extra 100-1000 learning rate warm-up steps, where we linearly increase the learning rate from zero to the target learning rate.
We use a batch size of 1024 and uniformly sample the guidance strength $w \in [w_{min}=2, w_{max}=14]$ during training. %

\cref{fig:laionconvergence} provides a convergence analysis of the different training setting described above. We observe that our method approaches DDIM sampling of the base model after a few thousand training iterations and outperforms it quickly in the 1- and 2-step regime. However, for maximum performance, longer training is required.

\paragraph{LAION-5B 512$\times$512}
Similarly, we first fine-tune the original $\epsilon$-prediction model to a $\rvv$-prediction model.
We start from the DDIM teacher model with 512 sampling steps, and use the output as the target to train our distilled model.
For stage-one, we train the model for 2000-5000 gradient updates with constant loss~\cite{kingma2021variational,salimans2022progressive}. For stage-two, we train the model with 2000-5000 gradient updates except when the sampling step equals to 1,2, or 4, where we train for 10000-50000 gradient updates. We train the second-stage model with SNR-trunction loss~\cite{kingma2021variational,salimans2022progressive}. For both stages, we train with extra 1000 learning rate warm-up steps, where we linearly increase the learning rate from zero to the target learning rate.
We use a batch size of 512 and uniformly sample the guidance strength $w \in [w_{min}=2, w_{max}=14]$ during training. %
\stabledpmplot
\laionqualitativesupp

\paragraph{Additional results}
\dpmddimfidtable
\dpmddimcliptable
Besides DDIM, we also compare our method here with DPM$++$-Solver~\cite{dpmsolver,dpmpp}, a state-of-the-art sampler that requires no additional training and has achieved good results for $\geq10$ sampling steps for latent diffusion models. Unlike our distilled model, this method, similar to DDIM, must use classifier-free guidance to achieve good results. This doubles the number of U-Net evaluations compared to our $w$-conditional approach. 

We provide a qualitative comparison of these sampling methods in \cref{fig:laionqualitativesupp}, where we clearly see the benefits of our distillation approach for low numbers of sampling steps: our method produces sharper and more coherent results than the training-free samplers.
This behavior is reflected by the quantitative FID and CLIP analysis in \cref{fig:stabledpmplot} and \cref{tab:dpmddimfidtable},~\cref{tab:dpmddimcliptable}.
While the speed-up here is not quite as significant as in pixel-space, our method still achieves very good results with 2 or 4 sampling steps. Our approach further reduces the maximum memory or denoising step by a half compared to existing methods due to $w$-conditioning (since here we no longer need to evaluate both the unconditional model and conditional model for classifier-free guidance, we only need one distilled $w$-conditional model).
We hope that our work will lead to progress in real-time applications of general high-resolution text-to-image systems.

\begin{figure*}[!htb]
     \centering
     \begin{subfigure}[b]{0.4\linewidth}
         \centering
         \includegraphics[width=\textwidth]{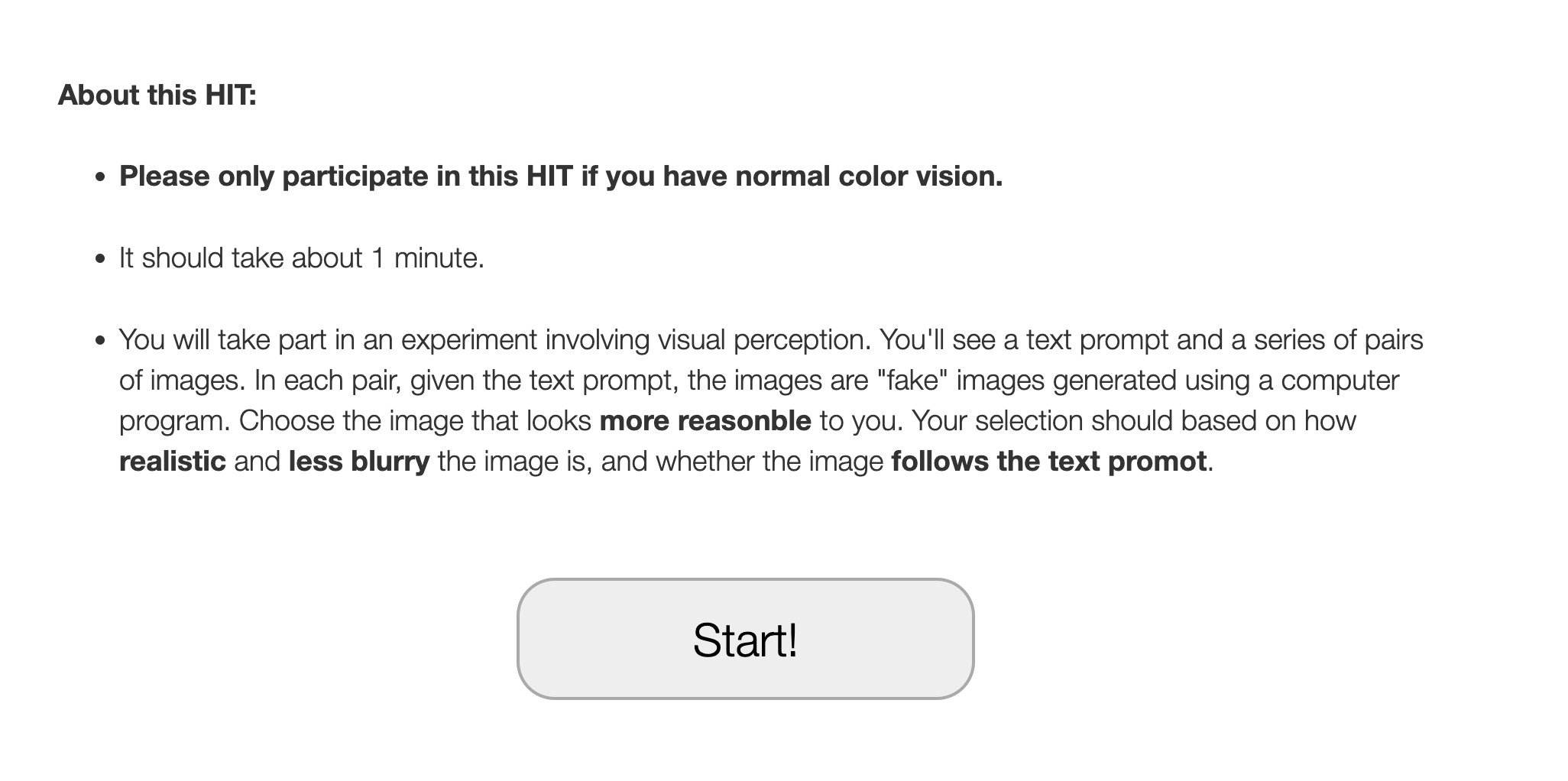}
         \caption{Instructions for the human evaluators on Amazon Mechanical Turk.}
     \end{subfigure}
     \begin{subfigure}[b]{0.4\linewidth}
         \centering
         \includegraphics[width=\textwidth]{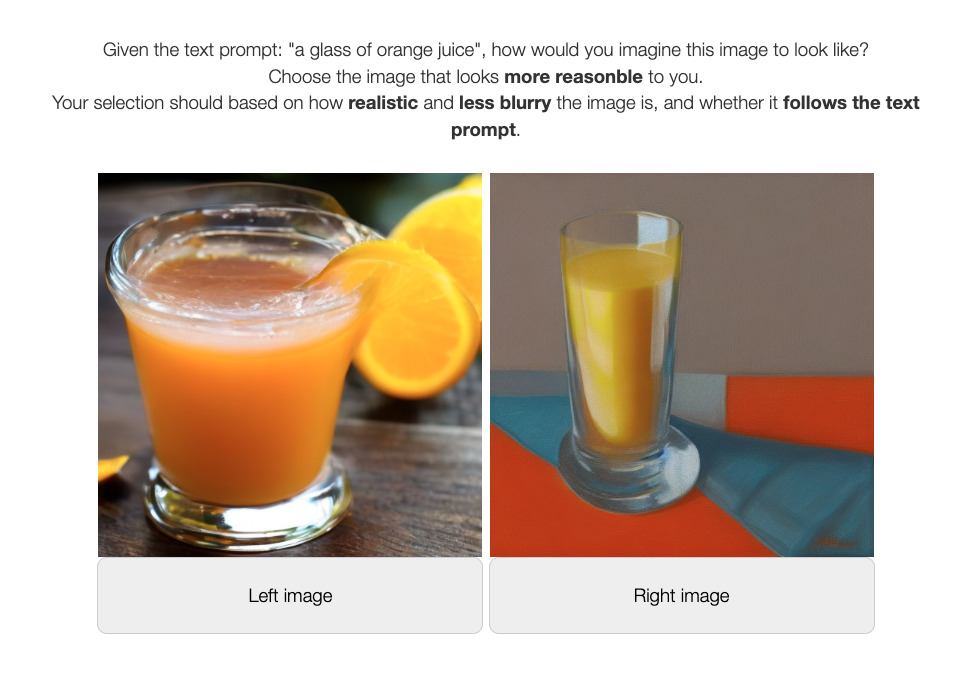}
         \caption{Images generated by our 4-step distillation model (left) and images generated by the 4$\times$2-step baseline (right).}
     \end{subfigure}
     
     \begin{subfigure}[b]{0.4\linewidth}
         \centering
         \includegraphics[width=\textwidth]{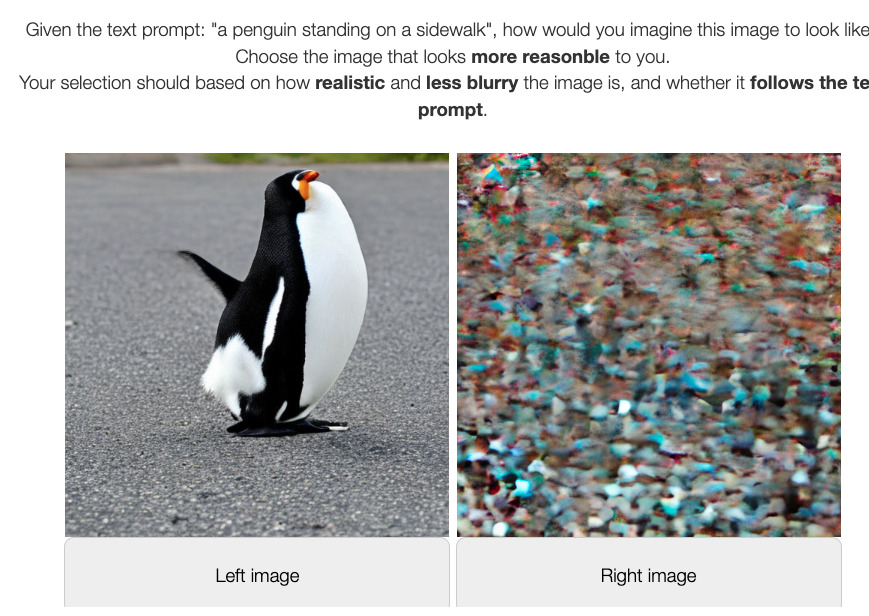}
         \caption{Images generated by our 2-step distillation model (left) and images generated by the 2$\times$2-step baseline (right).}
         \label{fig:app:human_eval_peguin}
     \end{subfigure}
     \begin{subfigure}[b]{0.4\linewidth}
         \centering
         \includegraphics[width=\textwidth]{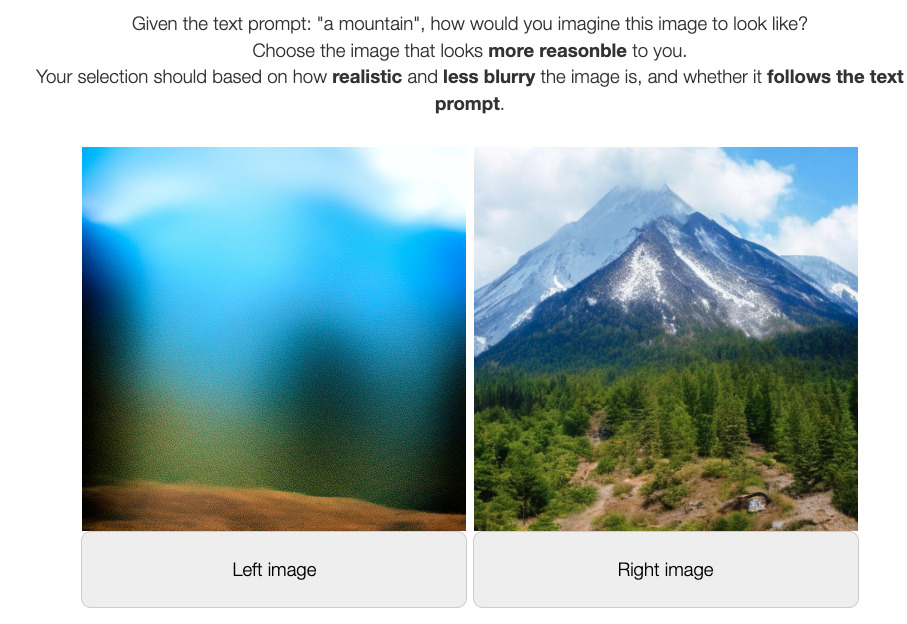}
         \caption{Images generated by the 2$\times$2-step baseline (left) and images generated by our 2-step distillation model (right).
         }
     \end{subfigure}
     
    \caption{A snapshot of the human evaluation interface we used on Amazon Mechanical Turk.}
    \label{fig:app:human_eval}
\end{figure*}

We also provide human evaluation results by leveraging Amazon Mechanical Turk. 
We generate images using text prompts from \cite{yu2022scaling}.
We compare our distilled model sampled using 2 or 4 denoising steps with DDIM and DPM$++$ solver sampled using  2$\times$2 or 4$\times$2 denoising steps. 
For each setting, we generate 100 HITs each with 17 pair-wise comparisons between samples generated with our approach and the baseline. In each of the question, the user is shown the text prompt used to generate the image and asked to select the image that looks better to them. 
We provide a snapshot of our user interface in \cref{fig:app:human_eval}. We provide the results in \cref{table:app:text2img_human_eval}. Although we observe noisy answers (for instance some user would prefer the right image to the left image in \cref{fig:app:human_eval_peguin}), our distilled model still consistently outperforms the baselines in all the settings we considered in \cref{table:app:text2img_human_eval}. To get higher-quality user feedback and reduce the noise in the answers, in the future work, we will perform a new human evaluation with a larger sample size and extra constraints to ensure the quality of the response. We will also build a framework to automatically ignore HITs with random selections.

\subsection{Text-guided image-to-image translation}
\subsubsection{Training details}
We use the model trained for text-guided image generation. The training details can be found in \cref{sec:app:text_guided}.

\subsubsection{Extra analysis}
\begin{figure*}[!htb]
     \centering
     \begin{subfigure}[b]{\linewidth}
         \centering
         \includegraphics[width=\textwidth]{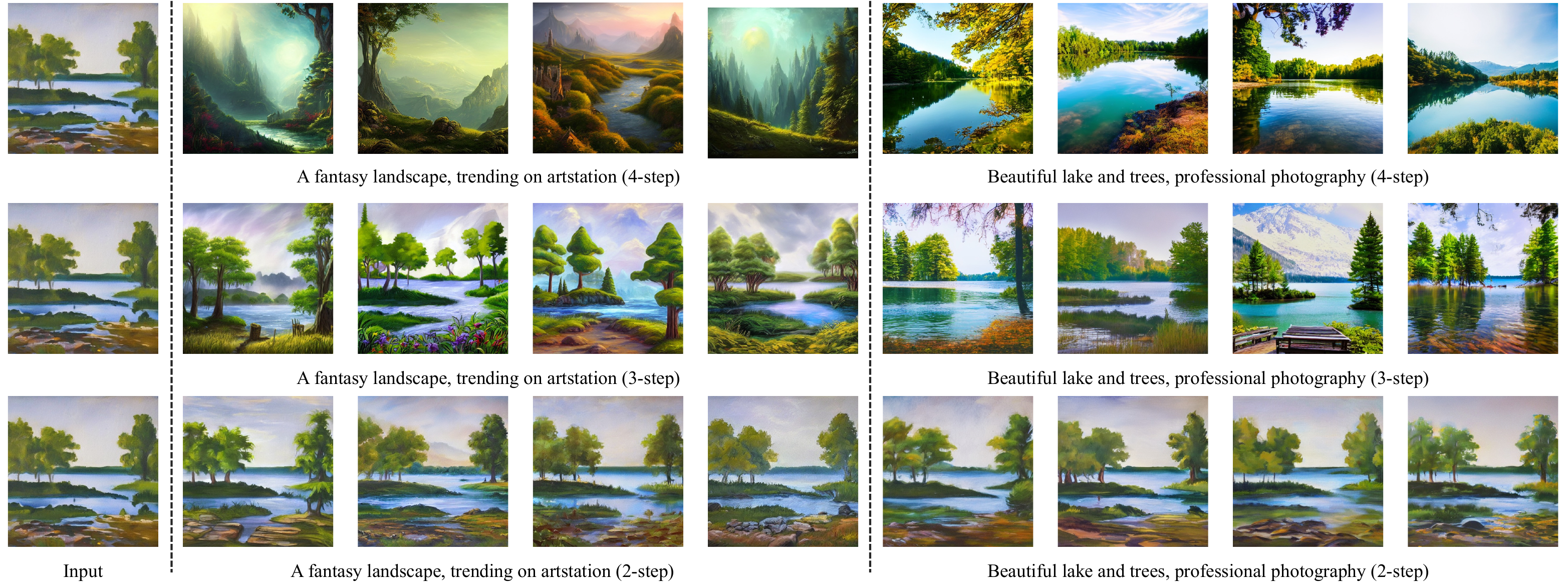}
     \end{subfigure}
    \caption{In this example, we study the trade-off between efficiency, realism, and controllability for guided image translation with SDEdit~\cite{meng2021sdedit}. We use a 4-step distilled text-guided image generation model trained on LAION-5B (512$\times$512). The training detail is discussed in \cref{sec:app:text_guided}. Given an input image (guide), we consider perturbing the input image with different noise level, with 2 denoising step corresponding to perturb the image with around 50\% noise, and 4 denoising step corresponding to perturb the image with around 100\% noise according to the DDIM noise schedule. We observe that the more noise we perturb, the more aligned the images are with the text prompt, but the less faithful they are to the input image.}
    \label{fig:app:img2img_tradeoff_4step_model}
\end{figure*}

\begin{figure*}[!htb]
     \centering
     \begin{subfigure}[b]{\linewidth}
         \centering
         \includegraphics[width=\textwidth]{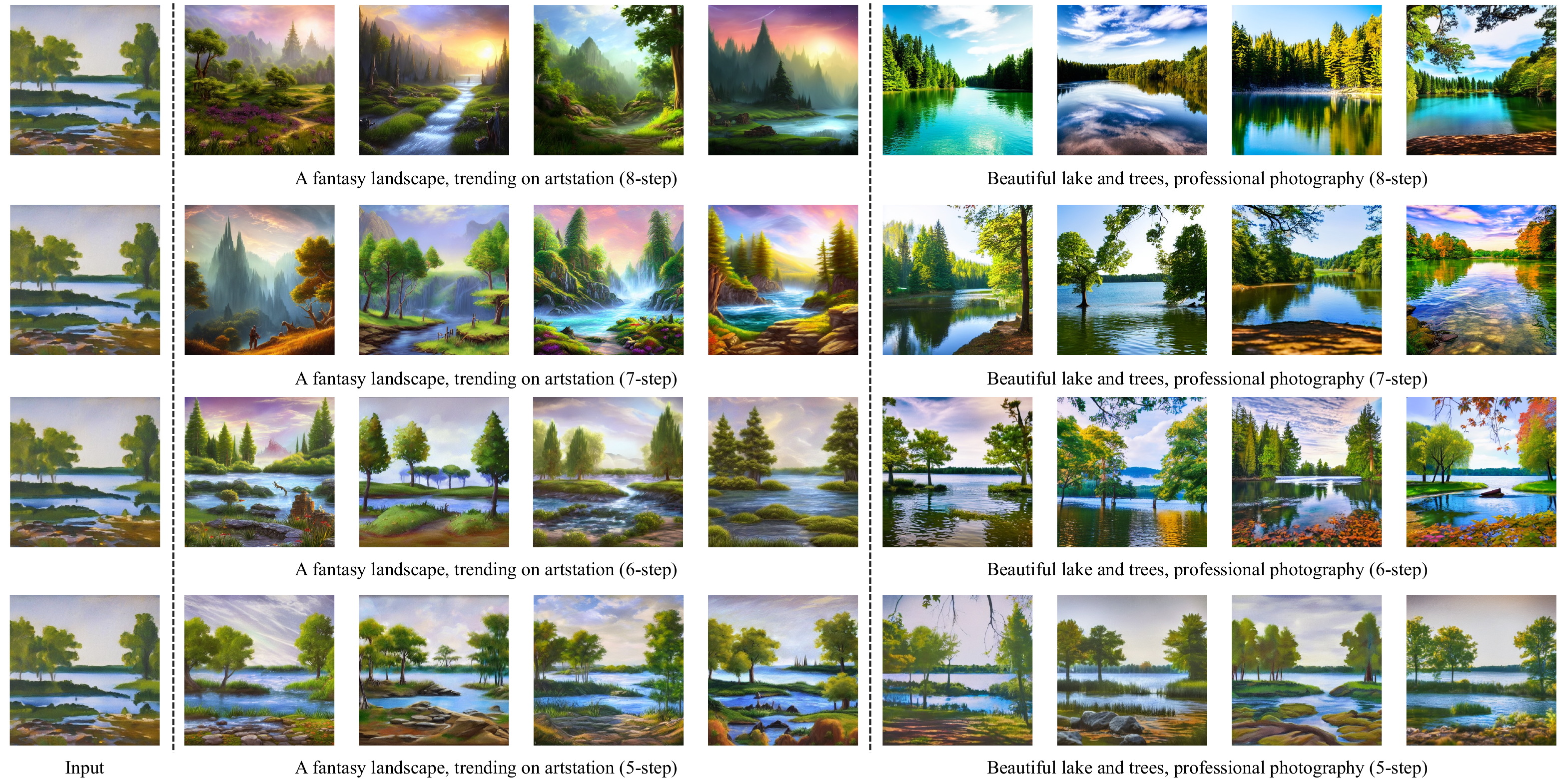}
     \end{subfigure}
    \caption{In this example, we study the trade-off between efficiency, realism, and controllability for guided image translation with SDEdit~\cite{meng2021sdedit}. We use a 8-step distilled text-guided image generation model trained on LAION-5B (512$\times$512). The training detail is discussed in \cref{sec:app:text_guided}. Given an input image (guide), we consider perturbing the input image with different noise level, with 5 denoising step corresponding to perturb the image with around 60\% noise, and 8 denoising step corresponding to perturb the image with around 100\% noise according to the DDIM noise schedule. We observe that the more noise we perturb, the more aligned the images are with the text prompt, but the less faithful they are to the input image.}
    \label{fig:app:img2img_tradeoff_8step_model}
\end{figure*}

We provide more analysis on the trade-off
between sample quality, controllability and efficiency in \cref{fig:app:img2img_tradeoff_4step_model} and \cref{fig:app:img2img_tradeoff_8step_model}. Similar to \cite{meng2021sdedit}, we also observe a trade-off between realism, controllability and faithfulness as we increase the initial perturbed noise level: the more noise we add, the more aligned the images are to the text prompt, but less faithful to the input image (see \cref{fig:app:img2img_tradeoff_4step_model} and \cref{fig:app:img2img_tradeoff_8step_model}).

\inpaintingtable

\begin{table}
\centering
\resizebox{\linewidth}{!}{
\begin{tabular}{ccccccc}
\Xhline{1\arrayrulewidth}
Ours &Baseline &Our method is better $(\uparrow)$\\
\Xhline{2\arrayrulewidth}
Distillation 2-step &DDIM 2$\times$2-step &66.32\% \\
Distillation 2-step &DPM$++$ 2$\times$2-step &68.97\% \\
Distillation 2-step &DDIM 4$\times$2-step &57.44\% \\
Distillation 2-step &DPM$++$ 4$\times$2-step &59.88\% \\
Distillation 4-step &DDIM 4$\times$2-step &67.36\% \\
Distillation 4-step &DPM$++$ 4$\times$2-step &64.71\% \\
\Xhline{2\arrayrulewidth}
\end{tabular}
}
\caption{ 
Human evaluation on text-guided image generation. Here the model is trained on LAION-5B (512$\times$512). 
We leverage Amazon Mechanical Turk for human evaluation. We perform pairwise comparison between our method and the baselines. 
We compare our method using 2 or 4 denoising steps with DDIM~\cite{song2020denoising} and DPM$++$~\cite{dpmpp} samplers using 2$\times$2 or 4$\times$2 denoising steps. We use a guidance strength of 12.5 for all methods. 
For each setting, we distribute 100 HITs each with 17 pairwise comparison questions. 
We show MTurk workers the text prompt as well as the two generated images, and then ask them to select the one they think is better.
We provide a snapshot of the interface in \cref{fig:app:human_eval}.
In the table, we report the percentage that the MTurk workers think our method is better than the baseline. 
Although, we observe noise in the response (some user would prefer the right image to the left image in \cref{fig:app:human_eval_peguin}), 
our method still consistently outperform the baselines in all settings. For the future work, we will incorporate schemes to ignore invalid HITs with random answers. We will also perform another human evaluation study with larger sample sizes and more constraints to ensure high-quality responses.
}
\label{table:app:text2img_human_eval}
\end{table}

\inpaintingqualitative

\subsection{Image inpainting}
\subsubsection{Training details}
Similar to our previous experiments, we fine-tune the $\eps$-prediction model to a $\rvv$-prediction model, using the large mask generation scheme suggested in \emph{LAMA}~\cite{suvorov2022resolution} and train on LAION-5B at $512 \times 512$ resolution.
We start from the DDIM teacher model with 512 sampling steps, and use the output as the target to train our distilled model.
For stage-one, we train the model for 2000 gradient updates with constant loss~\cite{kingma2021variational,salimans2022progressive}. For stage-two, we train the model with 10000 gradient updates except when the sampling size equals to 1 or 2, where we train for 5000 gradient updates. We train the second stage model with SNR-trunction loss~\cite{kingma2021variational,salimans2022progressive}. For both stages, we train with extra 1000 learning rate warm-up steps, where we linearly increase the learning rate from zero to the target learning rate.
We use a batch size of 512 and uniformly sample the guidance strength $w \in [w_{min}=2, w_{max}=14]$ during training. %

\paragraph{Additional evaluation results}
A quantitative comparison with DDIM sampling at low sampling numbers of sampling steps can be found in \cref{tab:inpaintingtable}, additional samples are in \cref{fig:inpaintingqualitative}.

%% file: appendix_extra_samples.tex
\section{Extra samples for pixel-space distillation}
In this section, we provide extra samples for the pixel-space distillation models. We generate samples using the deterministic sampler (see~\cref{alg:app:stage2}) and the stochastic sampler (see~\cref{alg:stage2_stochastic}).

\clearpage
\begin{figure*}[h!]
\centering
\begin{subfigure}{0.22\textwidth}
    \adjustbox{width=\linewidth, trim={.0\width} {.5\height} {0.5\width} {0\height},clip}{\includegraphics[width=\linewidth]{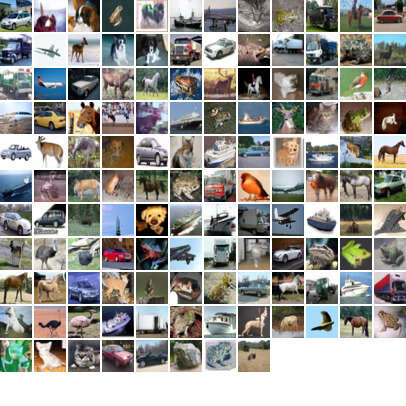}}
    \caption{$w=0$}
\end{subfigure}
\hfill
\begin{subfigure}{0.22\textwidth}
    \adjustbox{width=\linewidth, trim={.0\width} {.5\height} {0.5\width} {0\height},clip}{\includegraphics[width=\linewidth]{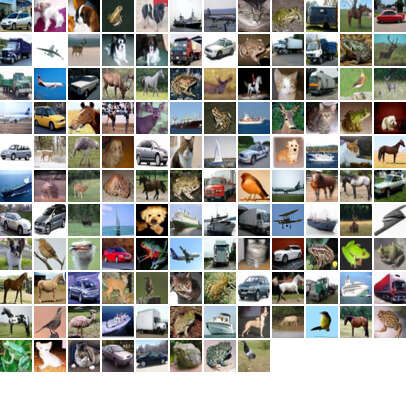}}
    \caption{$w=1$}
\end{subfigure}
\hfill
\begin{subfigure}{0.22\textwidth}
     \adjustbox{width=\linewidth, trim={.0\width} {.5\height} {0.5\width} {0\height},clip}{\includegraphics[width=\linewidth]{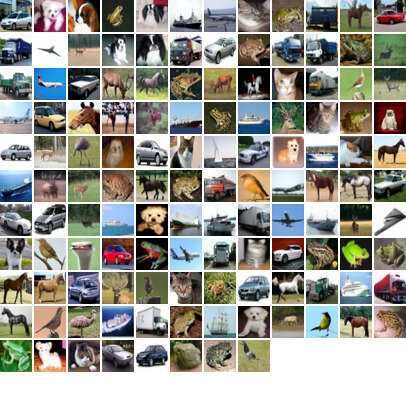}}
    \caption{$w=2$}
\end{subfigure}
\hfill
\begin{subfigure}{0.22\textwidth}
    \adjustbox{width=\linewidth, trim={.0\width} {.5\height} {0.5\width} {0\height},clip}{\includegraphics[width=\linewidth]{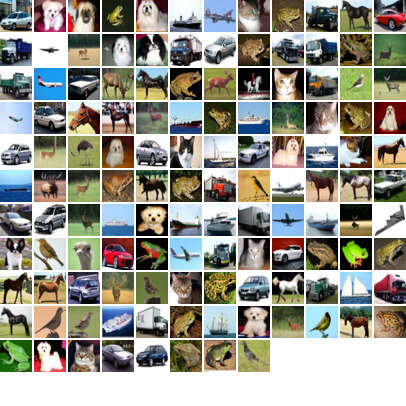}}
    \caption{$w=4$}
\end{subfigure}        
\caption{Ours (deterministic in pixel-space) on CIFAR-10. Distilled 256 sampling steps.}
\label{fig:figures}
\end{figure*}

\begin{figure*}[!ht]
\centering
\begin{subfigure}{0.22\textwidth}
    \adjustbox{width=\linewidth, trim={.0\width} {.5\height} {0.5\width} {0\height},clip}{\includegraphics[width=\linewidth]{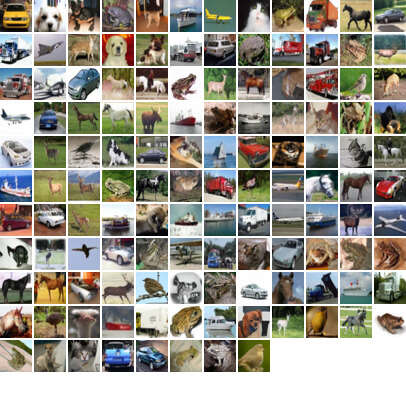}}
    \caption{$w=0$}
\end{subfigure}
\hfill
\begin{subfigure}{0.22\textwidth}
    \adjustbox{width=\linewidth, trim={.0\width} {.5\height} {0.5\width} {0\height},clip}{\includegraphics[width=\linewidth]{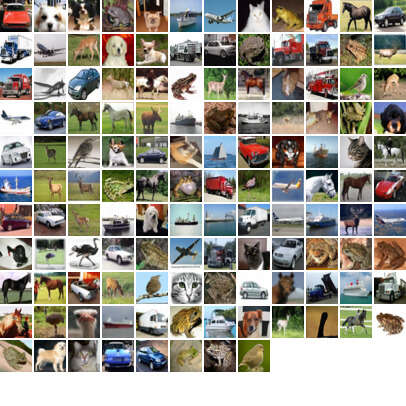}}
    \caption{$w=1$}
\end{subfigure}
\hfill
\begin{subfigure}{0.22\textwidth}
     \adjustbox{width=\linewidth, trim={.0\width} {.5\height} {0.5\width} {0\height},clip}{\includegraphics[width=\linewidth]{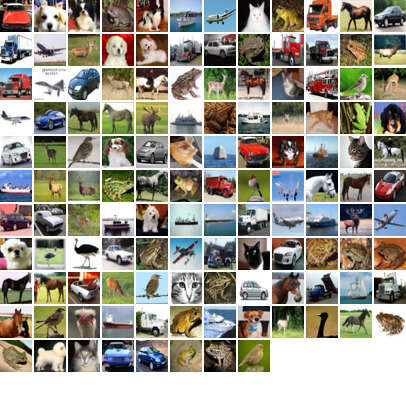}}
    \caption{$w=2$}
\end{subfigure}
\hfill
\begin{subfigure}{0.22\textwidth}
    \adjustbox{width=\linewidth, trim={.0\width} {.5\height} {0.5\width} {0\height},clip}{\includegraphics[width=\linewidth]{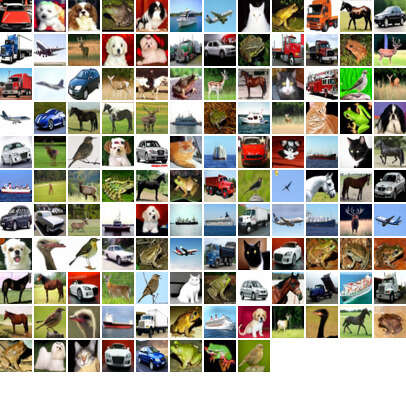}}
    \caption{$w=4$}
\end{subfigure}        
\caption{Ours (stochastic in pixel-space) on CIFAR-10. Distilled 256 sampling steps.}
\label{fig:figures}
\end{figure*}

\begin{figure*}[h!]
\centering
\begin{subfigure}{0.22\textwidth}
    \adjustbox{width=\linewidth, trim={.0\width} {.5\height} {0.5\width} {0\height},clip}{\includegraphics[width=\linewidth]{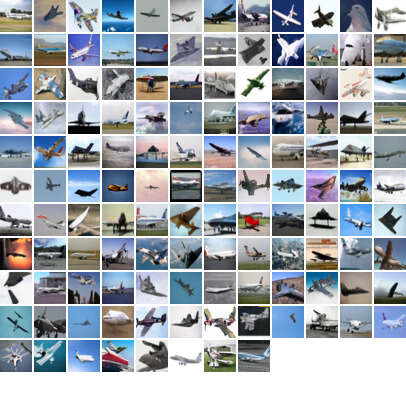}}
    \caption{$w=0$}
\end{subfigure}
\hfill
\begin{subfigure}{0.22\textwidth}
    \adjustbox{width=\linewidth, trim={.0\width} {.5\height} {0.5\width} {0\height},clip}{\includegraphics[width=\linewidth]{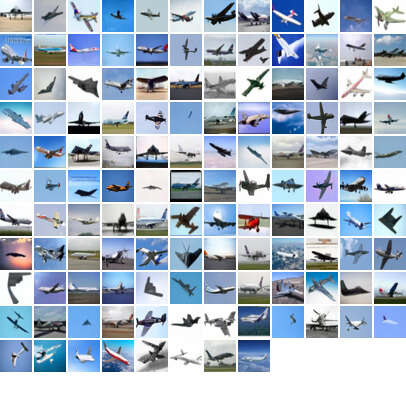}}
    \caption{$w=1$}
\end{subfigure}
\hfill
\begin{subfigure}{0.22\textwidth}
     \adjustbox{width=\linewidth, trim={.0\width} {.5\height} {0.5\width} {0\height},clip}{\includegraphics[width=\linewidth]{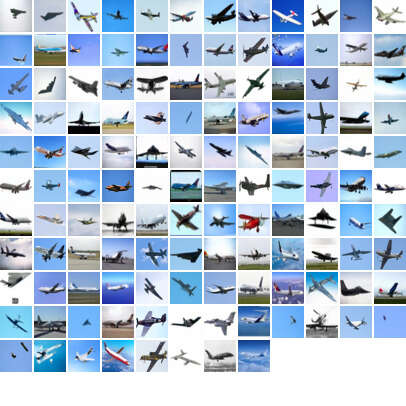}}
    \caption{$w=2$}
\end{subfigure}
\hfill
\begin{subfigure}{0.22\textwidth}
    \adjustbox{width=\linewidth, trim={.0\width} {.5\height} {0.5\width} {0\height},clip}{\includegraphics[width=\linewidth]{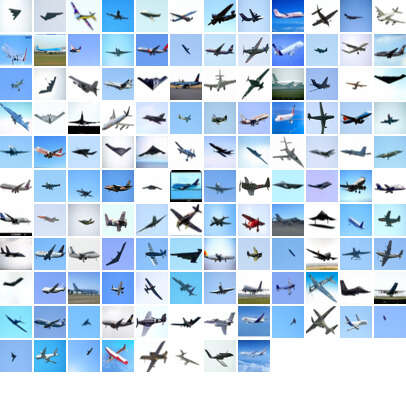}}
    \caption{$w=4$}
\end{subfigure}        
\caption{Ours (deterministic in pixel-space) on CIFAR-10. Distilled 256 sampling steps. Class-conditioned samples.}
\label{fig:figures}
\end{figure*}

\begin{figure*}[h!]
\centering
\begin{subfigure}{0.22\textwidth}
    \adjustbox{width=\linewidth, trim={.0\width} {.5\height} {0.5\width} {0\height},clip}{\includegraphics[width=\linewidth]{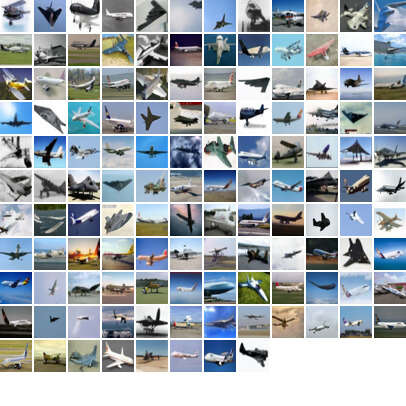}}
    \caption{$w=0$}
\end{subfigure}
\hfill
\begin{subfigure}{0.22\textwidth}
    \adjustbox{width=\linewidth, trim={.0\width} {.5\height} {0.5\width} {0\height},clip}{\includegraphics[width=\linewidth]{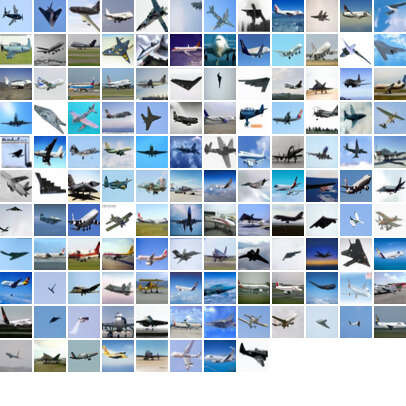}}
    \caption{$w=1$}
\end{subfigure}
\hfill
\begin{subfigure}{0.22\textwidth}
     \adjustbox{width=\linewidth, trim={.0\width} {.5\height} {0.5\width} {0\height},clip}{\includegraphics[width=\linewidth]{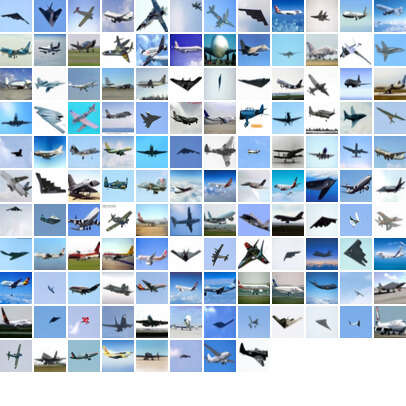}}
    \caption{$w=2$}
\end{subfigure}
\hfill
\begin{subfigure}{0.22\textwidth}
    \adjustbox{width=\linewidth, trim={.0\width} {.5\height} {0.5\width} {0\height},clip}{\includegraphics[width=\linewidth]{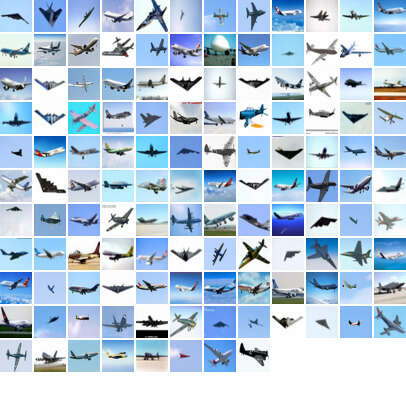}}
    \caption{$w=4$}
\end{subfigure}        
\caption{Ours (stochastic in pixel-space) on CIFAR-10. Distilled 256 sampling steps. Class-conditioned samples.}
\label{fig:figures}
\end{figure*}

\begin{figure*}[!htb]
\centering
\begin{subfigure}{0.22\textwidth}
    \adjustbox{width=\linewidth, trim={.0\width} {.5\height} {0.5\width} {0\height},clip}{\includegraphics[width=\linewidth]{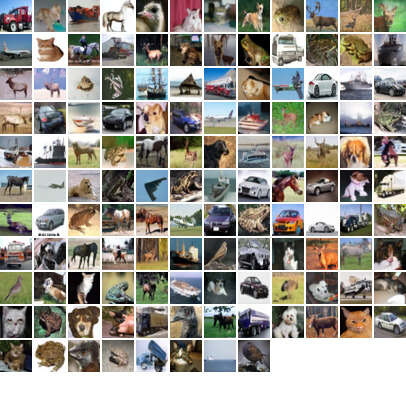}}
    \caption{$w=0$}
\end{subfigure}
\hfill
\begin{subfigure}{0.22\textwidth}
    \adjustbox{width=\linewidth, trim={.0\width} {.5\height} {0.5\width} {0\height},clip}{\includegraphics[width=\linewidth]{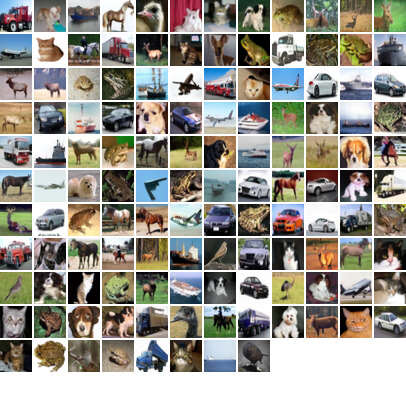}}
    \caption{$w=1$}
\end{subfigure}
\hfill
\begin{subfigure}{0.22\textwidth}
     \adjustbox{width=\linewidth, trim={.0\width} {.5\height} {0.5\width} {0\height},clip}{\includegraphics[width=\linewidth]{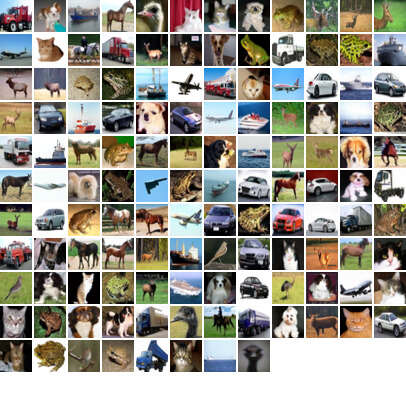}}
    \caption{$w=2$}
\end{subfigure}
\hfill
\begin{subfigure}{0.22\textwidth}
    \adjustbox{width=\linewidth, trim={.0\width} {.5\height} {0.5\width} {0\height},clip}{\includegraphics[width=\linewidth]{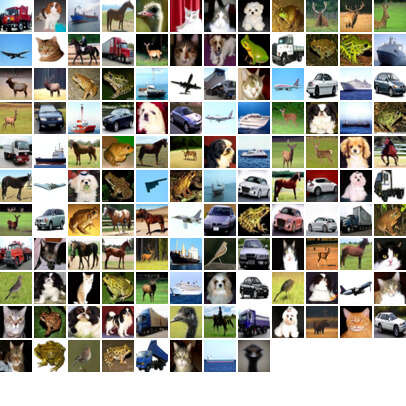}}
    \caption{$w=4$}
\end{subfigure}        
\caption{Ours (deterministic in pixel-space) on CIFAR-10. Distilled 4 sampling steps.}
\label{fig:figures}
\end{figure*}

\begin{figure*}[!hb]
\centering
\begin{subfigure}{0.22\textwidth}
    \adjustbox{width=\linewidth, trim={.0\width} {.5\height} {0.5\width} {0\height},clip}{\includegraphics[width=\linewidth]{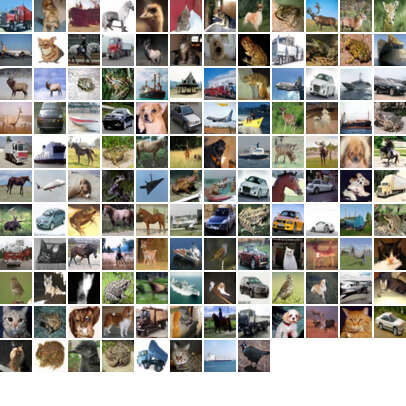}}
    \caption{$w=0$}
\end{subfigure}
\hfill
\begin{subfigure}{0.22\textwidth}
    \adjustbox{width=\linewidth, trim={.0\width} {.5\height} {0.5\width} {0\height},clip}{\includegraphics[width=\linewidth]{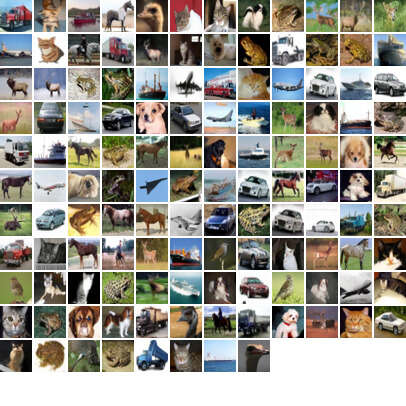}}
    \caption{$w=1$}
\end{subfigure}
\hfill
\begin{subfigure}{0.22\textwidth}
     \adjustbox{width=\linewidth, trim={.0\width} {.5\height} {0.5\width} {0\height},clip}{\includegraphics[width=\linewidth]{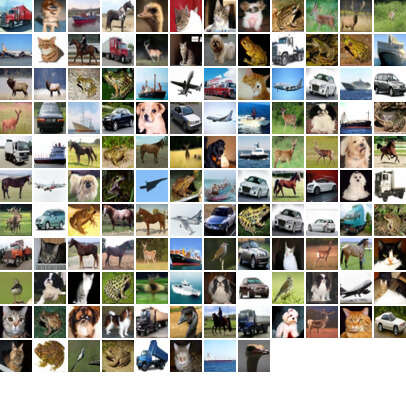}}
    \caption{$w=2$}
\end{subfigure}
\hfill
\begin{subfigure}{0.22\textwidth}
    \adjustbox{width=\linewidth, trim={.0\width} {.5\height} {0.5\width} {0\height},clip}{\includegraphics[width=\linewidth]{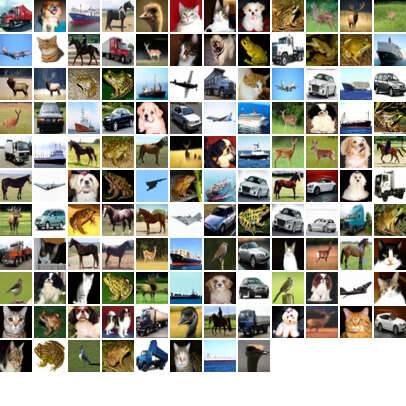}}
    \caption{$w=4$}
\end{subfigure}        
\caption{Ours (stochastic in pixel-space) on CIFAR-10. Distilled 4 sampling steps.}
\label{fig:figures}
\end{figure*}
 
\begin{figure*}[!hb]
\centering
\begin{subfigure}{0.22\textwidth}
    \adjustbox{width=\linewidth, trim={.0\width} {.5\height} {0.5\width} {0\height},clip}{\includegraphics[width=\linewidth]{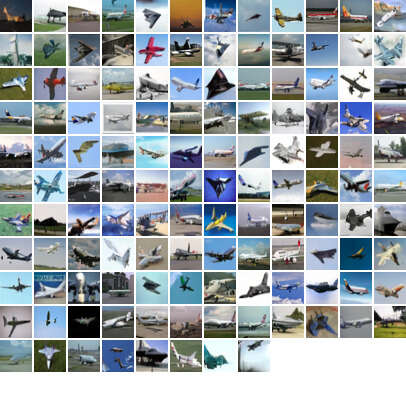}}
    \caption{$w=0$}
\end{subfigure}
\hfill
\begin{subfigure}{0.22\textwidth}
    \adjustbox{width=\linewidth, trim={.0\width} {.5\height} {0.5\width} {0\height},clip}{\includegraphics[width=\linewidth]{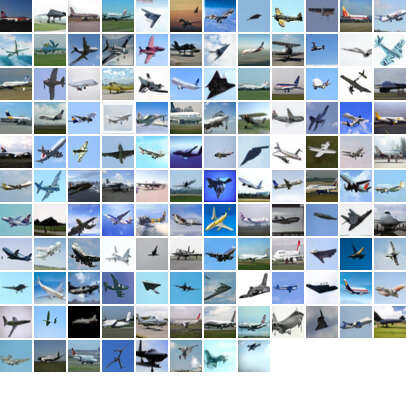}}
    \caption{$w=1$}
\end{subfigure}
\hfill
\begin{subfigure}{0.22\textwidth}
     \adjustbox{width=\linewidth, trim={.0\width} {.5\height} {0.5\width} {0\height},clip}{\includegraphics[width=\linewidth]{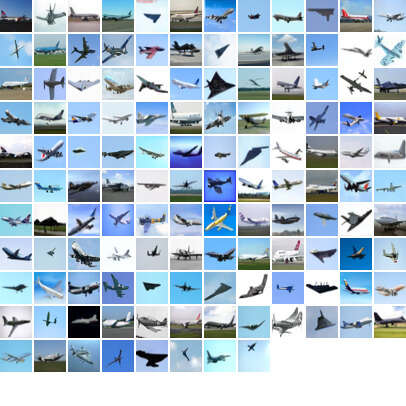}}
    \caption{$w=2$}
\end{subfigure}
\hfill
\begin{subfigure}{0.22\textwidth}
    \adjustbox{width=\linewidth, trim={.0\width} {.5\height} {0.5\width} {0\height},clip}{\includegraphics[width=\linewidth]{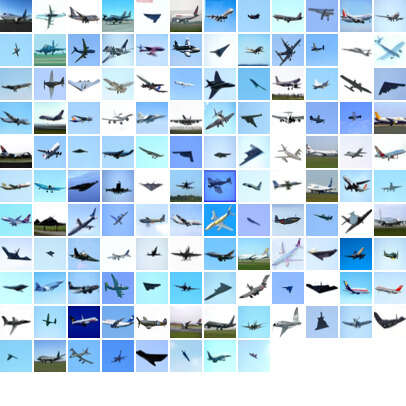}}
    \caption{$w=4$}
\end{subfigure}        
\caption{Ours (deterministic in pixel-space) on CIFAR-10. Distilled 4 sampling steps. Class-conditioned samples.}
\label{fig:figures}
\end{figure*}

\begin{figure*}[!hb]
\centering
\begin{subfigure}{0.22\textwidth}
    \adjustbox{width=\linewidth, trim={.0\width} {.5\height} {0.5\width} {0\height},clip}{\includegraphics[width=\linewidth]{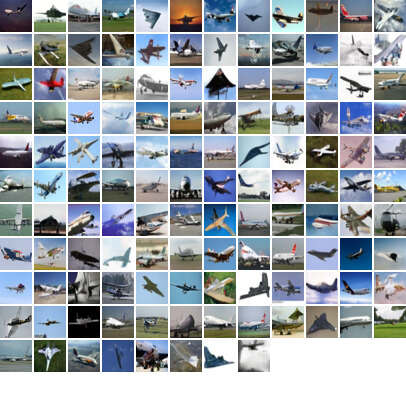}}
    \caption{$w=0$}
\end{subfigure}
\hfill
\begin{subfigure}{0.22\textwidth}
    \adjustbox{width=\linewidth, trim={.0\width} {.5\height} {0.5\width} {0\height},clip}{\includegraphics[width=\linewidth]{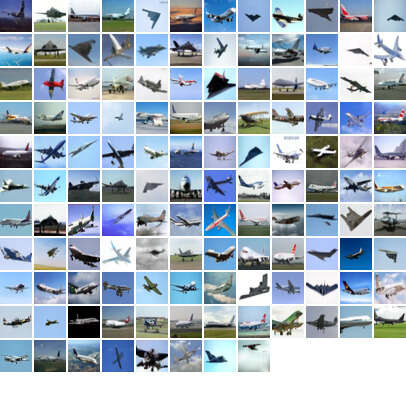}}
    \caption{$w=1$}
\end{subfigure}
\hfill
\begin{subfigure}{0.22\textwidth}
     \adjustbox{width=\linewidth, trim={.0\width} {.5\height} {0.5\width} {0\height},clip}{\includegraphics[width=\linewidth]{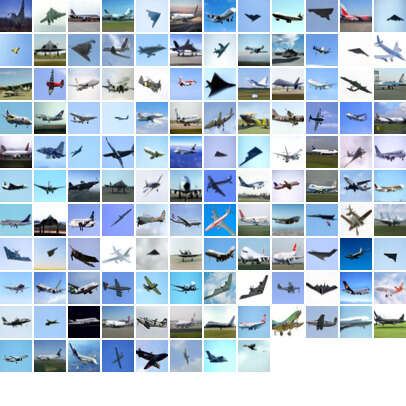}}
    \caption{$w=2$}
\end{subfigure}
\hfill
\begin{subfigure}{0.22\textwidth}
    \adjustbox{width=\linewidth, trim={.0\width} {.5\height} {0.5\width} {0\height},clip}{\includegraphics[width=\linewidth]{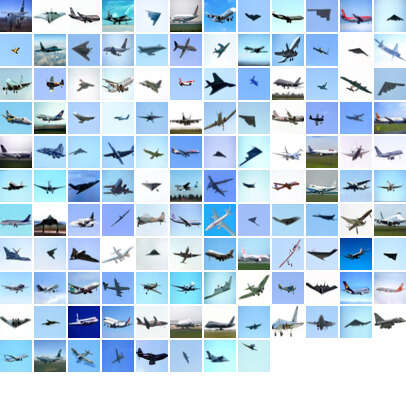}}
    \caption{$w=4$}
\end{subfigure}        
\caption{Ours (stochastic in pixel-space) on CIFAR-10. Distilled 4 sampling steps. Class-conditioned samples.}
\label{fig:figures}
\end{figure*}
 
 \begin{figure*}[!hb]
\centering
\begin{subfigure}{0.22\textwidth}
    \adjustbox{width=\linewidth, trim={.0\width} {.5\height} {0.5\width} {0\height},clip}{\includegraphics[width=\linewidth]{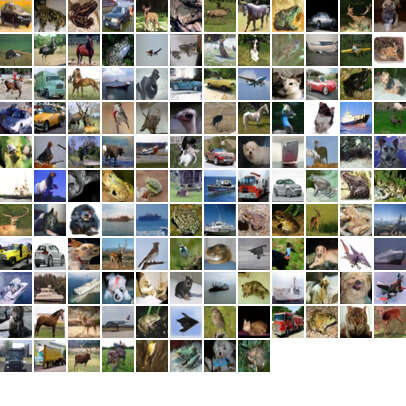}}
    \caption{$w=0$}
\end{subfigure}
\hfill
\begin{subfigure}{0.22\textwidth}
    \adjustbox{width=\linewidth, trim={.0\width} {.5\height} {0.5\width} {0\height},clip}{\includegraphics[width=\linewidth]{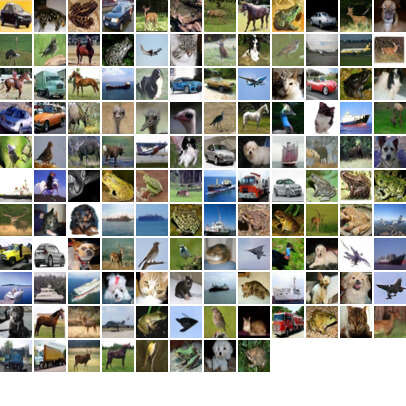}}
    \caption{$w=1$}
\end{subfigure}
\hfill
\begin{subfigure}{0.22\textwidth}
     \adjustbox{width=\linewidth, trim={.0\width} {.5\height} {0.5\width} {0\height},clip}{\includegraphics[width=\linewidth]{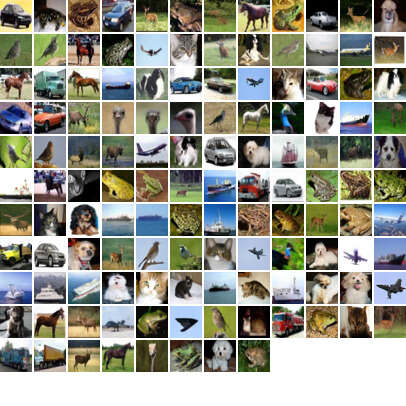}}
    \caption{$w=2$}
\end{subfigure}
\hfill
\begin{subfigure}{0.22\textwidth}
    \adjustbox{width=\linewidth, trim={.0\width} {.5\height} {0.5\width} {0\height},clip}{\includegraphics[width=\linewidth]{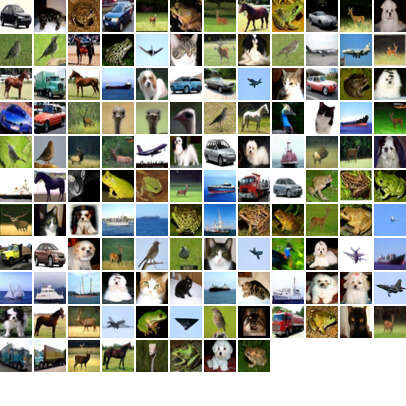}}
    \caption{$w=4$}
\end{subfigure}        
\caption{Ours (deterministic in pixel-space) on CIFAR-10. Distilled 2 sampling steps.}
\label{fig:figures}
\end{figure*}
 
 \begin{figure*}[!hb]
\centering
\begin{subfigure}{0.22\textwidth}
    \adjustbox{width=\linewidth, trim={.0\width} {.5\height} {0.5\width} {0\height},clip}{\includegraphics[width=\linewidth]{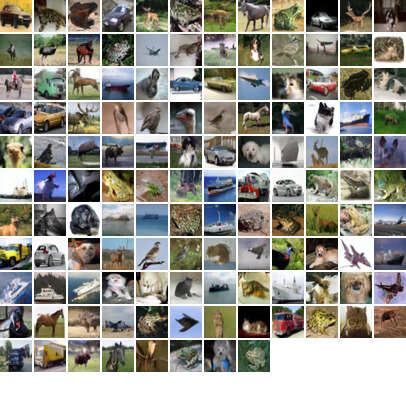}}
    \caption{$w=0$}
\end{subfigure}
\hfill
\begin{subfigure}{0.22\textwidth}
    \adjustbox{width=\linewidth, trim={.0\width} {.5\height} {0.5\width} {0\height},clip}{\includegraphics[width=\linewidth]{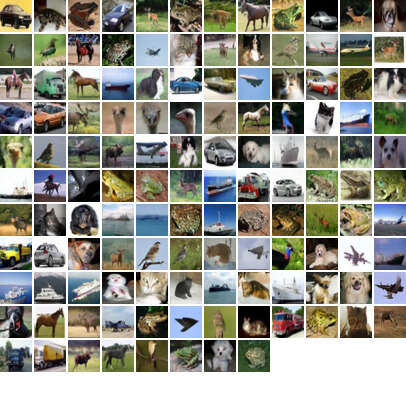}}
    \caption{$w=1$}
\end{subfigure}
\hfill
\begin{subfigure}{0.22\textwidth}
     \adjustbox{width=\linewidth, trim={.0\width} {.5\height} {0.5\width} {0\height},clip}{\includegraphics[width=\linewidth]{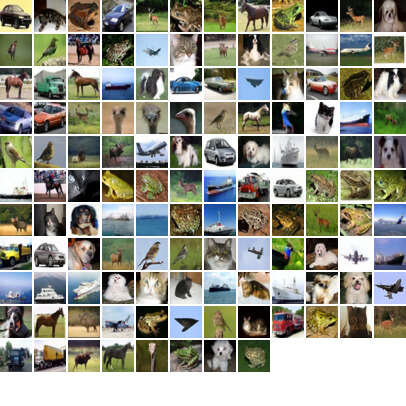}}
    \caption{$w=2$}
\end{subfigure}
\hfill
\begin{subfigure}{0.22\textwidth}
    \adjustbox{width=\linewidth, trim={.0\width} {.5\height} {0.5\width} {0\height},clip}{\includegraphics[width=\linewidth]{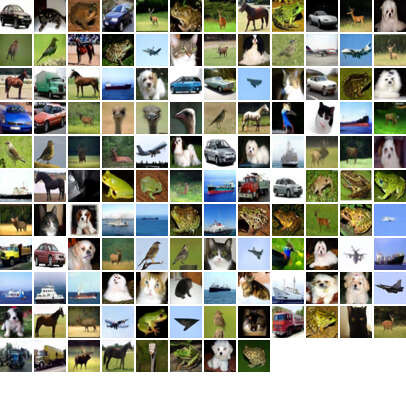}}
    \caption{$w=4$}
\end{subfigure}        
\caption{Ours (stochastic in pixel-space) on CIFAR-10. Distilled 2 sampling steps.}
\label{fig:figures}
\end{figure*}

\begin{figure*}[!hb]
\centering
\begin{subfigure}{0.22\textwidth}
    \adjustbox{width=\linewidth, trim={.0\width} {.5\height} {0.5\width} {0\height},clip}{\includegraphics[width=\linewidth]{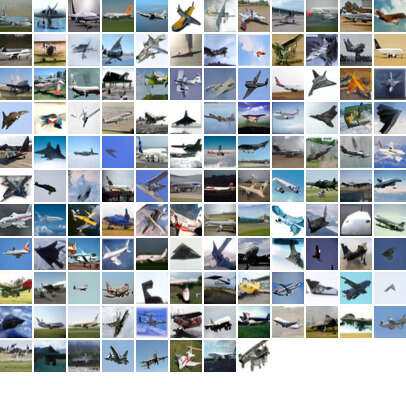}}
    \caption{$w=0$}
\end{subfigure}
\hfill
\begin{subfigure}{0.22\textwidth}
    \adjustbox{width=\linewidth, trim={.0\width} {.5\height} {0.5\width} {0\height},clip}{\includegraphics[width=\linewidth]{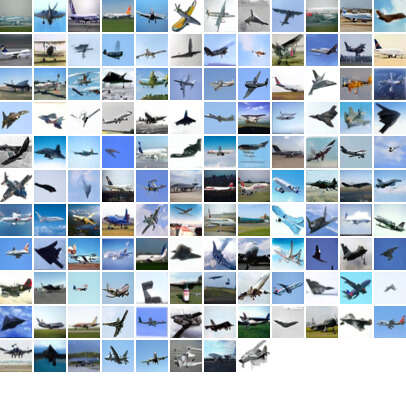}}
    \caption{$w=1$}
\end{subfigure}
\hfill
\begin{subfigure}{0.22\textwidth}
     \adjustbox{width=\linewidth, trim={.0\width} {.5\height} {0.5\width} {0\height},clip}{\includegraphics[width=\linewidth]{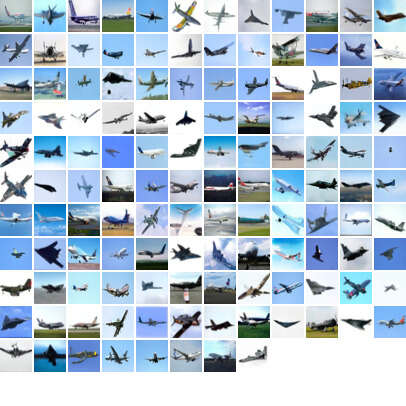}}
    \caption{$w=2$}
\end{subfigure}
\hfill
\begin{subfigure}{0.22\textwidth}
    \adjustbox{width=\linewidth, trim={.0\width} {.5\height} {0.5\width} {0\height},clip}{\includegraphics[width=\linewidth]{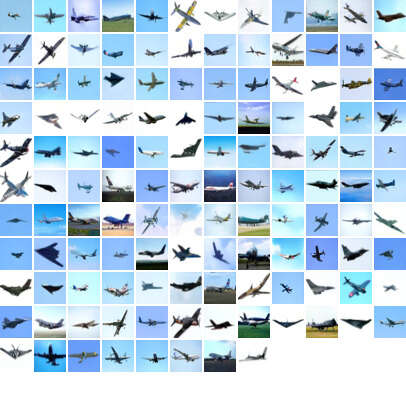}}
    \caption{$w=4$}
\end{subfigure}        
\caption{Ours (deterministic in pixel-space) on CIFAR-10. Distilled 2 sampling steps. Class-conditioned samples.}
\label{fig:figures}
\end{figure*}

\begin{figure*}[!hb]
\centering
\begin{subfigure}{0.22\textwidth}
    \adjustbox{width=\linewidth, trim={.0\width} {.5\height} {0.5\width} {0\height},clip}{\includegraphics[width=\linewidth]{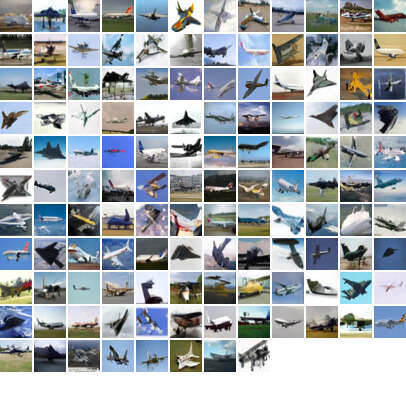}}
    \caption{$w=0$}
\end{subfigure}
\hfill
\begin{subfigure}{0.22\textwidth}
    \adjustbox{width=\linewidth, trim={.0\width} {.5\height} {0.5\width} {0\height},clip}{\includegraphics[width=\linewidth]{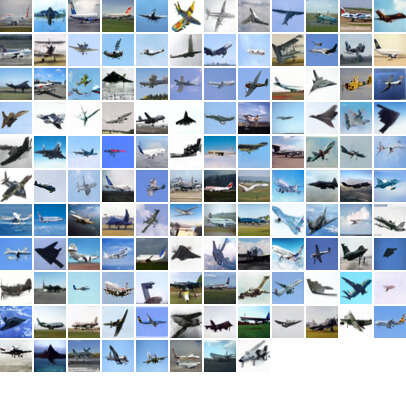}}
    \caption{$w=1$}
\end{subfigure}
\hfill
\begin{subfigure}{0.22\textwidth}
     \adjustbox{width=\linewidth, trim={.0\width} {.5\height} {0.5\width} {0\height},clip}{\includegraphics[width=\linewidth]{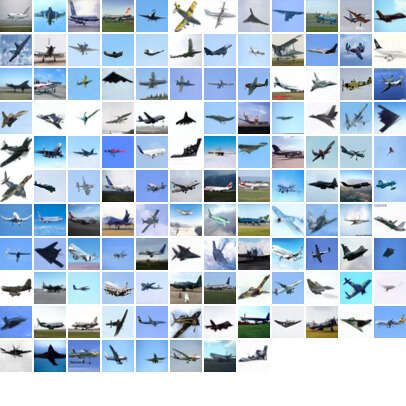}}
    \caption{$w=2$}
\end{subfigure}
\hfill
\begin{subfigure}{0.22\textwidth}
    \adjustbox{width=\linewidth, trim={.0\width} {.5\height} {0.5\width} {0\height},clip}{\includegraphics[width=\linewidth]{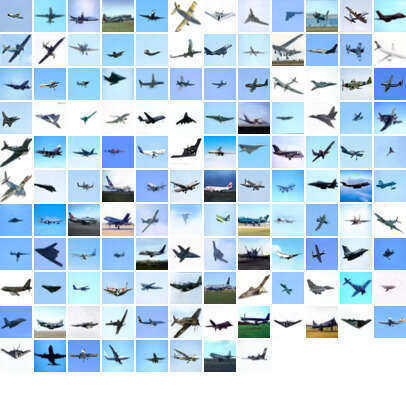}}
    \caption{$w=4$}
\end{subfigure}        
\caption{Ours (stochastic in pixel-space) on CIFAR-10. Distilled 2 sampling steps. Class-conditioned samples.}
\label{fig:figures}
\end{figure*}
 
 \begin{figure*}[!hb]
\centering
\begin{subfigure}{0.22\textwidth}
    \adjustbox{width=\linewidth, trim={.0\width} {.5\height} {0.5\width} {0\height},clip}{\includegraphics[width=\linewidth]{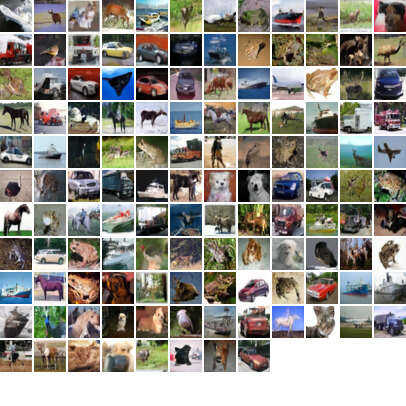}}
    \caption{$w=0$}
\end{subfigure}
\hfill
\begin{subfigure}{0.22\textwidth}
    \adjustbox{width=\linewidth, trim={.0\width} {.5\height} {0.5\width} {0\height},clip}{\includegraphics[width=\linewidth]{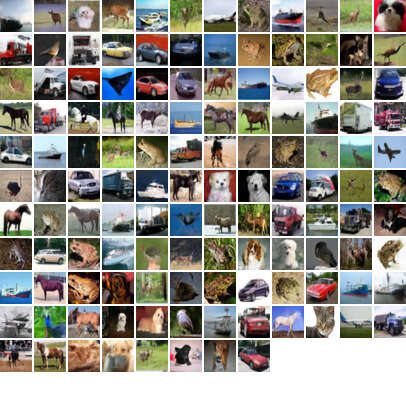}}
    \caption{$w=1$}
\end{subfigure}
\hfill
\begin{subfigure}{0.22\textwidth}
     \adjustbox{width=\linewidth, trim={.0\width} {.5\height} {0.5\width} {0\height},clip}{\includegraphics[width=\linewidth]{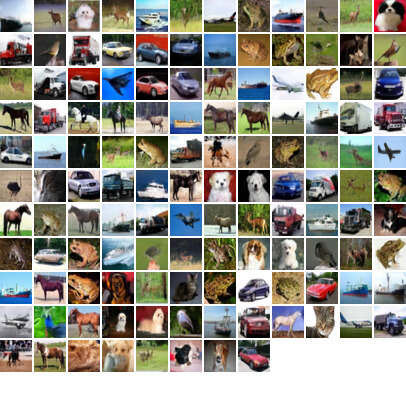}}
    \caption{$w=2$}
\end{subfigure}
\hfill
\begin{subfigure}{0.22\textwidth}
    \adjustbox{width=\linewidth, trim={.0\width} {.5\height} {0.5\width} {0\height},clip}{\includegraphics[width=\linewidth]{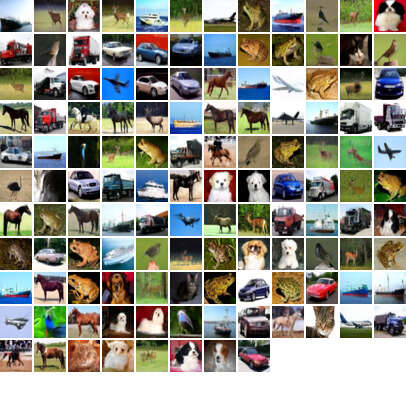}}
    \caption{$w=4$}
\end{subfigure}        
\caption{Ours (deterministic in pixel-space) on CIFAR-10. Distilled 1 sampling step.}
\label{fig:figures}
\end{figure*}
 
 \begin{figure*}[!hb]
\centering
\begin{subfigure}{0.22\textwidth}
    \adjustbox{width=\linewidth, trim={.0\width} {.5\height} {0.5\width} {0\height},clip}{\includegraphics[width=\linewidth]{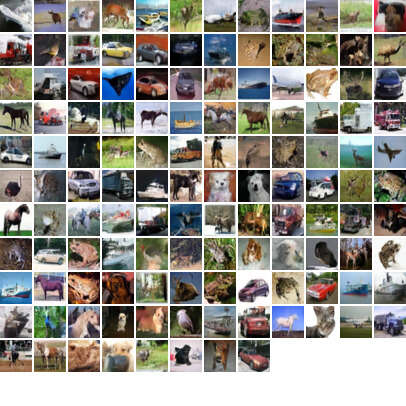}}
    \caption{$w=0$}
\end{subfigure}
\hfill
\begin{subfigure}{0.22\textwidth}
    \adjustbox{width=\linewidth, trim={.0\width} {.5\height} {0.5\width} {0\height},clip}{\includegraphics[width=\linewidth]{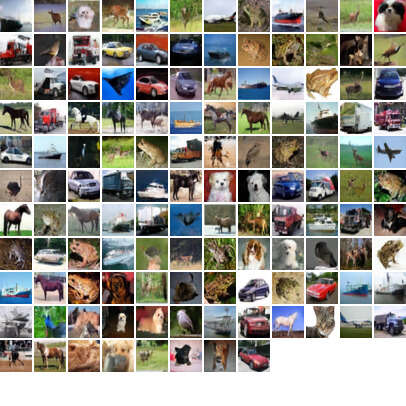}}
    \caption{$w=1$}
\end{subfigure}
\hfill
\begin{subfigure}{0.22\textwidth}
     \adjustbox{width=\linewidth, trim={.0\width} {.5\height} {0.5\width} {0\height},clip}{\includegraphics[width=\linewidth]{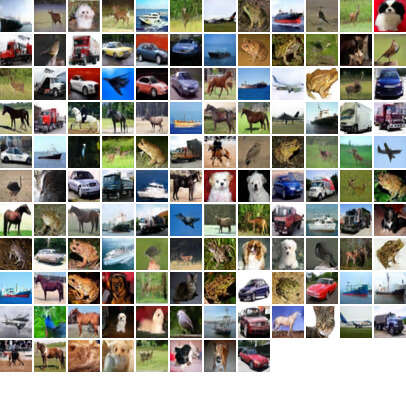}}
    \caption{$w=2$}
\end{subfigure}
\hfill
\begin{subfigure}{0.22\textwidth}
    \adjustbox{width=\linewidth, trim={.0\width} {.5\height} {0.5\width} {0\height},clip}{\includegraphics[width=\linewidth]{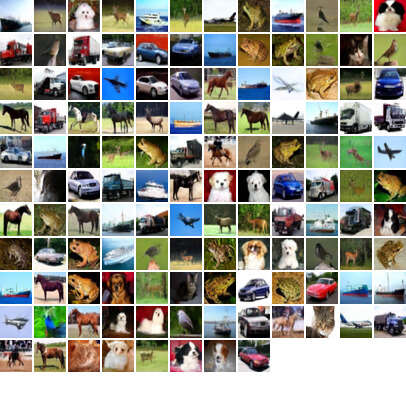}}
    \caption{$w=4$}
\end{subfigure}        
\caption{Ours (stochastic in pixel-space) on CIFAR-10. Distilled 1 sampling step.}
\label{fig:figures}
\end{figure*}

\begin{figure*}[!hb]
\centering
\begin{subfigure}{0.22\textwidth}
    \adjustbox{width=\linewidth, trim={.0\width} {.5\height} {0.5\width} {0\height},clip}{\includegraphics[width=\linewidth]{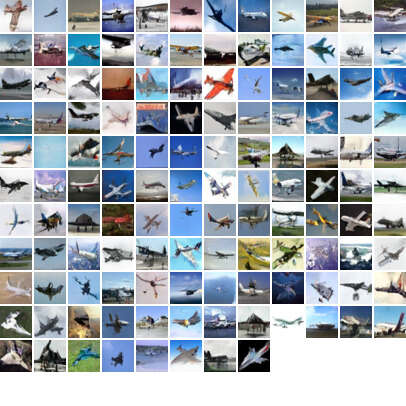}}
    \caption{$w=0$}
\end{subfigure}
\hfill
\begin{subfigure}{0.22\textwidth}
    \adjustbox{width=\linewidth, trim={.0\width} {.5\height} {0.5\width} {0\height},clip}{\includegraphics[width=\linewidth]{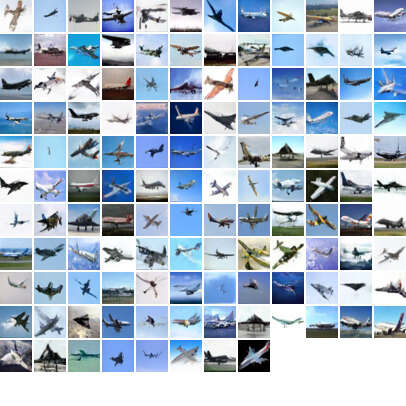}}
    \caption{$w=1$}
\end{subfigure}
\hfill
\begin{subfigure}{0.22\textwidth}
     \adjustbox{width=\linewidth, trim={.0\width} {.5\height} {0.5\width} {0\height},clip}{\includegraphics[width=\linewidth]{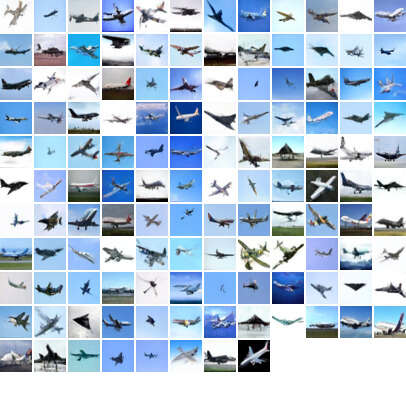}}
    \caption{$w=2$}
\end{subfigure}
\hfill
\begin{subfigure}{0.22\textwidth}
    \adjustbox{width=\linewidth, trim={.0\width} {.5\height} {0.5\width} {0\height},clip}{\includegraphics[width=\linewidth]{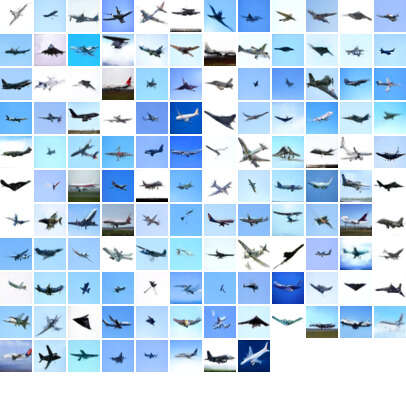}}
    \caption{$w=4$}
\end{subfigure}        
\caption{Ours (deterministic in pixel-space) on CIFAR-10. Distilled 1 sampling step. Class-conditioned samples.}
\label{fig:figures}
\end{figure*}

\begin{figure*}[!hb]
\centering
\begin{subfigure}{0.22\textwidth}
    \adjustbox{width=\linewidth, trim={.0\width} {.5\height} {0.5\width} {0\height},clip}{\includegraphics[width=\linewidth]{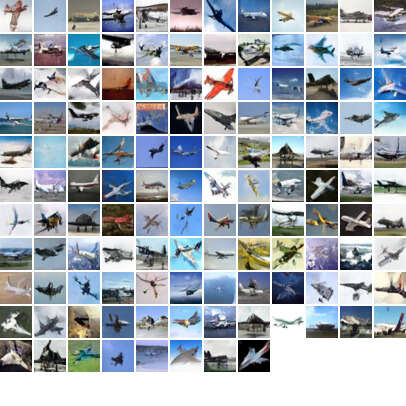}}
    \caption{$w=0$}
\end{subfigure}
\hfill
\begin{subfigure}{0.22\textwidth}
    \adjustbox{width=\linewidth, trim={.0\width} {.5\height} {0.5\width} {0\height},clip}{\includegraphics[width=\linewidth]{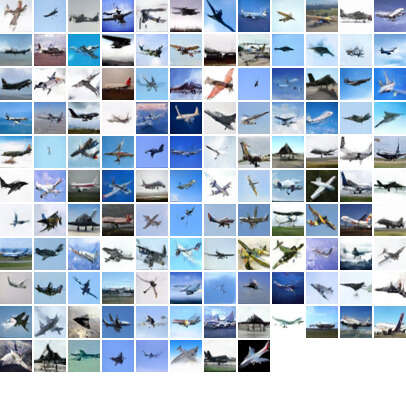}}
    \caption{$w=1$}
\end{subfigure}
\hfill
\begin{subfigure}{0.22\textwidth}
     \adjustbox{width=\linewidth, trim={.0\width} {.5\height} {0.5\width} {0\height},clip}{\includegraphics[width=\linewidth]{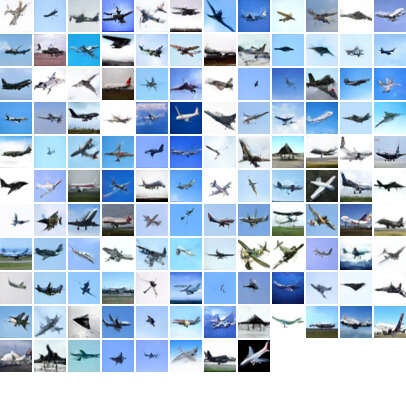}}
    \caption{$w=2$}
\end{subfigure}
\hfill
\begin{subfigure}{0.22\textwidth}
    \adjustbox{width=\linewidth, trim={.0\width} {.5\height} {0.5\width} {0\height},clip}{\includegraphics[width=\linewidth]{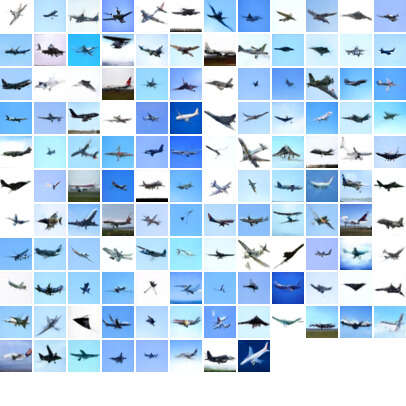}}
    \caption{$w=4$}
\end{subfigure}        
\caption{Ours (stochastic in pixel-space) on CIFAR-10. Distilled 1 sampling step. Class-conditioned samples.}
\label{fig:figures}
\end{figure*}

\begin{figure*}[!hb]
\centering
\begin{subfigure}{0.22\textwidth}
    \adjustbox{width=\linewidth, trim={.0\width} {.5\height} {0.5\width} {0\height},clip}{\includegraphics[width=\linewidth]{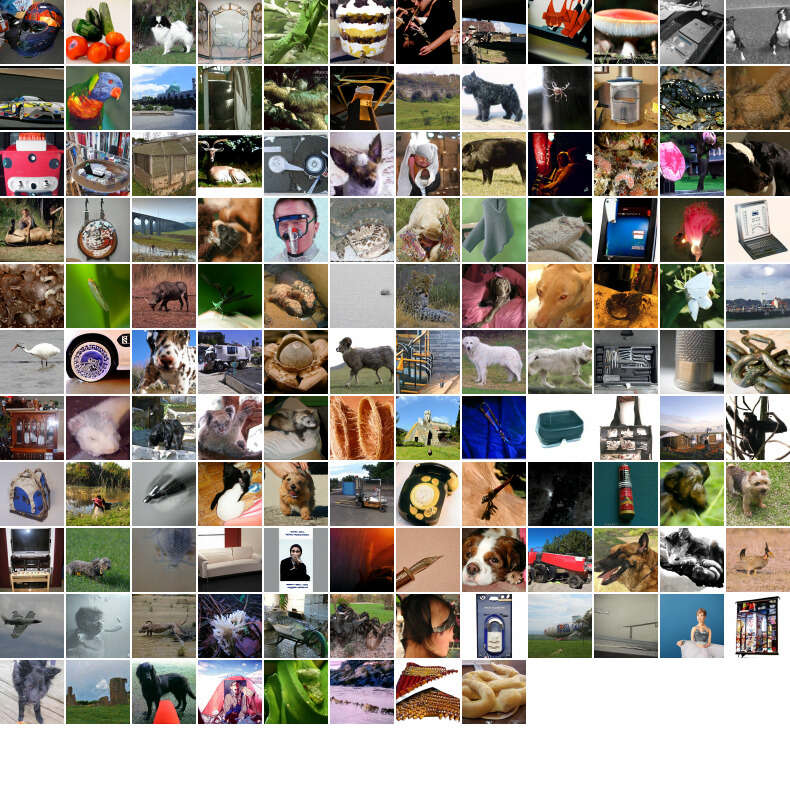}}
    \caption{$w=0$}
\end{subfigure}
\hfill
\begin{subfigure}{0.22\textwidth}
    \adjustbox{width=\linewidth, trim={.0\width} {.5\height} {0.5\width} {0\height},clip}{\includegraphics[width=\linewidth]{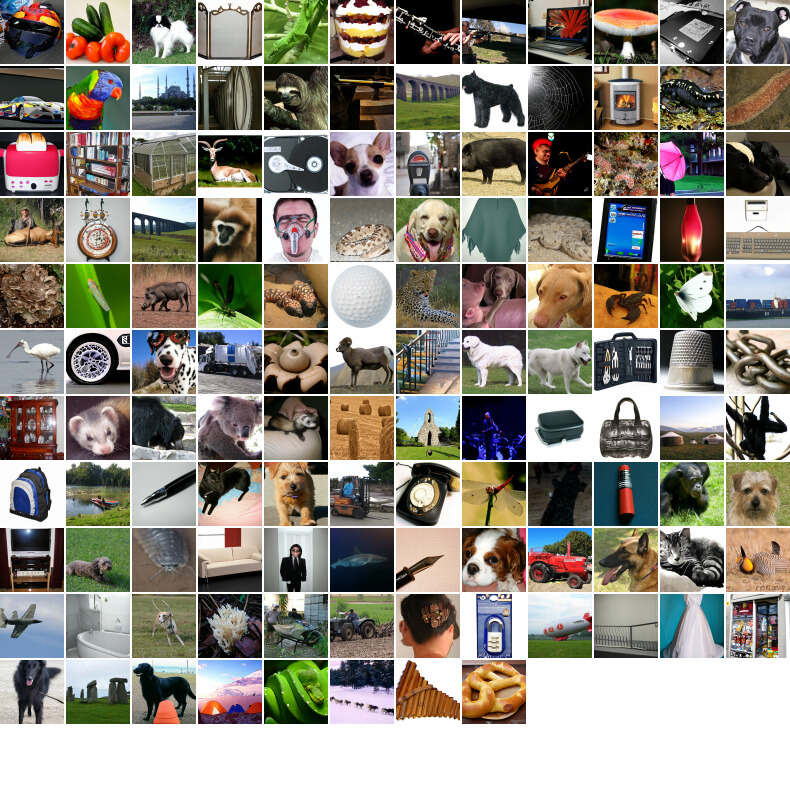}}
    \caption{$w=1$}
\end{subfigure}
\hfill
\begin{subfigure}{0.22\textwidth}
     \adjustbox{width=\linewidth, trim={.0\width} {.5\height} {0.5\width} {0\height},clip}{\includegraphics[width=\linewidth]{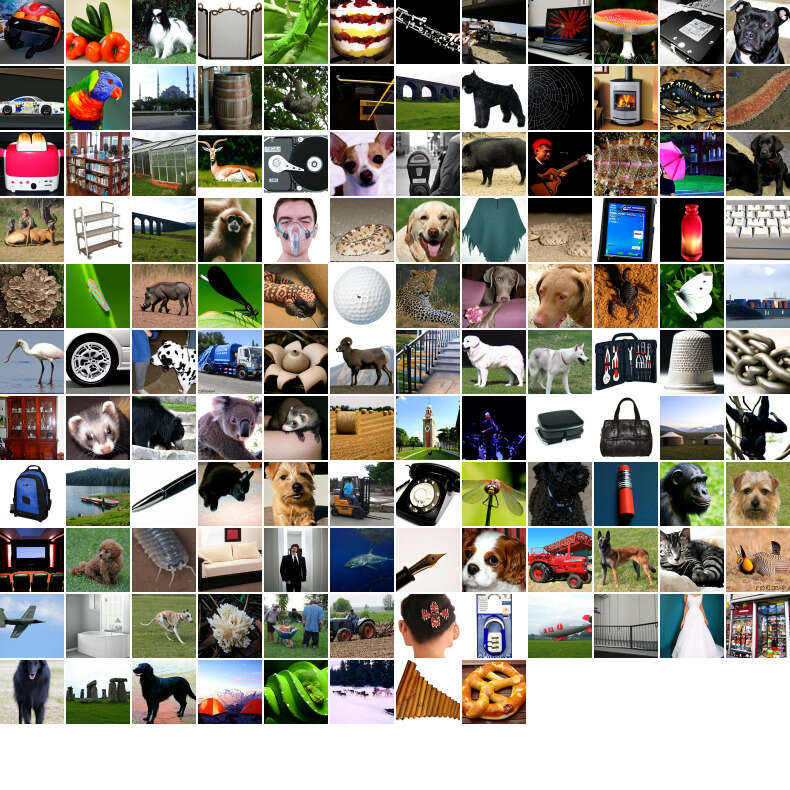}}
    \caption{$w=2$}
\end{subfigure}
\hfill
\begin{subfigure}{0.22\textwidth}
    \adjustbox{width=\linewidth, trim={.0\width} {.5\height} {0.5\width} {0\height},clip}{\includegraphics[width=\linewidth]{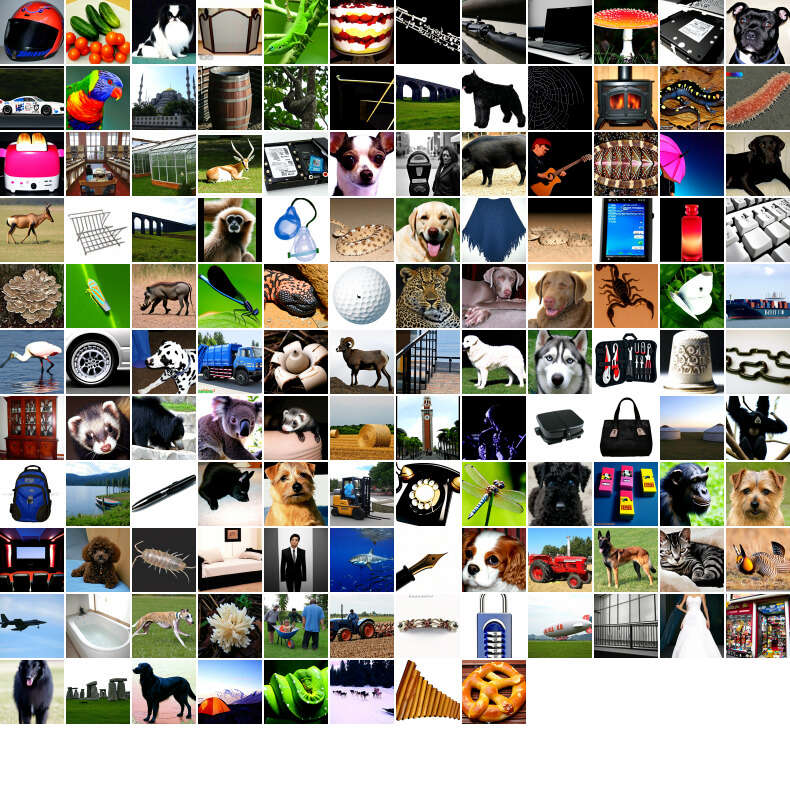}}
    \caption{$w=4$}
\end{subfigure}        
\caption{Ours (deterministic in pixel-space) on ImageNet 64x64. Distilled 256 sampling steps.}
\label{fig:figures}
\end{figure*}

\begin{figure*}[!hb]
\centering
\begin{subfigure}{0.22\textwidth}
    \adjustbox{width=\linewidth, trim={.0\width} {.5\height} {0.5\width} {0\height},clip}{\includegraphics[width=\linewidth]{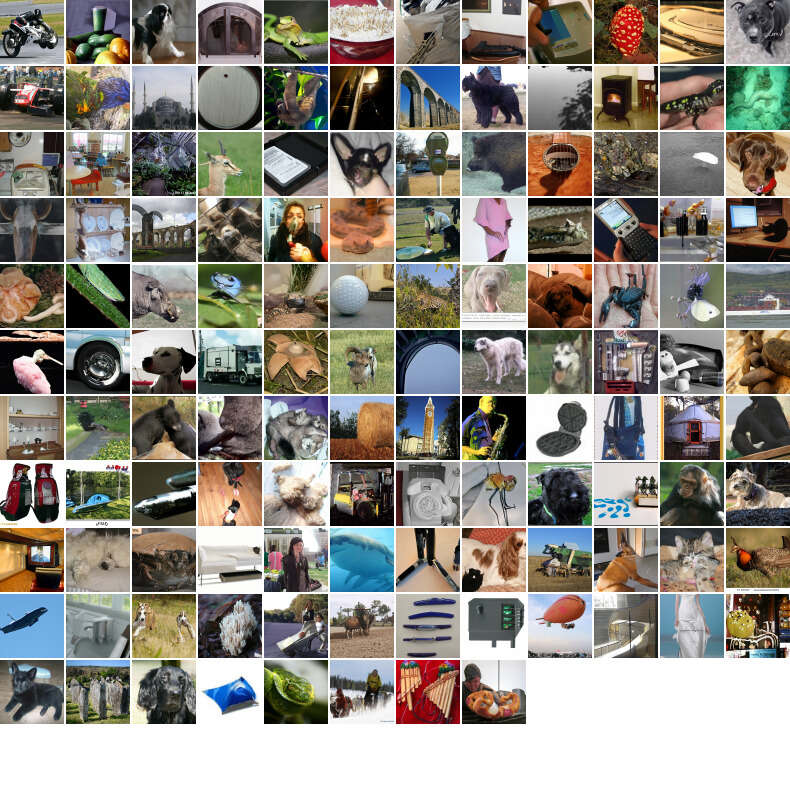}}
    \caption{$w=0$}
\end{subfigure}
\hfill
\begin{subfigure}{0.22\textwidth}
    \adjustbox{width=\linewidth, trim={.0\width} {.5\height} {0.5\width} {0\height},clip}{\includegraphics[width=\linewidth]{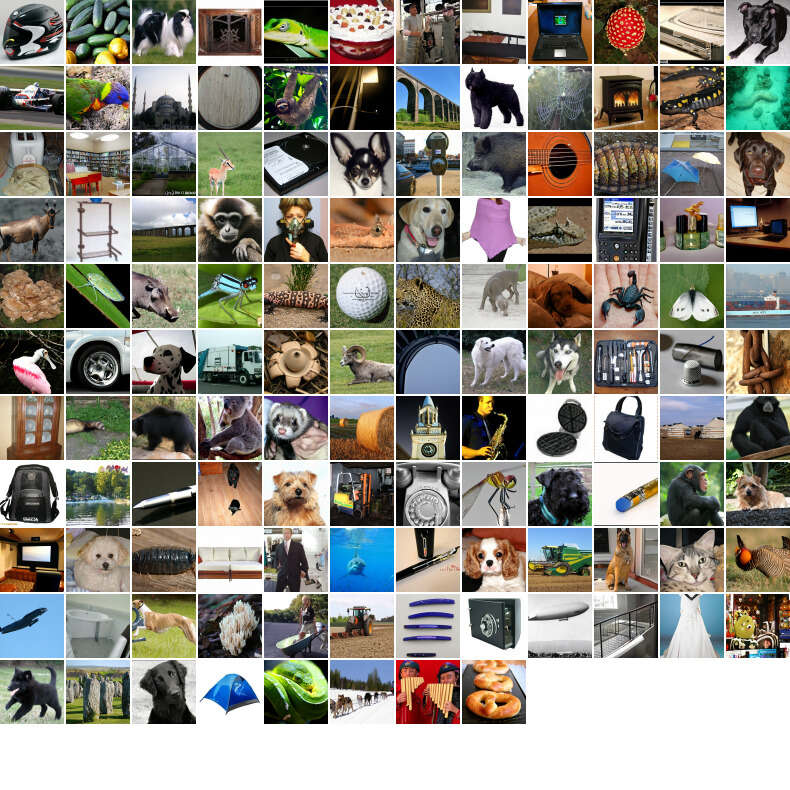}}
    \caption{$w=1$}
\end{subfigure}
\hfill
\begin{subfigure}{0.22\textwidth}
     \adjustbox{width=\linewidth, trim={.0\width} {.5\height} {0.5\width} {0\height},clip}{\includegraphics[width=\linewidth]{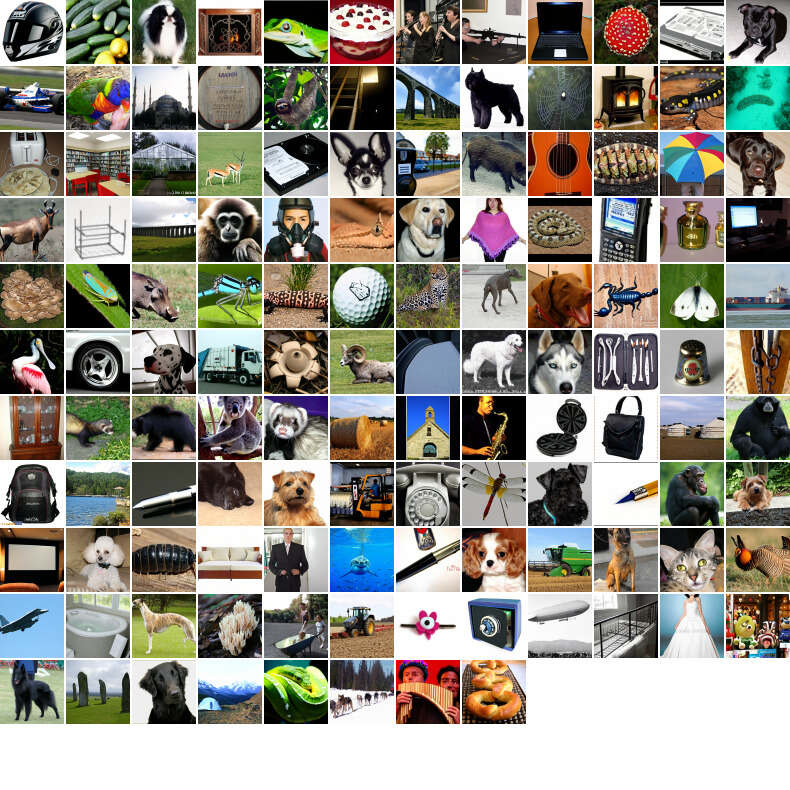}}
    \caption{$w=2$}
\end{subfigure}
\hfill
\begin{subfigure}{0.22\textwidth}
    \adjustbox{width=\linewidth, trim={.0\width} {.5\height} {0.5\width} {0\height},clip}{\includegraphics[width=\linewidth]{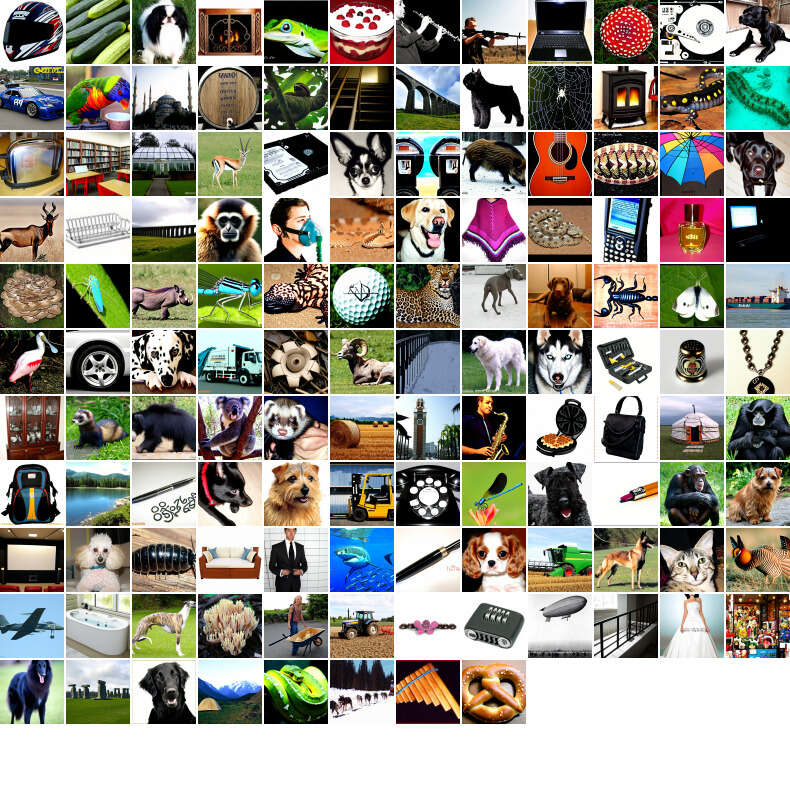}}
    \caption{$w=4$}
\end{subfigure}        
\caption{Ours (stochastic in pixel-space) on ImageNet 64x64. Distilled 256 sampling steps.}
\label{fig:figures}
\end{figure*}

\begin{figure*}[!hb]
\centering
\begin{subfigure}{0.22\textwidth}
    \adjustbox{width=\linewidth, trim={.0\width} {.5\height} {0.5\width} {0\height},clip}{\includegraphics[width=\linewidth]{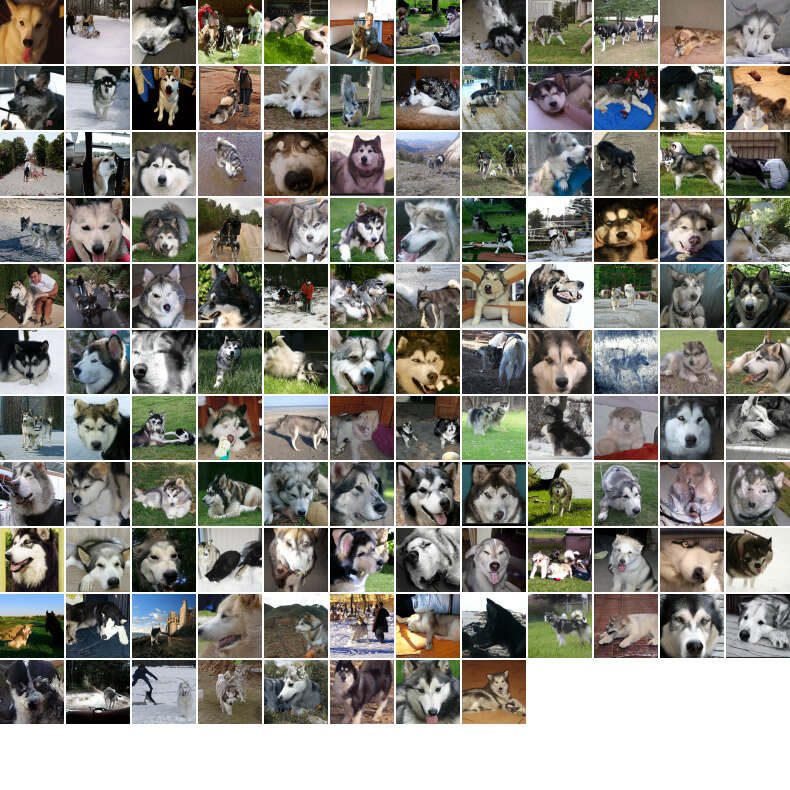}}
    \caption{$w=0$}
\end{subfigure}
\hfill
\begin{subfigure}{0.22\textwidth}
    \adjustbox{width=\linewidth, trim={.0\width} {.5\height} {0.5\width} {0\height},clip}{\includegraphics[width=\linewidth]{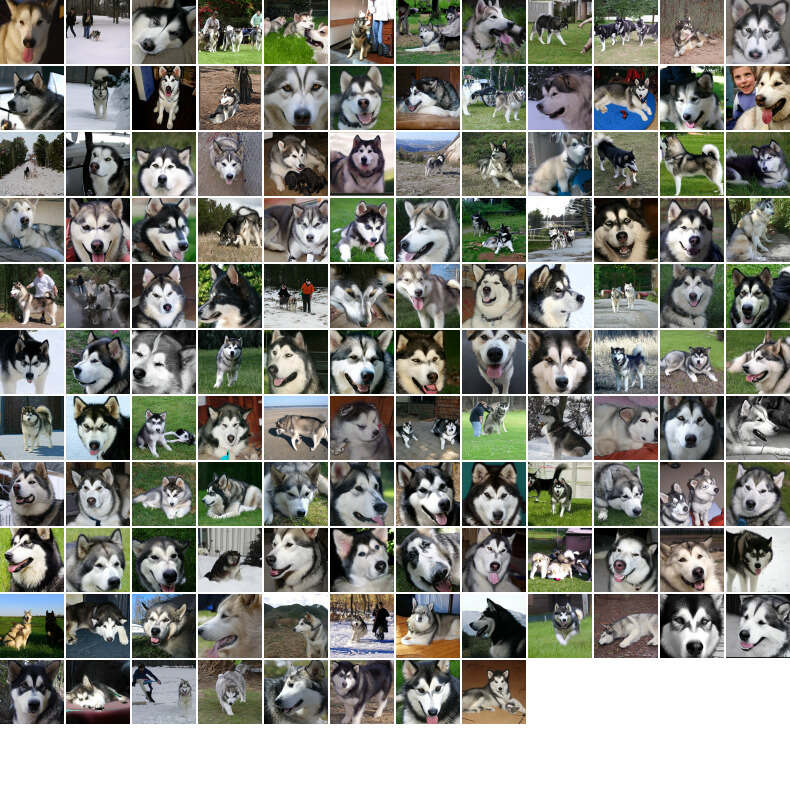}}
    \caption{$w=1$}
\end{subfigure}
\hfill
\begin{subfigure}{0.22\textwidth}
     \adjustbox{width=\linewidth, trim={.0\width} {.5\height} {0.5\width} {0\height},clip}{\includegraphics[width=\linewidth]{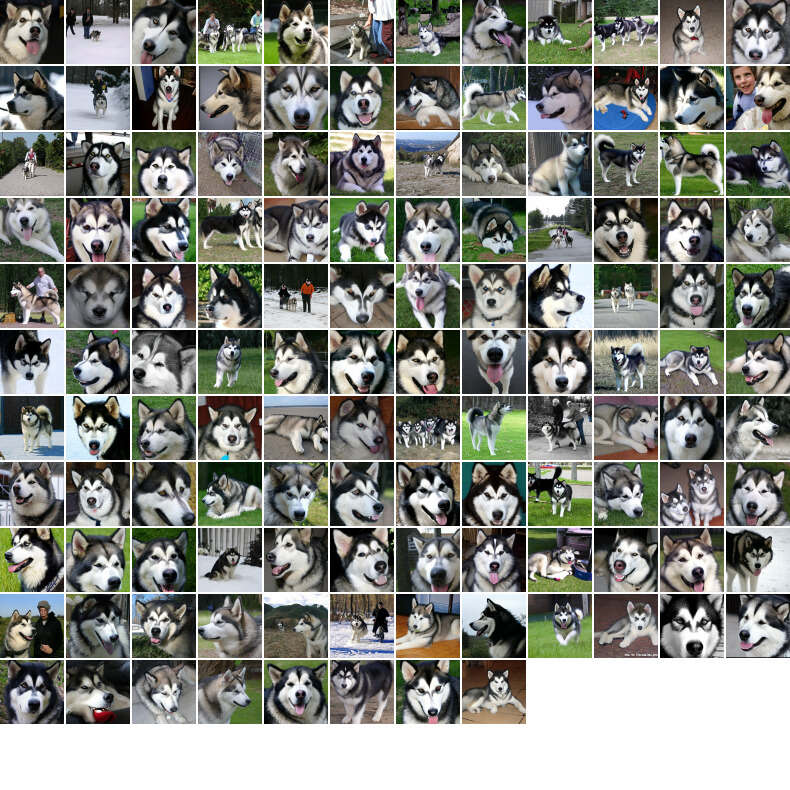}}
    \caption{$w=2$}
\end{subfigure}
\hfill
\begin{subfigure}{0.22\textwidth}
    \adjustbox{width=\linewidth, trim={.0\width} {.5\height} {0.5\width} {0\height},clip}{\includegraphics[width=\linewidth]{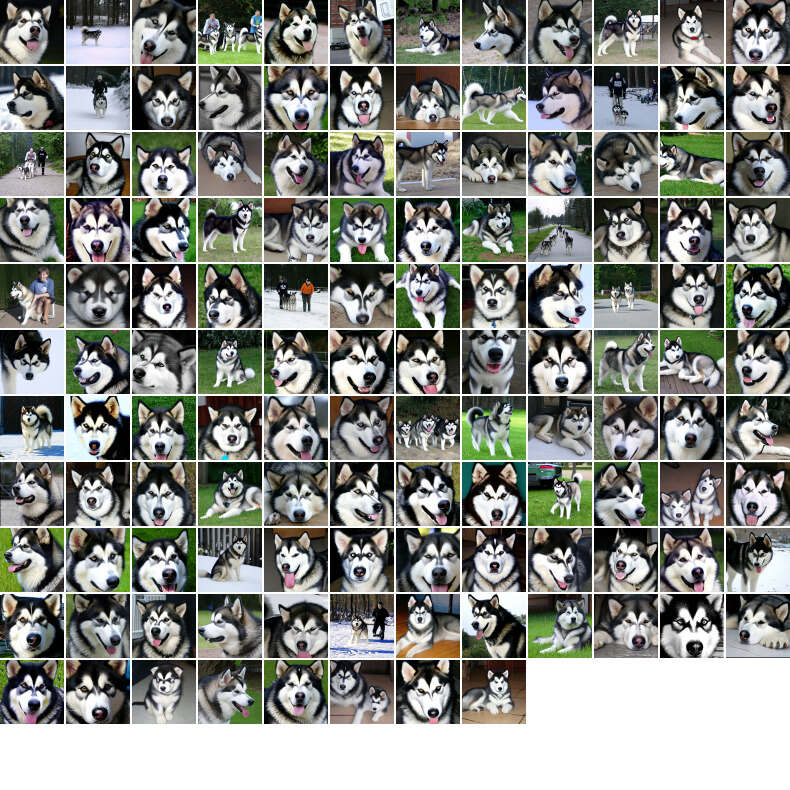}}
    \caption{$w=4$}
\end{subfigure}        
\caption{Ours (deterministic in pixel-space) on ImageNet 64x64. Distilled 256 sampling steps. Class-conditioned samples.}
\label{fig:figures}
\end{figure*}
 
\begin{figure*}[!hb]
\centering
\begin{subfigure}{0.22\textwidth}
    \adjustbox{width=\linewidth, trim={.0\width} {.5\height} {0.5\width} {0\height},clip}{\includegraphics[width=\linewidth]{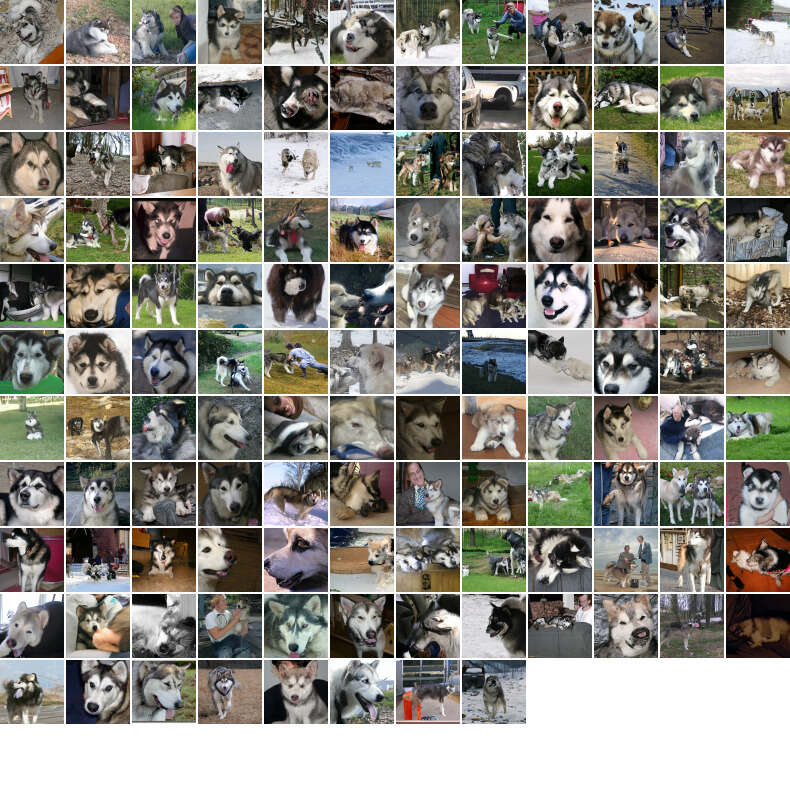}}
    \caption{$w=0$}
\end{subfigure}
\hfill
\begin{subfigure}{0.22\textwidth}
    \adjustbox{width=\linewidth, trim={.0\width} {.5\height} {0.5\width} {0\height},clip}{\includegraphics[width=\linewidth]{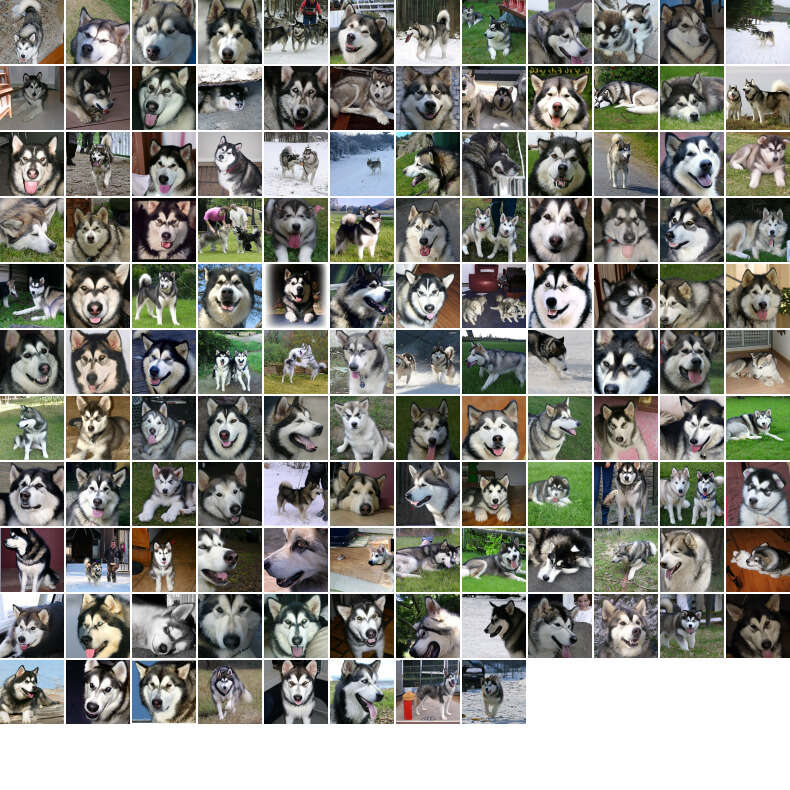}}
    \caption{$w=1$}
\end{subfigure}
\hfill
\begin{subfigure}{0.22\textwidth}
     \adjustbox{width=\linewidth, trim={.0\width} {.5\height} {0.5\width} {0\height},clip}{\includegraphics[width=\linewidth]{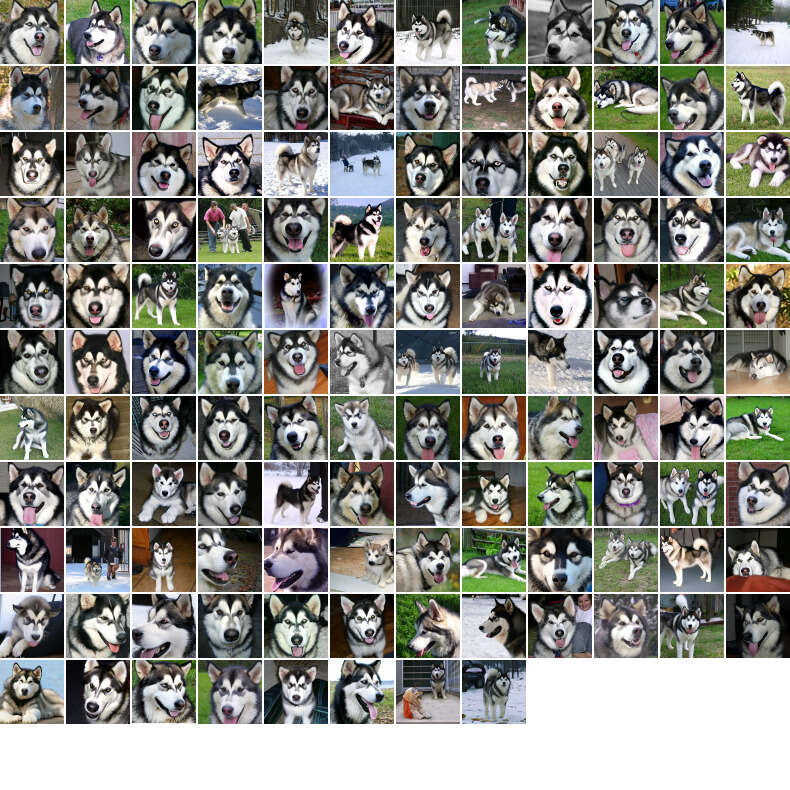}}
    \caption{$w=2$}
\end{subfigure}
\hfill
\begin{subfigure}{0.22\textwidth}
    \adjustbox{width=\linewidth, trim={.0\width} {.5\height} {0.5\width} {0\height},clip}{\includegraphics[width=\linewidth]{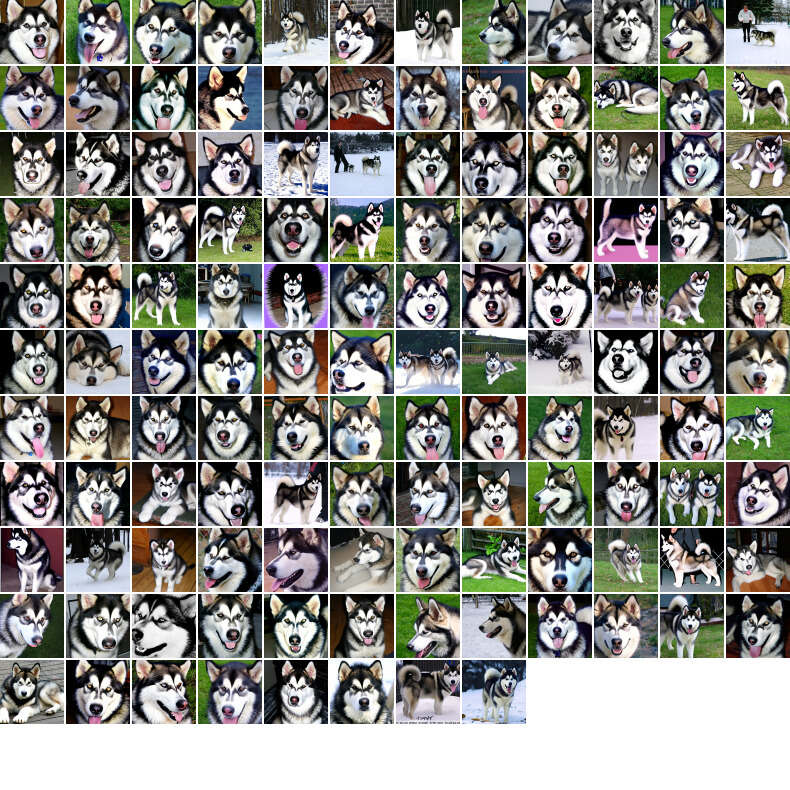}}
    \caption{$w=4$}
\end{subfigure}        
\caption{Ours (stochastic in pixel-space) on ImageNet 64x64. Distilled 256 sampling steps. Class-conditioned samples.}
\label{fig:figures}
\end{figure*}

\begin{figure*}[!hb]
\centering
\begin{subfigure}{0.22\textwidth}
    \adjustbox{width=\linewidth, trim={.0\width} {.5\height} {0.5\width} {0\height},clip}{\includegraphics[width=\linewidth]{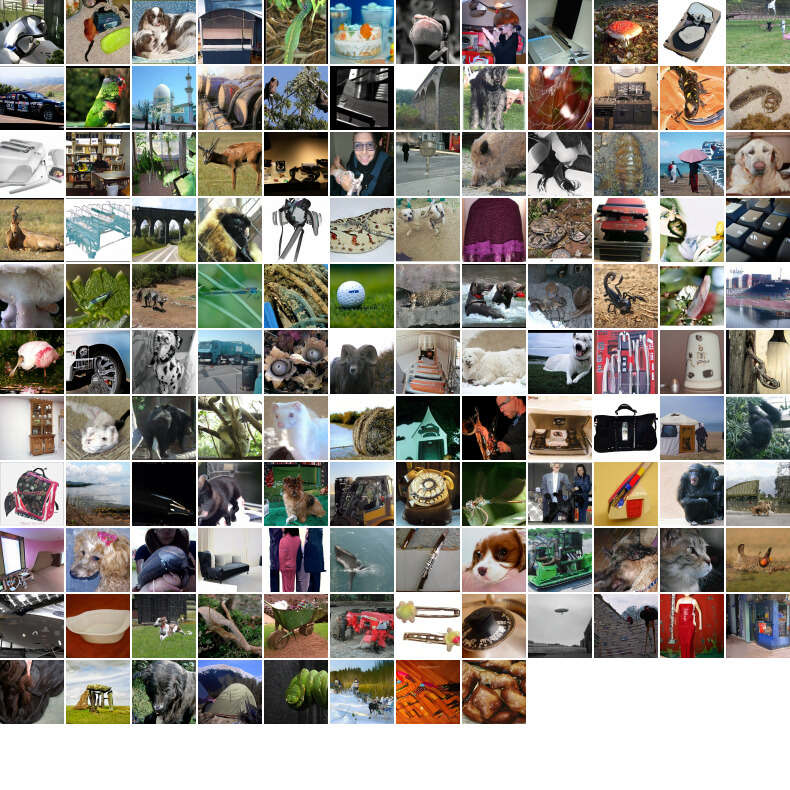}}
    \caption{$w=0$}
\end{subfigure}
\hfill
\begin{subfigure}{0.22\textwidth}
    \adjustbox{width=\linewidth, trim={.0\width} {.5\height} {0.5\width} {0\height},clip}{\includegraphics[width=\linewidth]{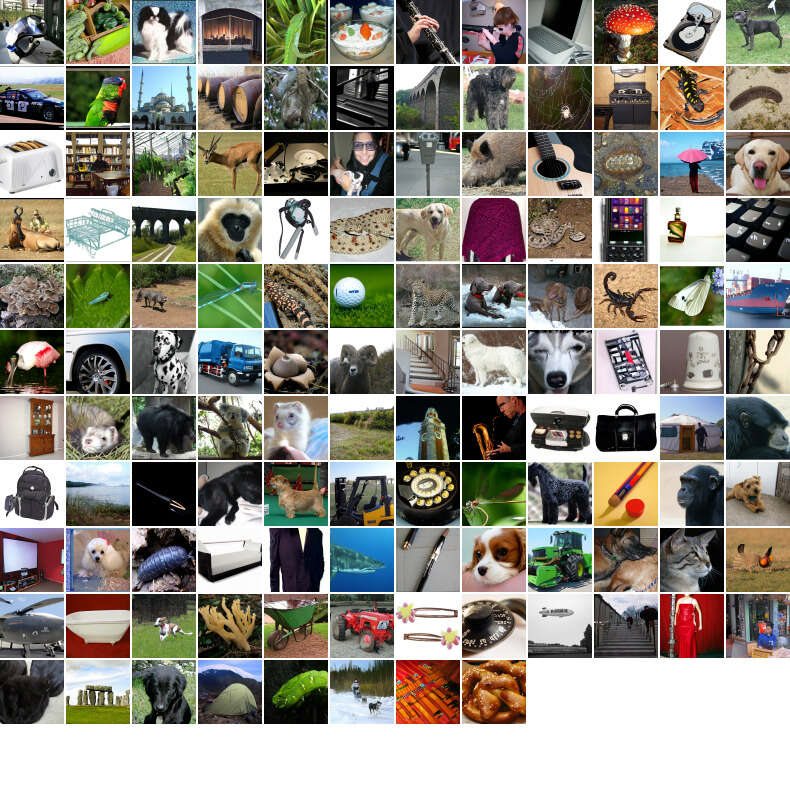}}
    \caption{$w=1$}
\end{subfigure}
\hfill
\begin{subfigure}{0.22\textwidth}
     \adjustbox{width=\linewidth, trim={.0\width} {.5\height} {0.5\width} {0\height},clip}{\includegraphics[width=\linewidth]{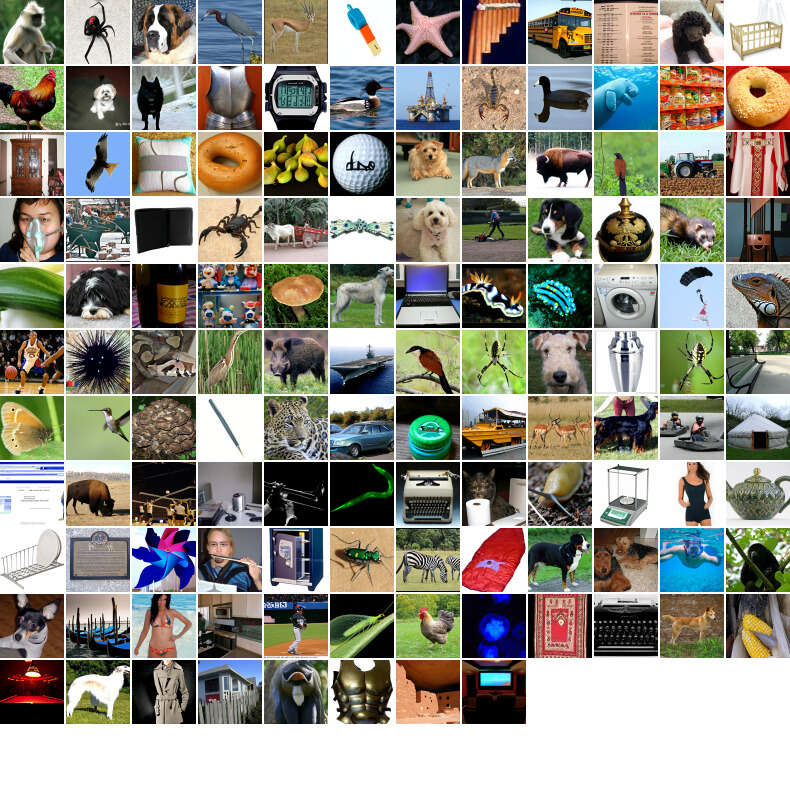}}
    \caption{$w=2$}
\end{subfigure}
\hfill
\begin{subfigure}{0.22\textwidth}
    \adjustbox{width=\linewidth, trim={.0\width} {.5\height} {0.5\width} {0\height},clip}{\includegraphics[width=\linewidth]{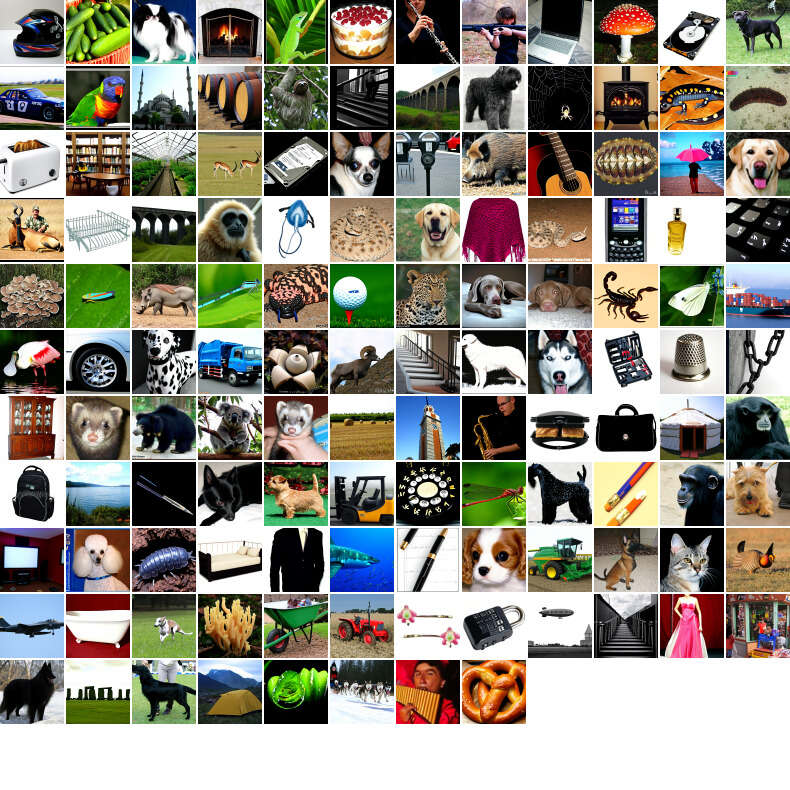}}
    \caption{$w=4$}
\end{subfigure}        
\caption{Ours (deterministic in pixel-space) on ImageNet 64x64. Distilled 8 sampling step. 
}
\label{fig:figures}
\end{figure*}

\begin{figure*}[!hb]
\centering
\begin{subfigure}{0.22\textwidth}
    \adjustbox{width=\linewidth, trim={.0\width} {.5\height} {0.5\width} {0\height},clip}{\includegraphics[width=\linewidth]{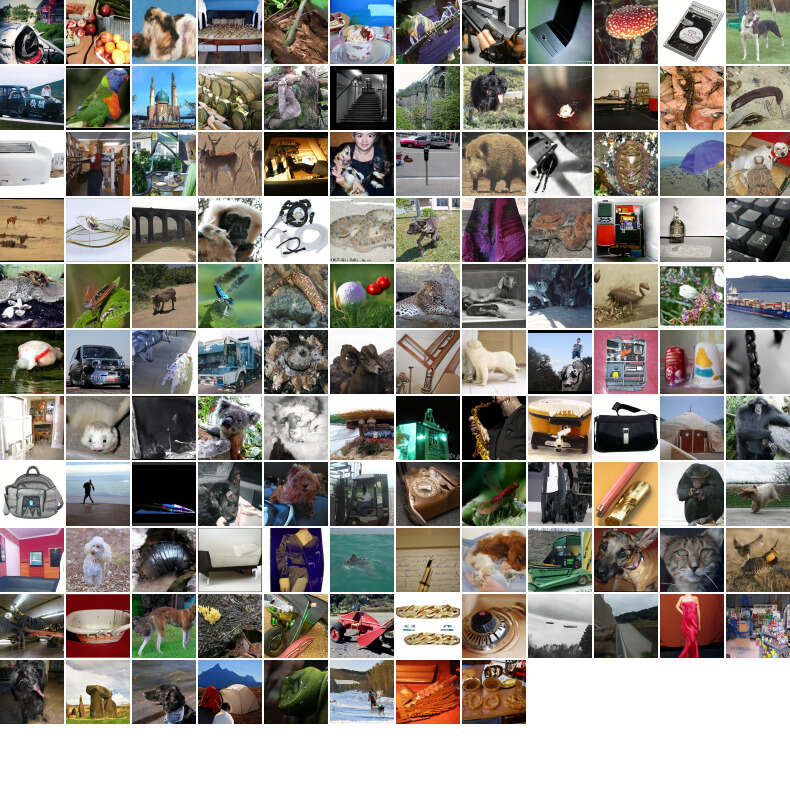}}
    \caption{$w=0$}
\end{subfigure}
\hfill
\begin{subfigure}{0.22\textwidth}
    \adjustbox{width=\linewidth, trim={.0\width} {.5\height} {0.5\width} {0\height},clip}{\includegraphics[width=\linewidth]{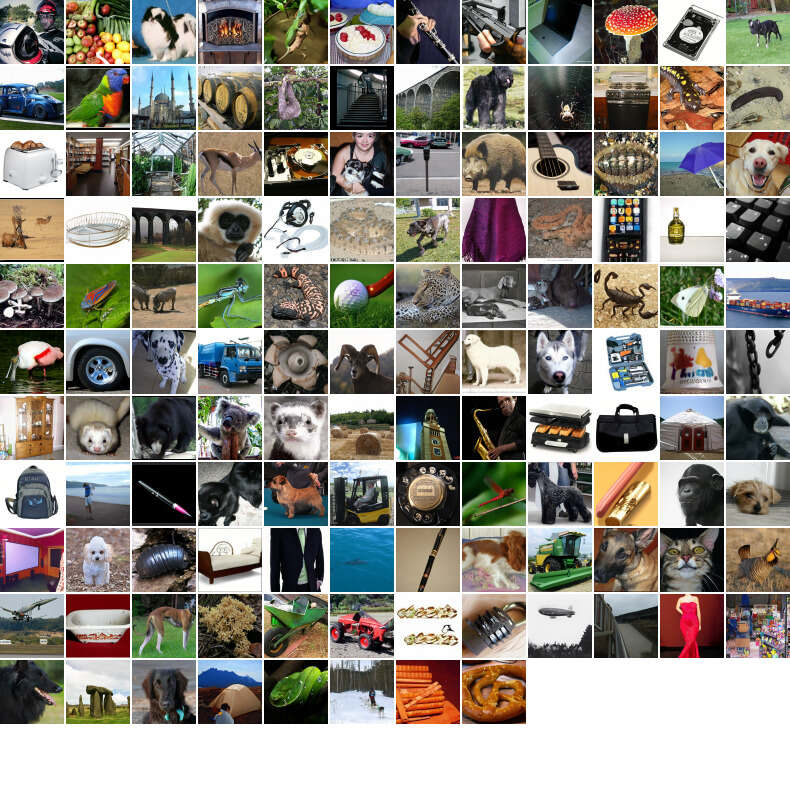}}
    \caption{$w=1$}
\end{subfigure}
\hfill
\begin{subfigure}{0.22\textwidth}
     \adjustbox{width=\linewidth, trim={.0\width} {.5\height} {0.5\width} {0\height},clip}{\includegraphics[width=\linewidth]{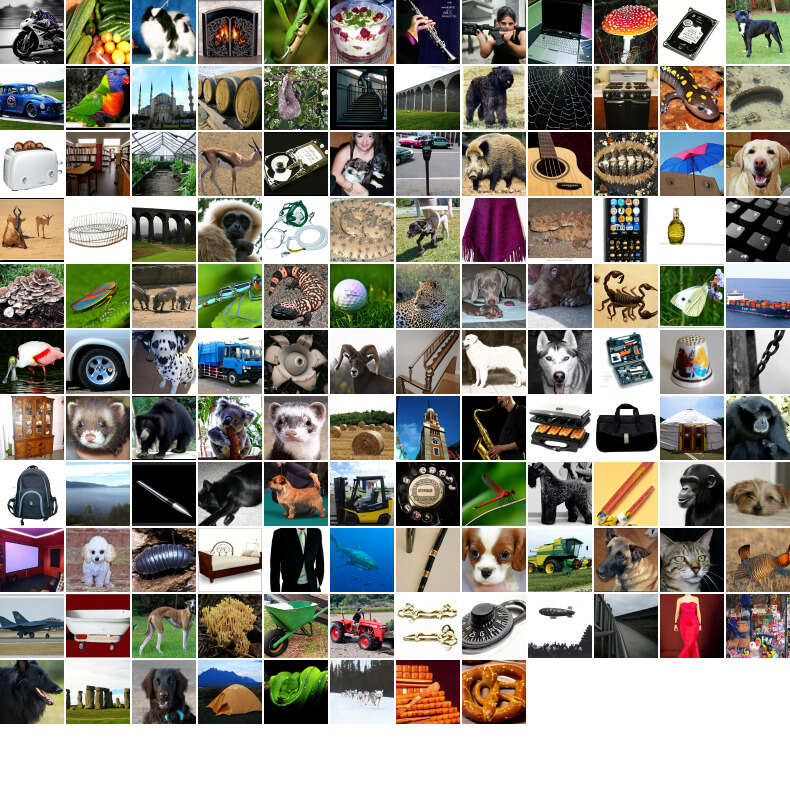}}
    \caption{$w=2$}
\end{subfigure}
\hfill
\begin{subfigure}{0.22\textwidth}
    \adjustbox{width=\linewidth, trim={.0\width} {.5\height} {0.5\width} {0\height},clip}{\includegraphics[width=\linewidth]{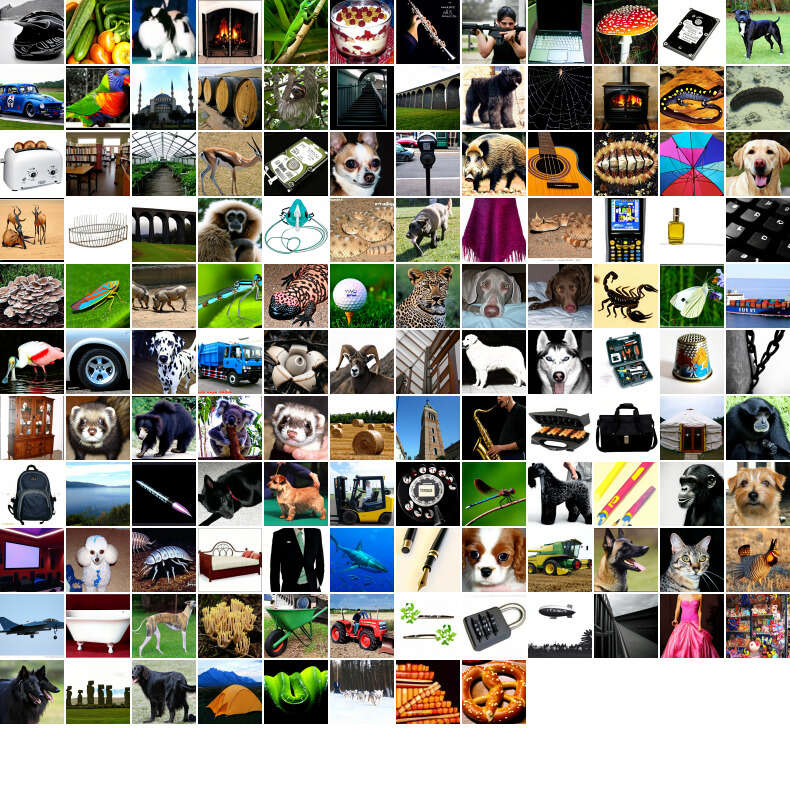}}
    \caption{$w=4$}
\end{subfigure}        
\caption{Ours (stochastic in pixel-space) on ImageNet 64x64. Distilled 8 sampling step.
}
\label{fig:figures}
\end{figure*}

\begin{figure*}[!hb]
\centering
\begin{subfigure}{0.22\textwidth}
    \adjustbox{width=\linewidth, trim={.0\width} {.5\height} {0.5\width} {0\height},clip}{\includegraphics[width=\linewidth]{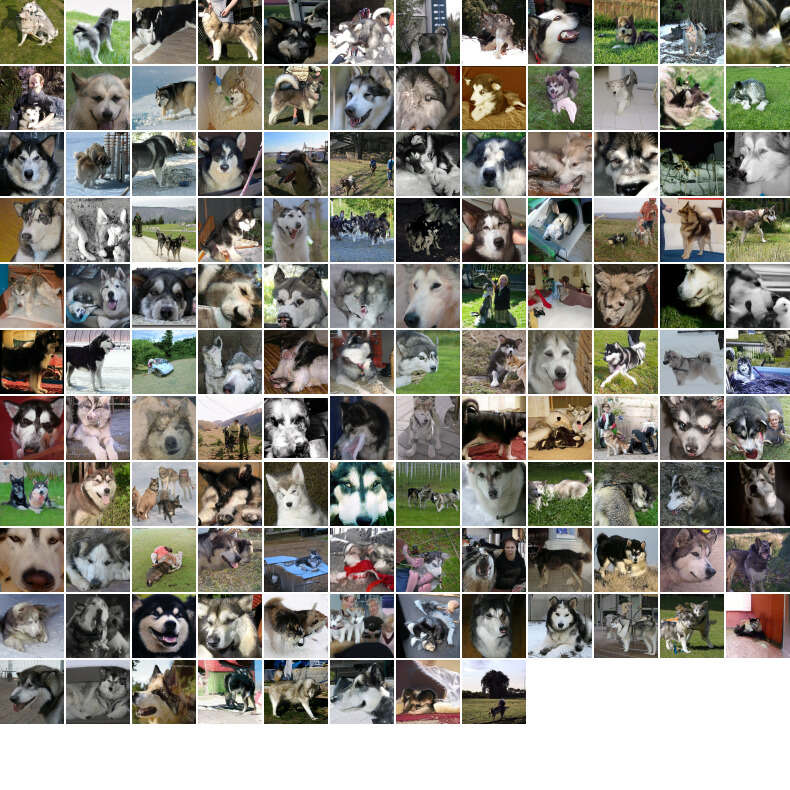}}
    \caption{$w=0$}
\end{subfigure}
\hfill
\begin{subfigure}{0.22\textwidth}
    \adjustbox{width=\linewidth, trim={.0\width} {.5\height} {0.5\width} {0\height},clip}{\includegraphics[width=\linewidth]{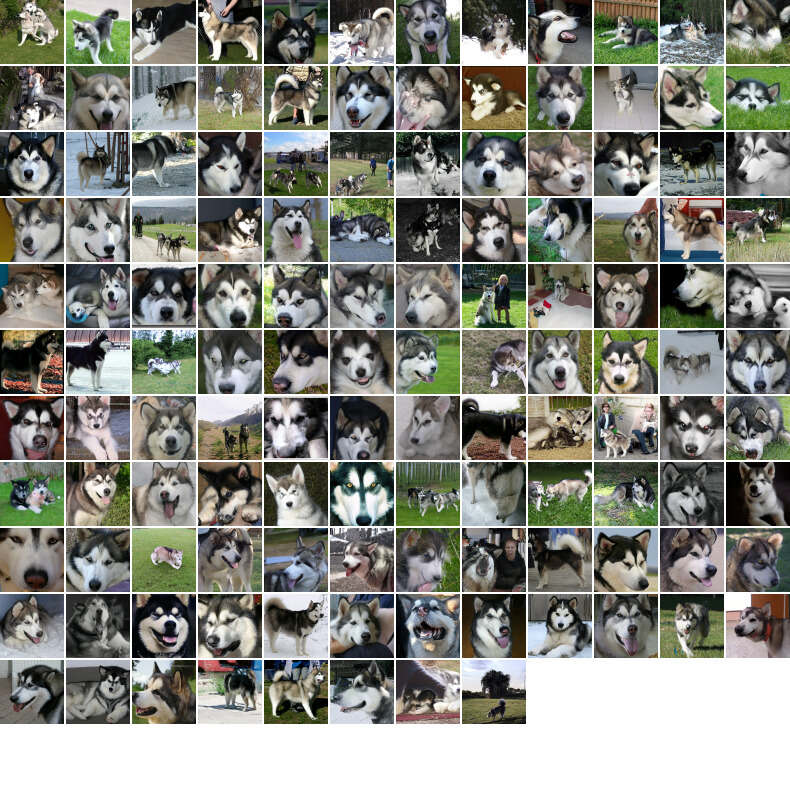}}
    \caption{$w=1$}
\end{subfigure}
\hfill
\begin{subfigure}{0.22\textwidth}
     \adjustbox{width=\linewidth, trim={.0\width} {.5\height} {0.5\width} {0\height},clip}{\includegraphics[width=\linewidth]{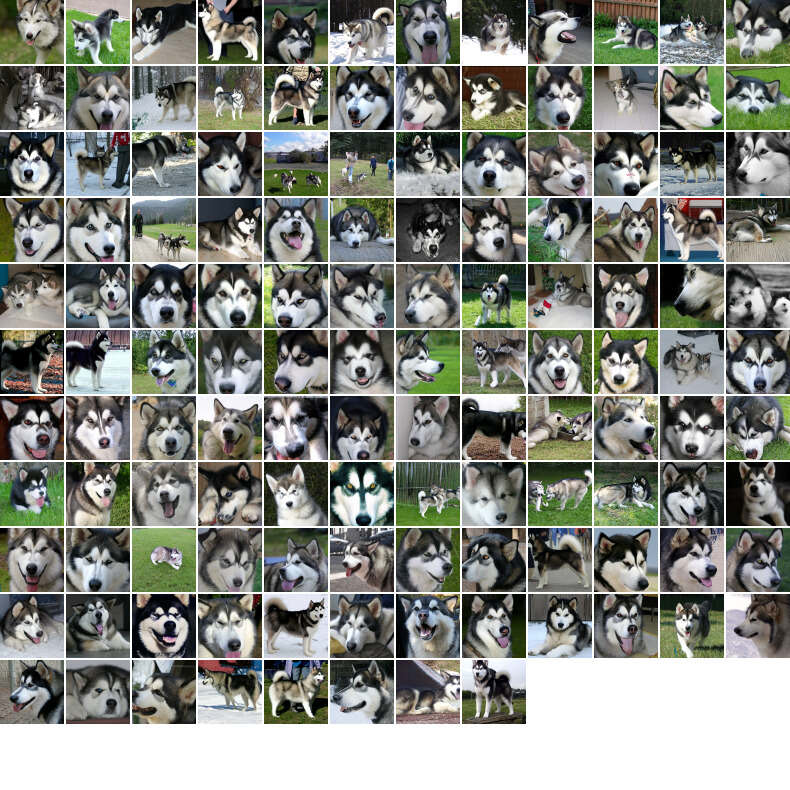}}
    \caption{$w=2$}
\end{subfigure}
\hfill
\begin{subfigure}{0.22\textwidth}
    \adjustbox{width=\linewidth, trim={.0\width} {.5\height} {0.5\width} {0\height},clip}{\includegraphics[width=\linewidth]{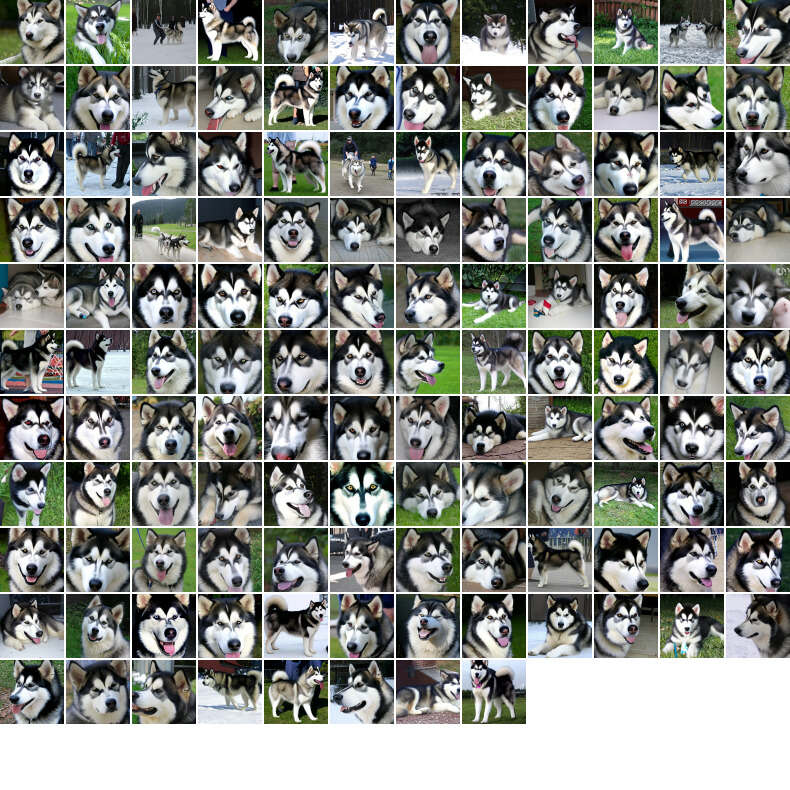}}
    \caption{$w=4$}
\end{subfigure}        
\caption{Ours (deterministic in pixel-space) on ImageNet 64x64. Distilled 8 sampling step. Class-conditioned samples.}
\label{fig:figures}
\end{figure*}

\begin{figure*}[!hb]
\centering
\begin{subfigure}{0.22\textwidth}
    \adjustbox{width=\linewidth, trim={.0\width} {.5\height} {0.5\width} {0\height},clip}{\includegraphics[width=\linewidth]{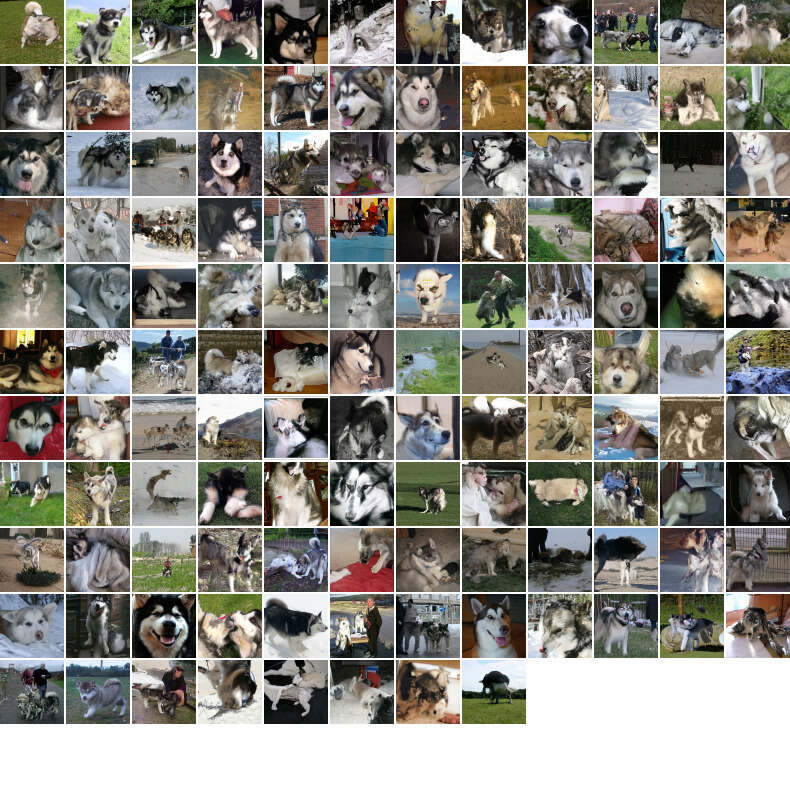}}
    \caption{$w=0$}
\end{subfigure}
\hfill
\begin{subfigure}{0.22\textwidth}
    \adjustbox{width=\linewidth, trim={.0\width} {.5\height} {0.5\width} {0\height},clip}{\includegraphics[width=\linewidth]{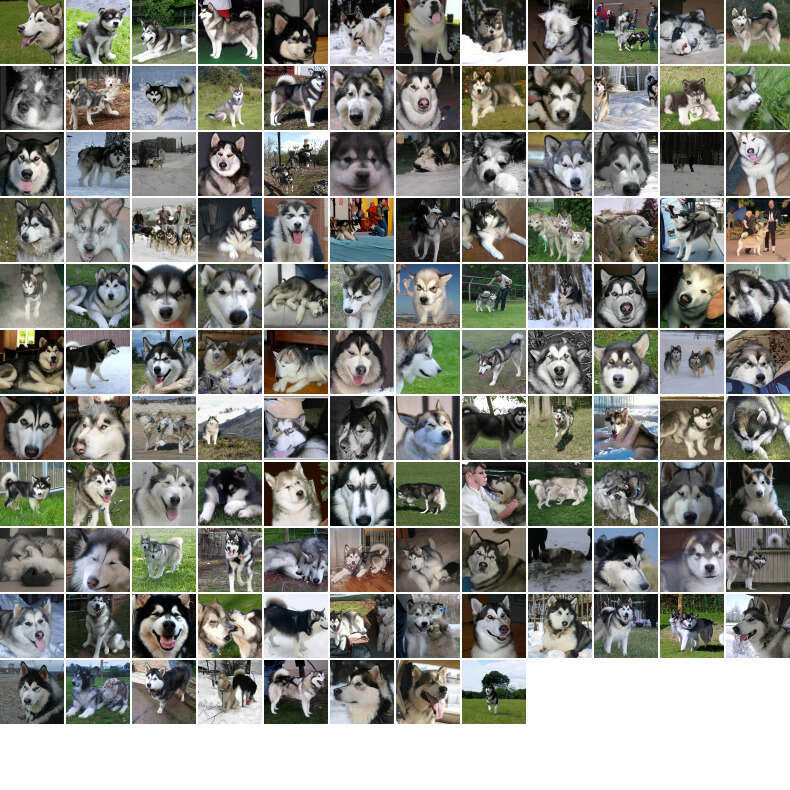}}
    \caption{$w=1$}
\end{subfigure}
\hfill
\begin{subfigure}{0.22\textwidth}
     \adjustbox{width=\linewidth, trim={.0\width} {.5\height} {0.5\width} {0\height},clip}{\includegraphics[width=\linewidth]{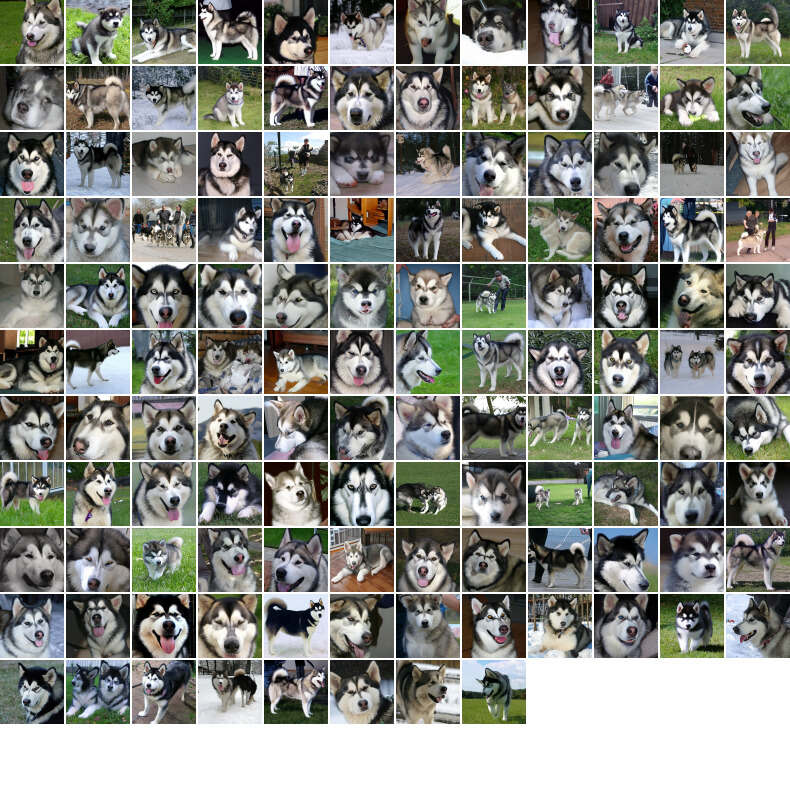}}
    \caption{$w=2$}
\end{subfigure}
\hfill
\begin{subfigure}{0.22\textwidth}
    \adjustbox{width=\linewidth, trim={.0\width} {.5\height} {0.5\width} {0\height},clip}{\includegraphics[width=\linewidth]{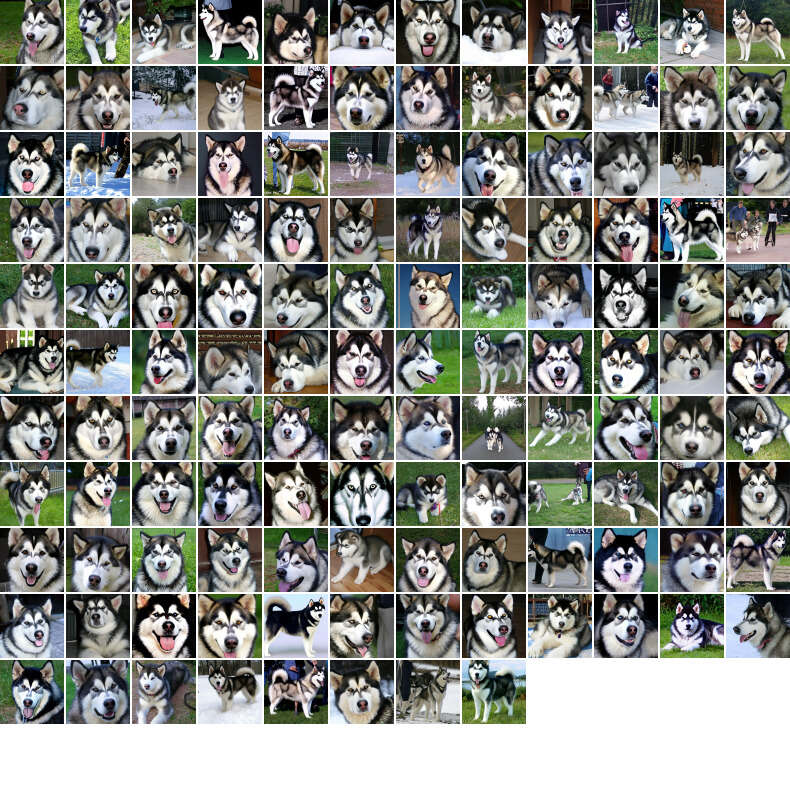}}
    \caption{$w=4$}
\end{subfigure}        
\caption{Ours (stochastic in pixel-space). Distilled 8 sampling step. Class-conditioned samples. 
}
\label{fig:figures}
\end{figure*}

 \begin{figure*}[!hb]
\centering
\begin{subfigure}{0.22\textwidth}
    \adjustbox{width=\linewidth, trim={.0\width} {.5\height} {0.5\width} {0\height},clip}{\includegraphics[width=\linewidth]{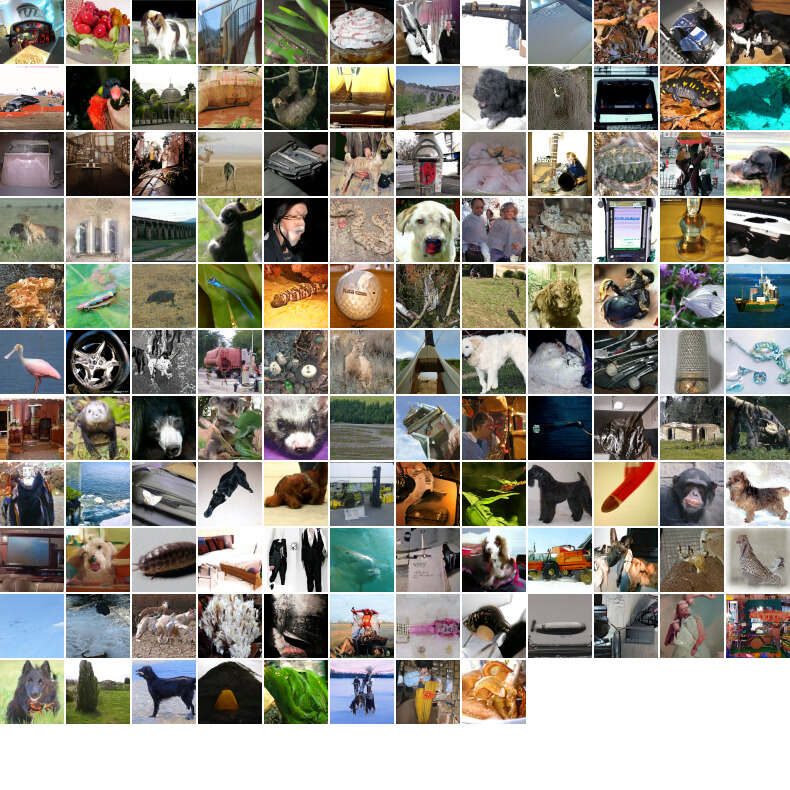}}
    \caption{$w=0$}
\end{subfigure}
\hfill
\begin{subfigure}{0.22\textwidth}
    \adjustbox{width=\linewidth, trim={.0\width} {.5\height} {0.5\width} {0\height},clip}{\includegraphics[width=\linewidth]{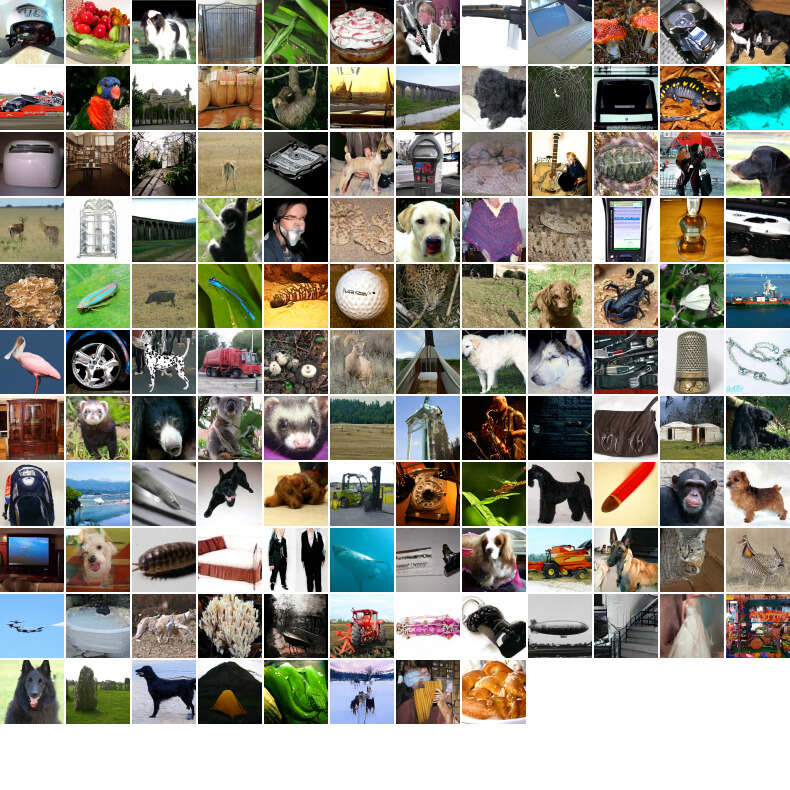}}
    \caption{$w=1$}
\end{subfigure}
\hfill
\begin{subfigure}{0.22\textwidth}
     \adjustbox{width=\linewidth, trim={.0\width} {.5\height} {0.5\width} {0\height},clip}{\includegraphics[width=\linewidth]{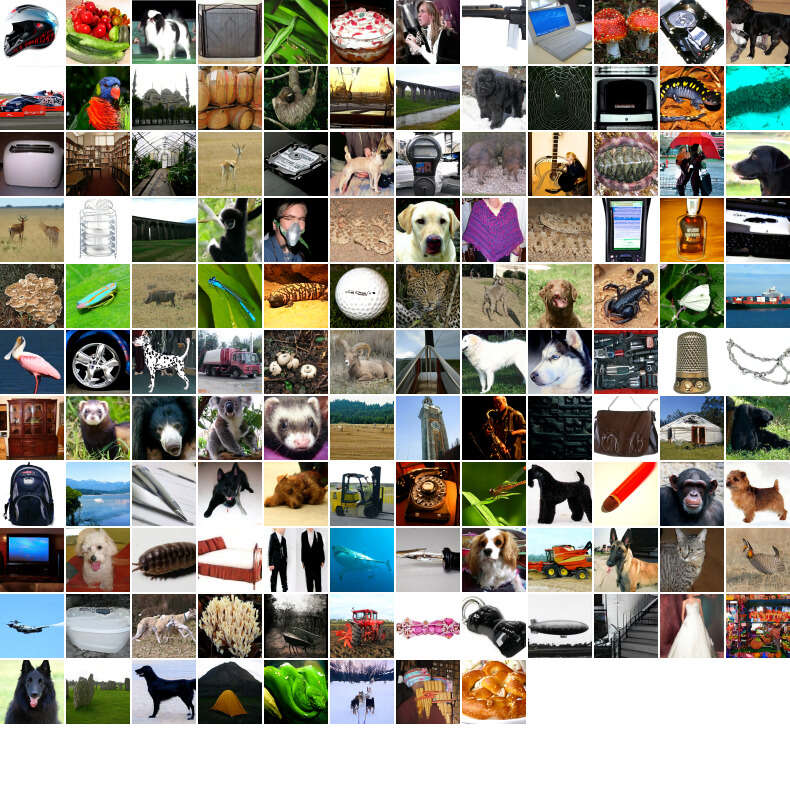}}
    \caption{$w=2$}
\end{subfigure}
\hfill
\begin{subfigure}{0.22\textwidth}
    \adjustbox{width=\linewidth, trim={.0\width} {.5\height} {0.5\width} {0\height},clip}{\includegraphics[width=\linewidth]{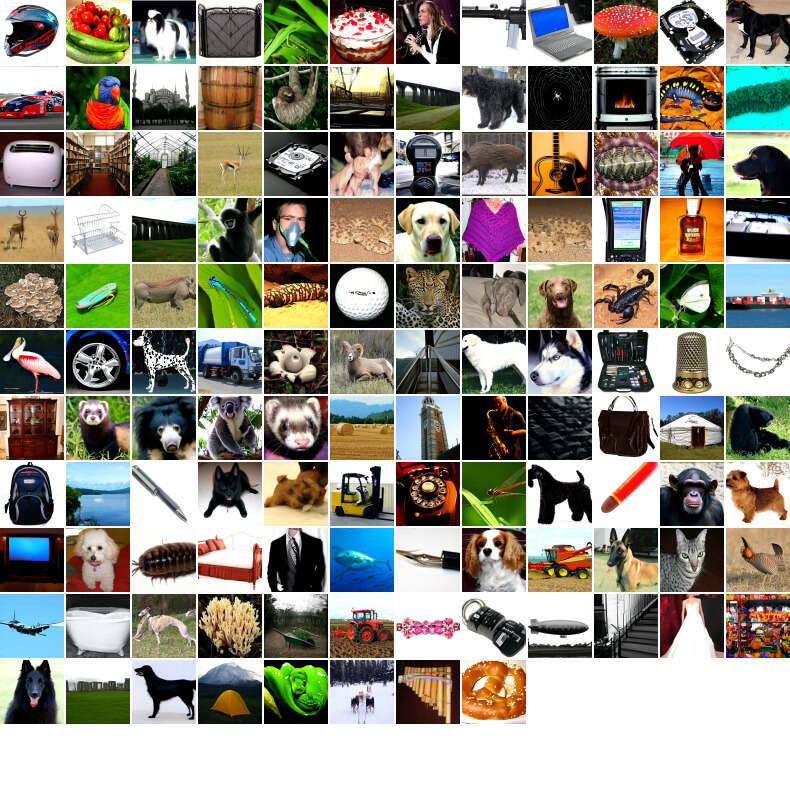}}
    \caption{$w=4$}
\end{subfigure}        
\caption{Ours (deterministic in pixel-space) on ImageNet 64x64. Distilled 2 sampling steps.}
\label{fig:figures}
\end{figure*}

 \begin{figure*}[!hb]
\centering
\begin{subfigure}{0.22\textwidth}
    \adjustbox{width=\linewidth, trim={.0\width} {.5\height} {0.5\width} {0\height},clip}{\includegraphics[width=\linewidth]{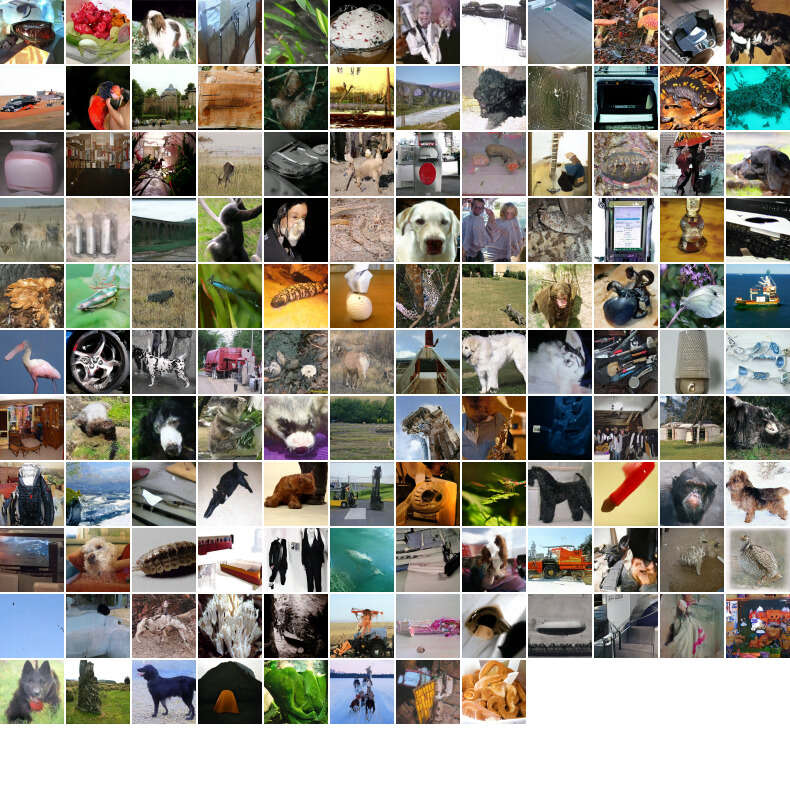}}
    \caption{$w=0$}
\end{subfigure}
\hfill
\begin{subfigure}{0.22\textwidth}
    \adjustbox{width=\linewidth, trim={.0\width} {.5\height} {0.5\width} {0\height},clip}{\includegraphics[width=\linewidth]{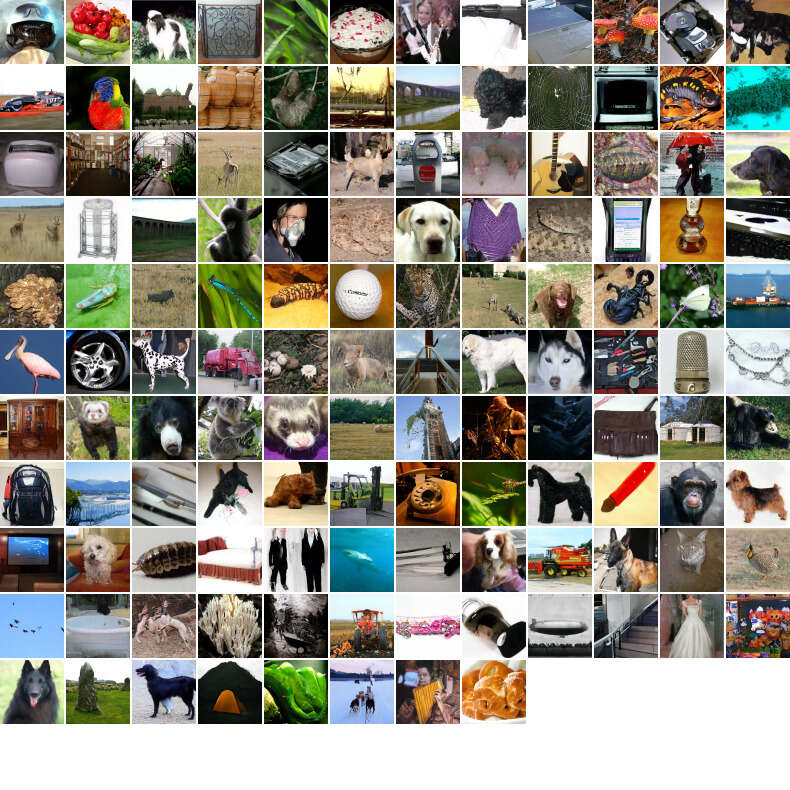}}
    \caption{$w=1$}
\end{subfigure}
\hfill
\begin{subfigure}{0.22\textwidth}
     \adjustbox{width=\linewidth, trim={.0\width} {.5\height} {0.5\width} {0\height},clip}{\includegraphics[width=\linewidth]{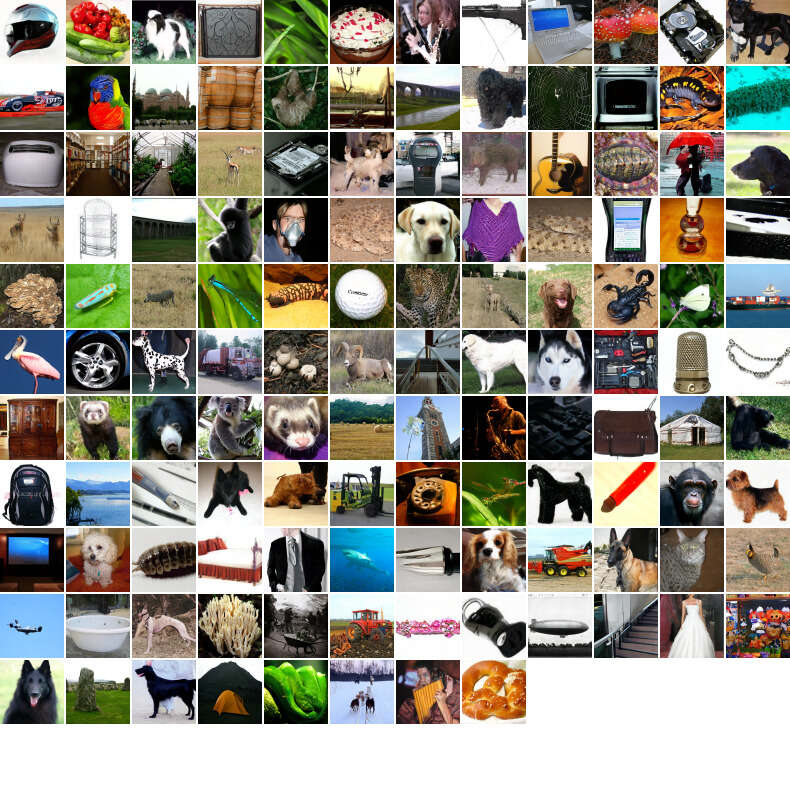}}
    \caption{$w=2$}
\end{subfigure}
\hfill
\begin{subfigure}{0.22\textwidth}
    \adjustbox{width=\linewidth, trim={.0\width} {.5\height} {0.5\width} {0\height},clip}{\includegraphics[width=\linewidth]{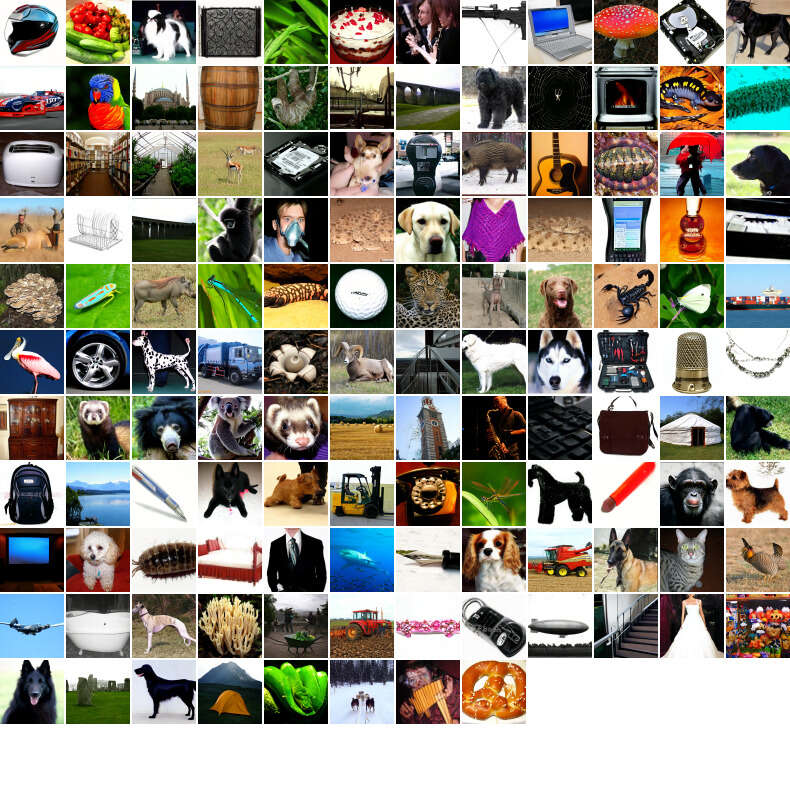}}
    \caption{$w=4$}
\end{subfigure}        
\caption{Ours (stochastic in pixel-space) on ImageNet 64x64. Distilled 2 sampling steps.}
\label{fig:figures}
\end{figure*}

\begin{figure*}[!hb]
\centering
\begin{subfigure}{0.22\textwidth}
    \adjustbox{width=\linewidth, trim={.0\width} {.5\height} {0.5\width} {0\height},clip}{\includegraphics[width=\linewidth]{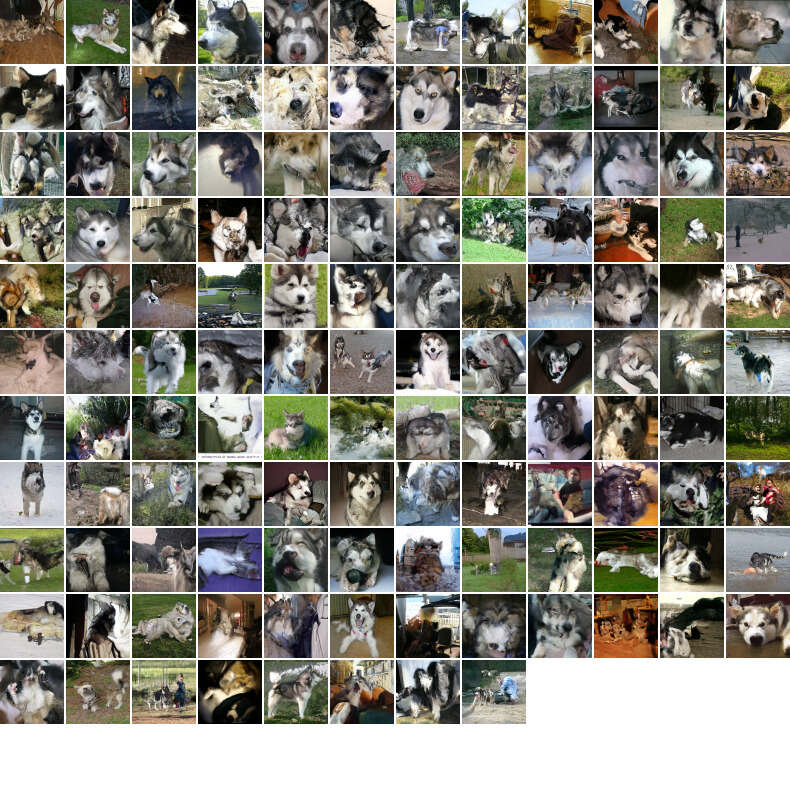}}
    \caption{$w=0$}
\end{subfigure}
\hfill
\begin{subfigure}{0.22\textwidth}
    \adjustbox{width=\linewidth, trim={.0\width} {.5\height} {0.5\width} {0\height},clip}{\includegraphics[width=\linewidth]{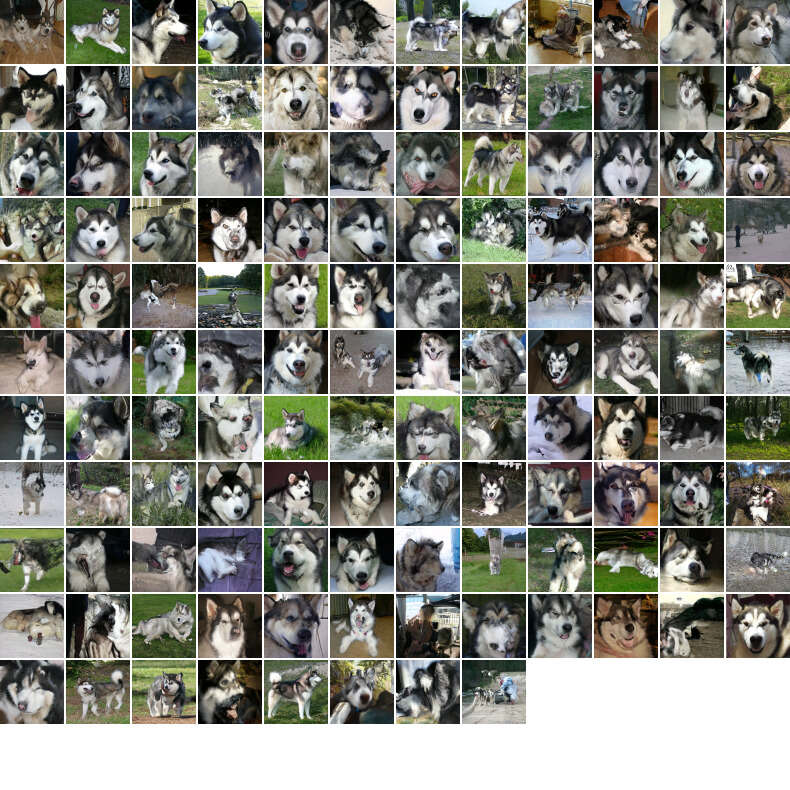}}
    \caption{$w=1$}
\end{subfigure}
\hfill
\begin{subfigure}{0.22\textwidth}
     \adjustbox{width=\linewidth, trim={.0\width} {.5\height} {0.5\width} {0\height},clip}{\includegraphics[width=\linewidth]{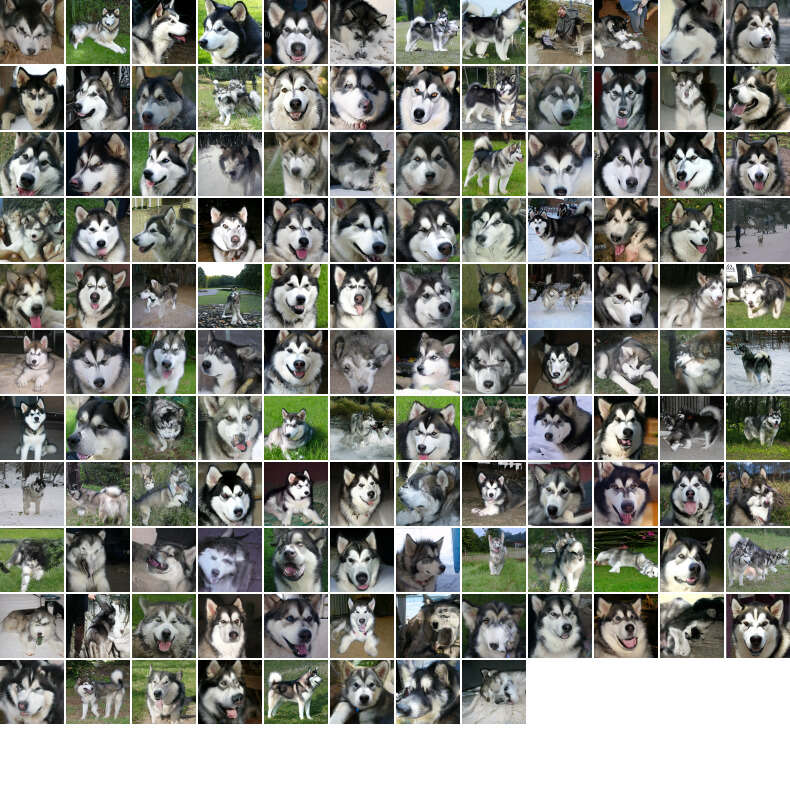}}
    \caption{$w=2$}
\end{subfigure}
\hfill
\begin{subfigure}{0.22\textwidth}
    \adjustbox{width=\linewidth, trim={.0\width} {.5\height} {0.5\width} {0\height},clip}{\includegraphics[width=\linewidth]{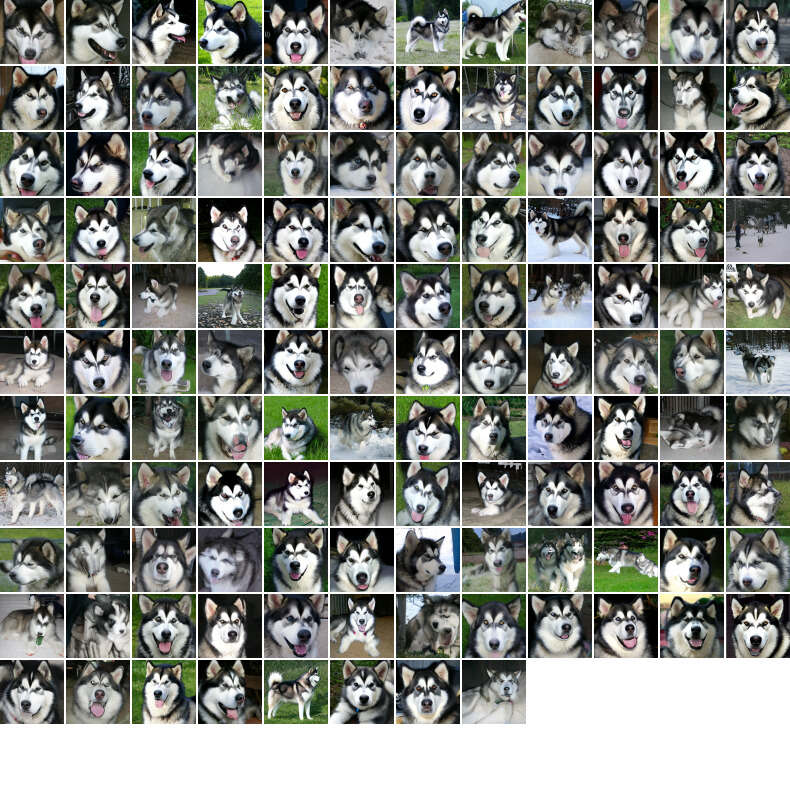}}
    \caption{$w=4$}
\end{subfigure}        
\caption{Ours (deterministic in pixel-space) on ImageNet 64x64. Distilled 2 sampling steps. Class-conditioned samples.}
\label{fig:figures}
\end{figure*}

\begin{figure*}[!hb]
\centering
\begin{subfigure}{0.22\textwidth}
    \adjustbox{width=\linewidth, trim={.0\width} {.5\height} {0.5\width} {0\height},clip}{\includegraphics[width=\linewidth]{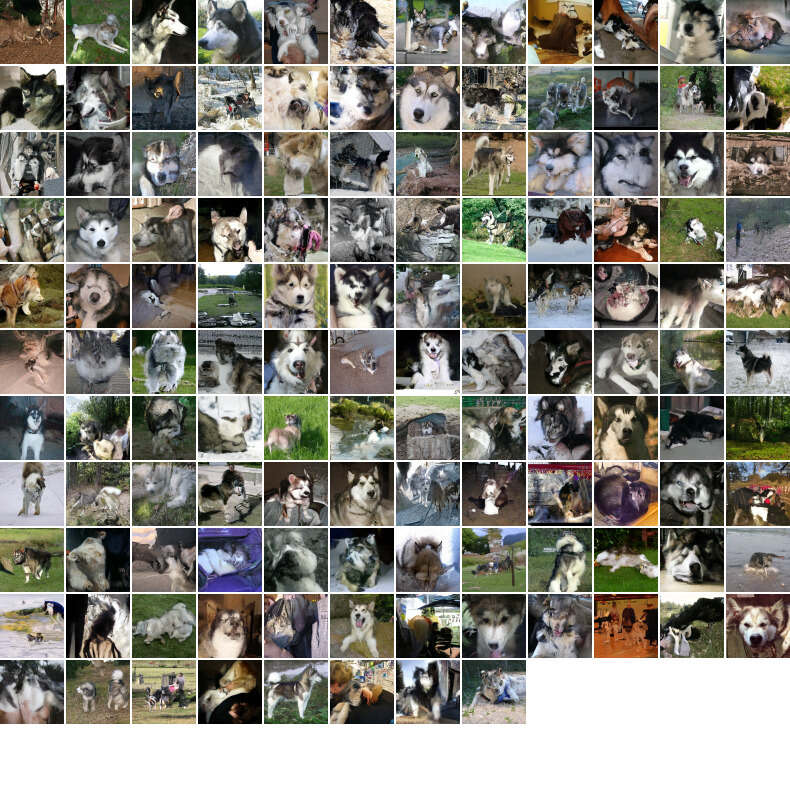}}
    \caption{$w=0$}
\end{subfigure}
\hfill
\begin{subfigure}{0.22\textwidth}
    \adjustbox{width=\linewidth, trim={.0\width} {.5\height} {0.5\width} {0\height},clip}{\includegraphics[width=\linewidth]{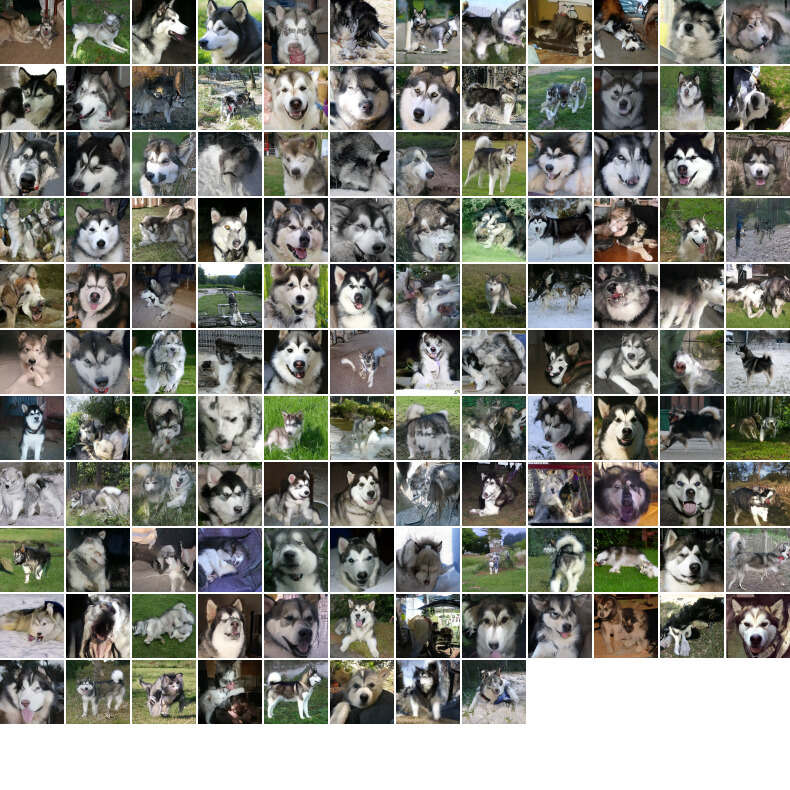}}
    \caption{$w=1$}
\end{subfigure}
\hfill
\begin{subfigure}{0.22\textwidth}
     \adjustbox{width=\linewidth, trim={.0\width} {.5\height} {0.5\width} {0\height},clip}{\includegraphics[width=\linewidth]{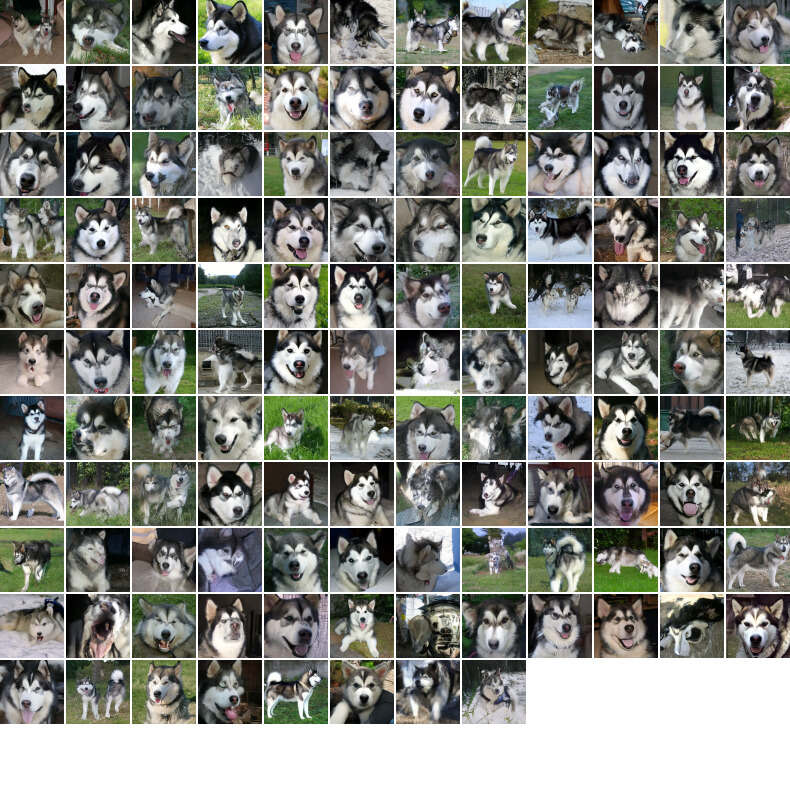}}
    \caption{$w=2$}
\end{subfigure}
\hfill
\begin{subfigure}{0.22\textwidth}
    \adjustbox{width=\linewidth, trim={.0\width} {.5\height} {0.5\width} {0\height},clip}{\includegraphics[width=\linewidth]{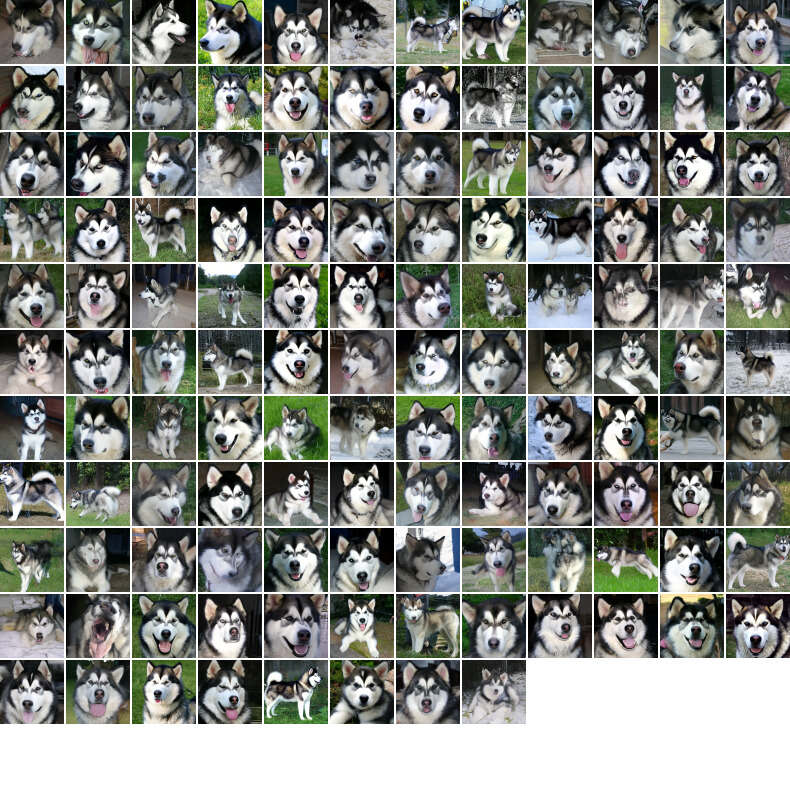}}
    \caption{$w=4$}
\end{subfigure}        
\caption{Ours (stochastic in pixel-space) on ImageNet 64x64. Distilled 2 sampling steps. Class-conditioned samples.}
\label{fig:figures}
\end{figure*}

 \begin{figure*}[!hb]
\centering
\begin{subfigure}{0.22\textwidth}
    \adjustbox{width=\linewidth, trim={.0\width} {.5\height} {0.5\width} {0\height},clip}{\includegraphics[width=\linewidth]{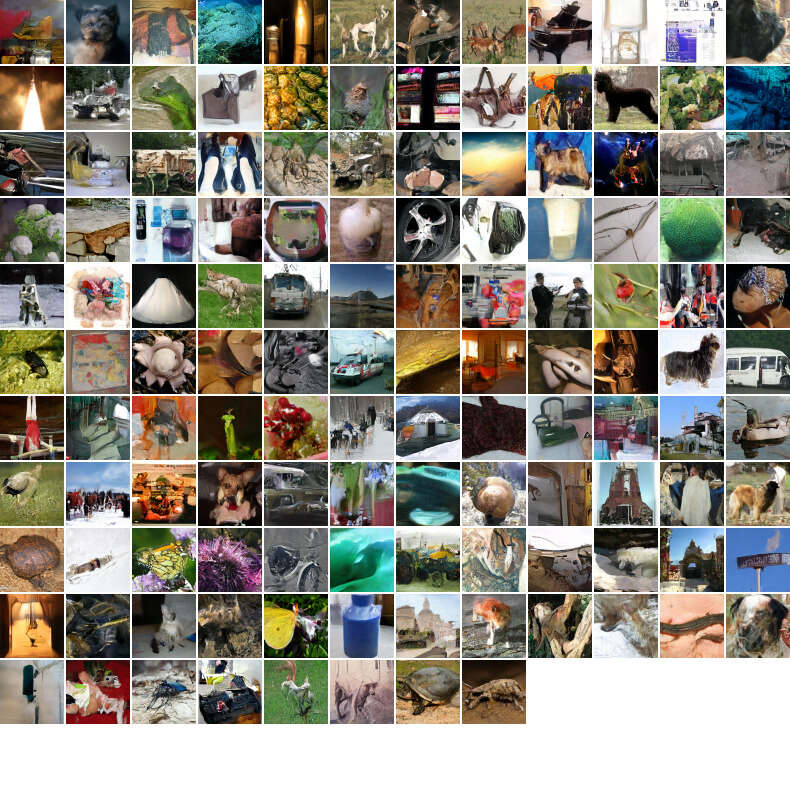}}
    \caption{$w=0$}
\end{subfigure}
\hfill
\begin{subfigure}{0.22\textwidth}
    \adjustbox{width=\linewidth, trim={.0\width} {.5\height} {0.5\width} {0\height},clip}{\includegraphics[width=\linewidth]{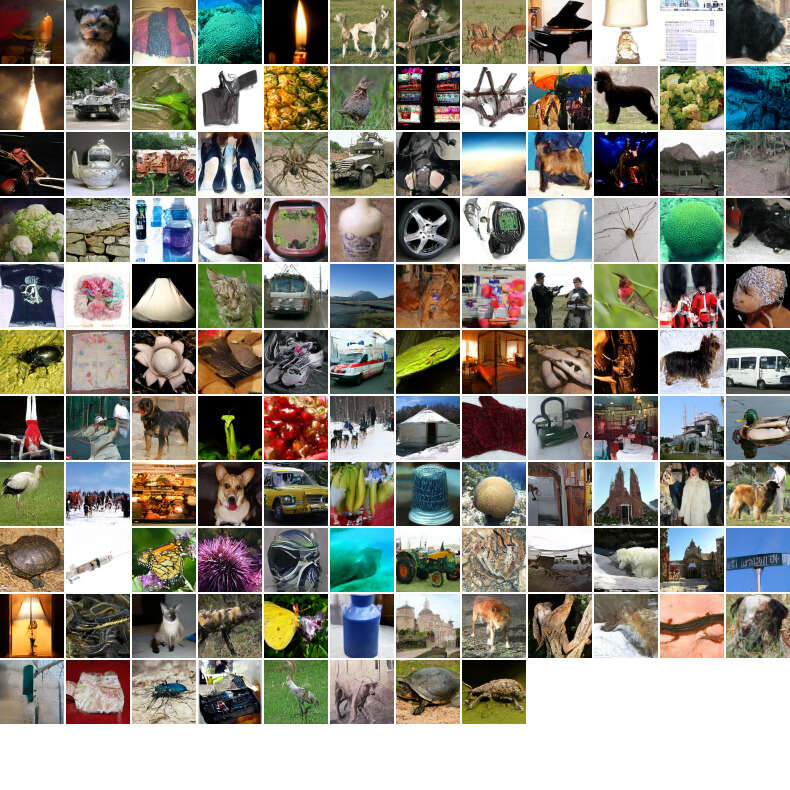}}
    \caption{$w=1$}
\end{subfigure}
\hfill
\begin{subfigure}{0.22\textwidth}
     \adjustbox{width=\linewidth, trim={.0\width} {.5\height} {0.5\width} {0\height},clip}{\includegraphics[width=\linewidth]{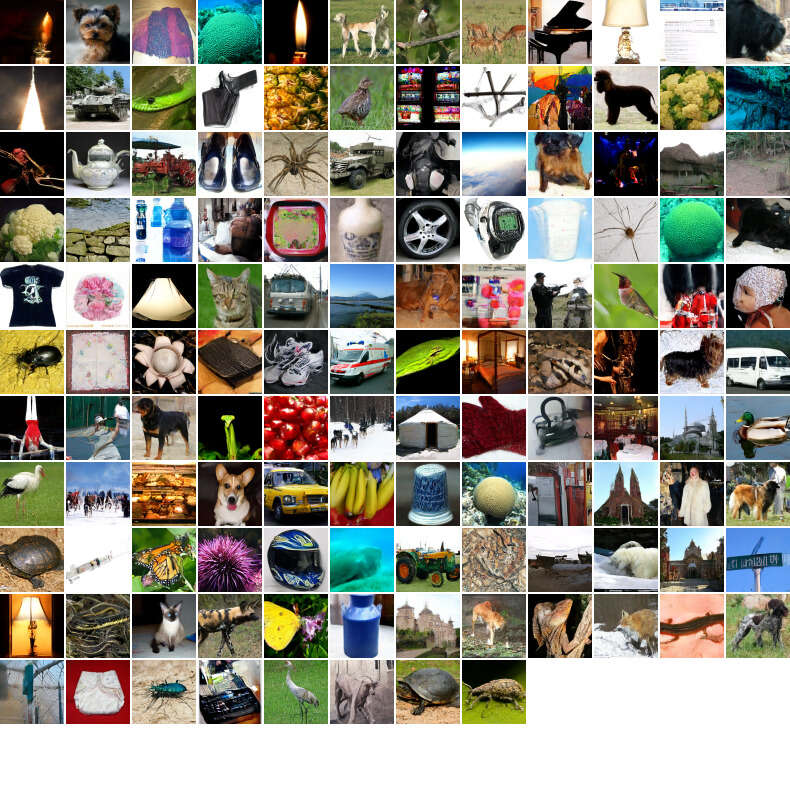}}
    \caption{$w=2$}
\end{subfigure}
\hfill
\begin{subfigure}{0.22\textwidth}
    \adjustbox{width=\linewidth, trim={.0\width} {.5\height} {0.5\width} {0\height},clip}{\includegraphics[width=\linewidth]{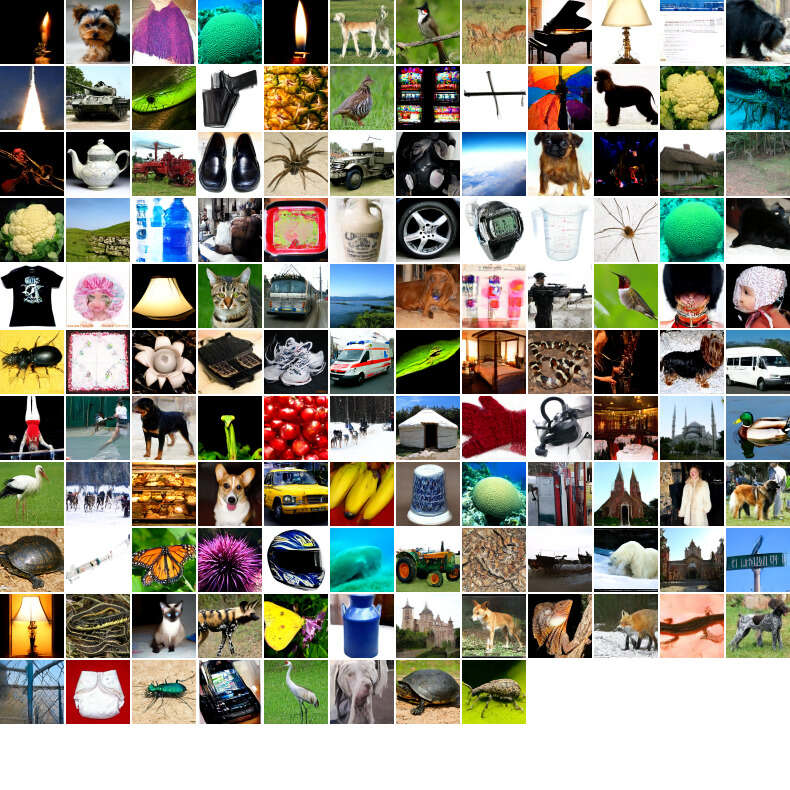}}
    \caption{$w=4$}
\end{subfigure}        
\caption{Ours (deterministic in pixel-space) on ImageNet 64x64. Distilled 1 sampling step.}
\label{fig:figures}
\end{figure*}

 \begin{figure*}[!hb]
\centering
\begin{subfigure}{0.22\textwidth}
    \adjustbox{width=\linewidth, trim={.0\width} {.5\height} {0.5\width} {0\height},clip}{\includegraphics[width=\linewidth]{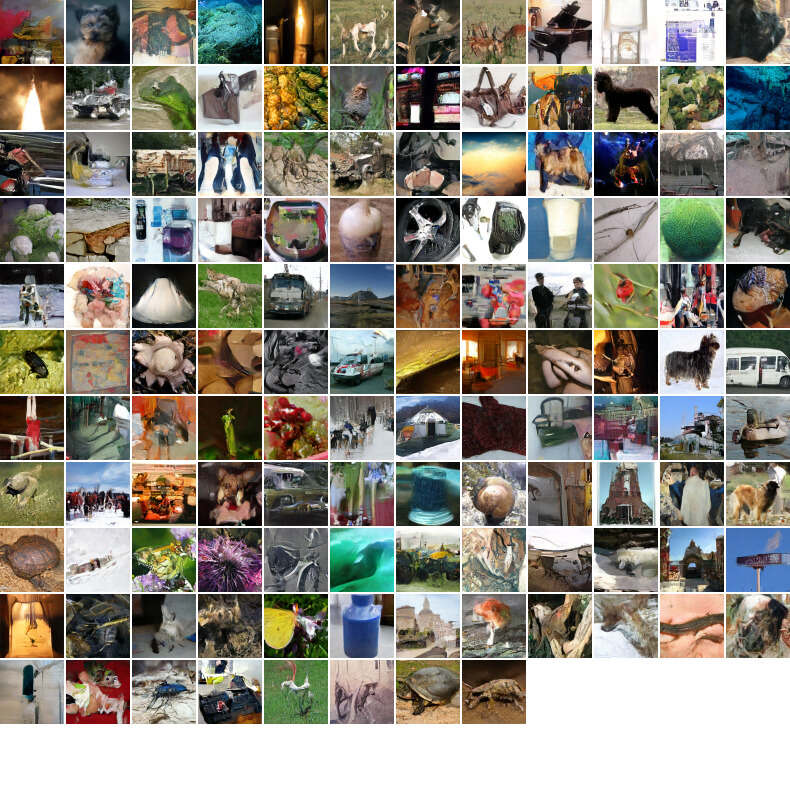}}
    \caption{$w=0$}
\end{subfigure}
\hfill
\begin{subfigure}{0.22\textwidth}
    \adjustbox{width=\linewidth, trim={.0\width} {.5\height} {0.5\width} {0\height},clip}{\includegraphics[width=\linewidth]{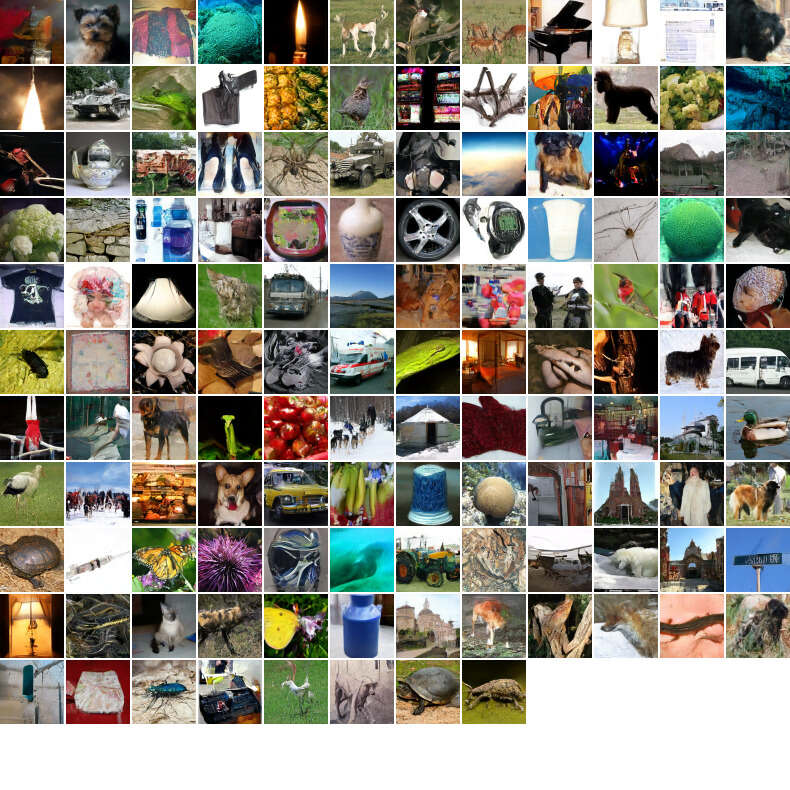}}
    \caption{$w=1$}
\end{subfigure}
\hfill
\begin{subfigure}{0.22\textwidth}
     \adjustbox{width=\linewidth, trim={.0\width} {.5\height} {0.5\width} {0\height},clip}{\includegraphics[width=\linewidth]{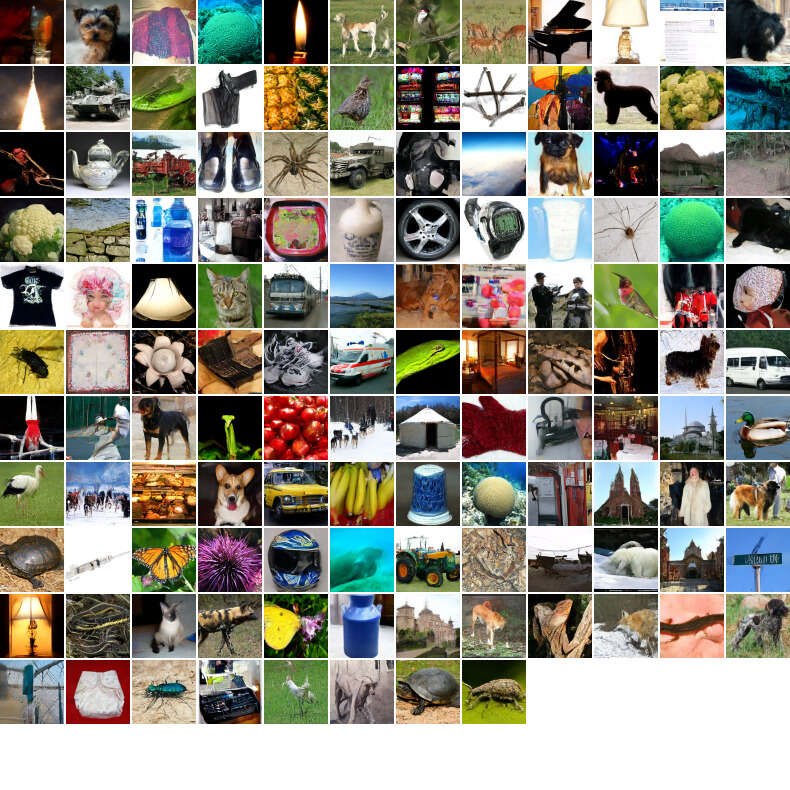}}
    \caption{$w=2$}
\end{subfigure}
\hfill
\begin{subfigure}{0.22\textwidth}
    \adjustbox{width=\linewidth, trim={.0\width} {.5\height} {0.5\width} {0\height},clip}{\includegraphics[width=\linewidth]{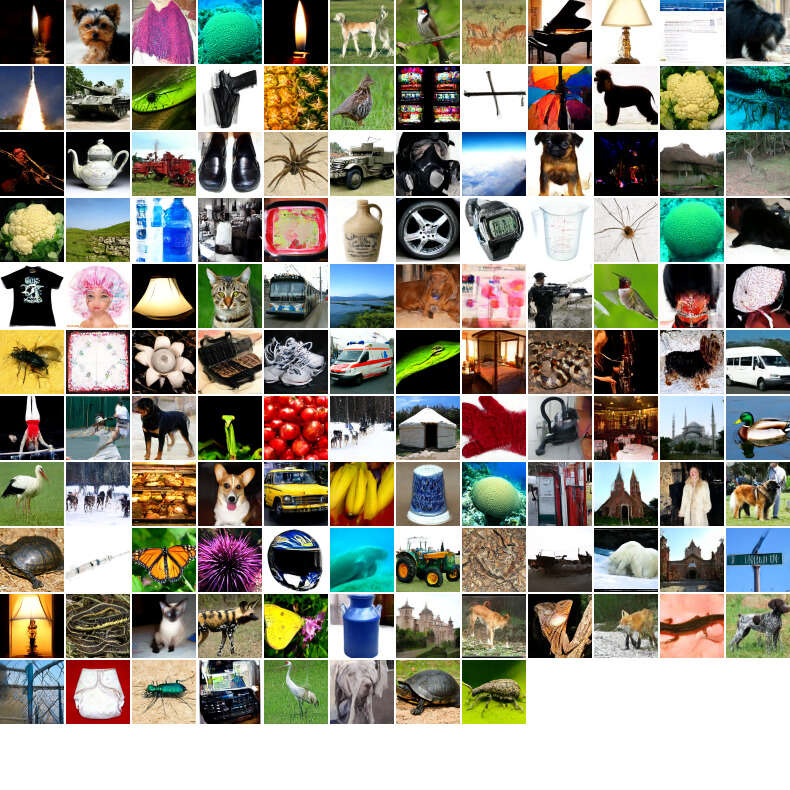}}
    \caption{$w=4$}
\end{subfigure}        
\caption{Ours (stochastic in pixel-space) on ImageNet 64x64. Distilled 1 sampling step.}
\label{fig:figures}
\end{figure*}

\begin{figure*}[!hb]
\centering
\begin{subfigure}{0.22\textwidth}
    \adjustbox{width=\linewidth, trim={.0\width} {.5\height} {0.5\width} {0\height},clip}{\includegraphics[width=\linewidth]{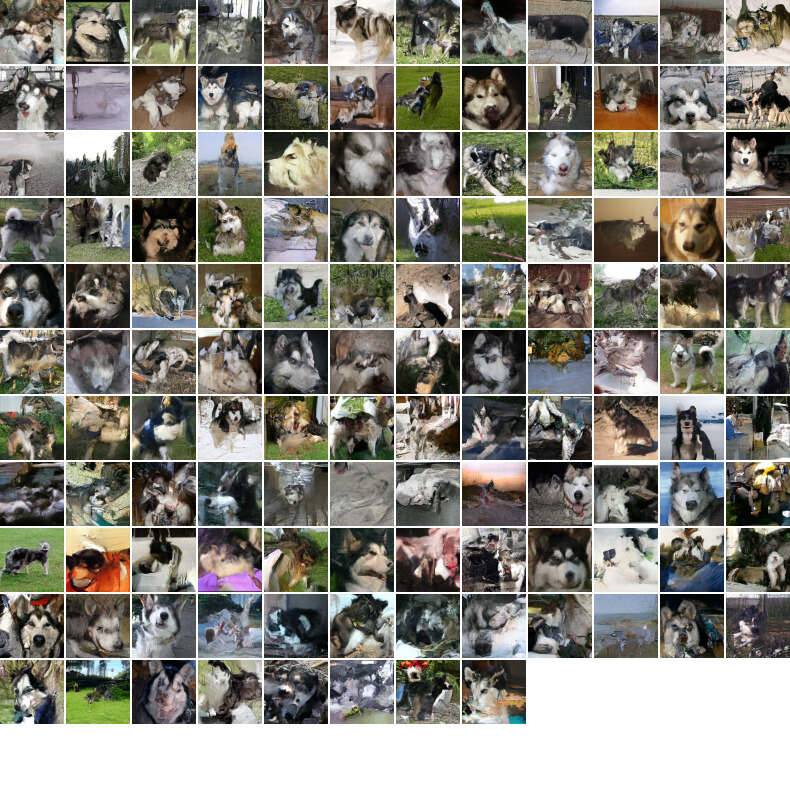}}
    \caption{$w=0$}
\end{subfigure}
\hfill
\begin{subfigure}{0.22\textwidth}
    \adjustbox{width=\linewidth, trim={.0\width} {.5\height} {0.5\width} {0\height},clip}{\includegraphics[width=\linewidth]{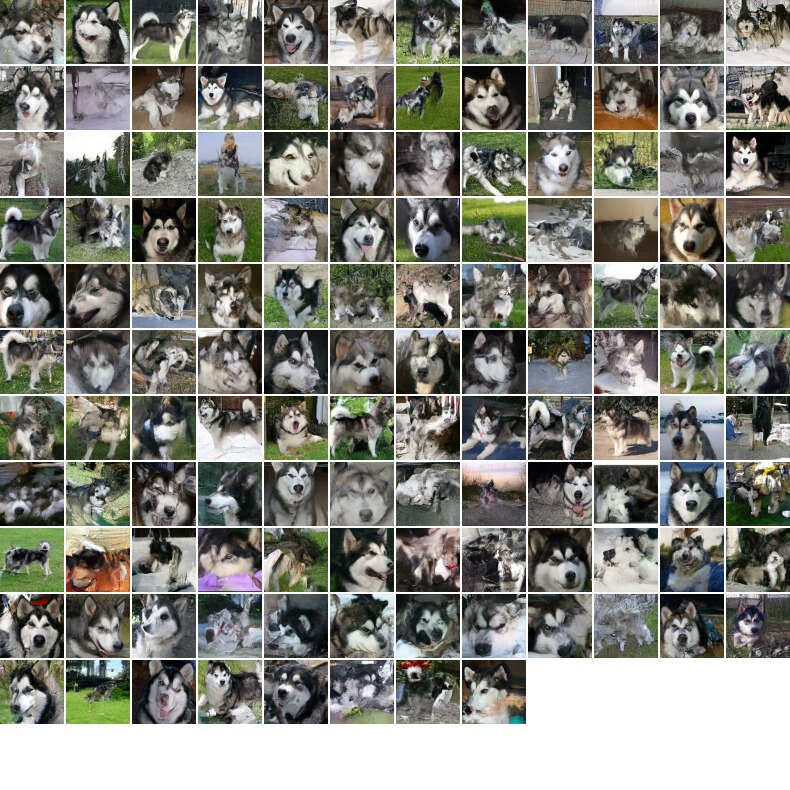}}
    \caption{$w=1$}
\end{subfigure}
\hfill
\begin{subfigure}{0.22\textwidth}
     \adjustbox{width=\linewidth, trim={.0\width} {.5\height} {0.5\width} {0\height},clip}{\includegraphics[width=\linewidth]{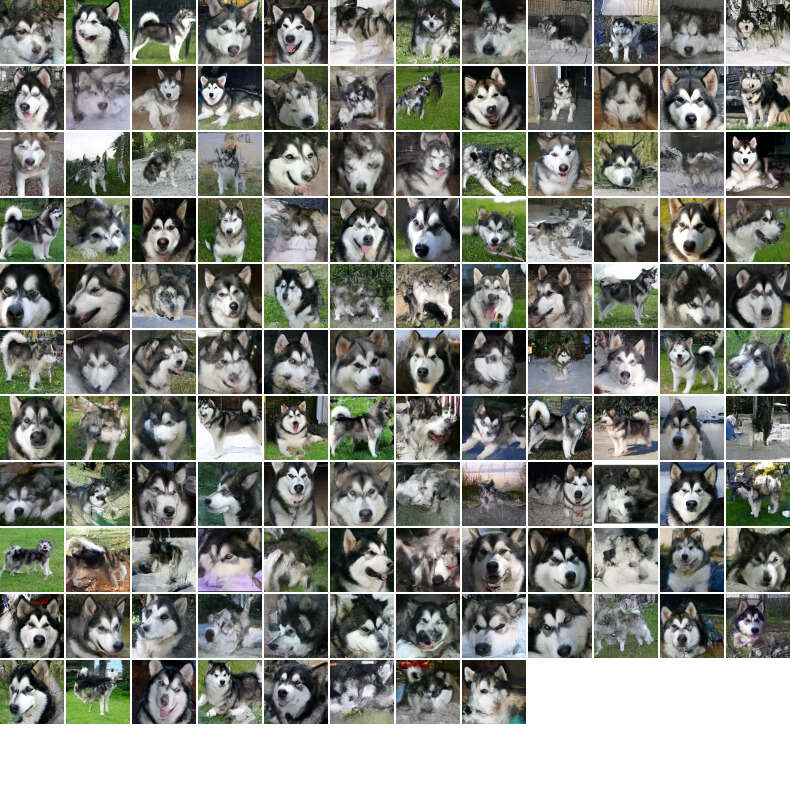}}
    \caption{$w=2$}
\end{subfigure}
\hfill
\begin{subfigure}{0.22\textwidth}
    \adjustbox{width=\linewidth, trim={.0\width} {.5\height} {0.5\width} {0\height},clip}{\includegraphics[width=\linewidth]{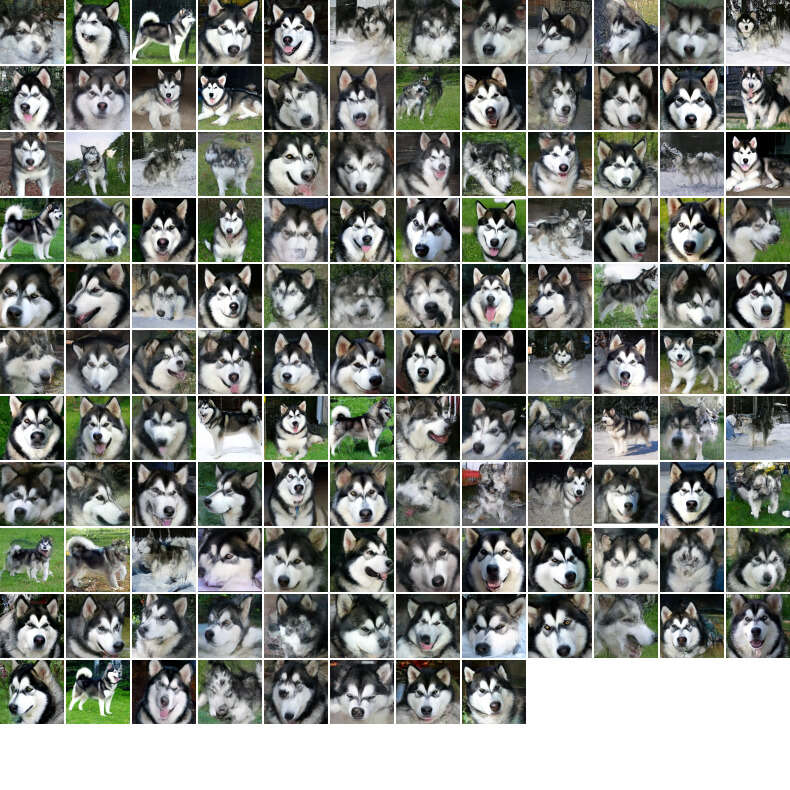}}
    \caption{$w=4$}
\end{subfigure}        
\caption{Ours (deterministic in pixel-space) on ImageNet 64x64. Distilled 1 sampling step. Class-conditioned samples.}
\label{fig:figures}
\end{figure*}

\begin{figure*}[!hb]
\centering
\begin{subfigure}{0.22\textwidth}
    \adjustbox{width=\linewidth, trim={.0\width} {.5\height} {0.5\width} {0\height},clip}{\includegraphics[width=\linewidth]{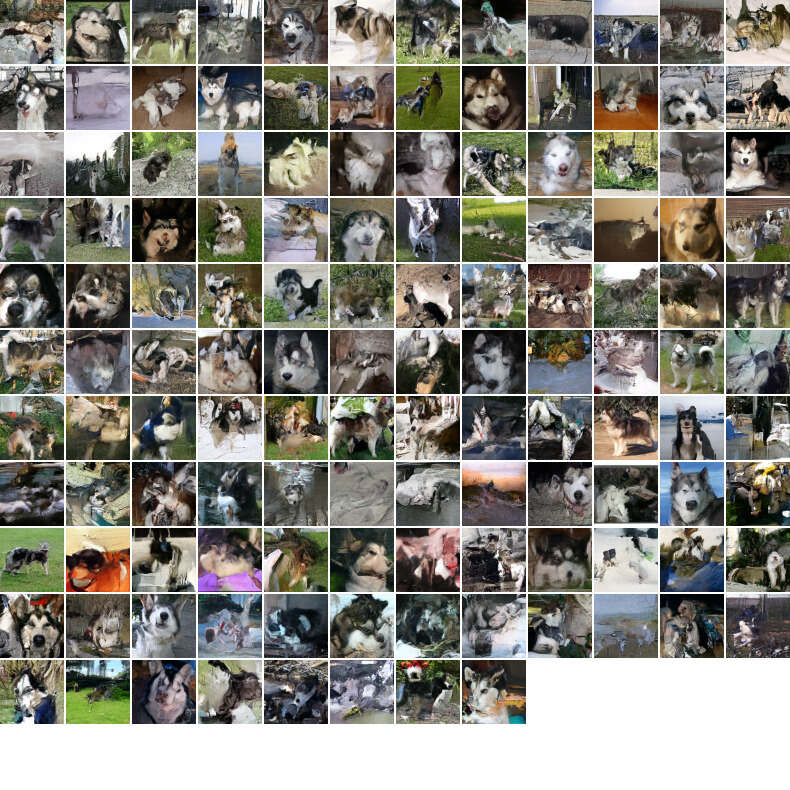}}
    \caption{$w=0$}
\end{subfigure}
\hfill
\begin{subfigure}{0.22\textwidth}
    \adjustbox{width=\linewidth, trim={.0\width} {.5\height} {0.5\width} {0\height},clip}{\includegraphics[width=\linewidth]{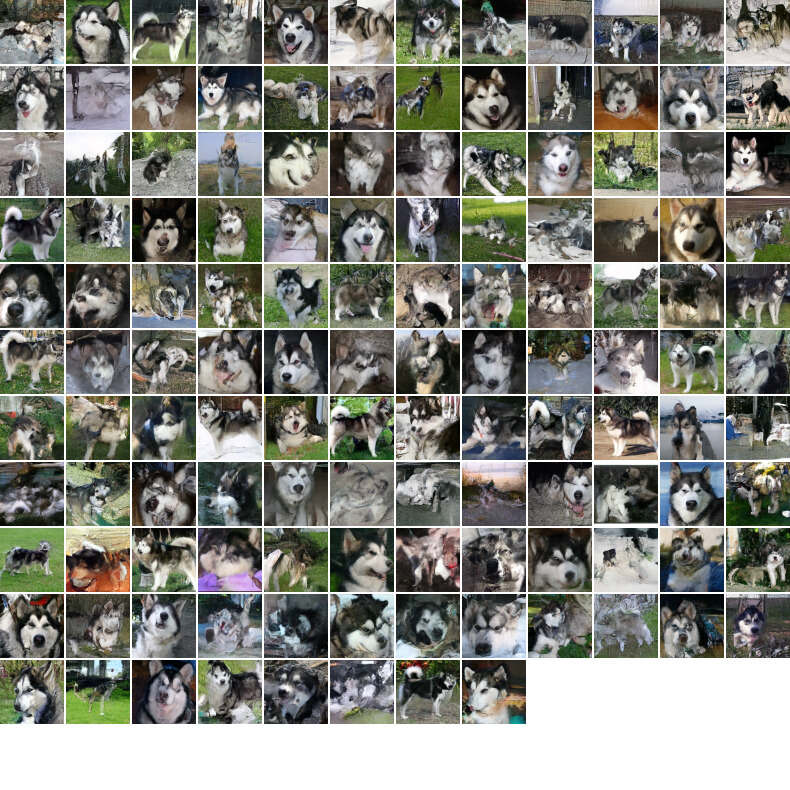}}
    \caption{$w=1$}
\end{subfigure}
\hfill
\begin{subfigure}{0.22\textwidth}
     \adjustbox{width=\linewidth, trim={.0\width} {.5\height} {0.5\width} {0\height},clip}{\includegraphics[width=\linewidth]{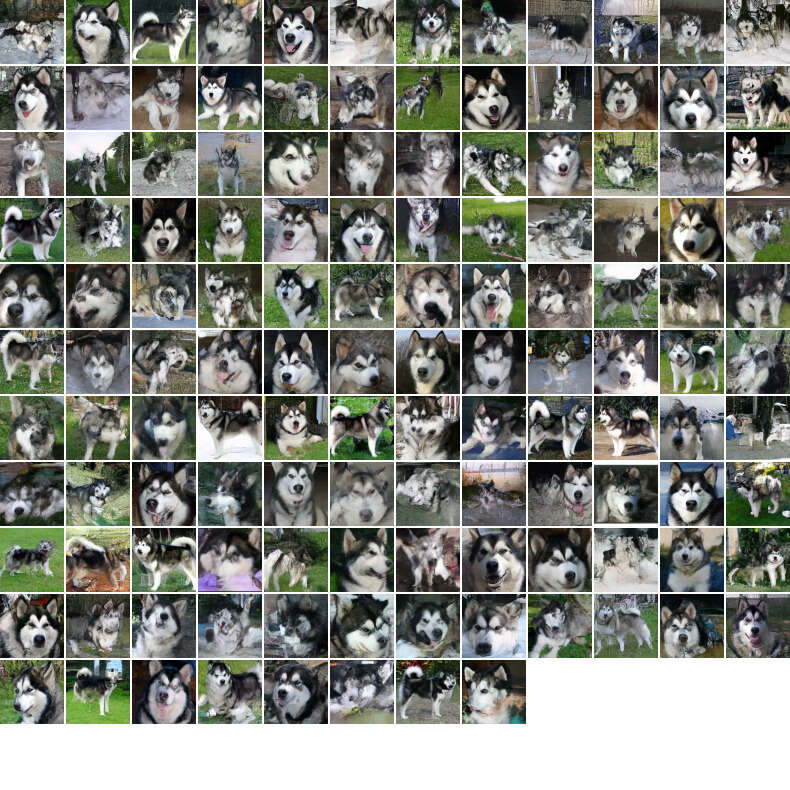}}
    \caption{$w=2$}
\end{subfigure}
\hfill
\begin{subfigure}{0.22\textwidth}
    \adjustbox{width=\linewidth, trim={.0\width} {.5\height} {0.5\width} {0\height},clip}{\includegraphics[width=\linewidth]{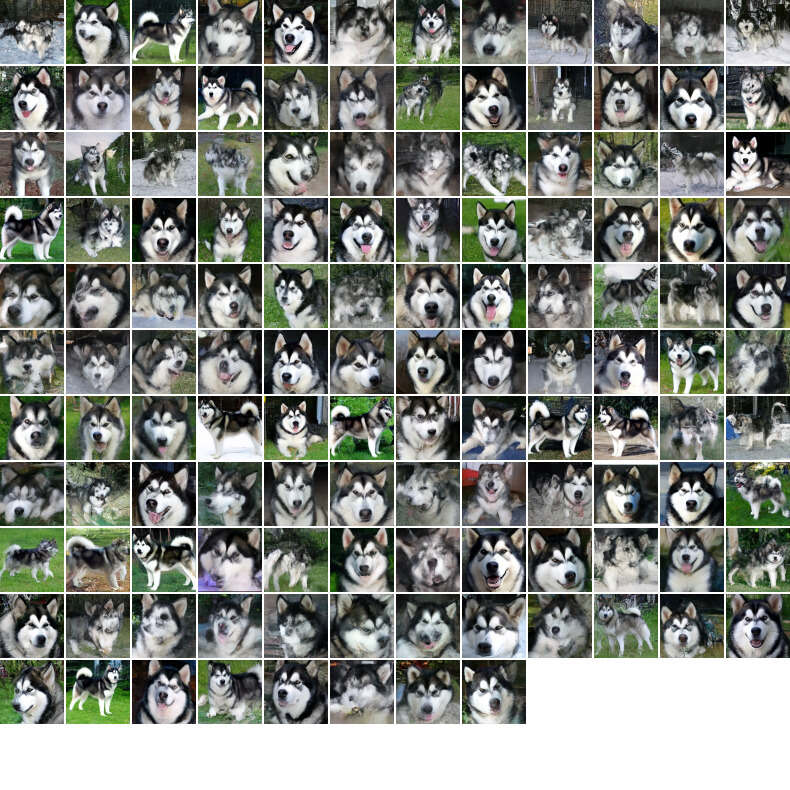}}
    \caption{$w=4$}
\end{subfigure}        
\caption{Ours (stochastic in pixel-space) on ImageNet 64x64. Distilled 1 sampling step. Class-conditioned samples.}
\label{fig:figures}
\end{figure*}